%% file: main.tex
\documentclass[runningheads]{llncs}
\usepackage{graphicx}
\usepackage{amsmath,amssymb} 
\usepackage{color}
\usepackage[width=122mm,left=12mm,paperwidth=146mm,height=193mm,top=12mm,paperheight=217mm]{geometry}

\usepackage{subcaption}
\usepackage{wrapfig}
\usepackage{enumitem}
\usepackage{booktabs}
\usepackage{tabularx}
\usepackage{array}
\usepackage{nicefrac}
\usepackage[export]{adjustbox}
\usepackage{hyperref}
\usepackage{cleveref}
\Crefname{figure}{Fig.}{Figs.}
\Crefname{equation}{Eq.}{Eqs.}
\Crefname{tabular}{Tab.}{Tabs.}
\Crefname{section}{Sec.}{Sections}
\input{macros.tex}
\newcommand{\projectpage}{\url{http://text2shape.stanford.edu/}}

\begin{document}
\pagestyle{headings}
\mainmatter
\def\ECCV18SubNumber{156}  

\title{Text2Shape: Generating Shapes from Natural Language by Learning Joint Embeddings}

\titlerunning{~}

\authorrunning{~}

\author{{Kevin Chen$^{1}$ ~ Christopher B. Choy$^{1}$ ~ Manolis Savva$^{2}$ ~ Angel X. Chang$^{2}$ ~ Thomas Funkhouser$^{2}$ ~ Silvio Savarese$^{1}$}}
\institute{\vspace{-0.5em}{$^{1}$Stanford University \quad\quad\quad $^{2}$Princeton University} \\ \projectpage}

\maketitle

\input{sec/abstract.tex}

\input{sec/intro.tex}

\input{sec/related.tex}
\input{sec/dataset.tex}

\input{sec/method.tex}

\input{sec/experiments.tex}

\input{sec/conclusion.tex}

\mypara{Acknowledgments.}
This material is based upon work supported by the National Science Foundation Graduate Research Fellowship Program under Grant No. DGE – 1147470. Any opinions, findings, and conclusions or recommendations expressed in this material are those of the author(s) and do not necessarily reflect the views of the National Science Foundation.
This work is supported by Google, Intel, and with the support of the Technical University of Munich--Institute for Advanced Study, funded by the German Excellence Initiative and the European Union Seventh Framework Programme under grant agreement n$^{o}$ 291763.

\section*{Appendix}

\appendix
\input{sec/appendix}

\bibliographystyle{splncs}
\bibliography{egbib}

\end{document}

%% file: macros.tex
\newcommand{\rulesep}{\unskip\ \vrule\ }

\newcommand{\todo}[1]{}
\renewcommand{\todo}[1]{{\color{red} TODO: {#1}}}
\newcommand{\mypara}{\vspace*{-3mm}\paragraph}
\newcommand{\denselist}{\setlength{\itemsep}{0pt}\setlength{\parskip}{0pt}\setlength{\parsep}{0pt}\setlength{\partopsep}{0pt}\vspace{-\topsep}}

\newcolumntype{L}[1]{>{\raggedright\let\newline\\\arraybackslash\hspace{0pt}}m{#1}}
\newcolumntype{C}[1]{>{\centering\let\newline\\\arraybackslash\hspace{0pt}}m{#1}}
\newcolumntype{R}[1]{>{\raggedleft\let\newline\\\arraybackslash\hspace{0pt}}m{#1}}

\newcolumntype{M}[1]{>{\centering\arraybackslash}m{#1}}

%% file: sec/abstract.tex
\begin{abstract}
We present a method for generating colored 3D shapes from natural language. To this end, we first learn joint embeddings of freeform text descriptions and colored 3D shapes.
Our model combines and extends learning by association and metric learning approaches to learn implicit cross-modal connections, and produces a joint representation that captures the many-to-many relations between language and physical properties of 3D shapes such as color and shape.
To evaluate our approach, we collect a large dataset of natural language descriptions for physical 3D objects in the ShapeNet dataset.
With this learned joint embedding we demonstrate text-to-shape retrieval that outperforms baseline approaches.
Using our embeddings with a novel conditional Wasserstein GAN framework, we generate colored 3D shapes from text.
Our method is the first to connect natural language text with realistic 3D objects exhibiting rich variations in color, texture, and shape detail.


\keywords{3D reconstruction, generative adversarial networks, text-to-shape retrieval, text-to-shape generation}
\end{abstract}

%% file: sec/intro.tex
\section{Introduction}


Language allows people to communicate thoughts, feelings, and ideas. Research in artificial intelligence has long sought to mimic this component of human cognition. One such goal is to bridge the natural language and visual modalities. Imagine describing a ``round glass coffee table with four wooden legs'', and having a matching colored 3D shape be retrieved or generated.  We refer to these tasks as i) \emph{text-to-shape retrieval} and ii) \emph{text-to-shape generation}.

Systems with such capabilities have many applications in computational design, fabrication, augmented reality, and education. For example, text-to-shape retrieval systems can be useful for querying 3D shape databases (TurboSquid, 3D Warehouse, Yobi3D, etc.) without reliance on human annotation of individual shapes. Likewise, text-to-shape generation can facilitate 3D design. Currently, 3D model designers rely on expensive modeling software (e.g., Maya, 3DS Max, Blender) with steep learning curves, and tedious, time-consuming manual design. A text-to-shape generation system can initialize shapes with basic attributes defined using natural language descriptions. Such a technology can save time and money, and allow inexperienced users to design their own shapes for fabrication.

\input{fig/pull.tex}

To achieve this goal, we need a system that understands natural language and 3D shapes. One way to connect language with shapes is to use a joint embedding space for text and shapes.  While there has been prior work on text to image embedding~\cite{gong2014improving,kiros2014unifying,klein2015associating,reed2016learning,wang2016learning} and image to shape embedding~\cite{li2015joint}, there has been no prior work on text to 3D shape embedding to the best of our knowledge. Moreover, prior approaches in learning text-image representations rely on fine-grained, category-level class or attribute labels~\cite{reed2016learning,reed2016generating,zhang2016stackgan,reed2016learning-what-where}. Such annotations are not only expensive, but also ill-defined: should we classify objects by color, material, or style?  Ideally, we would like to learn a joint embedding of text and 3D shapes directly from natural language descriptions, without relying on fine-grained category or attribute annotations.
However, connecting natural language to 3D shapes is challenging because there is no simple one-to-one mapping between text and 3D shapes (e.g., both ``round table'' and ``circular table'' can describe similar real objects).  Given a shape there are many ways to describe it, and given a natural language description there are many possible shapes that match the description.  


In this paper, we first present a method for learning a joint text and shape representation space directly from natural language descriptions of 3D shape instances, followed by our text-to-shape generation framework.  Unlike related work in text-to-image synthesis~\cite{reed2016learning,reed2016generative,zhang2016stackgan}, we do not rely on fine-grained category-level class labels or pre-training on large datasets.  Furthermore, we train the text and shape encoding components jointly in an end-to-end fashion, associating similar points in our data both within a modality (text-to-text or shape-to-shape) and between the two modalities (text-to-shape).

To do this, we take inspiration from recent work on learning by association~\cite{haeusser2017learning} to establish implicit cross-modal links between similar descriptions and shape instances, and combine this with a metric learning scheme~\cite{song2016metric} which strengthens links between similar instances in each modality (see \Cref{fig:pull}(b)).  Our approach leverages only instance-level correspondences between a text description and a 3D shape to cluster similar descriptions, and induces an attribute-based clustering of similar 3D shapes.  Thus, we obviate the need for expensive fine-grained category or attribute annotations.

We apply our method to text-to-shape retrieval and text-to-shape generation (see \Cref{fig:pull}(c)).  The retrieval task allows us to evaluate the quality of our jointly learned text-shape embedding against baselines from prior work.  The text-to-shape generation task is a challenging new task that we propose. We focus on colored shape generation because most descriptions of shapes involve color or material properties.  To address this task, we combine our joint embedding model with a novel conditional Wasserstein GAN framework, providing greater output quality and diversity compared to a conditional GAN formulation. Lastly, we use vector embedding arithmetic and our generator to manipulate shape attributes.

For realistic and challenging evaluation, we collect 75K natural language descriptions for 15K chair and table shapes in the ShapeNet~\cite{shapenet2015} dataset. To facilitate controlled evaluation, we also introduce a dataset of procedurally generated colored primitives (spheres, pyramids, boxes, etc.) with synthetic text captions.  Our experimental results on these datasets show that our model outperforms the baselines by a large margin for both the retrieval and generation tasks. In summary, our contributions are as follows:
\begin{itemize}\denselist
    \item We propose an end-to-end instance-level association learning framework for cross-modal associations (text and 3D shapes).
    \item We demonstrate that our joint embedding of text and 3D shapes can be used for text-to-shape retrieval, outperforming baseline methods.
    \item We introduce the task of text to colored shape generation and address it with our learned joint embeddings and a novel Conditional Wasserstein GAN.
    \item We create two new datasets of 3D shape color voxelizations and corresponding text: i) ShapeNet objects with natural language descriptions, and ii) procedurally generated geometric primitives with synthetic text descriptions.
\end{itemize}

%% file: fig/pull.tex
\begin{figure}
    \centering
    \includegraphics[width=\textwidth]{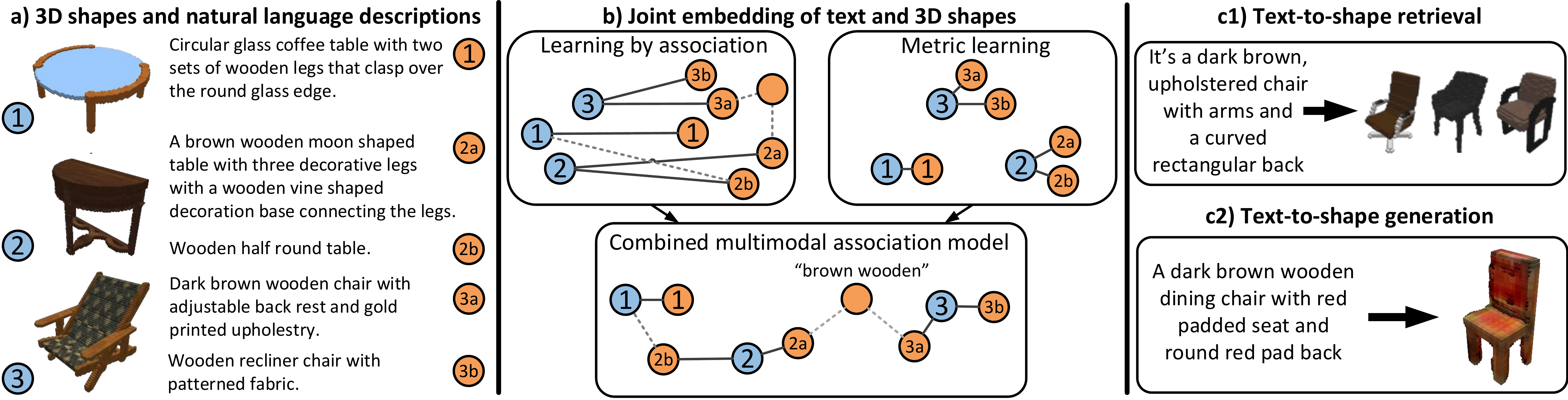}
    \vspace{-0.5em}
    \caption{By leveraging a new dataset of paired natural language descriptions and colored 3D shapes \textbf{(a)}, our method extends learning by association and metric learning to jointly learn a text and 3D shape embedding that clusters similar shapes and descriptions, establishing implicit semantic connections (dashed lines) \textbf{(b)}. We apply our learned embedding to two tasks: text-to-shape retrieval \textbf{(c1)}, where 3D shapes matching a description are retrieved from a dataset, and the challenging new task of text-to-shape generation \textbf{(c2)}, where a novel shape is generated from text.}
    \label{fig:pull}
    \vspace{-1.5em}
\end{figure}

%% file: sec/related.tex
\section{Related Work}

We review work in related areas such as multimodal representation learning, text-to-image synthesis, and 3D shape generation.

\textbf{Learning representations for visual descriptions}.
Reed et al.~\cite{reed2016learning} learn text embeddings which capture visual information. This work was the basis for several text-to-image synthesis works~\cite{zhang2016stackgan,reed2016learning-what-where,reed2016generative}. While this method~\cite{reed2016learning} was applied to the text-to-image problem~\cite{zhang2016stackgan,reed2016learning-what-where,reed2016generative}, it utilizes pre-training on large image datasets. More importantly, it relies on fine-grained category-level labels for each image (e.g. bird species), which are expensive to collect and not always well-defined (e.g., should 3D chairs be classified by color, material, or style?). In contrast, we do not rely on fine-grained category or attribute labels and use no pre-training.

\textbf{Learning multimodal representations}.
Ngiam et al.~\cite{ngiam2011multimodal} present a stacked multimodal autoencoder to learn joint audio and video representations. Srivastava and Salakhutdinov~\cite{srivastava2012multimodal} train a deep Boltzmann machine to model images and text tags. Sohn et al.~\cite{sohn_nips_14} introduce a generative model of multimodal data by hallucinating the missing data modality and minimizing the variation of information. Another class of methods learns joint representations using Canonical Correlation Analysis (CCA)~\cite{hardoon2004canonical}, with many extensions~\cite{Gong:2014:MES:2584252.2584265,gong2014improving,klein2015associating}, including deep CCA~\cite{andrew2013deep}. 
However, Ma et al.~\cite{ma2015finding} point out that this does not scale well to large amounts of data, such as our text and 3D shapes dataset. 
More recent work in text-image joint embeddings~\cite{kiros2014unifying,tsai2017learning} focuses on other aspects such as decoding distributed representations and using unsupervised learning on unlabeled data.
 
\textbf{Metric learning}.
Metric learning using deep neural networks has gained significant attention in recent years. 
Common methods include using a contrastive loss~\cite{chopra2005learning}, triplet loss~\cite{weinberger2006distance,schroff2015facenet}, or N-pair loss~\cite{sohn2016improved}. 
We extend the method of Song et al.~\cite{song2016metric} in our joint representation learning approach. Since our problem is learning a joint metric space across different modalities, our framework is similar in spirit to prior cross-modal metric learning formulations~\cite{Wang_2015_CVPR,wang2016learning,Gordo_2017_CVPR} in that we compute distances both within modality (text-text) and across modalities (text-shape). However, we use a smoothed, triplet-based similarity loss.


\textbf{Generative models and generative adversarial networks (GANs)}.
Generative models are particularly relevant to our text-to-shape generation task. Recent works have shown the capability of neural networks to understand semantics~\cite{dosovitskiy2015learning} and generate realistic images~\cite{goodfellow2014generative,radford2015unsupervised}.
Reed et al.~\cite{reed2016generative} explored the challenging problem of text to image synthesis using GANs. They separately train a text embedding~\cite{reed2016learning}, and then use GANs to generate images. More recent work has improved text-to-image generation~\cite{nguyen2016plug,zhang2016stackgan,reed2016learning-what-where}. StackGAN~\cite{zhang2016stackgan}, for example, focuses on an orthogonal aspect of the problem and can in fact be combined with our approach.
Unlike these works, we do not use fine-grained labels, making our method more scalable. 

\textbf{3D voxel generation}.
Common approaches for 3D voxel generation using deep learning rely on losses such as voxel-wise cross entropy~\cite{choy20163d,sharma2016vconv,girdhar2016learning,tatarchenko2017octree}. GANs have also been used to generate 3D shape voxelizations from noise or shape inputs~\cite{3dgan,liu2017interactive,li2017grass}. Our framework builds upon 3D GANs but performs colored voxel generation from \emph{freeform text} and uses a Wasserstein GAN formulation~\cite{arjovsky2017wasserstein}. Tulsiani et al.~\cite{tulsiani2017multi} explored the related task of image to colored voxel generation, but to our knowledge we are the first to address \emph{text to colored voxel} generation. 

%% file: sec/dataset.tex
\section{Datasets}
\label{sec:dataset}

We address the problem of learning joint representations for freeform text descriptions and 3D shapes (CAD models). To train and evaluate methods on this task, we introduce two new datasets: a dataset of human-designed 3D CAD objects from ShapeNet~\cite{shapenet2015}, augmented with natural language descriptions, and a controlled, procedurally generated dataset of 3D geometric primitives (\Cref{fig:datasets}). 

\mypara{Data representation.}
Our datasets consist of text description -- 3D shape voxel grid pairs $(x, s)$ where $x$ is the textual description of a shape $s \in [0, 1]^{v^3 \cdot c}$. Here, $x$ is an arbitrary length vector of word indices where each word index corresponds to a word in the vocabulary, $v$ is the maximum number of voxels along any single dimension, and $c$ is the number of channels per voxel (4 for RGB + occupancy). 

\input{fig/datasets.tex}

\mypara{ShapeNet objects and natural language descriptions.}
To create a realistic dataset with real 3D objects and natural language descriptions, we use the ShapeNet~\cite{shapenet2015} table and chair object categories (with 8,447 and 6,591 instances, respectively).  These 3D shapes were created by human designers to accurately represent real objects.  We choose the table and chair categories because they contain many instances with fine-grained attribute variations in geometry, color and material.  We augment this shape dataset with 75,344 natural language descriptions (5 descriptions on average per shape) provided by people on the Amazon Mechanical Turk crowdsourcing platform (see \Cref{fig:pull}(a)).
We then produce color voxelizations of the CAD 3D meshes using a hybrid view-based and surface sampling method, followed by downsampling with a low-pass filter in voxel space.  The process for collecting natural language descriptions, the algorithm for producing color voxelizations, and the overall dataset construction are described in more detail in \Cref{sec:dataset-details}.  This large-scale dataset provides many challenging natural language descriptions paired with realistic 3D shapes.

\mypara{Primitive shapes.}
To enable systematic quantitative evaluation of our model, we create a dataset of 3D geometric primitives with corresponding text descriptions.  This data is generated by voxelizing 6 types of primitives (cuboids, ellipsoids, cylinders, cones, pyramids, and tori) in 14 color variations and 9 size variations.  The color and size variations are subjected to random perturbations generating 10 samples from each of 756 possible primitive configurations, thus creating 7560 voxelized shapes.  We then create corresponding text descriptions with a template-based approach that fills in attribute words for shape, size, and color in several orderings to produce sentences such as \emph{``a large red cylinder is narrow and tall''} (see \Cref{sec:dataset-details} for more details).  In total, we generate 192,602 descriptions, for an average of about 255 descriptions per primitive configuration.  Such synthetic text does not match natural language but it does allow for an easy benchmark with a clear mapping to the attributes of each primitive shape.




%% file: fig/datasets.tex
\begin{figure}[t]
	\vspace{-0.5em}
	\captionsetup[subfigure]{justification=centering,labelformat=empty}
	\centering
	\begin{subfigure}[t]{0.12\linewidth}
	    \includegraphics[width=\linewidth]{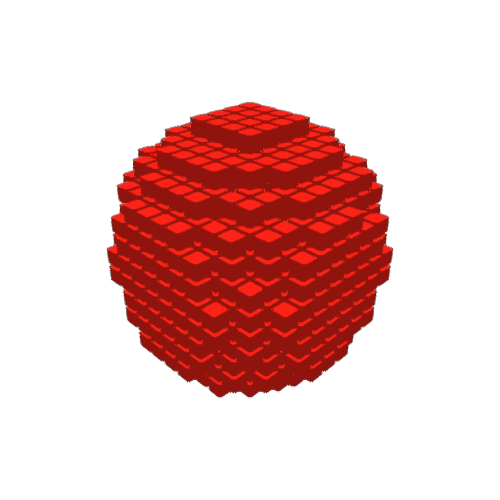}
		\caption{\scriptsize{\emph{A large red sphere}}}
	\end{subfigure}
	\begin{subfigure}[t]{0.14\linewidth}
        \includegraphics[width=\linewidth]{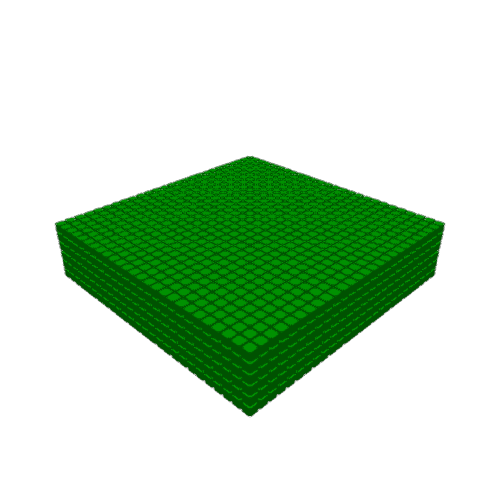}
        \caption{\scriptsize{\emph{A large short wide green box}}}
	\end{subfigure}
	\begin{subfigure}[t]{0.125\linewidth}
		\includegraphics[width=\linewidth]{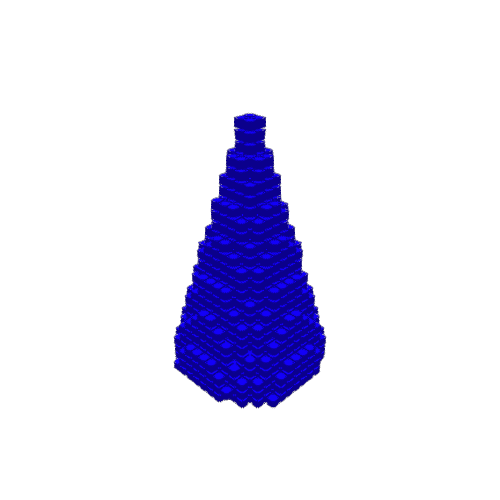}
		\caption{\scriptsize{\emph{The blue cone is large tall}}}
	\end{subfigure}
	\rulesep
	\begin{subfigure}[t]{0.19\linewidth}
		\centering\includegraphics[width=0.4\linewidth]{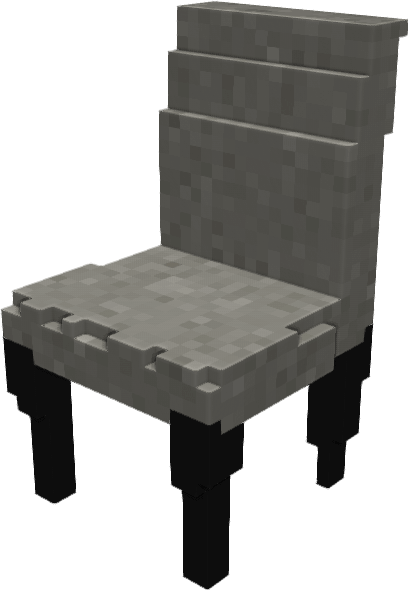}
		\caption{\scriptsize{\emph{A cushioned chair which is grey in color. The legs are small.}}}
	\end{subfigure}
	\begin{subfigure}[t]{0.18\linewidth}
        \centering\adjincludegraphics[width=0.6\linewidth,trim={{.1\width} {.1\height} {.1\width} {.1\height}},clip]{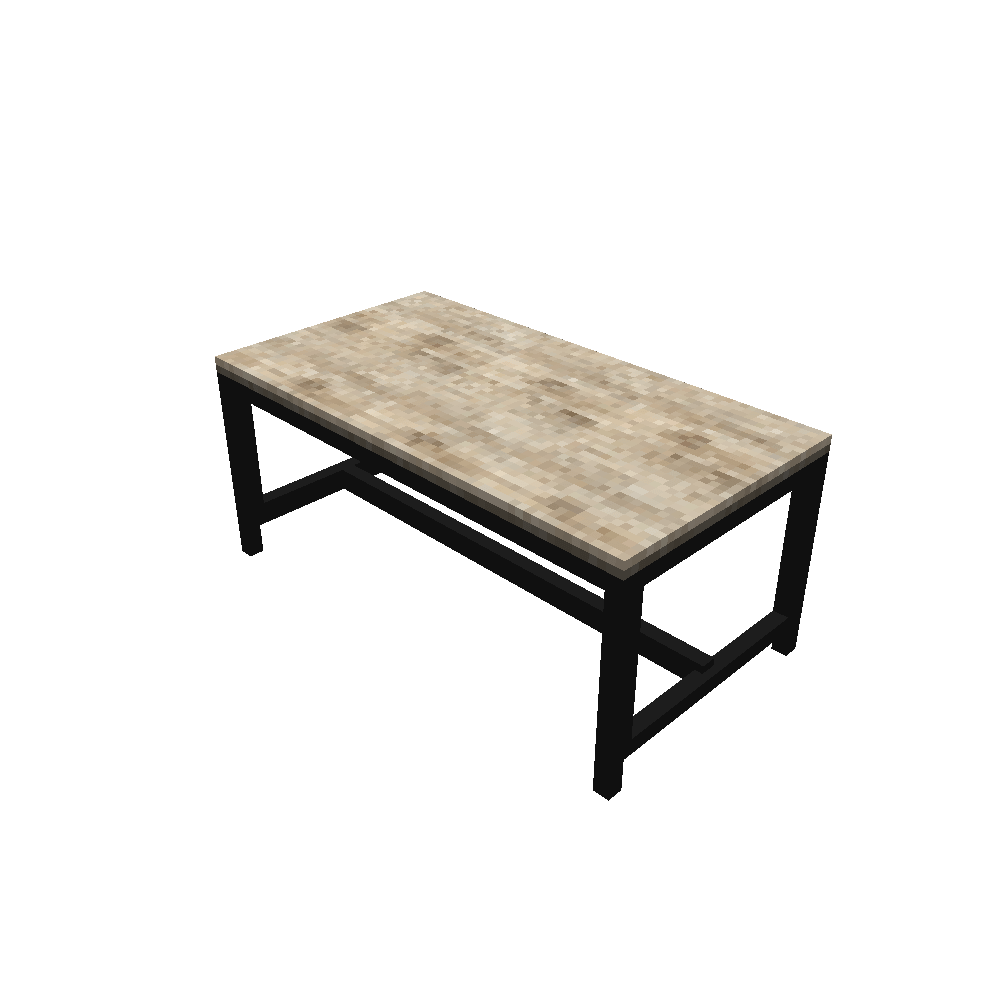}
        \caption{\scriptsize{\emph{A rectangular wooden coffee table with an iron base}}}
	\end{subfigure}
	\begin{subfigure}[t]{0.18\linewidth}
		\centering\includegraphics[width=0.6\linewidth]{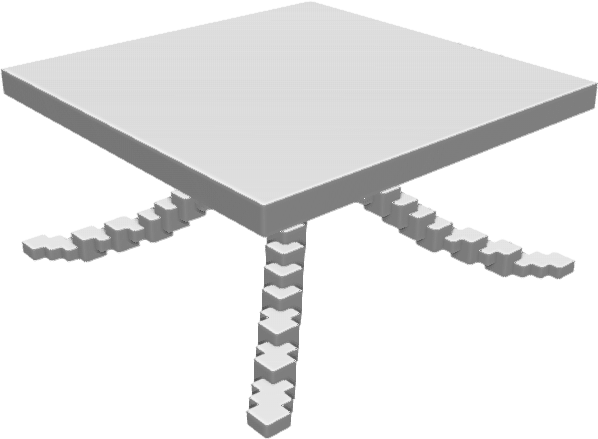}
		\caption{\scriptsize{\emph{White square table with four legs that curve out from the base}}}
	\end{subfigure}
	\vspace{-1em}
	\caption{\textbf{Our proposed 3D voxel -- text datasets.} \textbf{Left:} Procedurally generated primitives dataset with associated generated text descriptions. \textbf{Right:} Voxelizations of ShapeNet 3D CAD models with natural language descriptions.}
	\vspace{-1.5em}
	\label{fig:datasets}
\end{figure}

%% file: sec/method.tex
\section{Joint Text--3D Shape Representation Learning}
\label{sec:rep-learning}

\Cref{fig:pull}(b) shows an illustration of the components of our approach, with explicit text-shape links in solid lines and implicit similarity links in dashed lines.
We would like our joint embedding to: i) cluster similar text together and similar shapes together, ii) keep text descriptions close to their associated shape instance, and iii) separate text from shapes that are not similar.  To address i), we generalize the learning by association approach~\cite{haeusser2017learning} to handle multiple modalities (\Cref{sec:associations}) and to use instance-level associations between a description and a shape (\Cref{sec:instance-assoc}).  By using text-shape-text round trips we establish implicit connections between separate text and shape instances that are semantically similar (e.g. ``round table'' pulling shapes 1 and 2 together in \Cref{fig:pull}).
To enforce properties ii) and iii) in the embedding space, we jointly optimize an association learning objective with metric learning (\Cref{sec:metric-learning}). Our combined multimodal association model anchors text to associated shape instances while allowing latent similarities between different shapes and text to emerge.




\input{fig/mm_assoc.tex}

\subsection{Learning Cross-Modal Associations}
\label{sec:associations}


We generalize the same-modal data association approach of Haeusser et al.~\cite{haeusser2017learning} to learn \emph{cross-modal} text-to-shape associations (see \Cref{fig:mm-assoc}). 
Suppose we are given a batch of $n$ shapes and $m$ descriptions, with potentially multiple descriptions per shape. As shown in \Cref{fig:mm-assoc}(a), each shape and each text description is passed through a shape or text encoder, producing shape embeddings $S$ and text embeddings $T$, with each row representing an individual shape or text embedding. To compute the embeddings, we used a CNN + RNN (GRU) structure similar to Reed et al.~\cite{reed2016generative} for the text encoder, and a 3D-CNN for the shape encoder.

We define the text-shape similarity matrix $M_{ij} = T_i \cdot S_j$ where $T_i$ and $S_j$ are the $i$th and $j$th rows of the corresponding matrices. To convert this into the probability of description $i$ associating with shape $j$ (as opposed to another shape in $S$), we pass these through a softmax layer: $P^{TS}_{ij} = \nicefrac{e^{M_{ij}}}{\sum_{j'} e^{M_{ij'}}}$.
Similarly, we can compute the probability of associating shape $i$ to description $j$ by replacing $M$ with $M^\top$. The round-trip probability for associating description $i$ with certain shapes and then associating those shapes with description $j$ in $T$ is then: $P^{TST}_{ij} = (P^{TS} P^{ST})_{ij}$. We abbreviate this text-shape-text round trip as TST (see \Cref{fig:mm-assoc}(b) dashed box).
For a given description $i$, our goal is to have $P^{TST}_{ij}$ be uniform over the descriptions $j$ which are similar to description $i$. Thus, we define the round-trip loss $\mathcal{L}^{TST}_R$ as the cross-entropy between the distribution $P^{TST}$ and the target uniform distribution.

To associate text descriptions with all possible matching shapes, we also impose a loss on the probability of associating each shape with any description: $P^\text{visit}_j = \nicefrac{\sum_i P^{TS}_{ij}}{m}$.  We maximize the entropy on this distribution $P^H$, using an entropy loss $\mathcal{L}^{TST}_H$ computed from the cross entropy between the $P^\text{visit}_j$ and the uniform distribution over the shapes.  Thus, our cross-modal association loss is:
\begin{align}
    \label{eq:tst-loss}
    \mathcal{L}^{TST} = \mathcal{L}^{TST}_R + \lambda \mathcal{L}^{TST}_H
\end{align}






\subsection{Instance-Level Associations}
\label{sec:instance-assoc}

The approach described in Sec.~\ref{sec:associations} relies on fine-grained category-level classification labels for each text description (e.g. a label for all brown chairs and a different label for all blue chairs). These fine-grained annotations are expensive to obtain and are not necessarily well-defined for 3D shapes from ShapeNet~\cite{shapenet2015} where descriptions can contain multiple colors or materials. 
Thus, we extend the above approach to work without such fine-grained \emph{category-level} annotations, making the method more scalable and practical.

We assume each shape belongs to its own \emph{instance-level} class containing its corresponding descriptions. 
These instance-level labels are appropriate for the association learning approach (\Cref{sec:associations}) since it does not explicitly separate descriptions and shapes not labeled to correspond to each other. Thus, our cross-modal association learning is more relaxed than previous approaches~\cite{reed2016learning,Wang_2015_CVPR,wang2016learning} that use direct supervision over the cross-modal relationships, allowing us to learn cross-modal associations between descriptions and shapes that were not labeled to correspond with each other (e.g. two similar chairs, each with its own text descriptions).

Moreover, we extend the association learning approach with additional round trips as shown in \Cref{fig:mm-assoc}(b). Our experiments show this greatly improves the performance of our model for the multimodal problem. 
In addition to enforcing a loss over the text-shape-text (TST) round trip, we impose a shape-text-shape (STS) loss $\mathcal{L}^{STS}$ which takes a form similar to \Cref{eq:tst-loss}. This ensures that the shapes are associated with all of the text descriptions and provides additional supervision, which is beneficial when our dataset has only instance-level labels. 

\subsection{Multimodal Metric Learning}
\label{sec:metric-learning}

To facilitate learning cross-modal associations, we impose metric learning losses. This serves as a bootstrapping method which improves the cross-modal associations significantly compared to using either association learning~\cite{haeusser2017associative} or metric learning~\cite{song2016metric} (ML) alone.

As in \Cref{sec:associations}, we define similarity between embeddings with the dot product.
Given a triplet $(x_i, x_j, x_k)$ of text description embeddings in which $(x_i, x_j)$ belong to the same instance class (positive pair) and $(x_i, x_k)$ belong to different instance classes (negative pair), our metric learning constraint is:
$F(x_i; \theta) \cdot F(x_j; \theta) > F(x_i; \theta) \cdot F(x_k;\theta) + \alpha$, 
where $F$ maps to the metric space and $\alpha$ is the margin. We use a variation of metric learning that uses a soft max to approximate hard negative mining~\cite{song2016metric} and combine it with the similarity constraint above in our text-to-text loss: 
\begin{equation}
    \label{eq:inverted_loss}
    \mathcal{L}^{TT}_{ML} = 
    \frac{1}{2 | \mathcal{P} |} \sum_{(i, j) \in \mathcal{P}} \left[ \log \left(V_i + V_j \right) - m_{i, j} \right]_+^2
\end{equation}
where $m_{i,j}$ denotes the similarity between $x_i$ and $x_j$, and $V_l = \sum_{k \in \mathcal{N}_l} \exp\{ \alpha + m_{l, k} \}$. Here, $\mathcal{N}_l$ is a negative set, defined as the set of indices that belong to an instance class other than the class $l$ is in, and $\mathcal{P}$ is a positive set in which both indices $i$ and $j$ belong to the same instance class.

The above approach can cluster similar text descriptions together and separate dissimilar text descriptions. To extend this for cross-modal similarities, we define cross-modal constraints with anchors from each modality (see \Cref{fig:mm-assoc}(c)): $F(x_i; \theta) \cdot F(y_j; \theta) > F(x_i; \theta) \cdot F(y_k;\theta) + \alpha$, 
where $x$ represents text embeddings and $y$ represents shape embeddings. The corresponding text-to-shape loss $\mathcal{L}_{ML}^{TS}$ can be derived similarly to \Cref{eq:inverted_loss}. 







\subsection{Full Multimodal Loss}
\label{sec:combined}

We combine the association losses with the metric learning losses to form the final loss function used to train the text and shape encoders:
\begin{equation}
    \mathcal{L}_{total} = \mathcal{L}^{TST} + \mathcal{L}^{STS} + \gamma (\mathcal{L}^{TT}_{ML} + \mathcal{L}^{TS}_{ML})
\end{equation}
Note that the embeddings are not restricted to have unit norm, since this can make optimization difficult~\cite{sohn2016improved}. To avoid degeneracy, we use a variation of the approach used by Sohn et al.~\cite{sohn2016improved} and regularize the $L^2$ norm of the embedding vectors only when they exceed a certain threshold. 

\section{Generating Colored 3D Shapes from Text}
\label{section:t2s}

We apply our joint representation learning approach to text-to-shape generation using generative adversarial networks. 
The model is comprised of three main components: the text encoder, the generator, and the critic. The text encoder maps the text into latent representations. The latent vectors are concatenated with a noise vector and then passed to the generator, which constructs the output shapes. Lastly, the critic determines both how realistic the generated outputs look and how closely they match with the corresponding text descriptions. Details for the architecture are in \Cref{sec:model-details}.







GANs are ideal for text-to-shape generation because they encourage the model to generate outputs with the correct properties while avoiding the use of per-voxel constraints. This is crucial because the same description should be able to generate different shapes which all capture the specified attributes. Penalizing the model for generating shapes which capture the correct attributes but do not match a specific model instance would be incorrect. Thus, the conditional GAN plays a key role in our text-to-shape framework.

Our formulation is based on the Wasserstein GAN~\cite{arjovsky2017wasserstein,gulrajani2017improved} which improves output diversity while avoiding mode collapse issues with traditional GANs. To our knowledge, we present the first conditional Wasserstein GAN. Rather than using the traditional conditional GAN~\cite{mirza2014conditional} approach, we sample batches of matching and mismatching text descriptions and shapes, in line with Reed et al.~\cite{reed2016generative}. In particular, we sample matching text-shape pairs, mismatching text-shape pairs, and text for the generator from the marginal distribution $p_\mathcal{T}$. The critic evaluates not only how realistic the shapes look but also how well they correspond to the descriptions. Thus, our objective is the following:
\begin{equation}
\begin{split}
    \mathcal{L}_\text{CWGAN} = &\mathbb{E}_{t \sim p_\mathcal{T}}[D(t, G(t))] + \mathbb{E}_{(\tilde t, \tilde s) \sim p_{\text{mis}}}[D(\tilde t, \tilde s)] \\
    &- 2 \mathbb{E}_{(\hat t, \hat s) \sim p_\text{mat}}[D(\hat t, \hat s)] + \lambda_{GP} \mathcal{L}_{GP}
\end{split}
\end{equation}
\begin{equation}
\begin{split}
    \mathcal{L}_{GP} = & \mathbb{E}_{(\bar t, \bar s) \sim p_{GP}}[(\lVert \nabla_{\bar t} D(\bar t, \bar s) \rVert_2 - 1)^2 + (\lVert \nabla_{\check s} D(\bar t, \bar s) \rVert_2 - 1)^2]
    \end{split}
\end{equation}
where $D$ is the critic, $G$ is the generator, and $p_{\text{mat}}$ and $p_{\text{mis}}$ are matching text-shape and mismatching text-shape pairs, respectively.  For enforcing the gradient penalty, rather than sampling along straight lines as in Guljarani et al.~\cite{gulrajani2017improved}, we randomly choose samples ($p_{GP}$) from the real distribution $p_{mat}$ (with probability 0.5) or from the generated fake voxelizations. Note that $t$ refers to text embeddings (\Cref{sec:rep-learning}) concatenated with randomly sampled noise vectors.

%% file: fig/mm_assoc.tex
\begin{figure}[t]
    \vspace{-0.5em}
    \includegraphics[width=\linewidth]{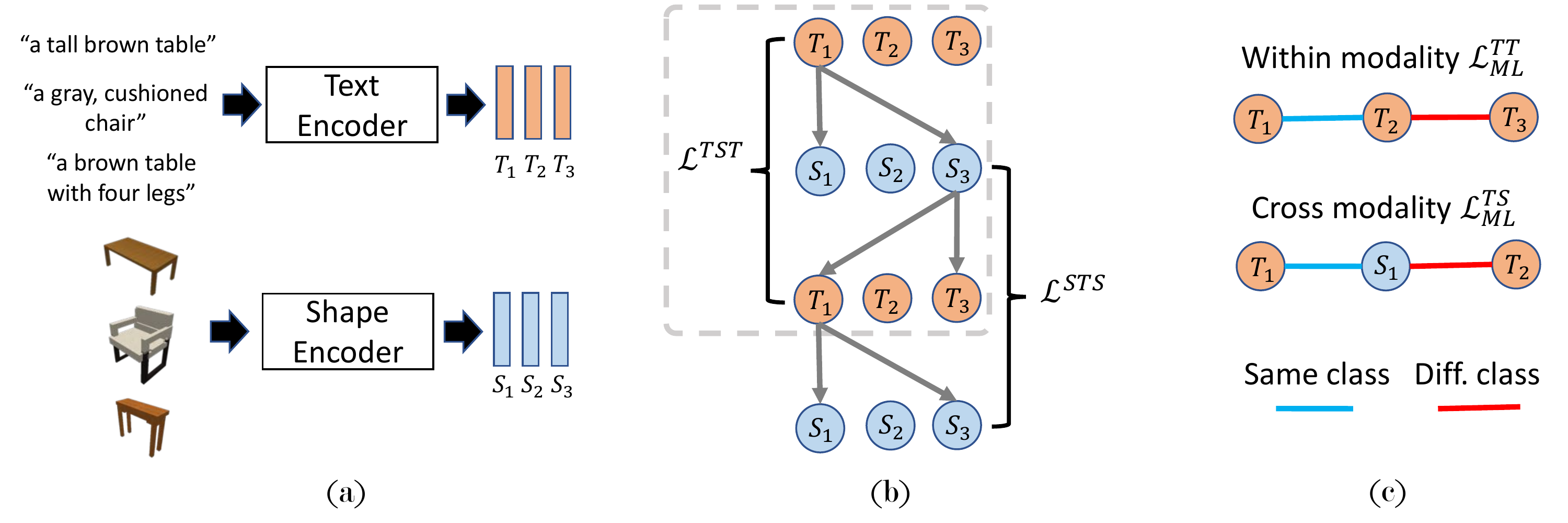}
    \vspace{-2.0em}
    \caption{\textbf{Overview of our joint representation learning approach.} We use text and shape encoders to compute embeddings (\textbf{a}). Instances of each modality make round trips back to similar instances of the same modality, establishing cross-modal associations (\textbf{b}). Metric learning similarities are computed within and across modalities, with embeddings belonging to the same class being pushed together and embeddings in different classes being pushed apart (\textbf{c}).
    }
    \label{fig:mm-assoc}
    \vspace{-1.5em}
\end{figure}

%% file: sec/experiments.tex
\section{Experiments}

We evaluate our learned joint representation through retrieval and generation experiments. In \Cref{sec:retrieval} we quantitatively evaluate the representations on the primitives dataset for text-to-shape retrieval, and show qualitative results on the ShapeNet dataset. In \Cref{sec:generation-results} we compare our text-to-shape generation results with baselines. Lastly, we visualize different dimensions of the learned embeddings and show generation results using vector arithmetic in \Cref{sec:analysis}.
We create train/val/test splits by random subsets of 80\%/10\%/10\% for both the primitives and the ShapeNet objects datasets.
We tokenize and lemmatize all input text using spaCy~\cite{spacy}.
For all experiments described here, we present results on the test split.
Our code, data, and pretrained models are published at \projectpage.

\subsection{Retrieval Task}
\label{sec:retrieval}

We evaluate on the text-to-shape retrieval task, where we are given a sentence and we retrieve the nearest shapes in the joint embedding space. Ideally, we would like to see that the retrieved shapes closely correspond with the query sentence. Here, we show results for the text-to-shape retrieval task. Additional experiments for text-to-text, shape-to-shape, and shape-to-text retrieval are in \Cref{sec:additional-retrieval-results}. Our goal in these experiments is to verify that our method produces reasonable embeddings that cluster semantically similar descriptions and shapes using only text-shape instance-level correspondences. 

The primitives dataset is constructed such that each shape-color-size configuration is a single class with text describing only that class and no others.
This admits a quantitative retrieval evaluation where the goal is to retrieve descriptions and shapes belonging to the same configuration.
For the ShapeNet dataset it is possible that a description could apply to multiple shapes (e.g. ``brown chair''), but we only have ground truth association for one shape.  Thus, we show quantitative results for the retrieval task on the primitives dataset (\Cref{tab:text-shape-retrieval}) and qualitative results for the ShapeNet dataset (\Cref{fig:cross-modal-retrieval}).  Quantitative and qualitative ShapeNet retrieval results comparing our method with baselines are included in \Cref{sec:additional-retrieval-results}.




\input{tab/retrieval-text-shape-wide.tex}

\input{fig/cross-modal-retrieval.tex}

We compare our model with the deep symmetric structured joint embedding (DS-SJE) by Reed et al.~\cite{reed2016learning} and the original formulation of learning by association~\cite{haeusser2017learning} with only TST round trips (LBA-TST). We also decompose our full model into the following: association learning only with both round trips (LBA-MM), metric learning (ML) only, combined LBA and ML with TST round trip only (Full-TST), and the full combined model (Full-MM). MM for ``multimodal'' denotes using both TST and STS losses. We measure retrieval performance using normalized discounted cumulative gain (NDCG)~\cite{jarvelin2002cumulated}, a commonly used information retrieval metric, and recall rate (RR@$k$)~\cite{song2016metric,Wang_2017_ICCV}, which considers a retrieval successful if at least one sample in the top $k$ retrievals is of the correct class.




\Cref{tab:text-shape-retrieval} shows that our full model achieves the best performance. We see that the DS-SJE method~\cite{reed2016learning}, which requires a pre-trained model, works well, as it directly optimizes for this objective. Our association learning--only approach with additional round trips (LBA-MM) performs significantly better, largely due to the simplistic nature of the primitives dataset. We found that for the ShapeNet dataset, the LBA-MM model struggled to learn a joint embedding space of text and shapes (see appendix for more details). Our full model (Full-MM) performs the best on both the primitives and ShapeNet dataset. \Cref{fig:cross-modal-retrieval} shows that the retrieved shapes generally match the query sentence. These examples show that our encoders cluster together similar descriptions and similar shapes despite the fact that retrieved neighbors come from different instances.

\subsection{Text-to-Shape Generation Task}
\label{sec:generation-results}

Can our learned joint representation be used for shape synthesis? To answer this question, we apply our learned embedding to the novel task of text-to-shape generation. We compare our representation learning and Wasserstein GAN approach (CWGAN) with the GAN-INT-CLS text-to-image synthesis method~\cite{reed2016generative}, which uses DS-SJE~\cite{reed2016learning} with the coarse chair/table labels for learning embeddings. We also compare with using our joint embeddings in conjunction with GAN-CLS (using the sampling method described in Sec.~\ref{section:t2s}), which we abbreviate as CGAN since it is a variant of the conditional GAN~\cite{mirza2014conditional}. We introduce the following metrics for quantitative evaluation of the generated shapes:
\begin{itemize}\denselist
    \item \textbf{Occupancy: IoU.} Mean intersection-over-union (IoU) between generated voxels and ground truth shape voxels, indicating the quality of the occupancy of the generated shapes, regardless of color. We threshold the outputs to be occupied if the probability is above $0.9$. 
    Models with similar shapes are more likely to have higher IoU than dissimilar models.
    \item \textbf{Realism: inception score.} We train a chair/table shape classifier and compute the inception score~\cite{salimans2016improved} to quantify how realistic the outputs look. 
    \item \textbf{Color: Earth Mover's Distance (EMD).} This metric measures how well generated colors match the ground truth. We downsample voxel colors in HSV space and compute the Earth Mover's Distance between the ground truth and the generated hue/saturation distributions using L1 as the ground distance. Only occupied voxels are used for computing the color distributions.
    \item \textbf{Color/occupancy: classification accuracy.} Accuracy of whether the generated shape class matches with the ground truth based on a shape classifier.
\end{itemize}

\Cref{tab:generation_results} shows that the CGAN model trained with our embedding outperforms the GAN-INT-CLS on all metrics. The CWGAN model further improves on all metrics except class accuracy. Qualitatively, \Cref{fig:text2shape_results} shows that the GAN-INT-CLS text-to-image synthesis method~\cite{reed2016generative} struggles to generate color and occupancy distributions which correspond with the text. This is largely attributed to poor embeddings learned using coarse chair/table labels. Our CGAN achieves comparable conditioning and generation results without relying on additional class labels, but there are still significant structural and color artifacts. Lastly, our CWGAN generates the most realistic shapes with more natural color distributions and better conditioning on the text compared to the baselines. For example, we correctly generate a chair with red padding on the seat and backrest in \Cref{fig:text2shape_results} (first row). Likewise, the table in the 4th row has multiple layers as desired. However, there is room for improvement: the table is not fully supported by its legs, and the circular table (row 2) has erratic coloring.

\input{tab/generation.tex}

\input{fig/text2shape_results.tex}

\mypara{Manipulating shape attributes with text.}
We demonstrate our method learns embeddings that connect natural language and 3D shape attributes. \Cref{fig:shape-editing} shows how output tables and chairs change to reflect different attributes in the input descriptions. As we vary the description to say ``white'' instead of ``wooden'', or ``rectangular'' instead of ``round'', the outputs change color and shape, respectively. Note that we do not enforce consistency between outputs; each result is independently generated. However, changing a single attribute often results in small changes to other attributes.

\input{fig/shape_editing.tex}




\subsection{Analysis of Learned Representation}
\label{sec:analysis}
\vspace{0.5em}
\mypara{Activation visualization.}
We visualize the representations learned by our model to demonstrate that they capture shape attributes. Using the text and shape encoders, we compute embeddings for ShapeNet test set objects to show that individual dimensions in our representation correlate with semantic attributes such as category, color, and shape.

\input{fig/activation_vis.tex}

\Cref{fig:activations} shows a saturated bar in each row representing the activation in a given dimension of the text embedding for the shape above it. 
In the top-left row, we see that one dimension is sensitive to the shape category (chair vs. table). In the subsequent rows, we see how other dimensions are sensitive to physical shape (rectangular vs. round tables) and color (white vs. brown vs. red). Further research in this direction to disentangle representations could enable generating shapes by composing attributes.

\input{fig/vector-arithmetic-combined.tex}

\mypara{Vector arithmetic.}
Mikolov et al.~\cite{mikolov2013distributed} demonstrated vector arithmetic on text embeddings (e.g., `King' - `Man' + `Woman' $\approx$ `Queen'). Following this, work on GANs (such as Radford et al.~\cite{radford2015unsupervised}) evaluated their models by vector arithmetic to show a meaningful representation space was learned.
We similarly show vector arithmetic results in \Cref{fig:vector-arithmetic}. The generator interprets the manipulated embeddings to produce shapes with the desired properties. For example, ``round'' is expressed by subtracting a rectangular object from a round one, and adding it to an existing rectangular shape (of a different color) produces a round version of the shape. Moreover, we can also perform arithmetic on a mixture of shape and text embeddings. These results illustrate our learned representation encodes shape attributes in a structured manner. Our model implicitly connects descriptions and shapes based on semantic attributes despite no explicit labeling of attribute relations between shape instances.






%% file: tab/retrieval-text-shape-wide.tex
\begin{table*}
\vspace{-2.5em}
\centering
\caption{\textbf{Text-to-shape retrieval evaluation on primitives dataset.} Our approach significantly outperforms baselines from prior work.}
\label{tab:text-shape-retrieval}
\footnotesize
\begin{tabular}{l|ccccccc}
\toprule
\textbf{Metric} & Random & DS-SJE~\cite{reed2016learning} & LBA-TST~\cite{haeusser2017learning} & ML & LBA-MM & Full-TST & Full-MM \\
\midrule
RR@1 & 0.24 & 81.77 & 5.06 & 25.93 & 91.13 & 94.24 & \textbf{95.07} \\
RR@5 & 0.76 & 90.70 & 15.29 & 57.24 & 98.27 & 97.55 & \textbf{99.08} \\
NDCG@5 & 0.27 & 81.29 & 5.92 & 25.00 & 91.90 & 95.20 & \textbf{95.51} \\
\bottomrule
\end{tabular}
\vspace{-1em}
\end{table*}

%% file: fig/cross-modal-retrieval.tex
\begin{figure}
    \vspace{-2.5em}
    \begin{center}
        \includegraphics[width=0.7\linewidth]{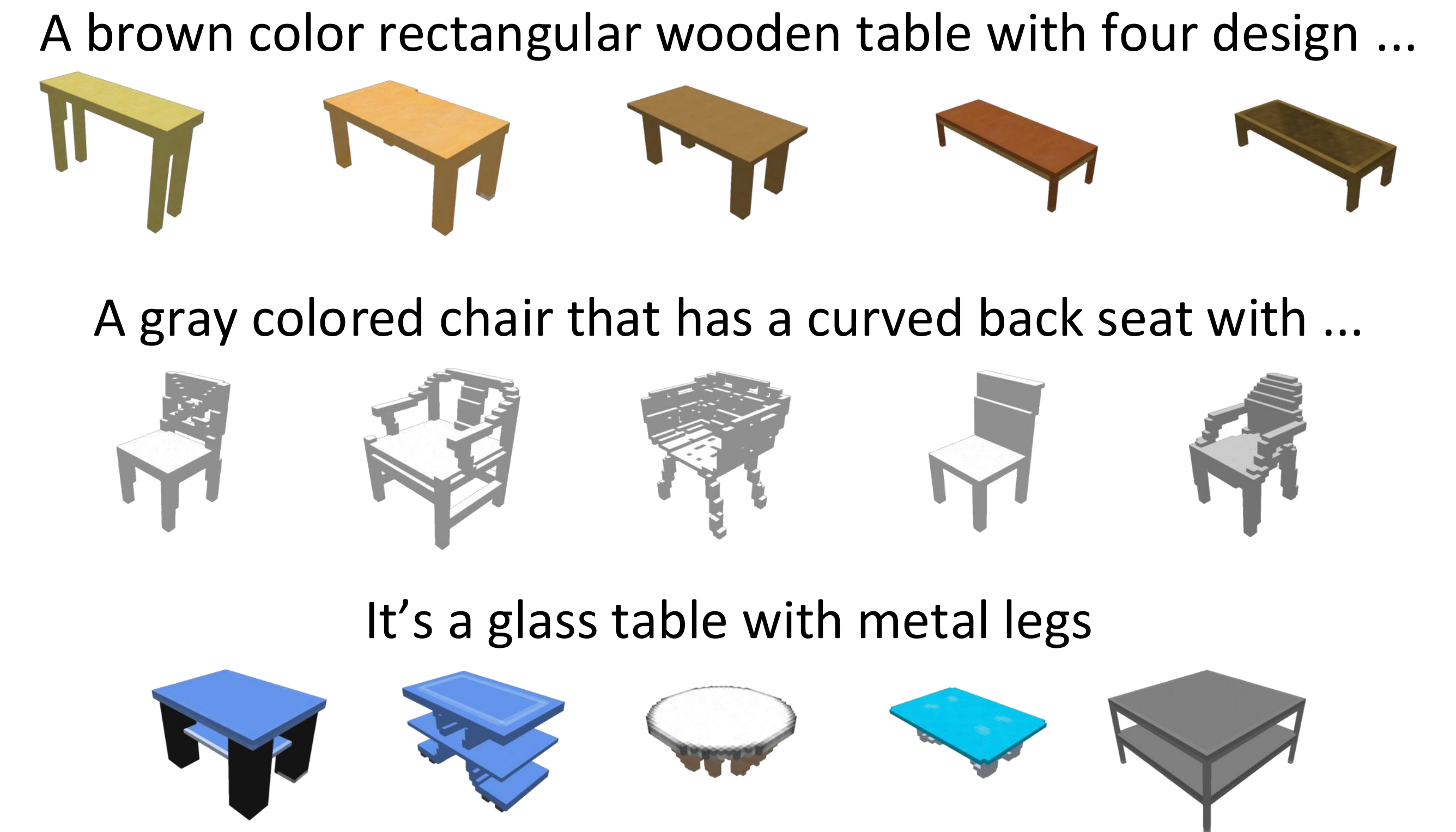}
        \vspace{-1em}
        \caption{\textbf{Text-to-shape retrieval.} Each row shows the five nearest neighbors to the text in our learned embedding, which match in category, color, and shape.}
        \label{fig:cross-modal-retrieval}
    \end{center}
    \vspace{-2.5em}
\end{figure}

%% file: tab/generation.tex
\begin{table}
\vspace{-2.5em}
\small
\begin{center}
\caption{\textbf{Text-to-shape generation evaluation on ShapeNet dataset.}}
\label{tab:generation_results}
\begin{tabular}{lcccc}
\toprule
Method & IoU$\uparrow$ & Inception$\uparrow$ & EMD$\downarrow$ & Class Acc.$\uparrow$ \\
\midrule
GAN-INT-CLS~\cite{reed2016generative} & 9.51 & 1.95 & 0.5039 & 95.57 \\
Ours (CGAN) & 6.06 & 1.95 & 0.4768 & \textbf{97.48} \\
Ours (CWGAN) & \textbf{9.64} & \textbf{1.96} & \textbf{0.4443} & 97.37 \\
\bottomrule
\end{tabular}
\end{center}
\vspace{-3em}
\end{table}


%% file: fig/text2shape_results.tex
\begin{figure}
    \vspace{-2em}
    \centering\scriptsize
    \begin{tabular}{m{6cm}M{1.3cm}M{1.3cm}M{1.3cm}M{1.3cm}}
    Input Text & GAN-INT-CLS~\cite{reed2016generative} & Ours CGAN & Ours CWGAN & GT
    \\\toprule
    Dark brown wooden dining chair with red padded seat and round red pad back. &
    \adjincludegraphics[width=\linewidth,trim={{.05\width} {.05\height} {.05\width} {.05\height}},clip]{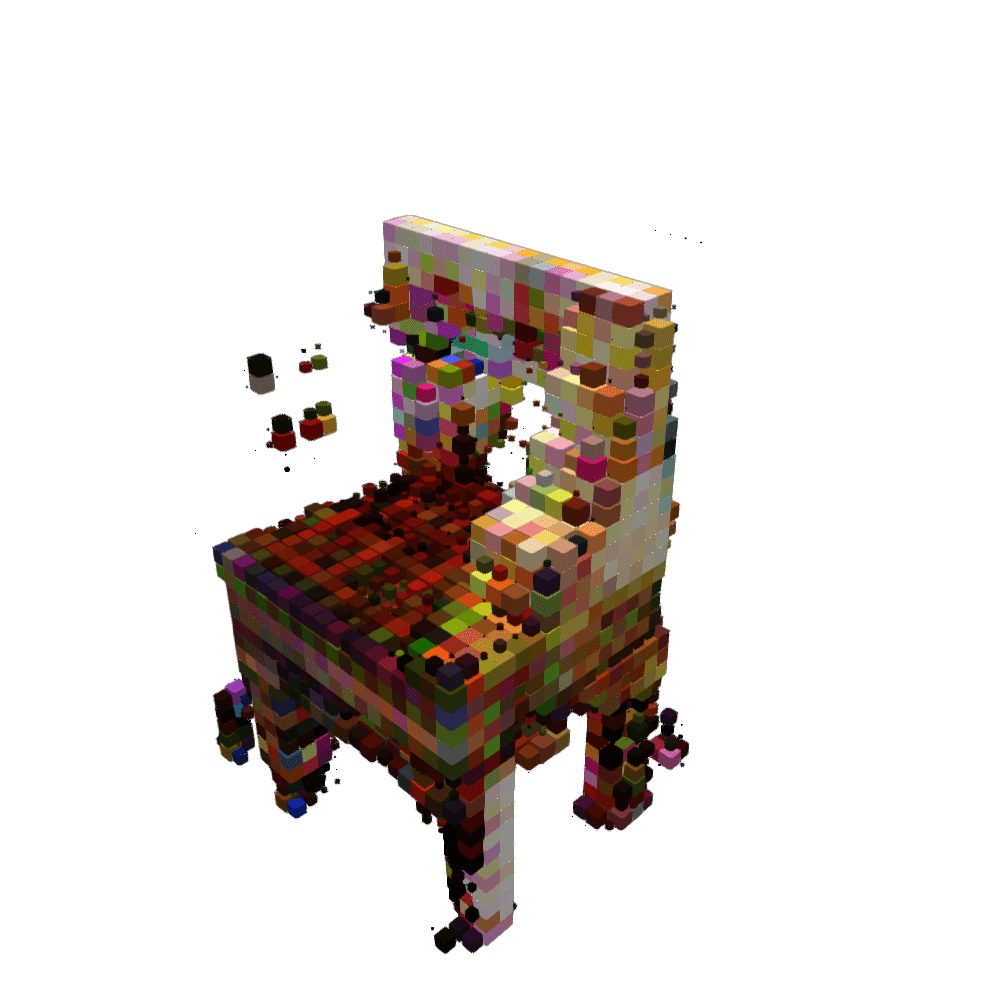} &
    \adjincludegraphics[width=\linewidth,trim={{.05\width} {.05\height} {.05\width} {.05\height}},clip]{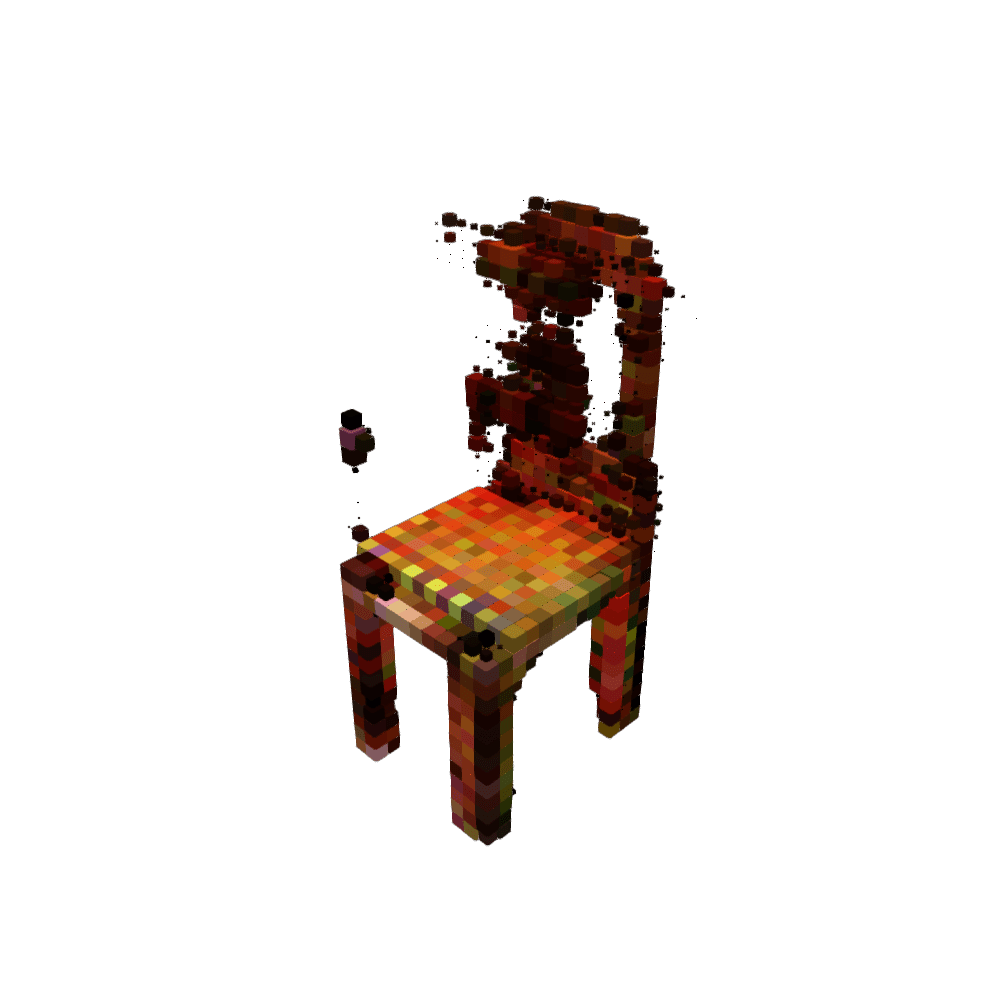} &
    \adjincludegraphics[width=\linewidth,trim={{.05\width} {.05\height} {.05\width} {.05\height}},clip]{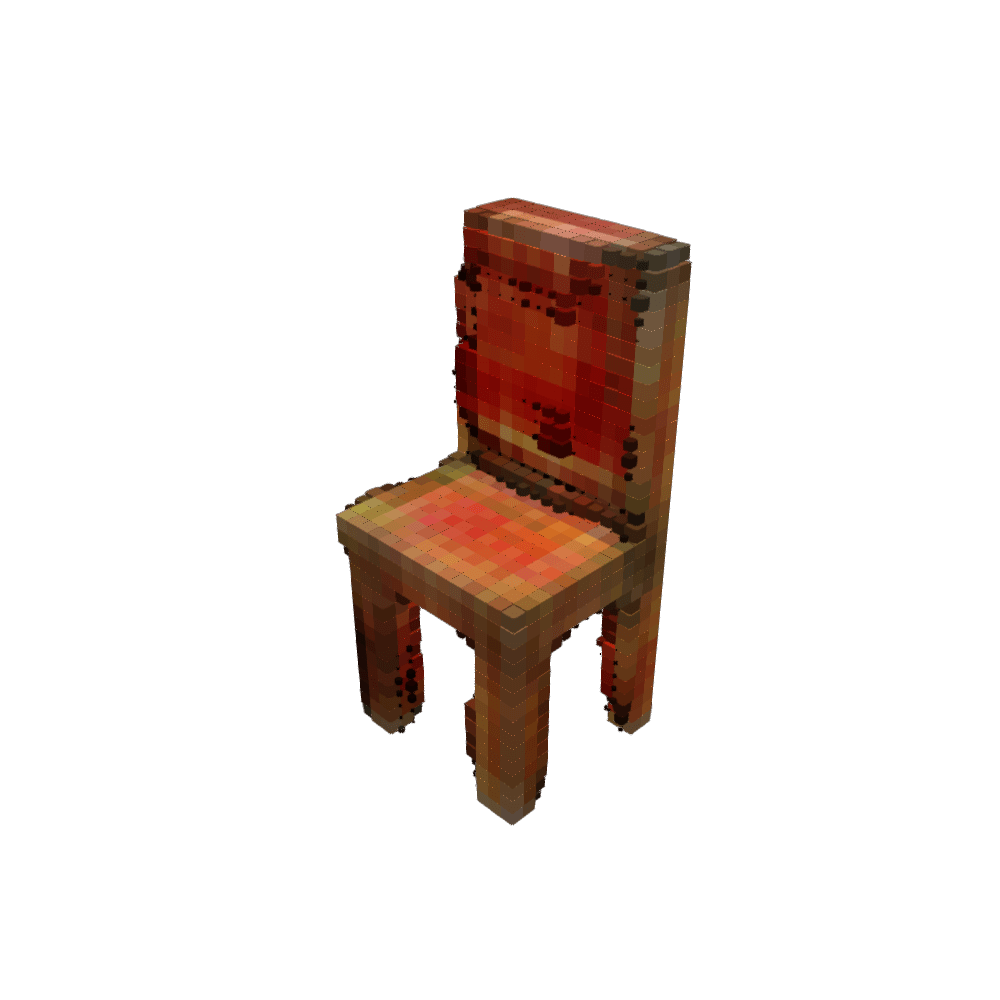} &
    \adjincludegraphics[width=\linewidth,trim={{.05\width} {.05\height} {.05\width} {.05\height}},clip]{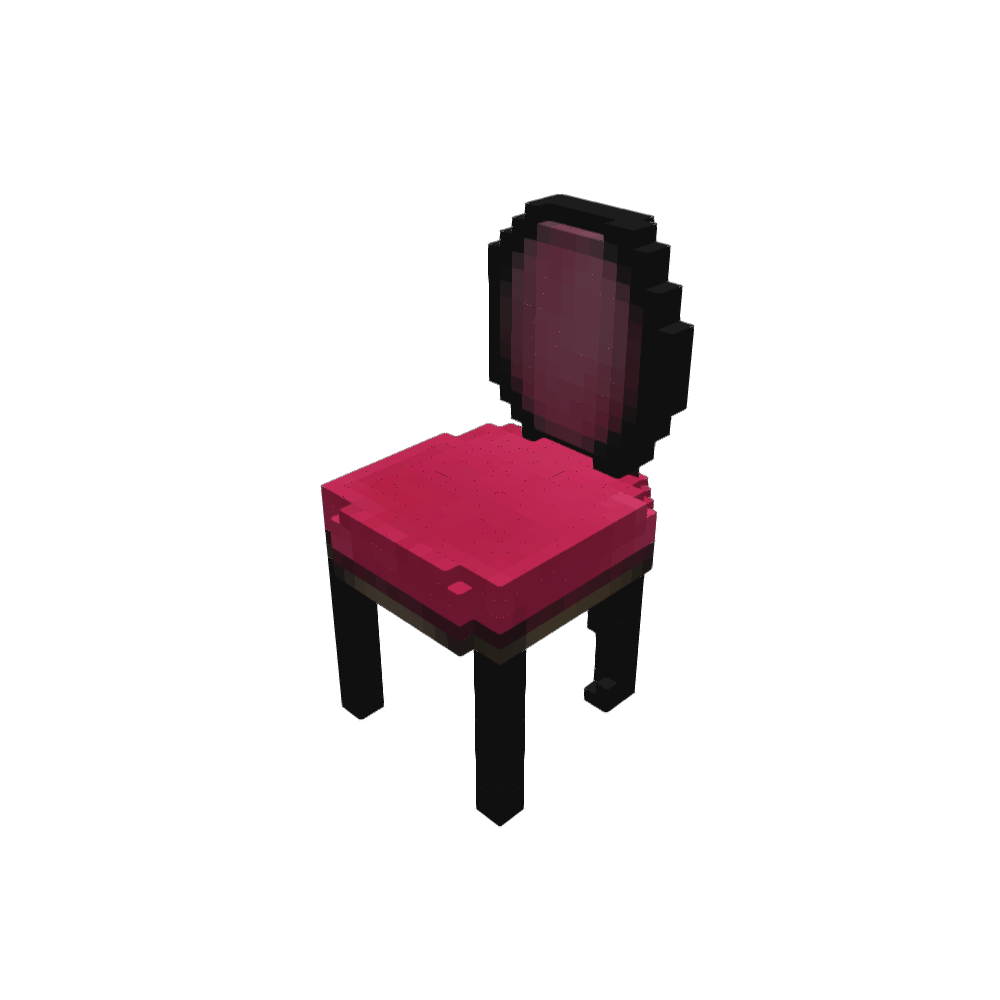}
    \\\midrule
    Circular table, I would expect to see couches surrounding this type of table. &
    \adjincludegraphics[width=\linewidth,trim={{.05\width} {.05\height} {.05\width} {.05\height}},clip]{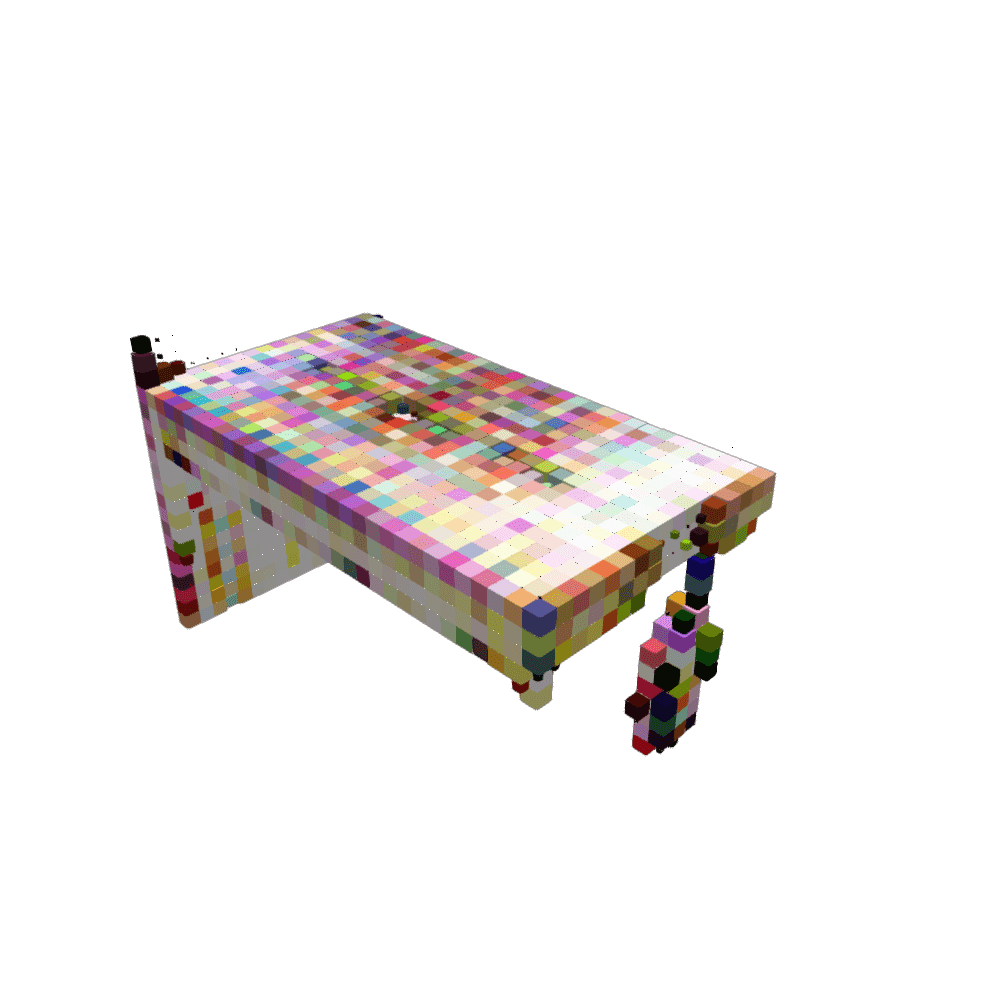} &
    \adjincludegraphics[width=\linewidth,trim={{.05\width} {.05\height} {.05\width} {.05\height}},clip]{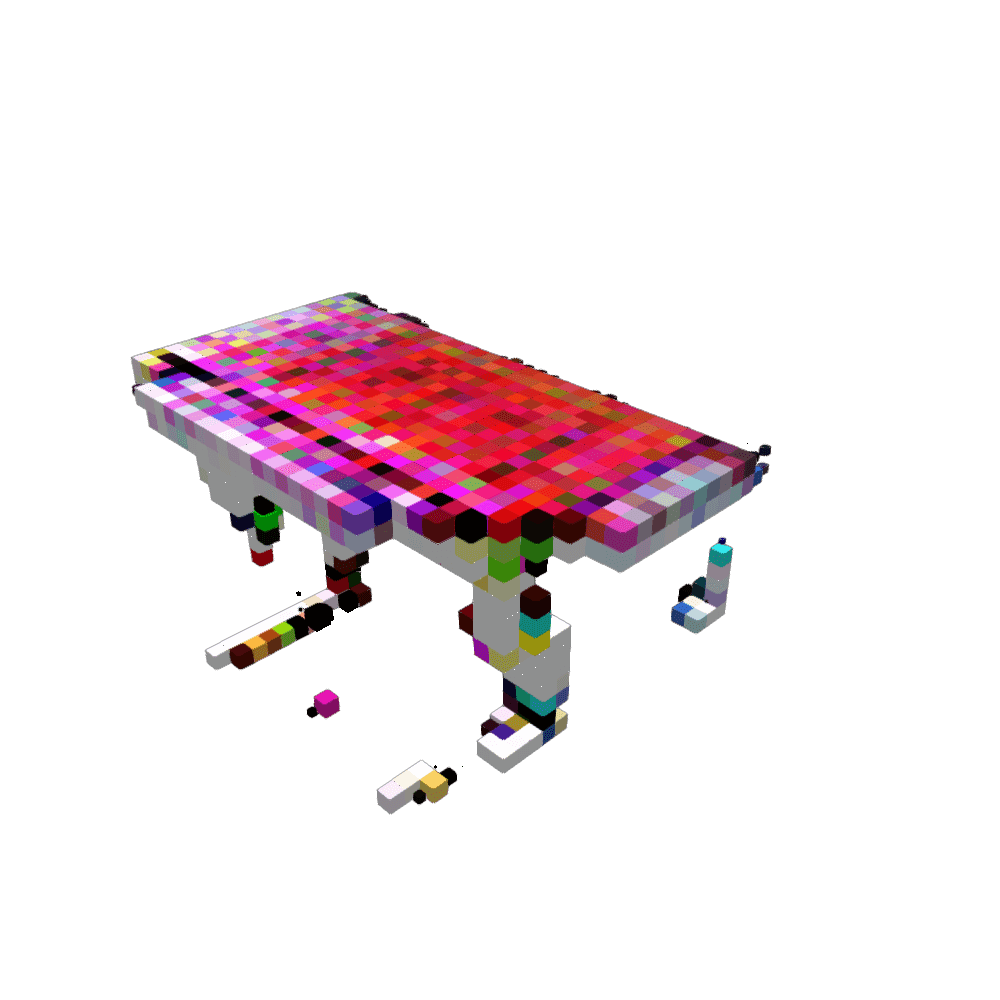} &
    \adjincludegraphics[width=\linewidth,trim={{.05\width} {.05\height} {.05\width} {.05\height}},clip]{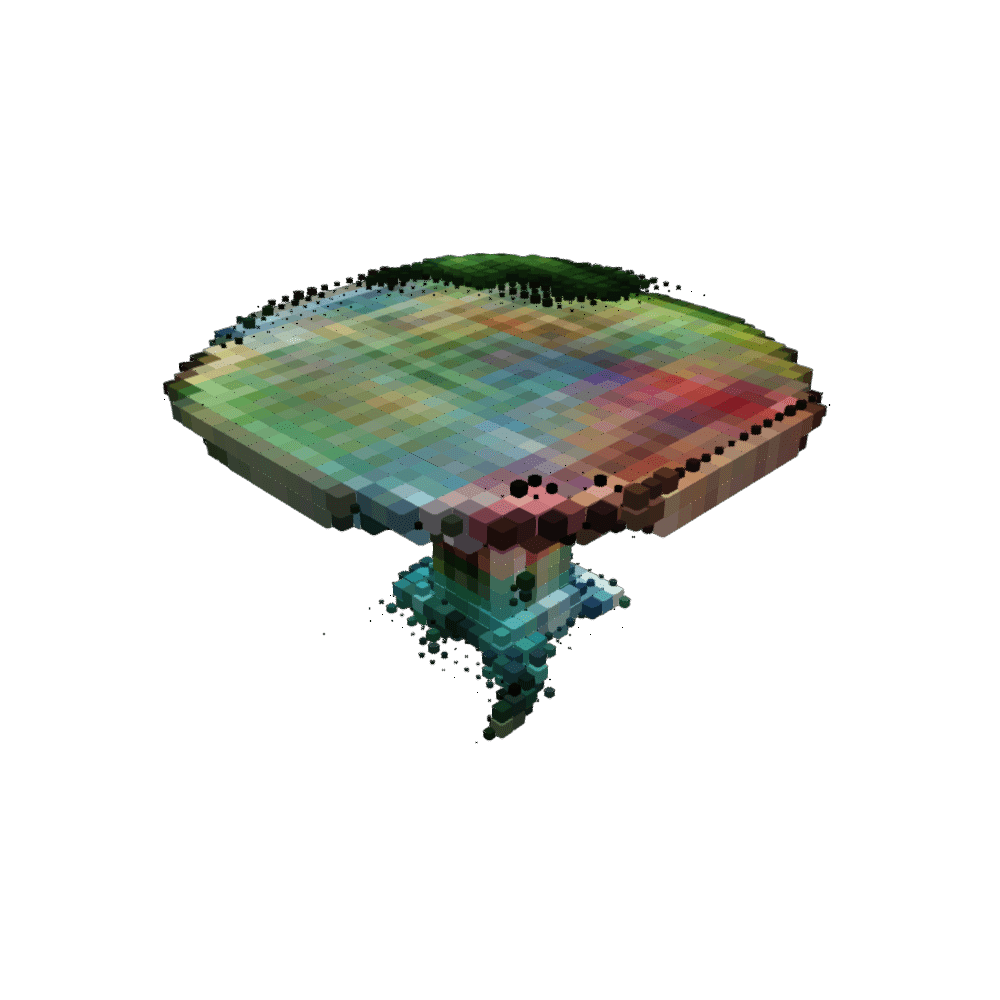} &
    \adjincludegraphics[width=\linewidth,trim={{.05\width} {.05\height} {.05\width} {.05\height}},clip]{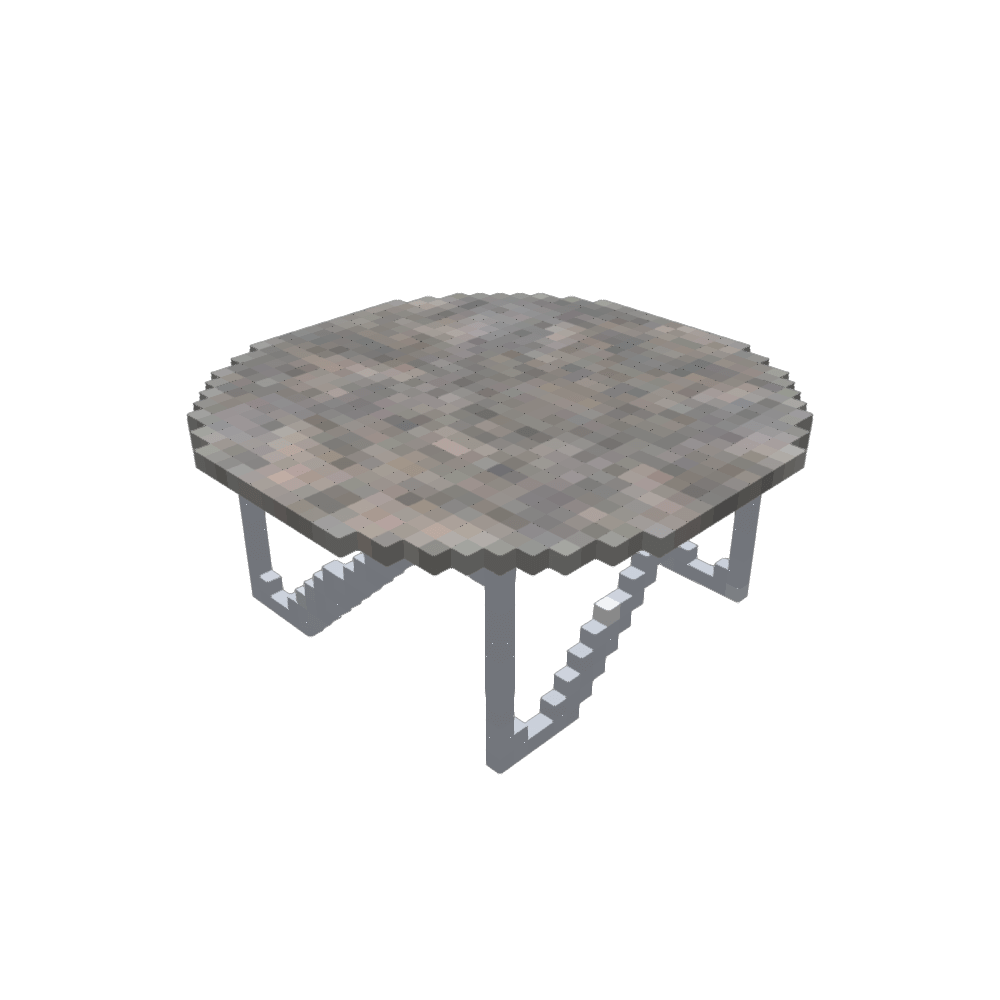}
    \\\midrule
    Waiting room chair leather legs and armrests are curved wood. &
    \adjincludegraphics[width=\linewidth,trim={{.05\width} {.05\height} {.05\width} {.05\height}},clip]{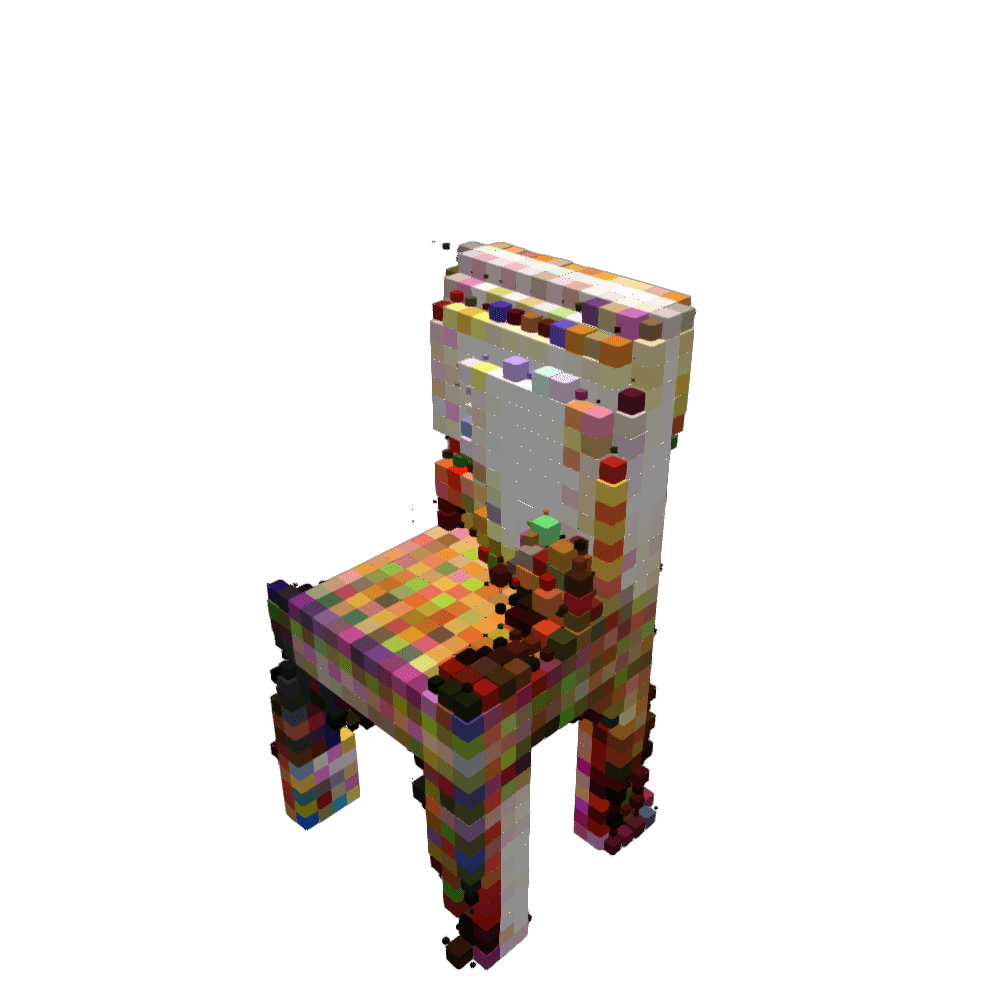} &
    \adjincludegraphics[width=\linewidth,trim={{.05\width} {.05\height} {.05\width} {.05\height}},clip]{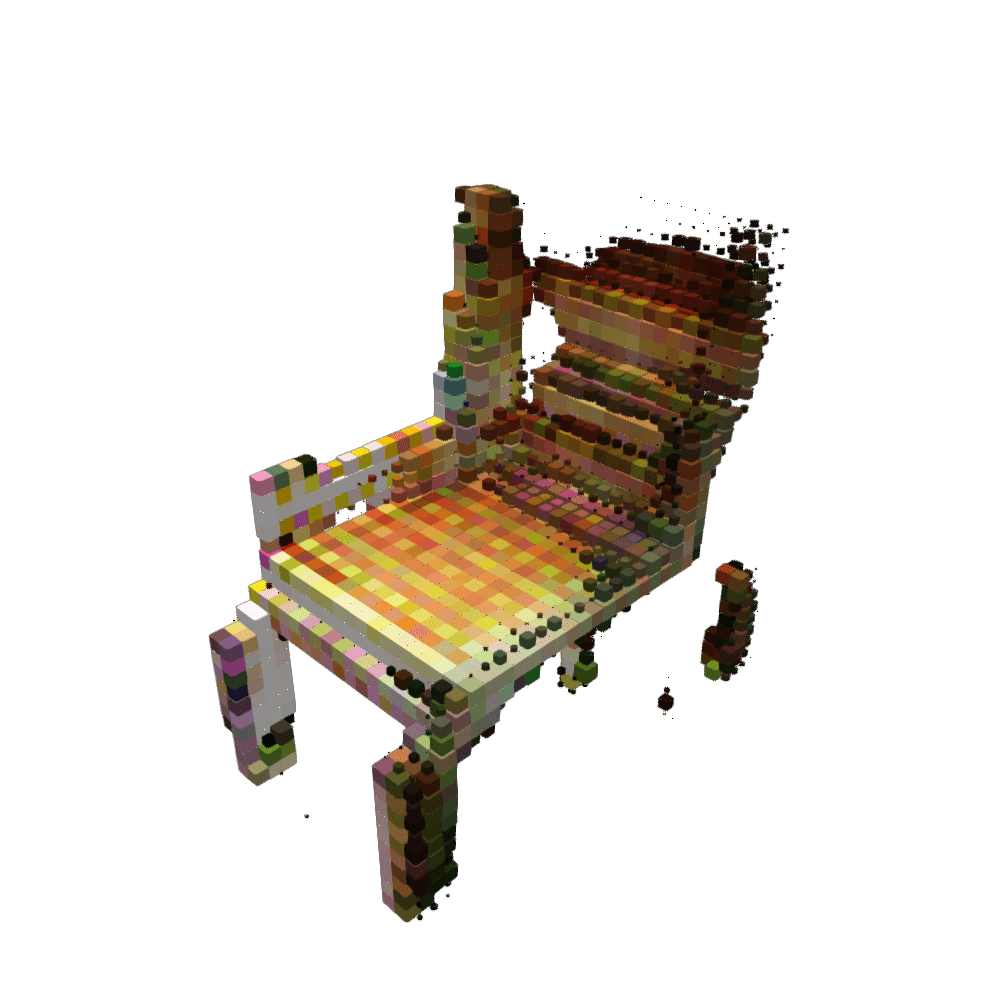} &
    \adjincludegraphics[width=\linewidth,trim={{.05\width} {.05\height} {.05\width} {.05\height}},clip]{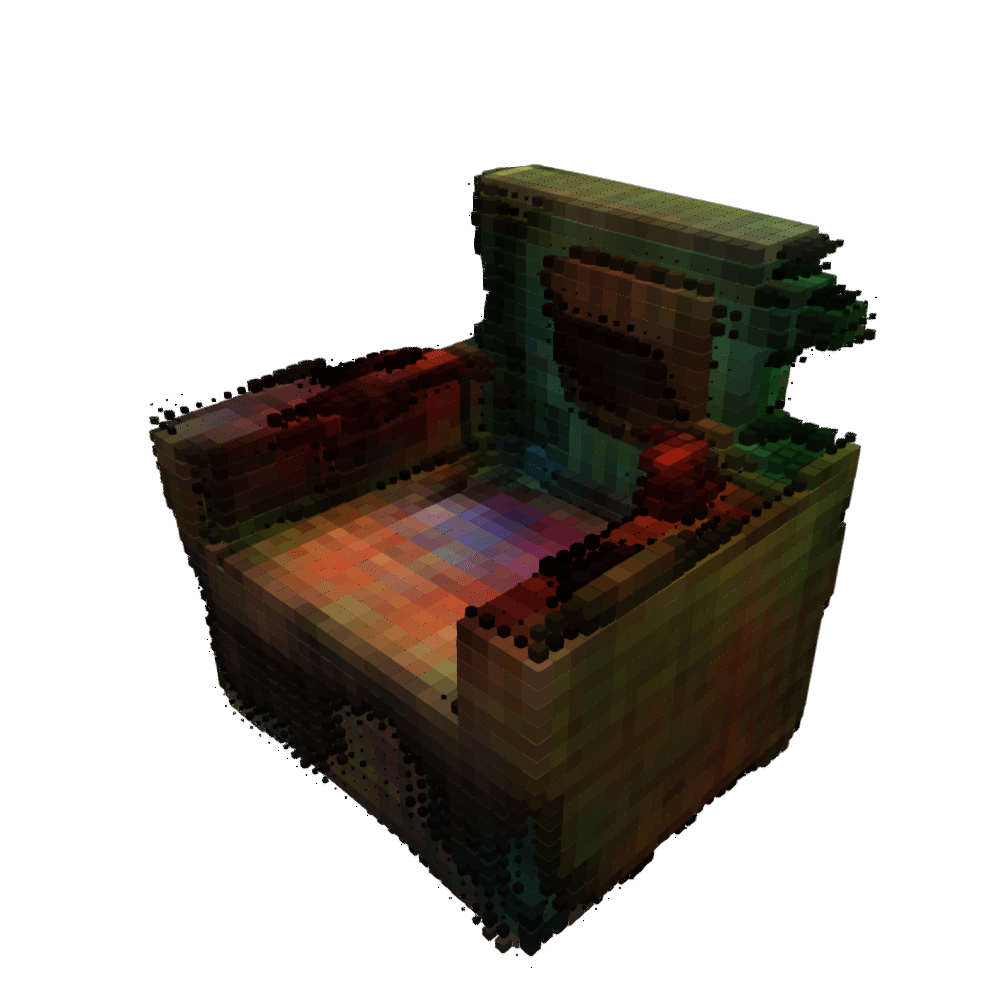} &
    \adjincludegraphics[width=\linewidth,trim={{.05\width} {.05\height} {.05\width} {.05\height}},clip]{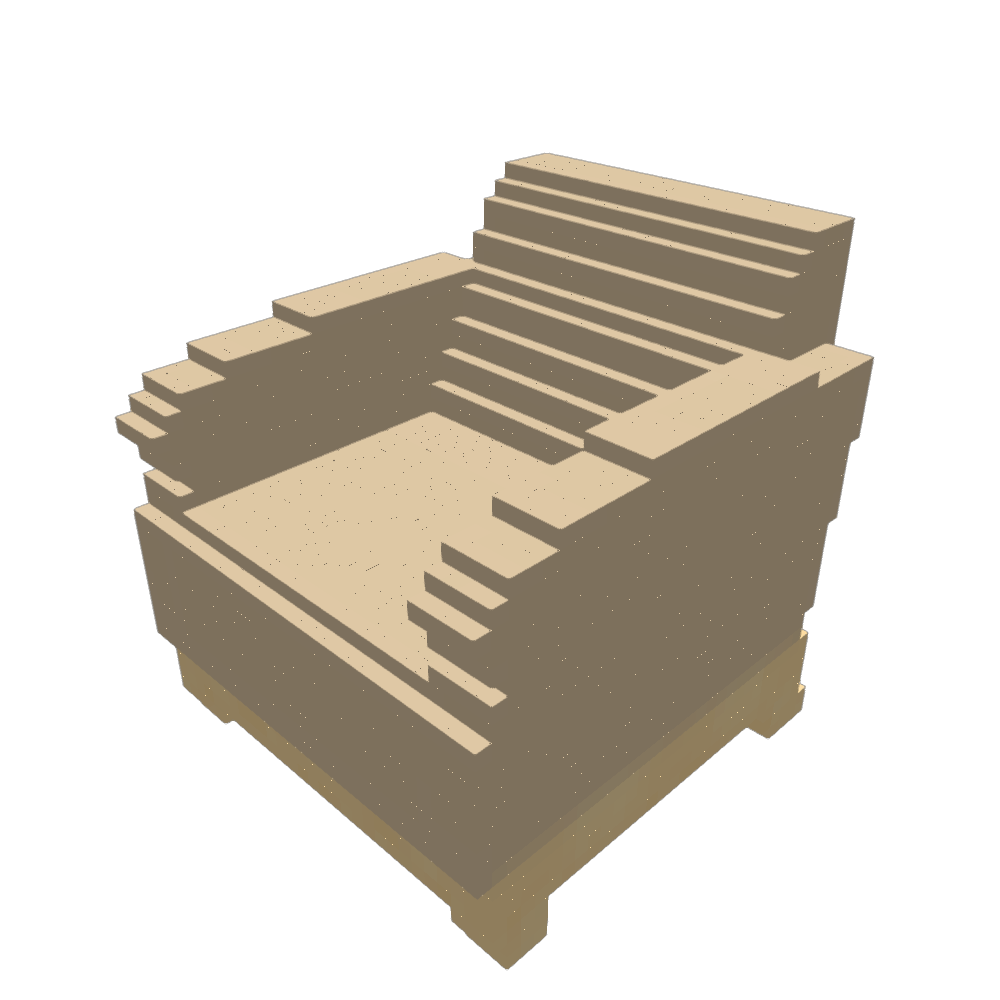}
    \\\midrule
    A multi-layered end table made of cherry wood. There is a rectangular surface with curved ends, and a square storage surface underneath that is slightly smaller. &
    \adjincludegraphics[width=\linewidth,trim={{.05\width} {.05\height} {.05\width} {.05\height}},clip]{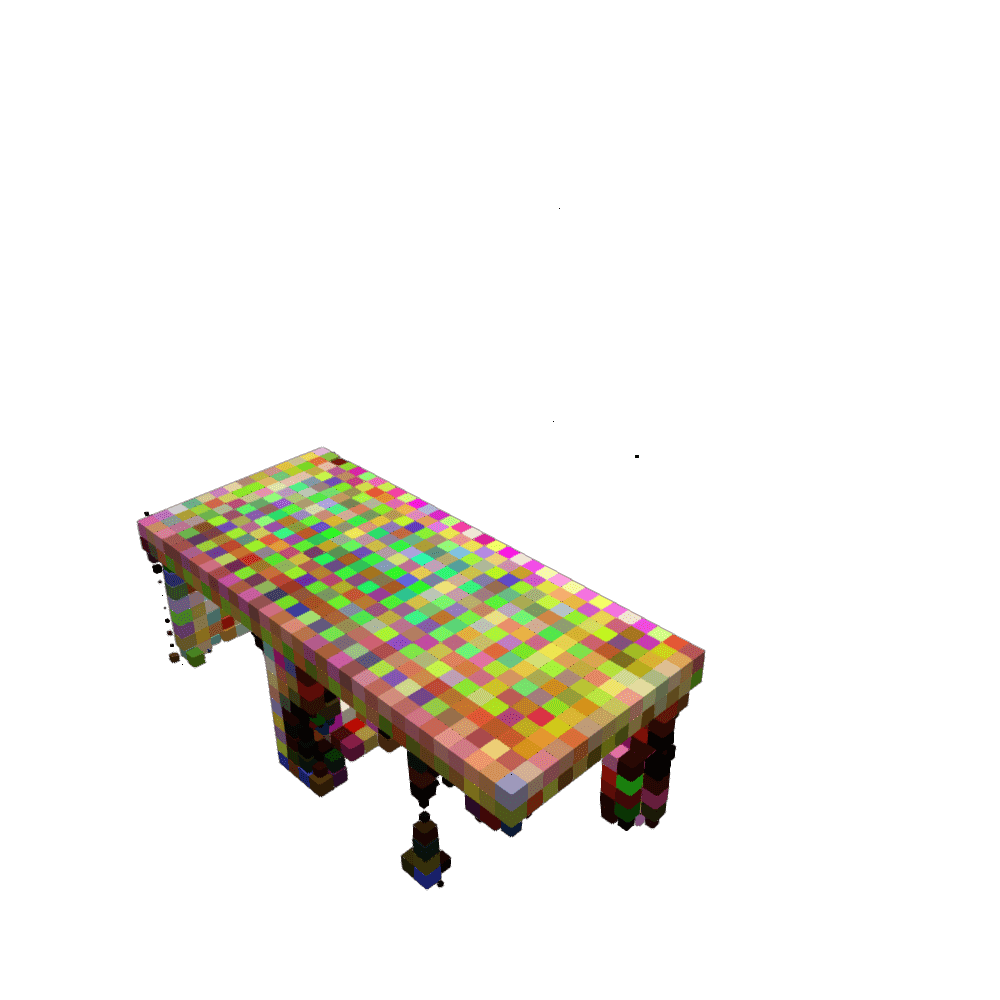} &
    \adjincludegraphics[width=\linewidth,trim={{.05\width} {.05\height} {.05\width} {.05\height}},clip]{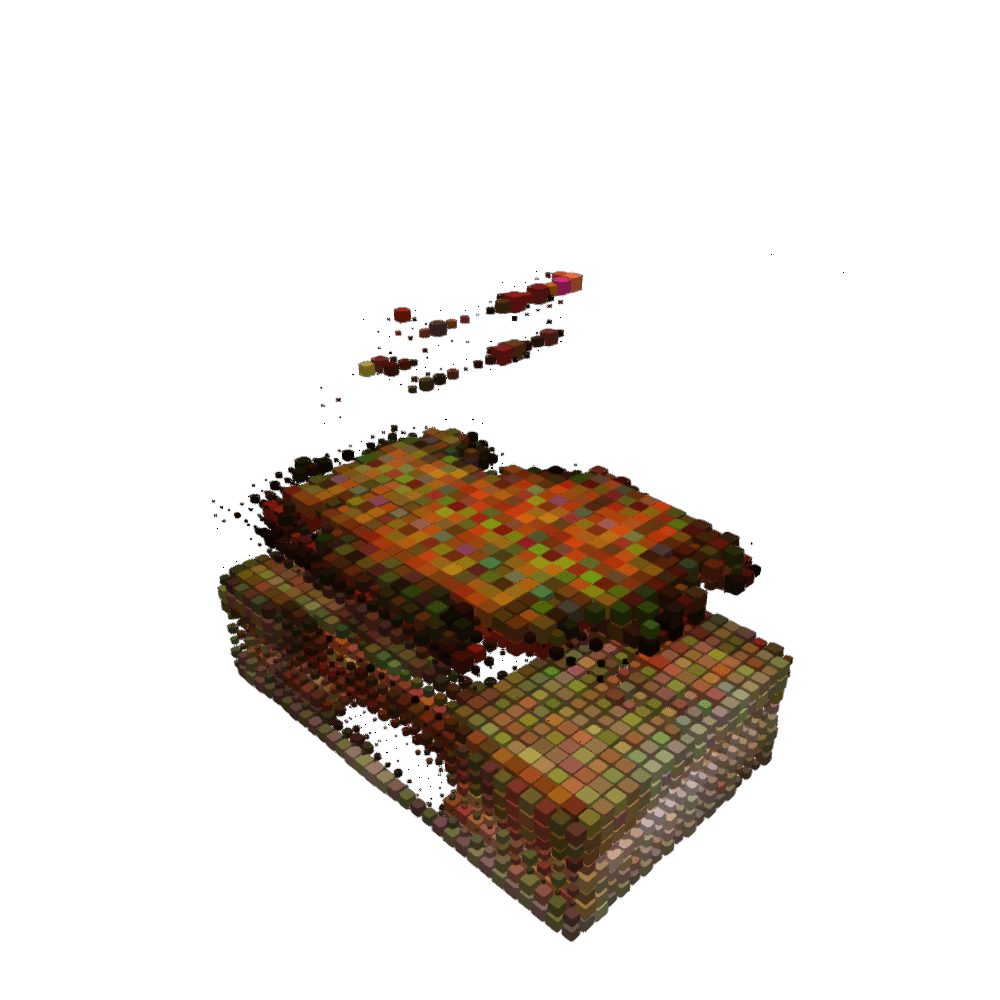} &
    \adjincludegraphics[width=\linewidth,trim={{.05\width} {.05\height} {.05\width} {.05\height}},clip]{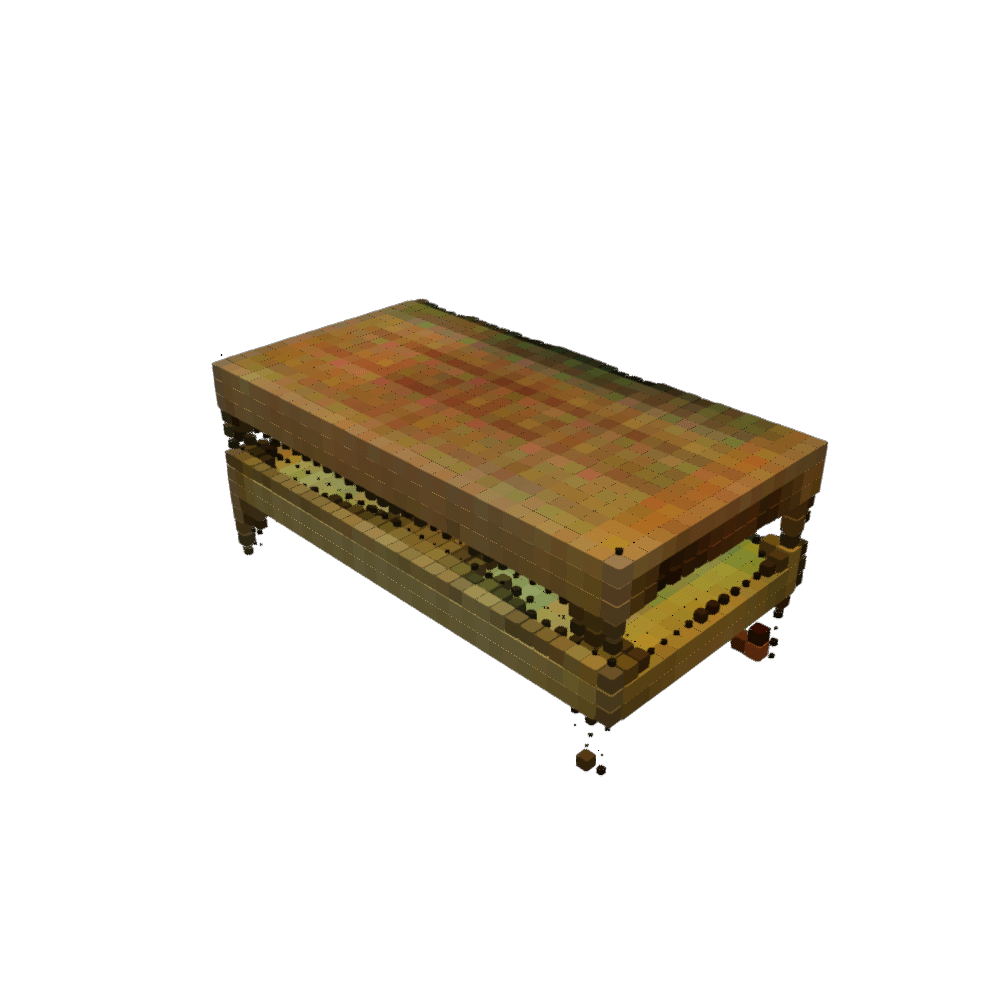} &
    \adjincludegraphics[width=\linewidth,trim={{.05\width} {.05\height} {.05\width} {.05\height}},clip]{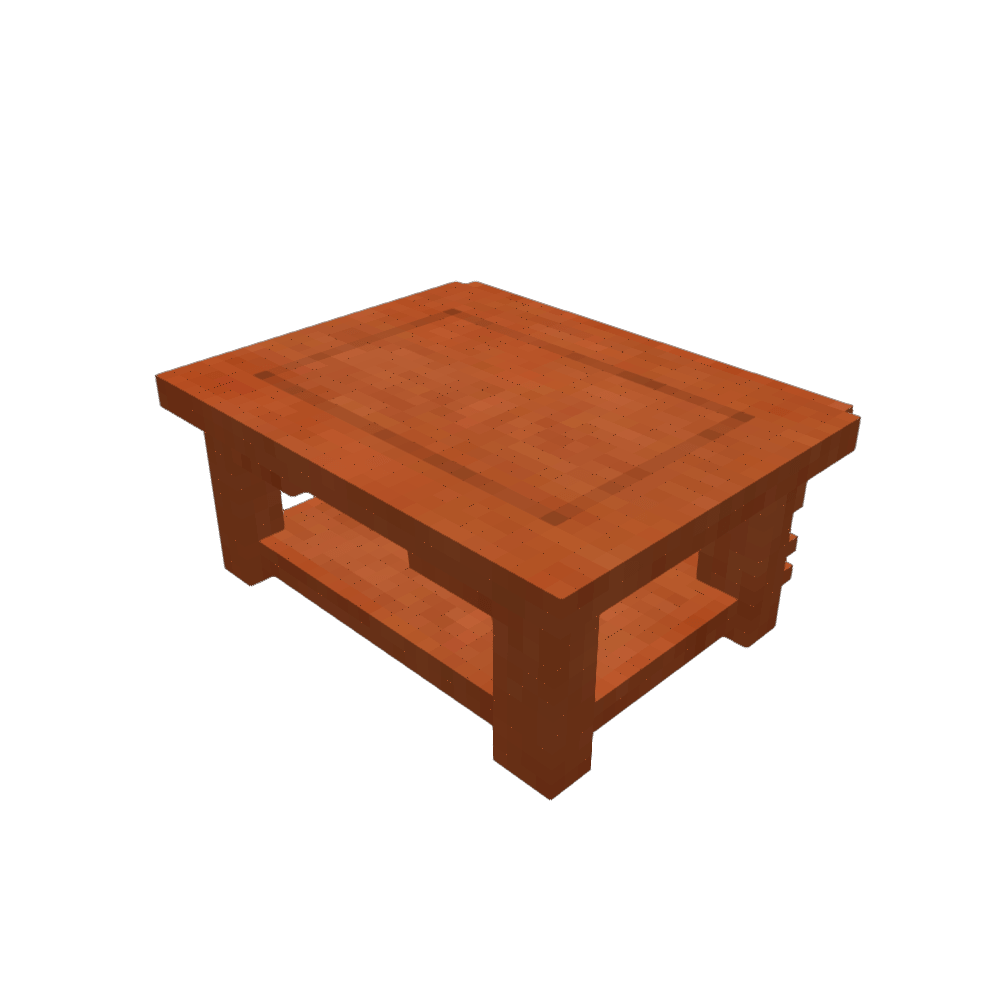}
    \\\midrule
    Brown colored dining table. It has four legs made of wood. &
    \adjincludegraphics[width=\linewidth,trim={{.05\width} {.05\height} {.05\width} {.05\height}},clip]{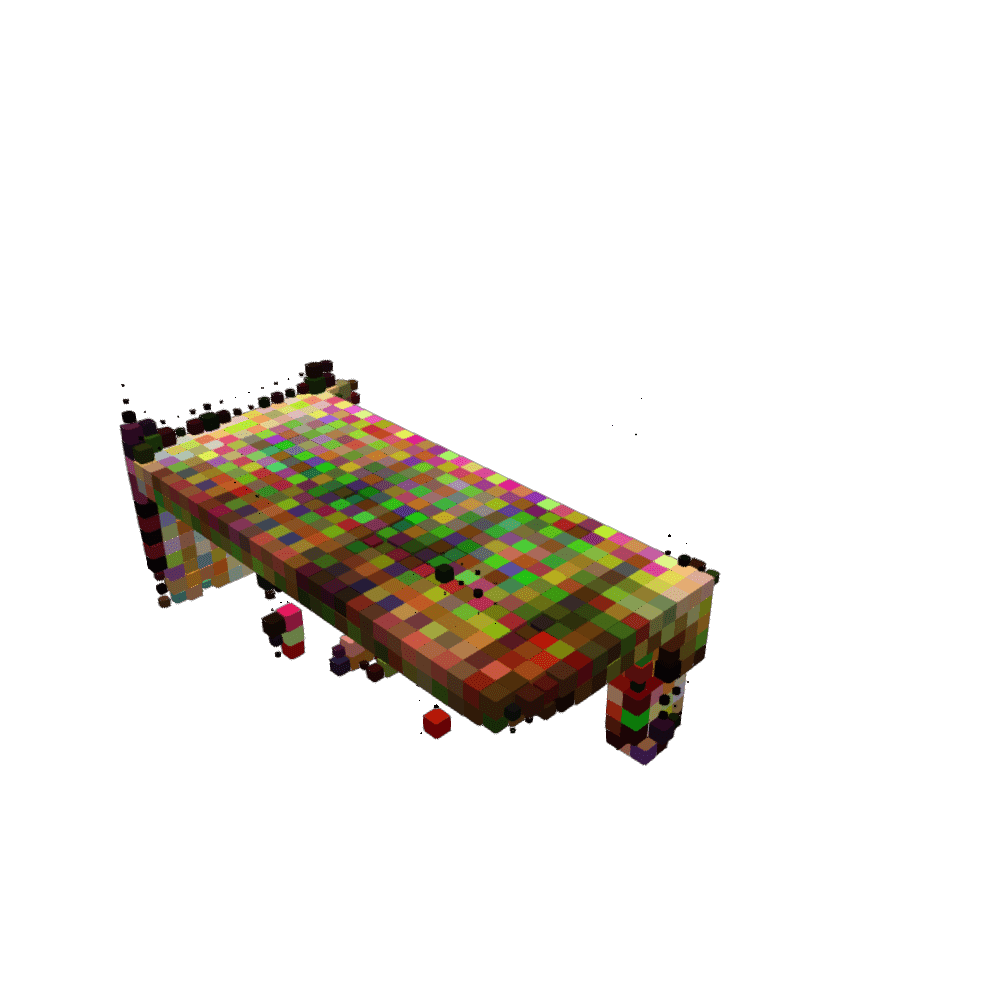} &
    \adjincludegraphics[width=\linewidth,trim={{.05\width} {.05\height} {.05\width} {.05\height}},clip]{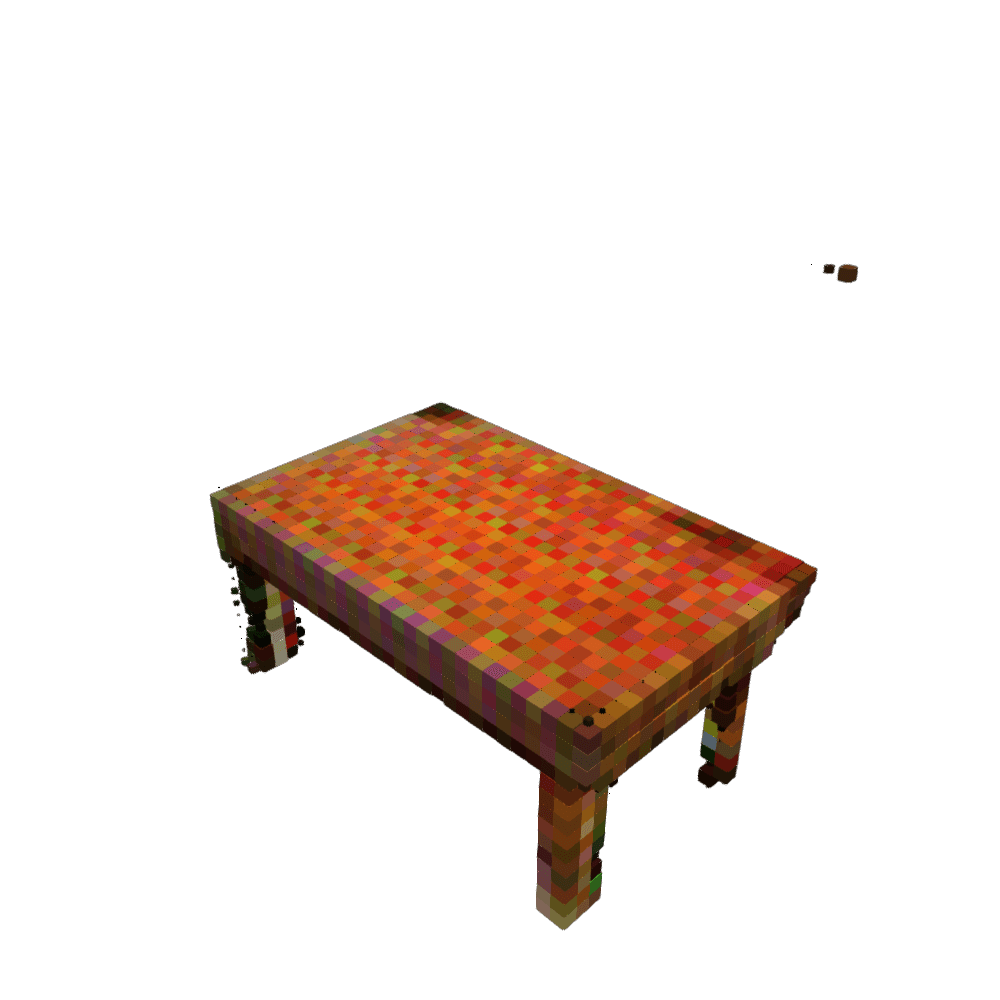} &
    \adjincludegraphics[width=\linewidth,trim={{.05\width} {.05\height} {.05\width} {.05\height}},clip]{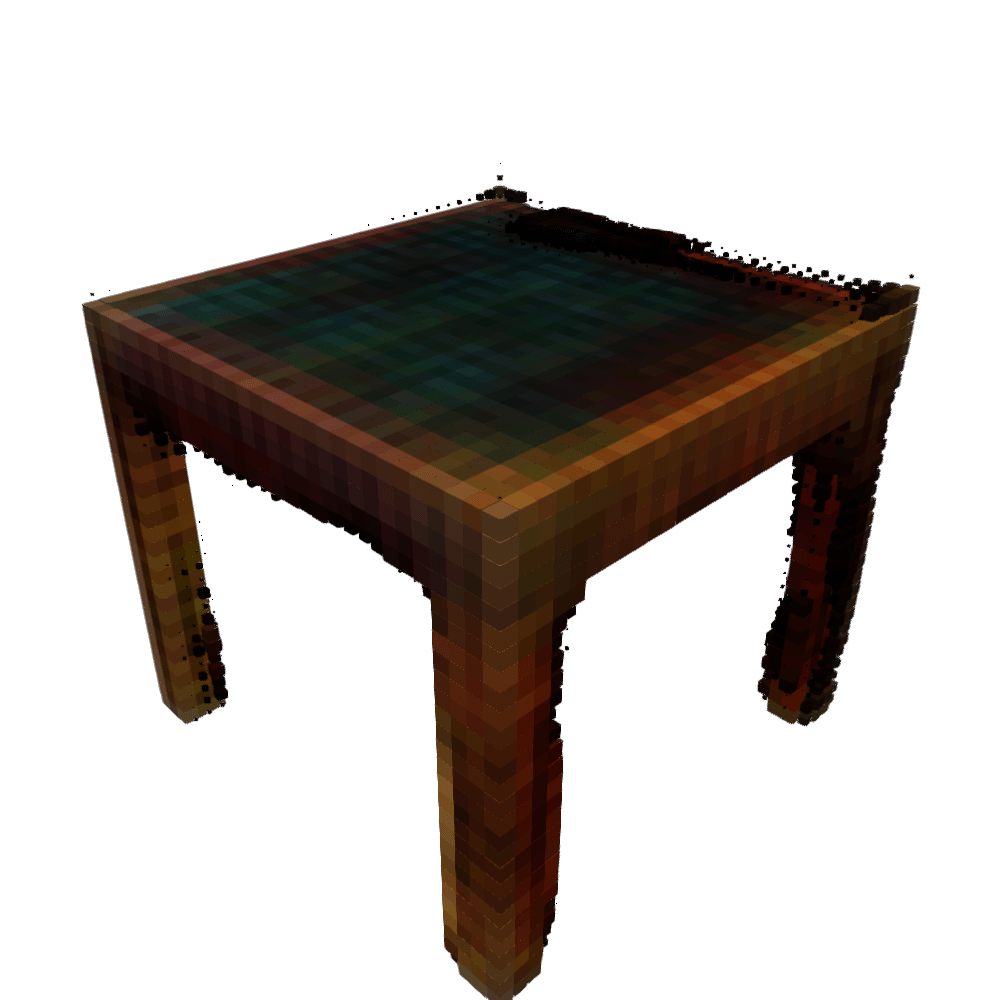} &
    \adjincludegraphics[width=\linewidth,trim={{.05\width} {.05\height} {.05\width} {.05\height}},clip]{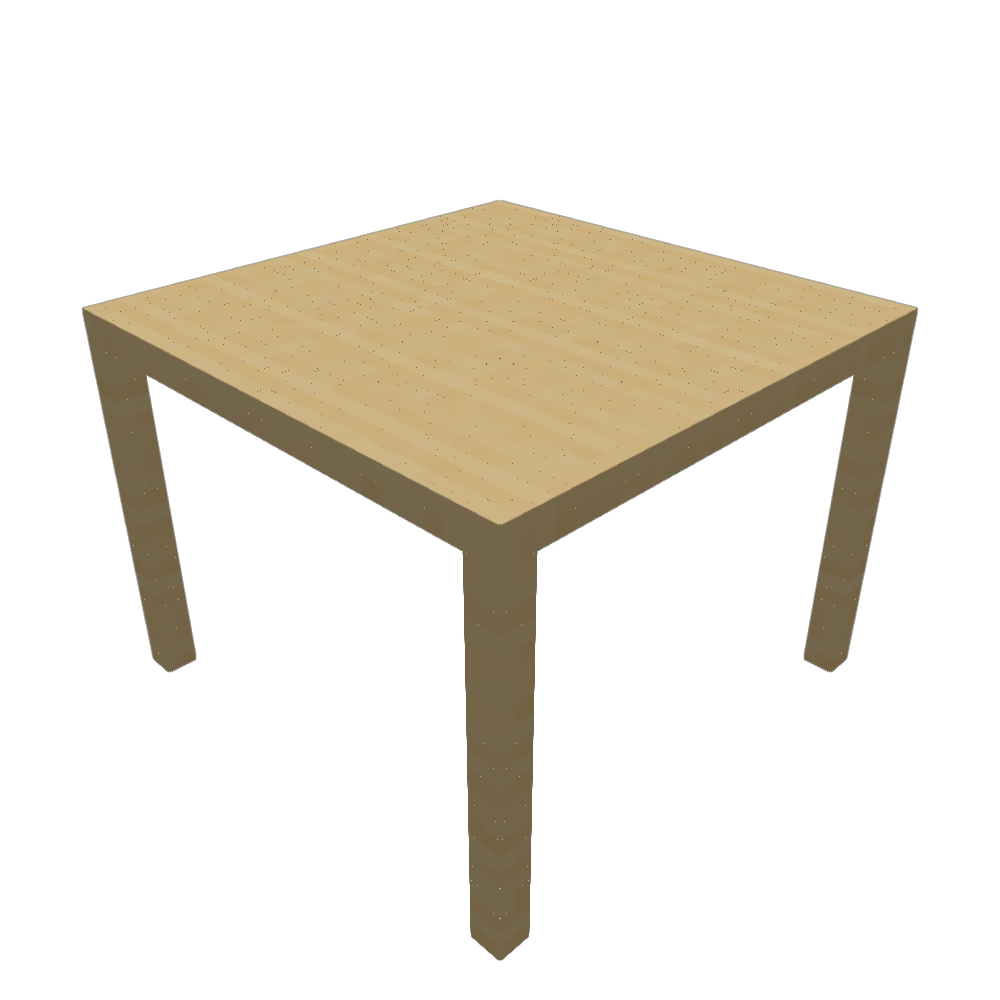}
    \end{tabular}
    \vspace{-1em}
    \caption{\textbf{Example text-to-shape generations}. Our CWGAN approach is correctly conditioned by the input text, generating shapes that match the described attributes. The baseline approach struggles to generate plausible shapes. Additional results can be found in the supplementary material.}
    \vspace{-1.5em}
    \label{fig:text2shape_results}
\end{figure}

%% file: fig/shape_editing.tex
\begin{figure}
    \vspace{-3.5em}
	\captionsetup[subfigure]{justification=centering,labelformat=empty}
	\centering
	\begin{subfigure}[t]{0.18\linewidth}
		\includegraphics[width=\linewidth]{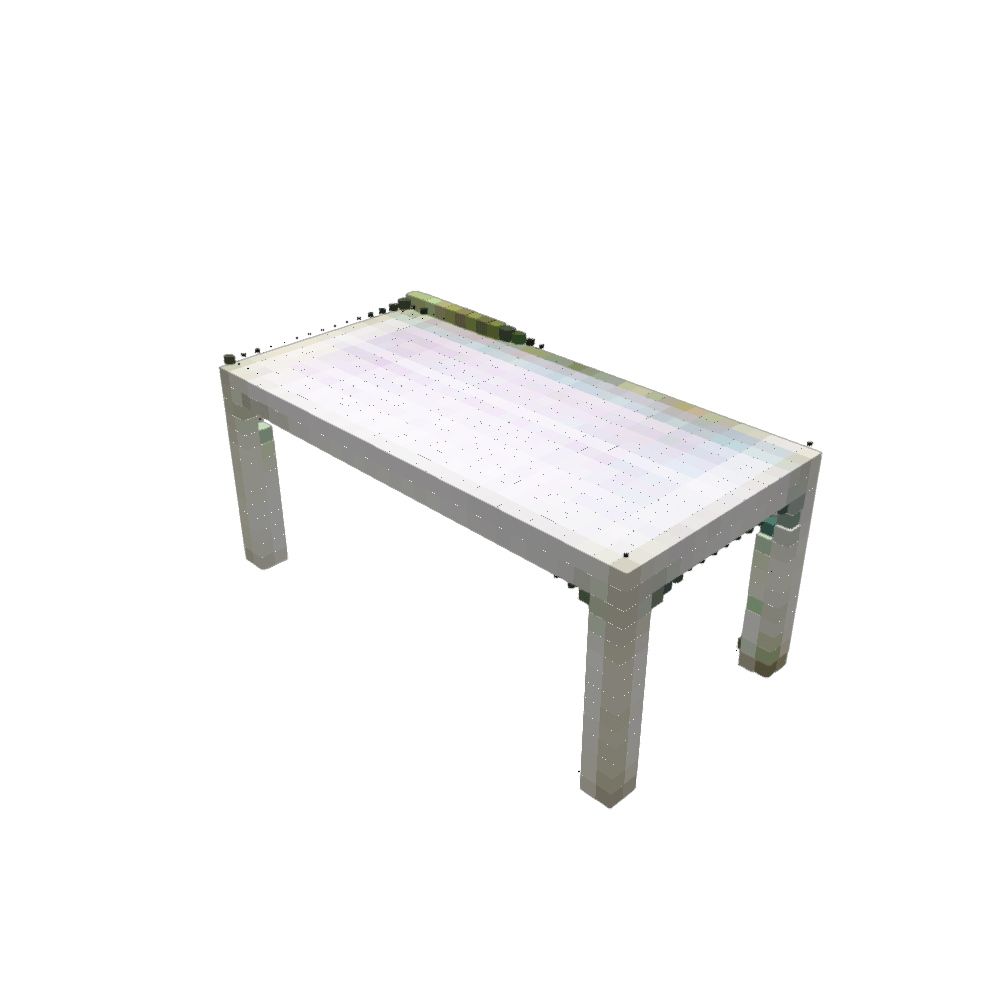}
		\caption{\scriptsize{\emph{White coffee table}}}
	\end{subfigure}
	\begin{subfigure}[t]{0.18\linewidth}
		\includegraphics[width=\linewidth]{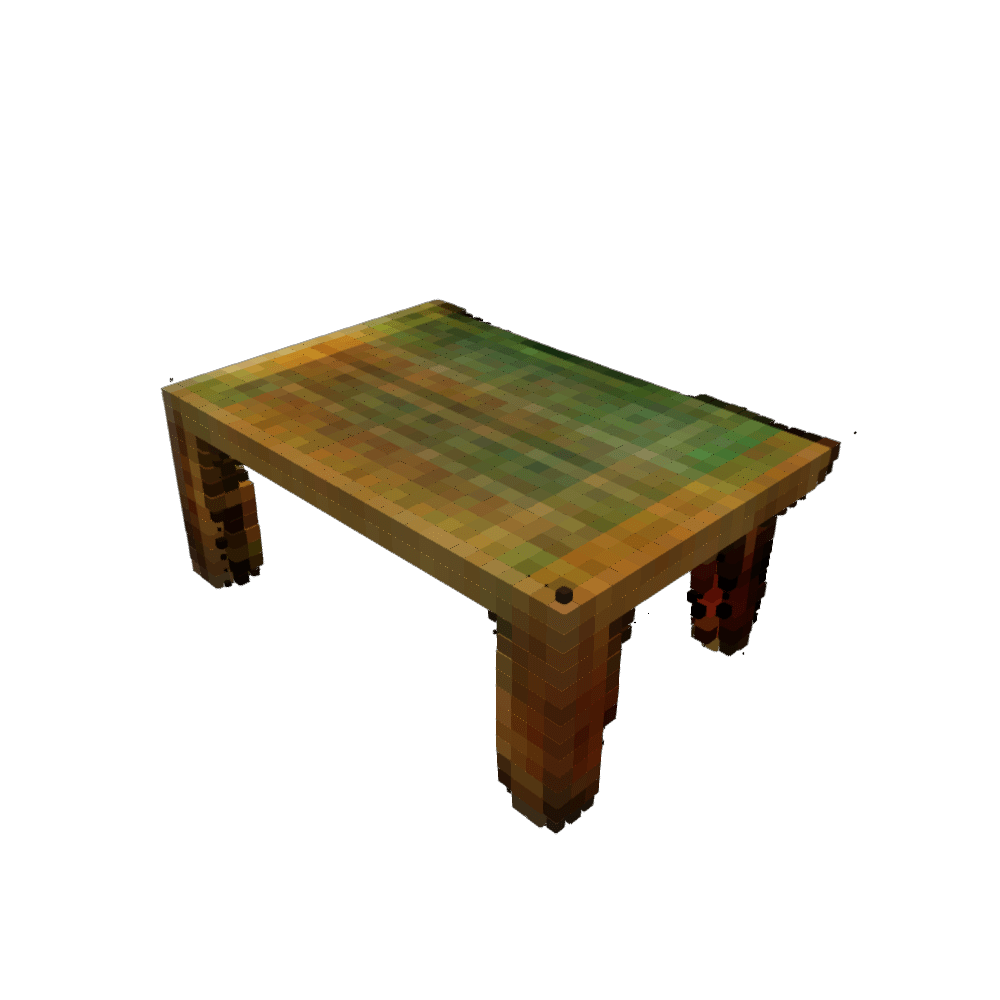}
		\caption{\scriptsize{\emph{Wooden coffee table}}}
	\end{subfigure}
	\begin{subfigure}[t]{0.18\linewidth}
        \includegraphics[width=\linewidth]{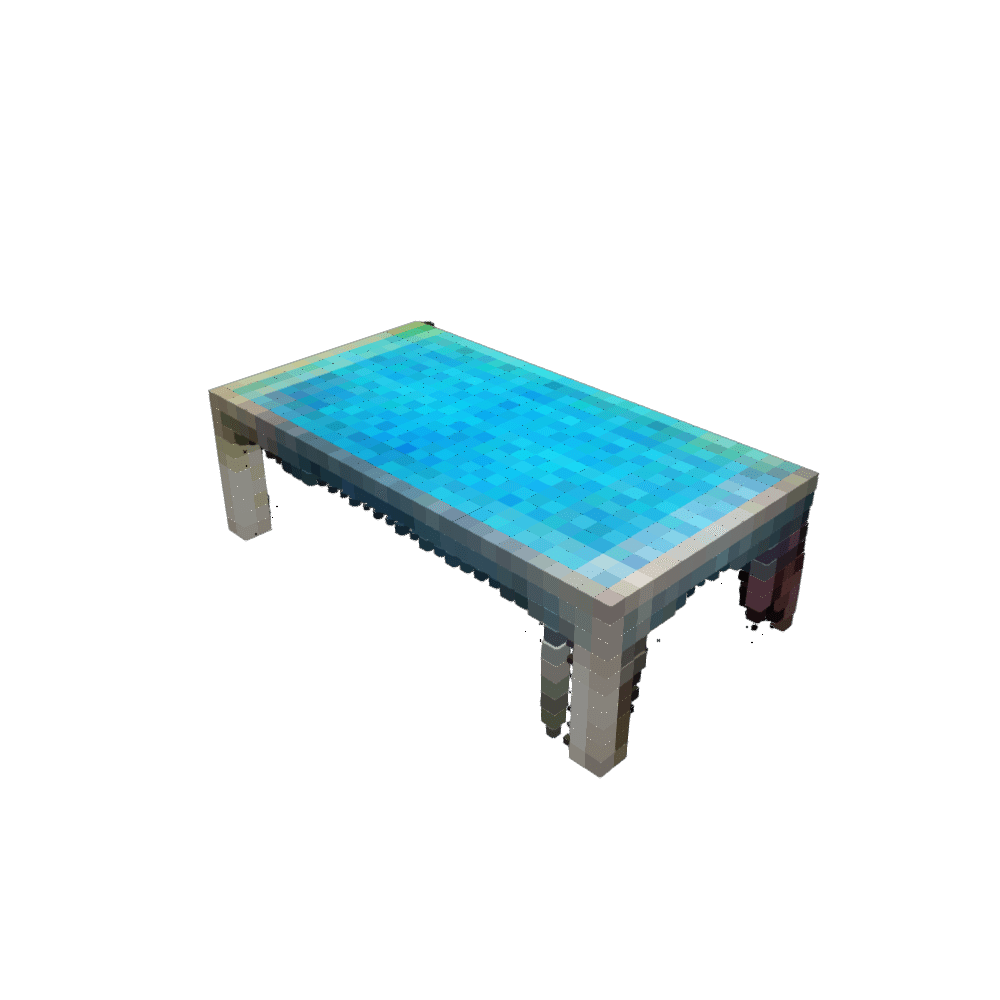}
        \caption{\scriptsize{\emph{Rectangular glass coffee table}}}
	\end{subfigure}
	\begin{subfigure}[t]{0.18\linewidth}
		\includegraphics[width=\linewidth]{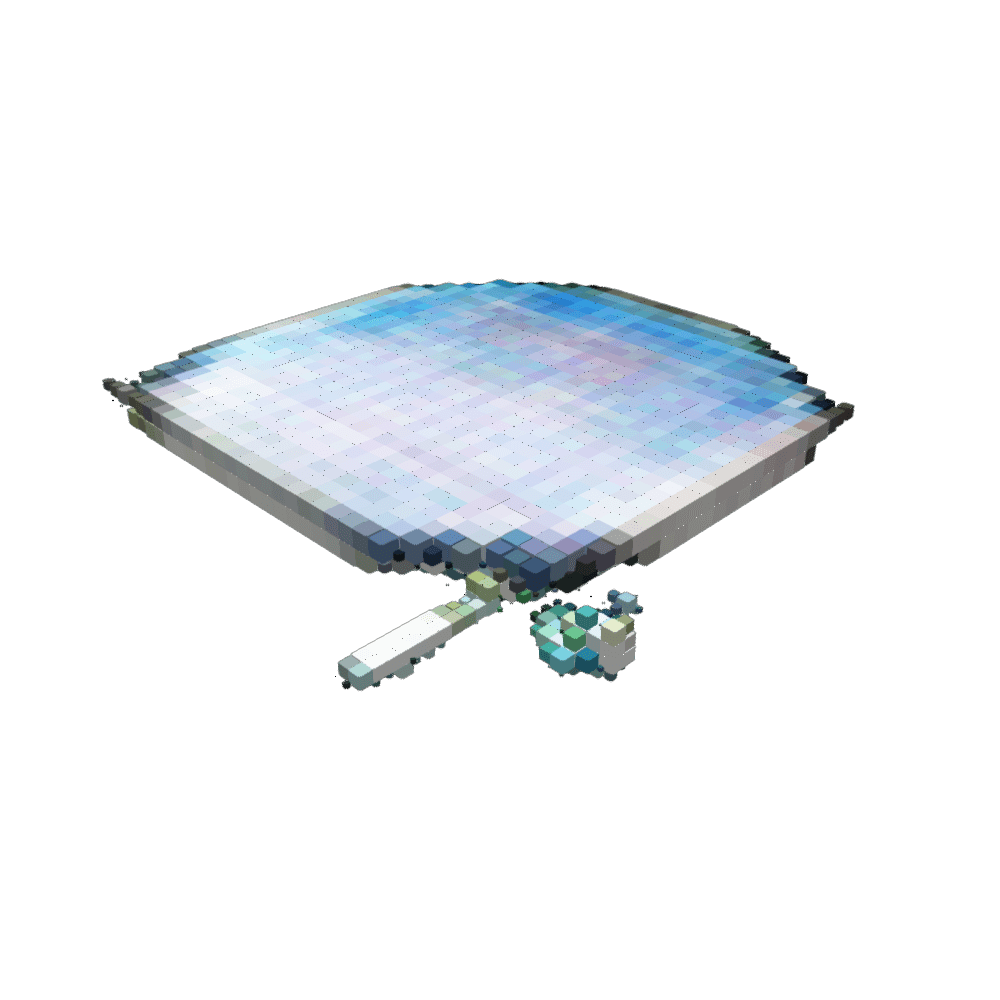}
		\caption{\scriptsize{\emph{Glass round coffee table}}}
	\end{subfigure}
	\begin{subfigure}[t]{0.18\linewidth}
		\includegraphics[width=\linewidth]{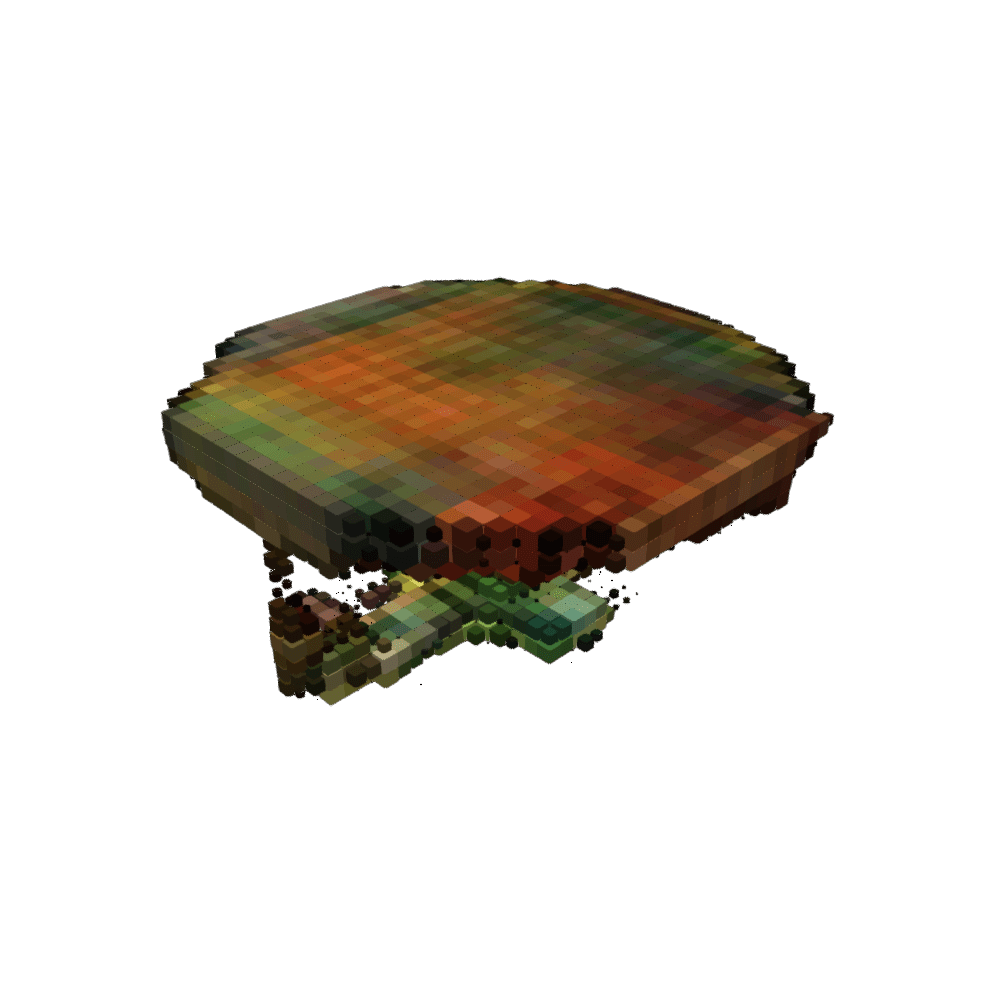}
		\caption{\scriptsize{\emph{Red round coffee table}}}
	\end{subfigure}
	
	\begin{subfigure}[t]{0.18\linewidth}
		\includegraphics[width=\linewidth]{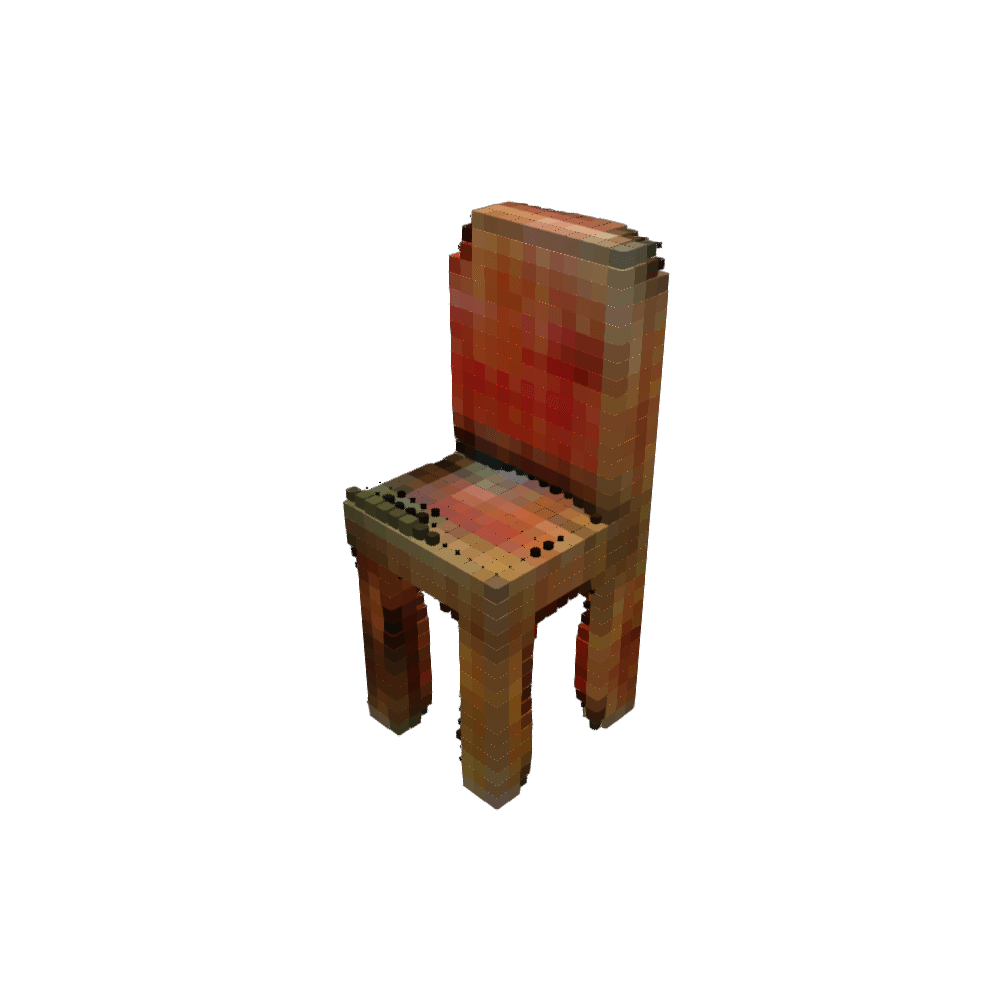}
		\caption{\scriptsize{\emph{Red chair}}}
	\end{subfigure}
	\begin{subfigure}[t]{0.18\linewidth}
		\includegraphics[width=\linewidth]{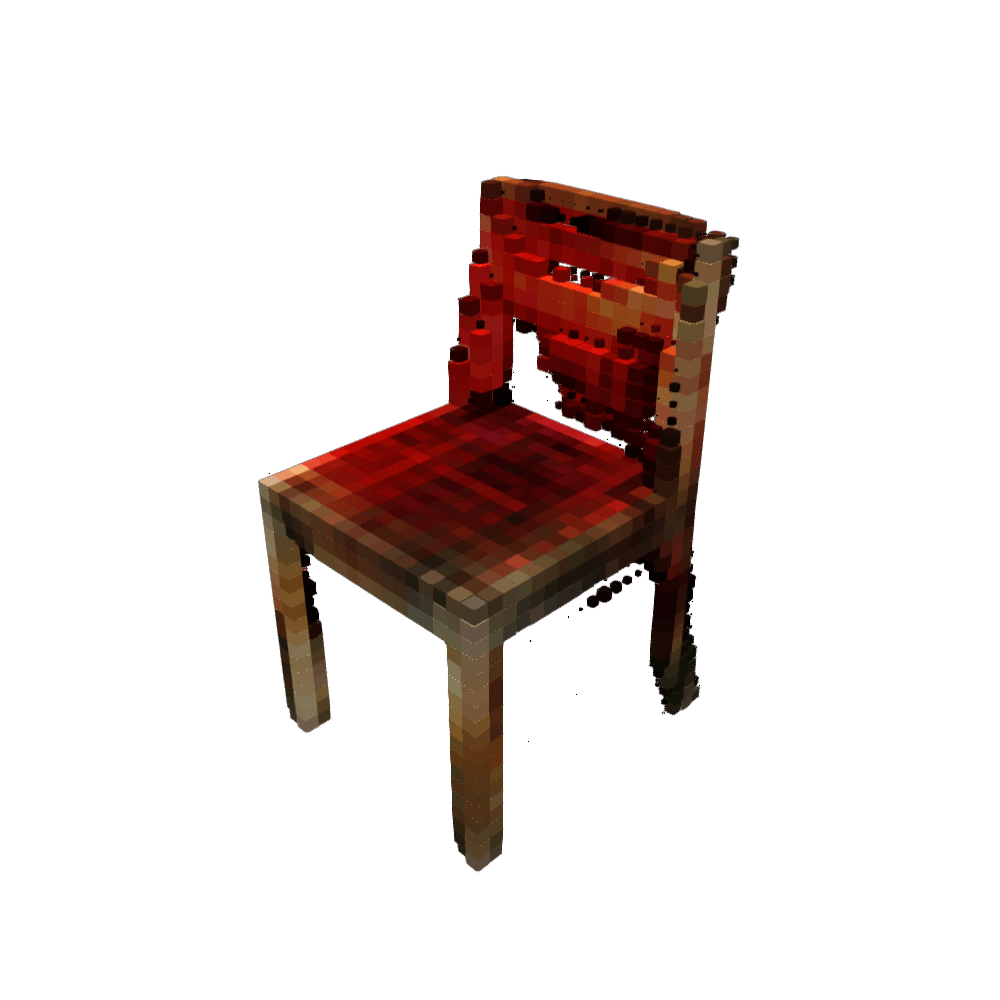}
		\caption{\scriptsize{\emph{Dining chair}}}
	\end{subfigure}
	\begin{subfigure}[t]{0.18\linewidth}
		\includegraphics[width=\linewidth]{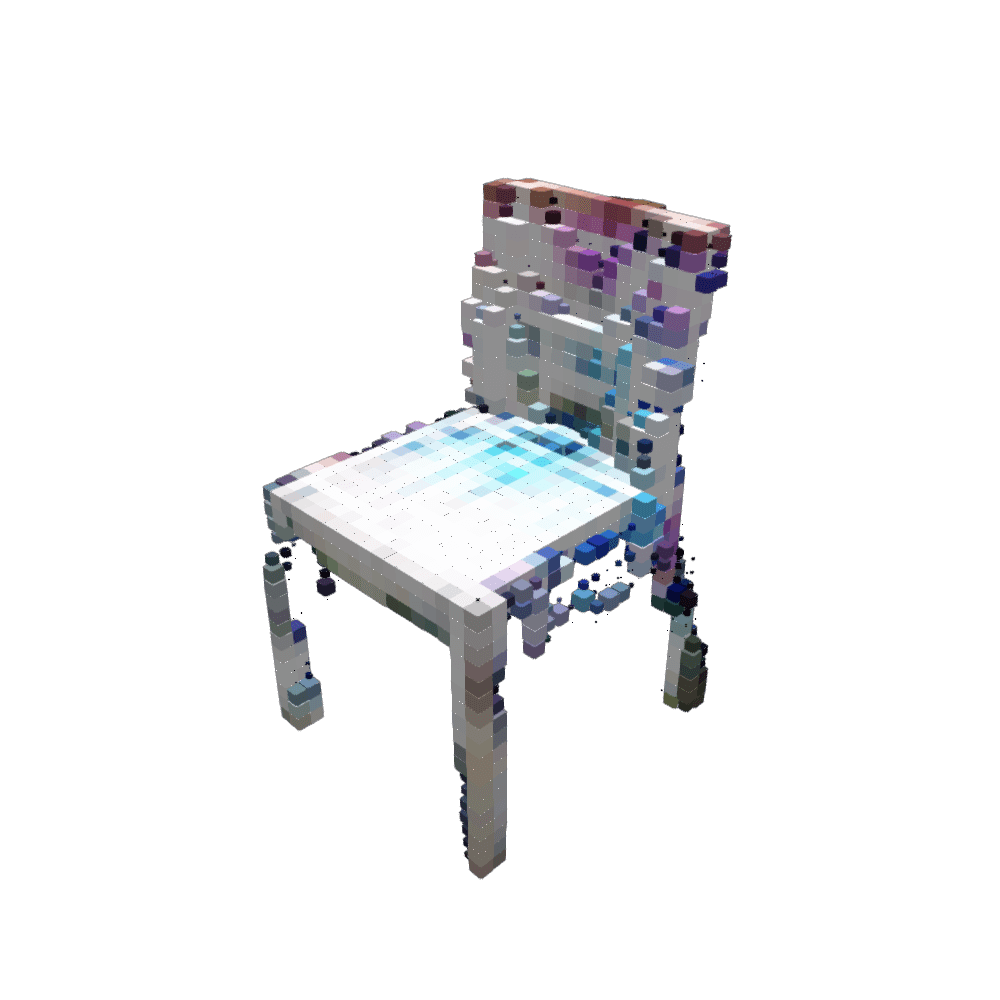}
		\caption{\scriptsize{\emph{Gray dining chair}}}
	\end{subfigure}
	\begin{subfigure}[t]{0.18\linewidth}
		\includegraphics[width=\linewidth]{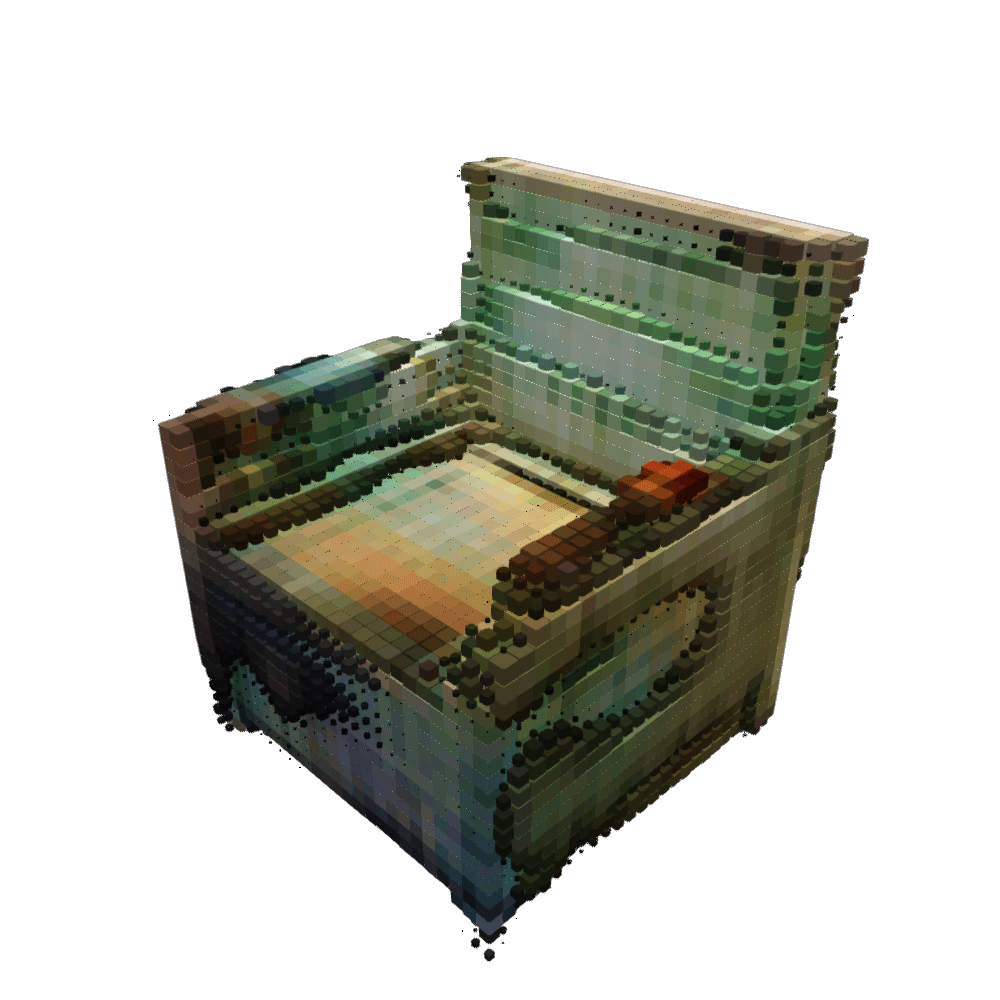}
		\caption{\scriptsize{\emph{Silver leather chair}}}
	\end{subfigure}
	\begin{subfigure}[t]{0.17\linewidth}
		\includegraphics[width=\linewidth]{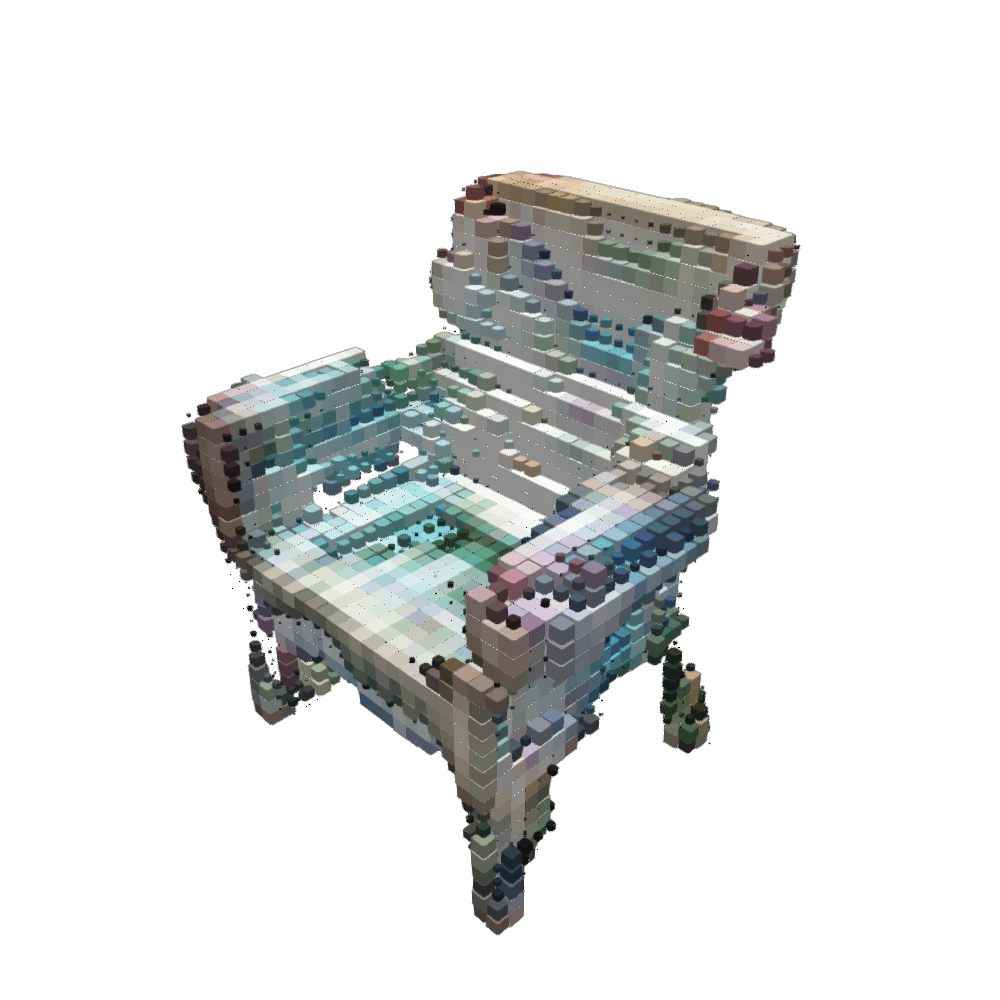}
		\caption{\scriptsize{\emph{Gray leather chair}}}
	\end{subfigure}
    \vspace{-1em}
    \caption{\textbf{Shape manipulation using text variations.} The attributes of the generated shapes can be controlled using text.}
    \label{fig:shape-editing}
    \vspace{-1.5em}
\end{figure}

%% file: fig/activation_vis.tex
\begin{figure}
    \vspace{-1em}
    \centering
    \includegraphics[width=\linewidth]{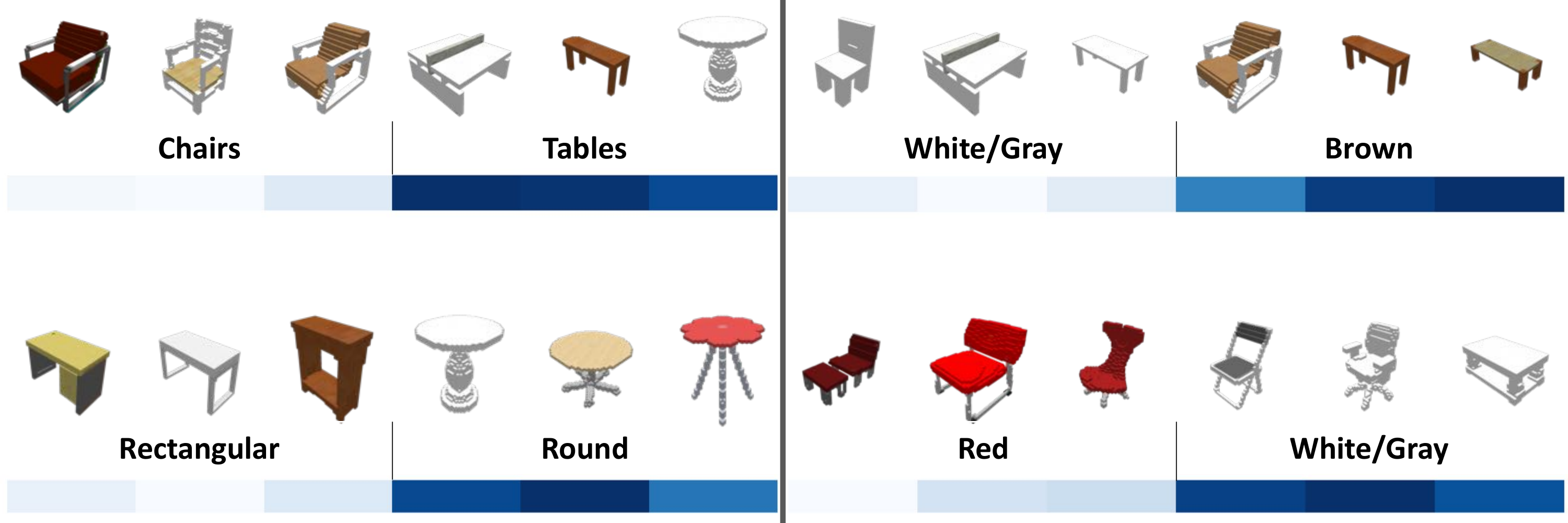}
    \vspace{-2em}
    \caption{\textbf{Visualizing learned representations.} Each of the 4 examples above visualizes the activation for a specific dimension, with darker colors indicating higher values. Individual dimensions correlate with category (top-left: chair vs. table), shape (bottom-left: rectangular vs. round), and color (top-right: white vs. brown, bottom-right: red vs. white/gray).
    }
    \label{fig:activations}
    \vspace{-1.5em}
\end{figure}

%% file: fig/vector-arithmetic-combined.tex
\begin{figure}
    \vspace{-2em}
	\captionsetup[subfigure]{justification=centering,labelformat=empty}
	\centering
	\begin{subfigure}[t]{0.49\linewidth}
	    \includegraphics[width=\linewidth]{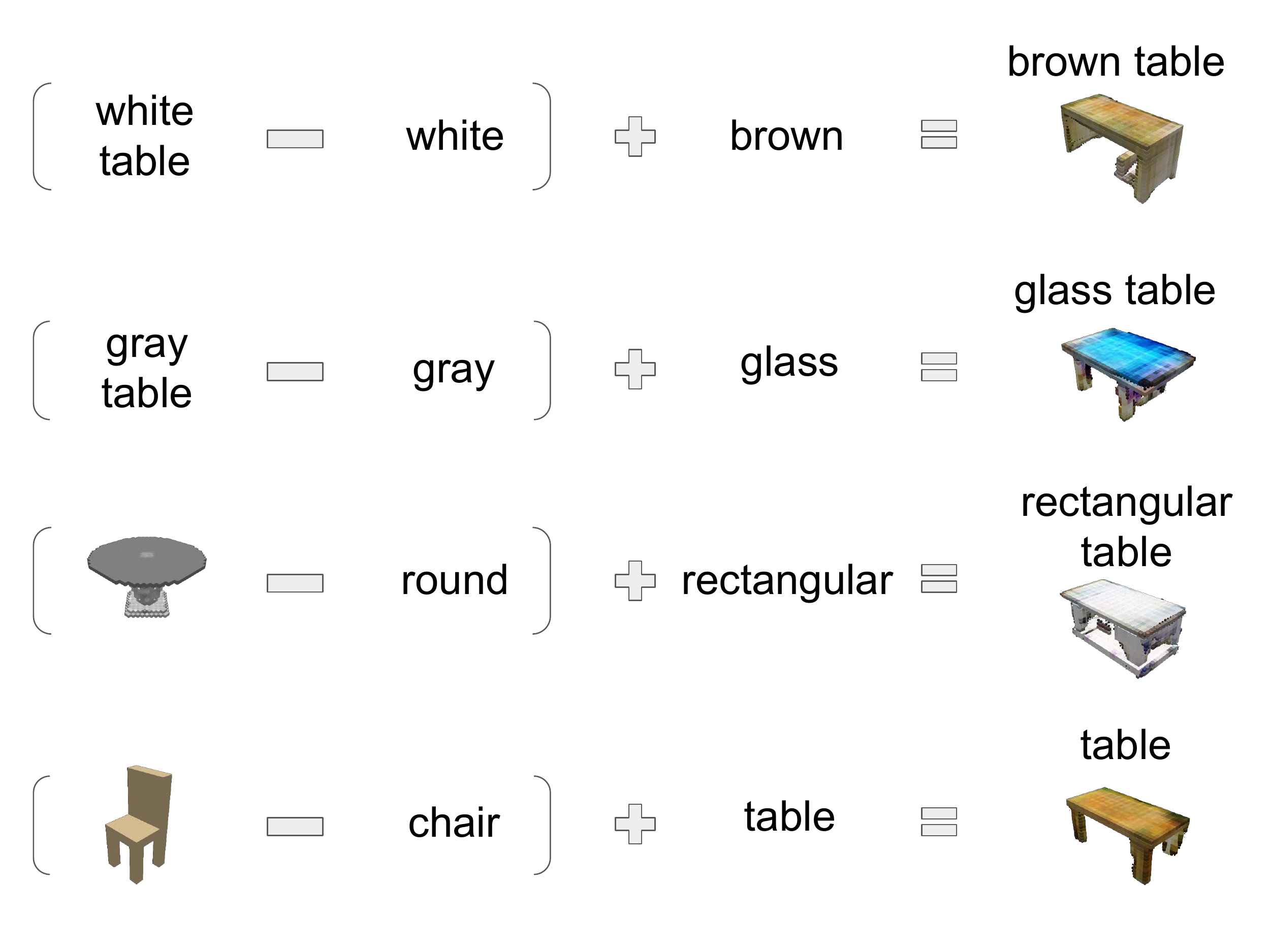}
	\end{subfigure}
	\rulesep
	\begin{subfigure}[t]{0.49\linewidth}
		\centering\includegraphics[width=\linewidth]{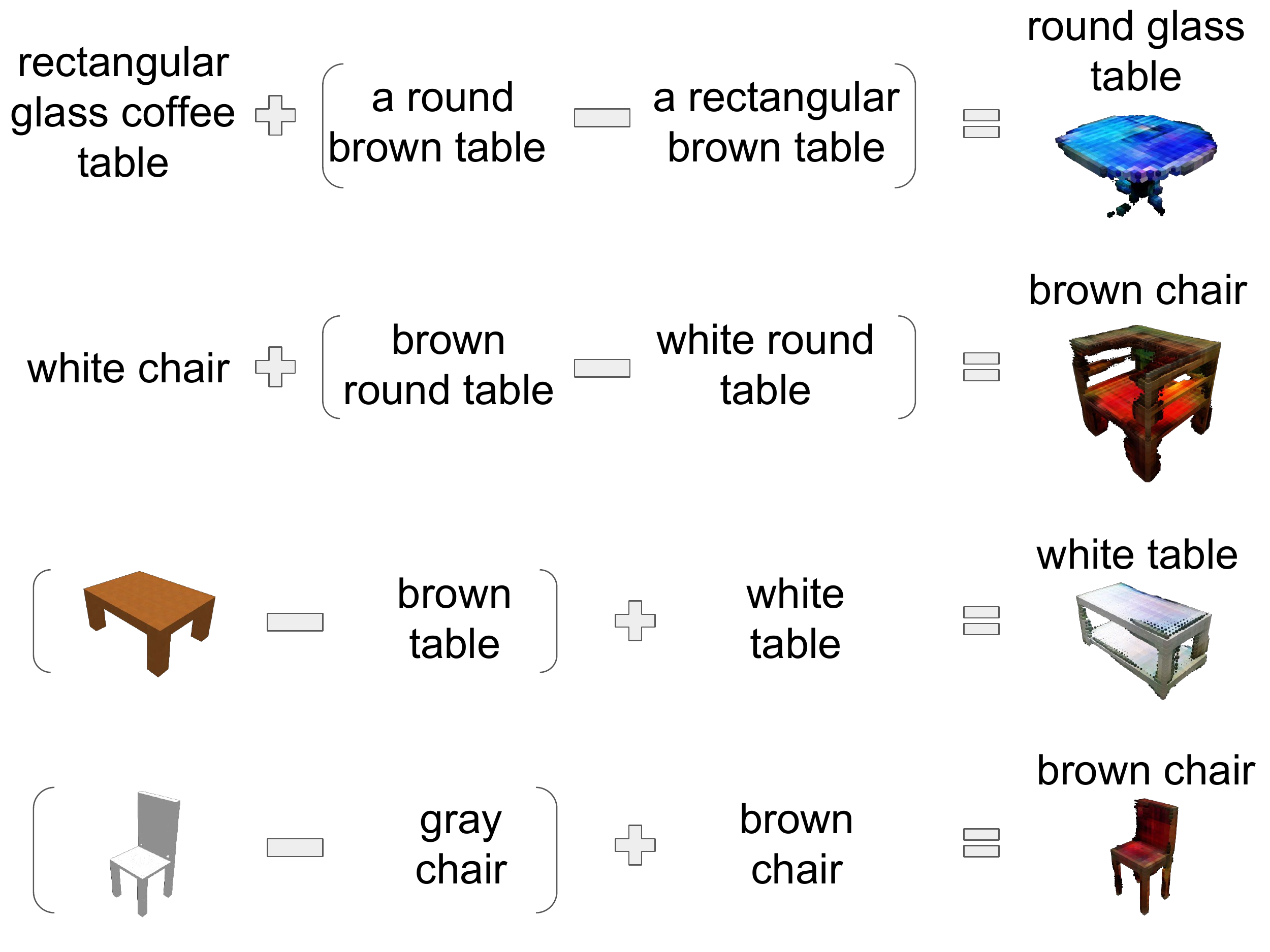}
	\end{subfigure}
	\vspace{-2em}
	\caption{\textbf{Vector arithmetic on joint text-shape embeddings}. Our embeddings admit composition and transfer of attributes to control shape generation.}
 	\vspace{-2em}
	\label{fig:vector-arithmetic}
\end{figure}

%% file: sec/conclusion.tex
\vspace{-0.5em}
\section{Conclusion}
\vspace{-0.5em}
We presented a method for learning joint embeddings of text and 3D shapes, trained end-to-end using only instance-level natural language descriptions of 3D shapes.
We showed that the learned embeddings enable retrieval between the text and shape modalities, outperforming approaches from prior work. We then combined our embedding with a conditional Wasserstein GAN formulation for the new task of text-to-shape generation. This is a challenging problem, and our approach is only a first step. To improve the quality of generated shapes we could use stronger priors to model real-world color distributions or leverage the bilateral symmetry common in physical objects. We hope that our work catalyzes more research in connecting natural language with realistic 3D objects exhibiting rich variations in color, texture, and shape detail.



%% file: sec/appendix.tex
\section{Model Details}
\label{sec:model-details}

In this section, we describe the details of our model architecture, its implementation, and training procedures. The architectures are outlined in \Cref{tab:architecture}. We plan to release our datasets and our code.

\mypara{Text encoder architecture.}
As stated in the main paper, we use a CNN RNN architecture. Each word is mapped to a random, trainable vector and passed to the CNN RNN, which has the structure outlined in \Cref{tab:architecture}. We use a GRU with 256 units. All layers except the last are followed by a ReLU operator and all weights have an L2 regularizer with weight 0.0005.

\input{tab/architectures.tex}

\mypara{Shape encoder architecture.}
The shape encoder takes the form of a standard 3D-CNN, utilizing 3D convolutions. All layers except the last are followed by a ReLU.

\mypara{Shape generation architecture.}
For upsampling, we use fractionally-strided convolutions. All layers except the last are followed by a ReLU. The last layer is followed by a sigmoid activation function.

For the critic, we use a leaky ReLU activation function with a leak rate of 0.2 in all layers except the last (which has a sigmoid activation for the non-Wasserstein GAN implementations). The text embeddings are passed through two 256-dimension FC layers (with leaky ReLU activations and batch normalization) and concatenated with the conv5 output, represented as the concatenation (concat) layer in \Cref{tab:architecture} (Critic). For the Wasserstein GAN critic, we do not use batch normalization. However, batch normalization is used in the non-Wasserstein GAN implementations.

\mypara{Text-conditional Wasserstein GAN.}
The text-to-shape generation model is illustrated in \Cref{fig:gan-overview}. As stated in the main paper, using the text encoder trained by our joint representation learning approach, the text is mapped to the latent space followed by concatenation with noise. The resulting vector is passed to the generator, which attempts to generate a plausible shape. This shape is then evaluated by the critic and the final loss is computed.

\input{fig/gan_overview.tex}

\mypara{Training details.}
For the joint representation learning, we use both learning by association and metric learning. For the association learning, we sample 100 unique shapes in each batch, and two matching captions per shape. We add a loss for when the embedding norm exceeds a threshold of 10.

For training the GAN, we use a learning rate of $5 \times 10^{-5}$ with a decay rate of 0.95 every $10000$ steps. In the first 25 training iterations, we run 100 critic training steps for every generator step. Afterwards, we train the critic 5 steps for every generator iteration. We use a gradient penalty coefficient of 10, similar to Guljarani et al.~\cite{gulrajani2017improved}. When training the GAN baselines, we used a procedure similar to Wu et al.~\cite{3dgan} in that we only update the discriminator weights if the accuracy in the previous batch is below 80\%. Note that we use the solid 32 resolution voxelizations described in \Cref{sec:shapenet-voxelization}.

\section{Dataset Details}
\label{sec:dataset-details}

We represent a shape as a colored (RGB) 3D voxel grid.  We use as training data, text description--3D voxel shape pairs $(x, s)$.  We create two such paired datasets to evaluate our method: a controlled, procedurally generated dataset of colored 3D geometric primitives, and a real 3D CAD object based on ShapeNet~\cite{shapenet2015}, augmented with natural language descriptions provided by people.

\mypara{Data representation.}
Each text description--3D shape voxel grid pair $(x, s)$ is composed of a textual description $x$ and a shape $s \in \mathbb{R}^{v^3 \cdot c}$. Here, $x$ is represented as an arbitrary length vector of word indices, where each word index corresponds to a particular word in the vocabulary, $n$ denotes the number of words in the description, $v$ is the number of voxels along the largest single dimension, and $c$ is the number of channels of the voxel grid (equal to 4 for RGB voxels).

\input{tab/dataset-statistics.tex}

\subsection{Primitive Shapes Dataset}

\paragraph{Primitive shape generation.}
We create a dataset of 3D geometric primitives with corresponding text descriptions (see \Cref{fig:primitives}).  This data is generated by voxelizing 6 types of primitives (cuboids, ellipsoids, cylinders, cones, pyramids, and tori), in 14 color variations and 9 size variations.  The color and size variations are subjected to random perturbations generating 10 samples from each of 756 possible primitive configurations, thus creating 7560 voxelized shapes.  The perturbations were implemented by applying a uniformly sampled noise for the color of the shape in HSV space (\%5 noise for hue, and 10\% for saturation and value), and 15\% noise for the height and radius scales of the shape.

\mypara{Sentence templates for synthetic dataset.}
We then create corresponding text descriptions with a template-based approach that fills in attribute words for shape, size, and color in several orderings to produce sentences such as \emph{``a large red cylinder is narrow and tall''}.  We use 10 template patterns to generate these text variations.  Two example templates are: ``\texttt{a \$\{largeness\} \$\{tallness\} \$\{wideness\} \$\{color\} \$\{shape\}}'' and ``\texttt{the \$\{shape\} is \$\{largeness\}} \\ \texttt{\$\{tallness\} \$\{wideness\} \$\{color\}}''.  Each of the variables are filled in with the term that applies to the specific primitive configuration.  In total, we generate 191,850 descriptions, for an average of about 255 descriptions per primitive configuration.  Such synthetic text does not match natural language but it does allow for an easy benchmark with a clear mapping to the attributes of each primitive shape.

\input{fig/primitives-examples.tex}

\mypara{Examples of generated primitives.}
See \Cref{fig:primitives} for some examples of generated primitive voxelizations and their corresponding text descriptions.

\subsection{ShapeNetCore Descriptions and Voxelizations}
\label{sec:shapenet-voxelization}

To create a more realistic dataset with real 3D objects and natural language descriptions, we use the ShapeNet~\cite{shapenet2015} table and chair object categories (with 8,447 and 6,591 instances correspondingly).  These 3D CAD models were created by human designers to realistically represent real objects.  We focus on only these two object categories to explore intra-category variation by getting detailed descriptions for many object instances.  We chose the table and chair categories from ShapeNetCore because they have significant variation in geometry and appearance.

We augment this dataset with 75,344 natural language descriptions (5 descriptions per CAD model) provided by people on the Amazon Mechanical Turk crowdsourcing platform (see \Cref{fig:voxelizations}).  The dataset construction is described in more detail in the following subsections.  This large-scale dataset provides much more challenging natural language descriptions paired with realistic 3D shapes.

\input{fig/description-ui.tex}

\mypara{Collecting shape descriptions.}
We collected natural language descriptions from people on the Amazon Mechanical Turk crowdsourcing platform.  The workers were shown a 3D animation of a rendered ShapeNet object (slowly rotating around the vertical axis to show a view of the object from all directions).  See \Cref{fig:description-ui} for a screenshot.  We had experimented with using a full 3D view of the object that allowed users to manipulate the camera but found that it was much slower for users to perform the task (mainly due to 3D data download size and the requirement for GPU hardware).  Most users did not manipulate the view using camera controls so we decided it was more efficient to show a pre-rendered rotating view as an animation.  One repercussion of this decision was that some people mentioned the fact that the object was rotating in their descriptions.

We instructed the workers to describe the appearance of the object using the prompt: ``describe the color, shape, material, and physical appearance of the object to another person''.  There was no restriction on the number of words and the number of sentences.  Many of the descriptions consisted of multiple sentences.

To understand what aspects of the shape people tend to describe, we sampled 100 text descriptions.  Based on this sample, 95\% of the descriptions described the type of object, 86\% described parts of the object and how they fit together, 83\% percent talked about color, 63\% material, and 47\% descibed the geometric shape of some part.  Some descriptions also mention other features such as functionality, comfort, and the aesthetics of the object.


\mypara{Color voxelization of ShapeNet models.}
We obtain color voxelizations from the ShapeNet 3D mesh models using a hybrid surface sampling and view-based approach.  This hybrid approach was designed to reliably sample all surfaces of an object even if they occur in concavities that are less likely to be observed from external viewpoints.  The task of computing color voxelizations for ShapeNet models is not trivial due to challenging properties of the 3D mesh representations: close and coplanar surfaces with different materials, thin structures, and significant variety in texture resolution and level of detail.

\input{fig/voxelizations.tex}

We perform our voxelization at a resolution of $256^3$ and then downsample to lower resolutions, with a low pass filter to remove sampling artifacts due to high frequency color variations in textures (see \Cref{fig:filtering}).  Additionally, we obtain two versions of the voxelization: one with voxels for only the surfaces, and one with the interior of the objects filled-in to represent the solid volume.  The solid voxelization is based on the surface voxelization, so we describe the method for surface voxelization first.

\mypara{Surface colored voxelization.}
To obtain a surface voxelization we first render the 3D mesh from a set of canonical views of canonical views orbiting around the centroid of the object.  We mark all surface triangles that are visible from these views.  We then perform two passes of surface point sampling of the 3D mesh.  In the first pass, we sample only from surface triangles that were marked visible, whereas in the second pass we sample all triangles except for triangles assigned to the material with lowest total visibility.  This two pass scheme helps to fill in unseen surfaces, while preventing samples from backward facing surfaces that usually have neutral white or gray material.  In total, we take two million sample points uniformly distributed on the surface of the 3D mesh, along with the associated color at each point.  We assign a higher weight to voxels that have surface normal pointing to the exterior of the shape.  When determining the color of a specific voxel, we take all samples with the maximal weight, and compute the color by taking the median sample in HSL space (sorting only on the hue channel).

\mypara{Solid colored voxelization.}
Solid color voxelizations are obtained by first performing a binary solid voxelization using the parity count and ray stabbing~\cite{nooruddin03} methods as implemented in binvox~\cite{binvox}.  Since internal colors are ill-defined, we choose to just propagate from the closest colored voxel to each uncolored voxel.  To check the quality of both solid and surface voxelizations, we render the resulting voxel grids and manually check approximately 10\% of the data.  \Cref{fig:filtering}, left shows an example of a solid slice from the interior of the solid voxelization.

\mypara{Low pass filtering.}
Sampling the colors using the method described above alone results in voxelizations with noisy colors, as shown in \Cref{fig:filtering}. These inaccurate colors are due to aliasing. To overcome this, we low-pass filter the high-resolution ($256 \times 256 \times 256$) colors using a Gaussian filter with a standard deviation of 0.5. To prevent any color shift, we element-wise divide the filtered voxels by a low pass filtered occupancy mask (of 0s and 1s). Lastly, we sample the high-resolution model to the desired resolution. \Cref{fig:filtering} shows that after filtering, we obtain much smoother voxelizations retaining textures and patterns even in low resolution voxels.  In our experiments, we use $32^3$ resolution solid voxelizations.

\input{fig/voxelization-examples.tex}

\mypara{Examples.}
\Cref{fig:voxelizations} shows several example ShapeNet object voxelizations along with the text of a description provided by a crowd worker.  Each shape has 5 such text descriptions.  \Cref{fig:filtering} shows all 5 descriptions for the armchair, which some descriptions being very succinct (a single ``armchair'') and other text describing in detail the color pattern and parts of the armchair.  Descriptions focus on different aspects of the object, and use various terms to refer to physical attributes.  For example, the crescent shaped table in \Cref{fig:voxelizations} first row, column 4 could be described as ``a T-shaped base connecting 3 legs'' rather than ``a wooden vine-shaped decoration base connecting the legs''.  Another difficult phenomenon is the use of negation.  For example the description for the blue chair in the last row, column 4 mentions the chair has ``no armrests''.  The diversity of language used to describe each object makes this a challenging dataset.  

\mypara{Limitations.}
This method for voxelization was designed to handle the specific data properties of ShapeNet 3D meshes.  It is however subject to some limitations.  Firstly, we did not attempt to reproduce transparent surfaces (see \Cref{fig:voxelizations} row 1, column 3 for an example of a voxelized glass coffee table).  It would be possible to extend our approach by tracking an alpha channel along with color, and assign opacity to each voxel.  This extension is possible since we use a hybrid surface sampling and view-based approach (it would be much harder with a purely view-based approach).  Secondly, the voxel colors only capture the diffuse component, and do not account for shading under illumination.  We discovered that this led to a disparity compared to our pre-rendered 3D mesh animations used for collecting natural language descriptions in the case of neutral white or light gray colors, which appeared darker under shading.  This led to a conflation of `white' with `gray' and `metallic' in the natural language descriptions (see \Cref{fig:voxelizations} bottom row, column 3 for an example).  To account for this we would either have to perform light-dependent shading in the sampled voxel colors, or render unshaded 3D meshes.  Finally, thin coplanar surfaces with different colors, or high frequency detail in textures are fundamentally challenging to capture using our fixed resolution voxel sampling scheme.

\subsection{Preprocessing of Natural Language Descriptions}

The text is lowercased, tokenized and lemmatized using the Spacy pipeline.\footnote{\url{https://spacy.io/}}  Since the natural language descriptions for ShapeNet had many mispellings, it was first spell-corrected using LanguageTool\footnote{\url{https://languagetool.org/}} (a combination of hunspell\footnote{\url{http://hunspell.github.io/}}, grammar rules, and custom spelling corrections for our dataset).  Extremely long descriptions with more than 96 tokens are discarded (of the 75360 descriptions we collected for ShapeNet, 16 were discarded due to length).  Low frequency terms (occurring $\leq 2$ times) were mapped to the UNK token and a dictionary representation was establishing using the remaining terms.  For the ShapeNet dataset, the total number of unique words was 8147, after dropping low frequency words and collapsing them into the UNK token, we are left with a vocabulary size of 3588.  The primitives dataset has no low frequency words, and has a small vocabulary size of 77 (with the UNK token).  \Cref{tab:dataset-stats} shows the overall statistics for the two datasets.  Note that the primitives dataset has many more descriptions per shape, but has much smaller language variation (with a much smaller vocabulary size and shorter descriptions).






\input{tab/primitives-same-modal-retrieval.tex}
\input{tab/primitives-cross-modal-retrieval.tex} 
\input{tab/shapenet-retrieval.tex}

\section{Additional Retrieval Results}
\label{sec:additional-retrieval-results}

We present additional retrieval experiment results.  These experiments are designed to demonstrate that our learned joint embedding of text and 3D shapes is structured to allow both within-modality and cross-modality retrieval. For within-modality retrieval, we provide results for text-to-text and shape-to-shape retrieval, in which we query an instance (e.g. description) and retrieve the nearest neighbors within the query's modality (e.g. other descriptions). For cross-modality retrieval, we query in one modality and retrieve in the other. We evaluate this through text-to-shape and shape-to-text retrieval tasks.

We first present quantitative results on all retrieval tasks (\Cref{sec:quantitative}). We then present qualitative results for text-to-text (\Cref{sec:text-text}), shape-to-shape (\Cref{sec:shape-shape}), and cross-modal retrieval (\Cref{sec:cross-modal}).

\subsection{Quantitative retrieval results}
\label{sec:quantitative}

We provide quantitative retrieval results for the primitives dataset in \Cref{tab:retrieval-same-modal}. For text-to-text retrieval, we see that our model is able to correctly retrieve all descriptions corresponding to the query configuration in the primitives dataset, significantly outperforming the baselines. 
Moreover, the association learning alone (LBA-TST) and metric learning alone (ML) fail to generalize well to the unseen configurations in the test set. However, when the two are trained jointly (Full-TST), we see a significant boost in performance. Lastly, in the original LBA work~\cite{haeusser2017learning}, the authors found that additional roundtrips do not boost performance for same-modal data. On the contrary, in working with multimodal data, we can see that using additional roundtrips in the MM model results in much higher retrieval performance. 

Similar to the text-to-text retrieval results, our model outperforms the baselines on both the shape-to-shape and text-to-shape retrieval tasks as shown in \Cref{tab:retrieval-same-modal} and \Cref{tab:retrieval-cross-modal}. Association learning and metric learning alone also underperform the DS-SJE baseline, but jointly training the two methods results in the best performing model.

Quantitative ShapeNet retrieval results are shown in \Cref{tab:shapenet-retrieval}. Since we only have instance-level labels, a retrieval is considered correct if the retrieved item (text description or shape) belongs to the same instance as the query. In other words, we categorize a retrieved sentence to be correct only if it came from the same model as the query sentence. If the retrieved sentence offers similar semantic meaning but came from a different model, the retrieval is marked as incorrect. Since there are numerous descriptions and shapes in the dataset which are semantically similar but are not labeled to be, the absolute numbers do not entirely capture the performance of the model. However, higher quantitative numbers still correlate with better performance. We do not report numbers on shape-to-shape retrieval since every shape is in its own instance-level class.

From \Cref{tab:shapenet-retrieval}, we can see that for text-to-text retrieval, our model outperforms the baseline methods, and that LBA alone and metric learning (ML) alone significantly underperform our full method (Full-MM). Moreover, utilizing additional round trips (MM) brings significant improvement compared to not using the round trips (TST). 

For the cross-modal retrieval, our TST model (Full-TST) actually performs slightly better than Full-MM on the shape-to-text retrieval task, though the performance of the two models are still comparable with each other for this task. Moreover, we see that our full model (Full-MM) outperforms all baselines for text-to-shape retrieval.  In the following subsection we give more examples for each retrieval task.







\subsection{Text-to-text Retrieval}
\label{sec:text-text}

Given a sentence in the test set, we retrieve the nearest sentences. Qualitative results of our model on the ShapeNet dataset are in \Cref{fig:text-text-retrieval}. Almost all of the retrieved descriptions come from different shapes, but they are semantically similar. For example, in the first row, we see that the model learns ``oval'' is similar to ``circular''. In the second example, all retrieved descriptions describe red chairs, and in the last example, the retrieved descriptions describe glass coffee tables.

\input{fig/text-text-retrieval.tex}




\subsection{Shape-to-shape Retrieval}
\label{sec:shape-shape}

\input{fig/shape-shape-retrieval.tex}


Given a shape, we retrieve the nearest shapes according to our learned embedding. We show qualitative results for shape-to-shape retrieval in \Cref{fig:shape-shape-retrieval}. The retrieved neighbors often correspond with the query in style and shape. For example, the retrieved neighbors of the brown armless chair are all also wooden armless chairs. When a sofa is queried, the nearest neighbors are also sofas. Likewise for tables, the retrieved neighbors have similar attributes (e.g. rectangular vs. round table). 

We compare our model against the baselines in \Cref{fig:shape-shape-retrieval-comparison}. While the LBA (TST) method is able to perform text-to-text retrieval fairly well, its performance is much worse for shape-to-shape retrieval. For example, when a chair is queried (second query of \Cref{fig:shape-shape-retrieval-comparison}), the LBA (TST) method retrieves tables. Similarly, the DS-SJE method does not perform well either, returning chairs which share almost no similar attributes. However, our method successfully retrieved chairs which look like they are of the same material, color, and style.

\input{fig/shape-shape-retrieval-comparison.tex}

\subsection{Cross-modal Retrieval}
\label{sec:cross-modal}


We provide additional results for the text-to-shape experiments described in the main paper. Qualitative results for ShapeNet text-to-shape retrieval are shown in \Cref{fig:cross-modal-retrieval-comparison}. The DS-SJE~\cite{reed2016learning} method seems to return the same results whenever a table caption is queried, regardless of the details in the description. This is because the representations learned by DS-SJE~\cite{reed2016learning} most likely do not capture information about the attributes. Rather, they capture category information (chair/table) only. The LBA (TST) method does not always retrieve shapes of the correct category (e.g. returns chairs instead of tables). Our method is able to understand certain details such as a ``blue glass top'' and retrieves shapes which indeed match the given description.

\input{fig/cross-modal-retrieval-comparison.tex}


\section{Additional Generation Results}
\label{sec:additional-generation-results}

In this section we present additional text-to-shape generation results (\Cref{sec:t2s-generation}) and vector arithmetic results (\Cref{sec:vector-arithmetic}).

\subsection{GAN generation results}
\label{sec:t2s-generation}

We show more generation results in \Cref{fig:text2shape-results-supplementary} and \Cref{fig:text2shape-results-supplementary-2}. In general, the GAN-INT-CLS approach~\cite{reed2016generative} struggles to properly condition on the text. For example, none of the glass or white tables (examples 4, 6, 7, 11, 14, 18, 19) are of the right color. Moreover, the the model exhibits mode collapse -- all of the chairs look the same and the tables in examples 4, 7, 8, 14, 17, 19, and 20 all look very similar. These poor generations are primarily due to the quality of the learned text embeddings, but also due to their GAN formulation as well. Because the representations are not trained to distinguish more than the chair/table categories, the generator and critic have a hard time learning how to properly condition on the input embeddings.

The CGAN method performs better than the GAN-INT-CLS baseline~\cite{reed2016generative}. Using our joint representation learning approach, the CGAN is able to properly condition on the text description such that the generated models usually have the correct category and color. However, we still see mode collapse, as examples 5, 8, 9, and 13 are very similar looking brown tables, and examples 1, 3, and 15 are very similar looking chairs. Furthermore, we see that the color distributions (such as in examples 3, 5, 9, 14, 18, and 19) look artificial and unsmooth.

Lastly, we see that our CWGAN produces the most realistic generations which also properly match the ground truth in terms of color and category. In example 6 for example, the CWGAN method is the only one to produce a table which is white. The other approaches fail by either generating a chair or generating a table of the wrong color. Moreover, we notice that the generations from the CWGAN model have much higher diversity compared to the other approaches -- a property that the Wasserstein GAN~\cite{arjovsky2017wasserstein,gulrajani2017improved} had over the the traditional GAN formulation~\cite{goodfellow2014generative}. We see large variations in the color, shape, and style of the generated CWGAN voxels.

From these results, we can also see that there is room for improvement. For example, the tables in examples 6 and 20 are rectangular rather than square. The round table in example 12 does not have three legs, and the desk in example 4 is not exactly an upside down U shape. Details in the descriptions such as these can be hard for the model to capture. Thus, while our model is able to significantly outperform the baselines, there is still much room for improvement in future work.

\input{fig/text2shape-results-supplementary.tex}

\input{fig/text2shape-results-supplementary-2.tex}

\subsection{Additional Vector Arithmetic Results}
\label{sec:vector-arithmetic}

Lastly, we provide additional results for vector arithmetic on the text and shape embeddings in \Cref{fig:vector-arithmetic-additional}. By performing simple operations on the learned representations, we can use our trained generator to produce novel shapes with specific attributes.

\input{fig/vector-arithmetic-additional.tex}

%% file: tab/architectures.tex

\begin{table}
\small\centering
\caption{Model architectures.}
\begin{tabular}{lcccc}
\toprule
Layer & Kernel & Channels & Stride & BN \\
\midrule
\multicolumn{5}{c}{Text encoder} \\
\midrule
conv1 & 3 & 128 & 1 & N \\
conv2 & 3 & 128 & 1 & Y \\
conv3 & 3 & 256 & 1 & N \\
conv4 & 3 & 256 & 1 & Y \\
GRU & -- & -- & -- & -- \\
fc5 & -- & 256 & -- & N \\
fc6 & -- & 128 & -- & N \\
\midrule
\multicolumn{5}{c}{Shape encoder} \\
\midrule
conv1 & 3 & 64 & 2 & Y \\
conv2 & 3 & 128 & 2 & Y \\
conv3 & 3 & 256 & 2 & Y \\
avg\_pool4 & -- & -- & -- & -- \\
fc5 & -- & 128 & -- & N \\
\bottomrule
\end{tabular}
\quad
\begin{tabular}{lcccc}
\toprule
Layer & Kernel & Channels & Stride & BN \\
\midrule
\multicolumn{5}{c}{Generator} \\
\midrule
fc1 & -- & 32768 & -- & Y \\
conv2\_trans & 4 & 512 & 1 & Y \\
conv3\_trans & 4 & 256 & 2 & Y \\
conv4\_trans & 4 & 128 & 2 & Y \\
conv5\_trans & 4 & 4 & 2 & N \\
\midrule
\multicolumn{5}{c}{Critic} \\
\midrule
conv1 & 4 & 64 & 2 & N \\
conv2 & 4 & 128 & 2 & N \\
conv3 & 4 & 256 & 2 & N \\
conv4 & 4 & 512 & 2 & N \\
conv5 & 2 & 256 & 2 & N \\
concat & -- & -- & -- & -- \\
fc6 & -- & 128 & -- & -- \\
fc7 & -- & 64 & -- & -- \\
fc8 & -- & 1 & -- & -- \\
\bottomrule
\end{tabular}
\vspace{-1em}
\label{tab:architecture}
\end{table}

%% file: fig/gan_overview.tex
\begin{figure}
    \includegraphics[width=\linewidth]{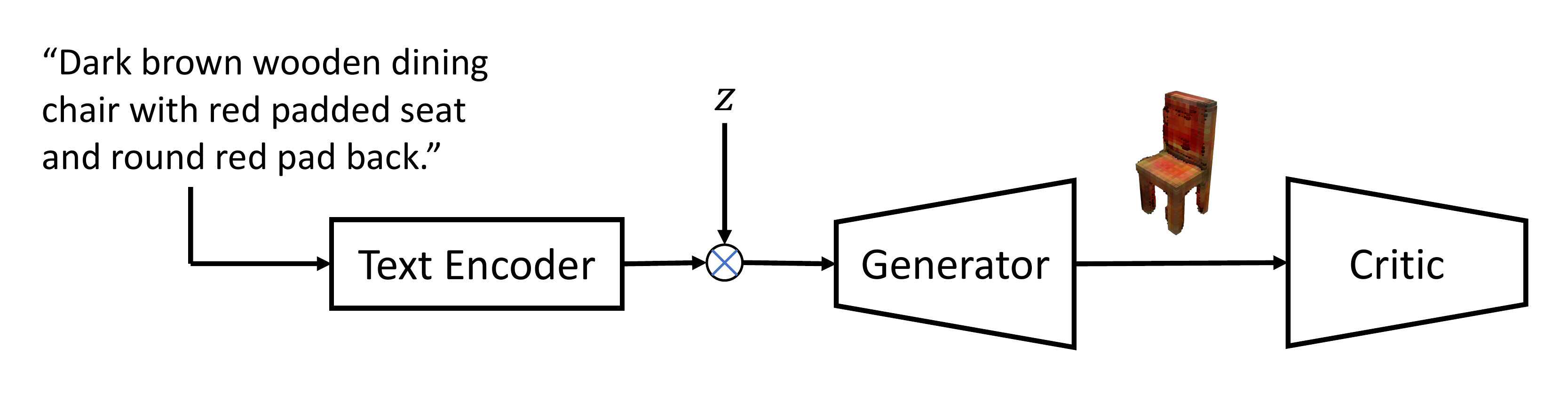}
    \caption{\textbf{Overview of our CWGAN framework.} We use a text encoder, trained using our joint representation learning method, to compute the latent embeddings. This is concatenated with noise $z \sim U[-0.5, 0.5]$ and fed into the generator. The output of the generator is input into a critic, which evaluates the plausibility of the generation given also the text embedding. Lastly, the loss is backpropagated through the networks.}
    \label{fig:gan-overview}
\end{figure}

%% file: tab/dataset-statistics.tex
\begin{table}
\small\centering
\caption{\textbf{Dataset statistics.}}
\begin{tabular}{lcc}
\toprule
 & Primitives & ShapeNet \\
\midrule
Shapes & 7,560 &  15,038 \\
Descriptions & 191,850 & 75,344 \\
Descriptions per shape & 255 & 5  \\
Total words & 1.3M & 1.2M \\
Ave words/description & 6.7 & 16.3  \\
Max words/description & 8 & 96 \\
Unique words & 76 & 8147 \\
Vocabulary size & 77 & 3,588 \\
\bottomrule
\end{tabular}
\label{tab:dataset-stats}
\end{table}

%% file: fig/primitives-examples.tex
\begin{figure}[t]
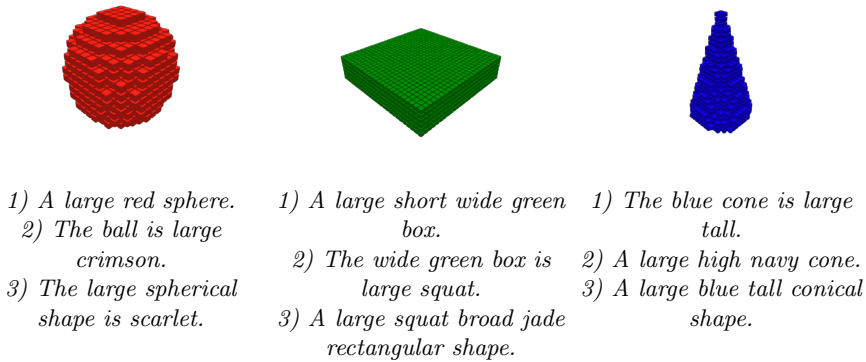

	\captionsetup[subfigure]{justification=centering,labelformat=empty}
	\footnotesize\centering
	\begin{subfigure}[t]{0.32\linewidth}
	    \centering
	    \includegraphics[width=0.7\linewidth]{fig/examples/sphere-red-h50-r50_0.png}
		\caption{\emph{1) A large red sphere.\\2) The ball is large crimson.\\3) The large spherical shape is scarlet.}}
	\end{subfigure}
	\begin{subfigure}[t]{0.32\linewidth}
	    \centering
        \includegraphics[width=0.7\linewidth]{fig/examples/box-green-h20-r100_0.png}
        \caption{\emph{1) A large short wide green box.\\2) The wide green box is large squat.\\3) A large squat broad jade rectangular shape.}}
	\end{subfigure}
	\begin{subfigure}[t]{0.32\linewidth}
	    \centering
		\includegraphics[width=0.7\linewidth]{fig/examples/cone-blue-h100-r50_0.png}
		\caption{\emph{1) The blue cone is large tall.\\2) A large high navy cone.\\3) A large blue tall conical shape.}}
	\end{subfigure}
	\caption{Example colored 3D voxel primitives with three generated text descriptions each.}
	\label{fig:primitives}
\end{figure}

%% file: fig/description-ui.tex
\begin{figure}
    \centering
    \includegraphics[width=0.7\linewidth]{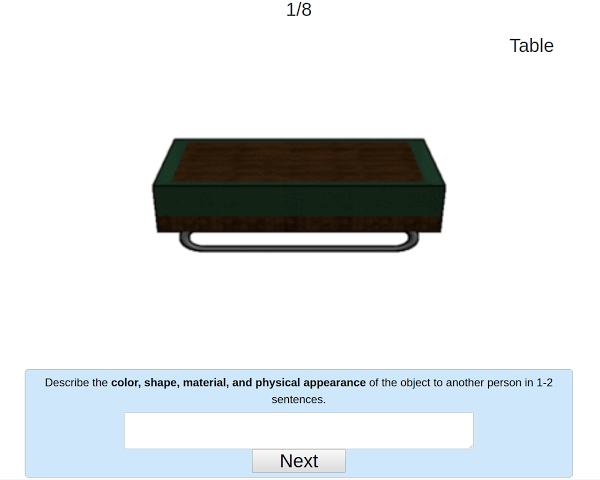}
	\caption{\emph{User interface we use for collecting natural language descriptions of 3D objects from crowdsourced workers.  The worker was shown a rotating view of the displayed object and asked to provide a freeform text description of the appearance of the object.}}
	\label{fig:description-ui}
\end{figure}

%% file: fig/voxelizations.tex
\begin{figure*}
    \small\centering
    \begin{tabular}{c|cccc}
        Original 3D mesh & $128^3$ voxelization & $64^3$ voxelization & $32^3$ voxelization\\
        \midrule
        \includegraphics[width=0.22\textwidth]{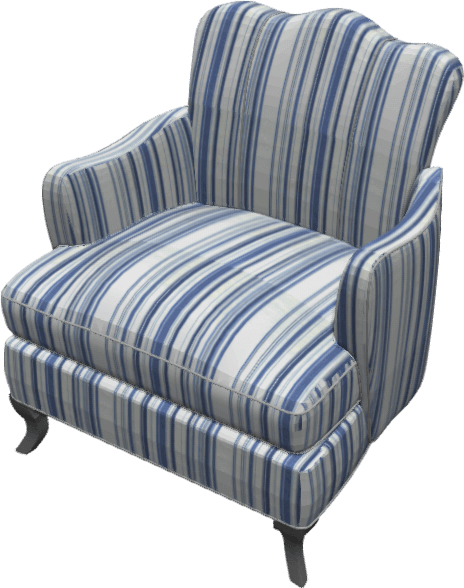} &
        \includegraphics[width=0.22\textwidth]{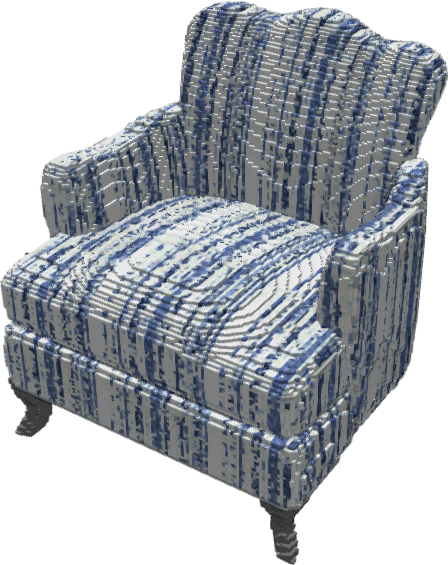} &
        \includegraphics[width=0.22\textwidth]{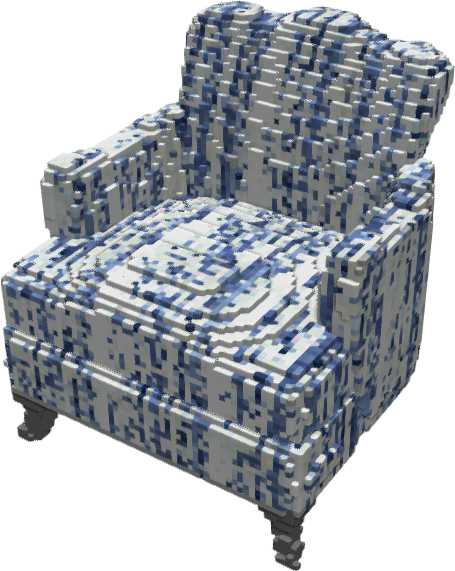} &
        \includegraphics[width=0.22\textwidth]{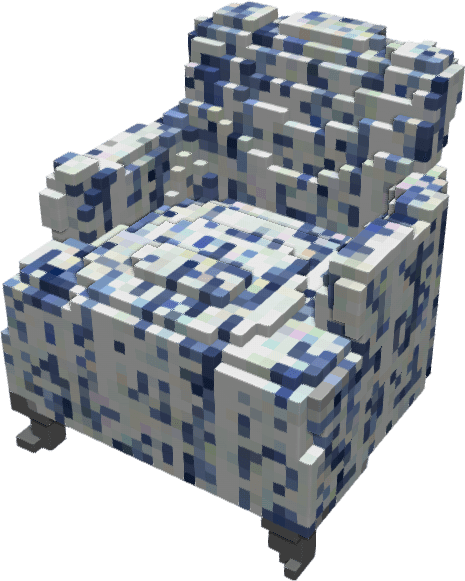}\\
        \midrule
        Interior voxelization & Filtered $128^3$ & Filtered $64^3$ & Filtered $32^3$ \\
         & voxelization & voxelization & voxelization\\
        \midrule
        \includegraphics[width=0.22\textwidth]{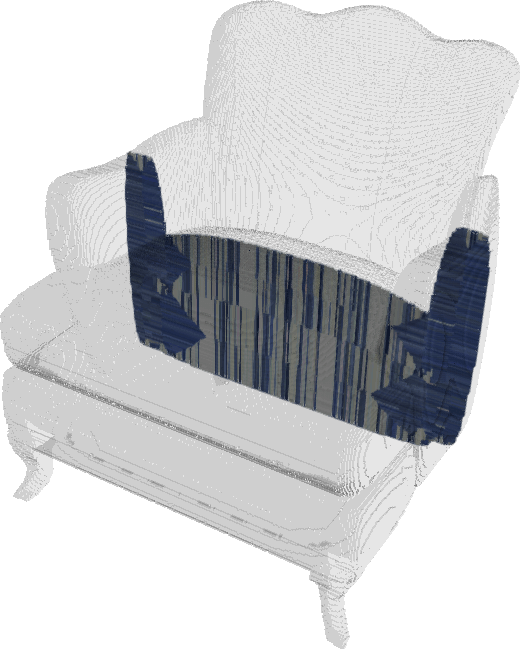} &
        \includegraphics[width=0.22\textwidth]{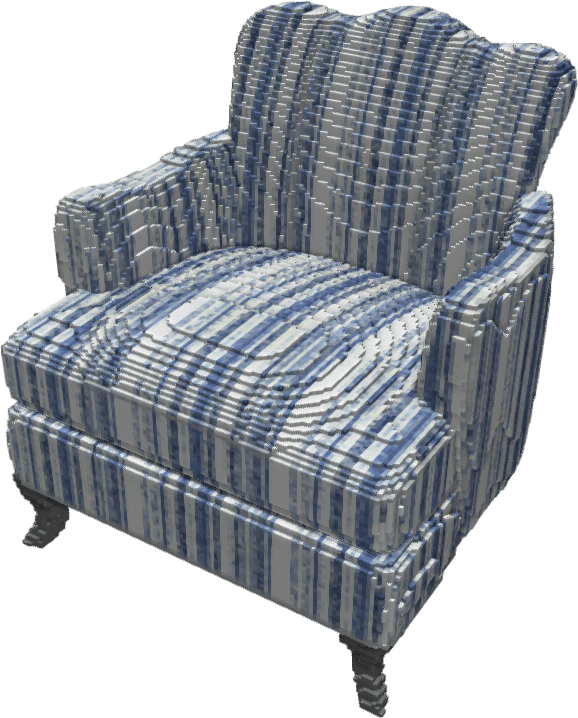} &
        \includegraphics[width=0.22\textwidth]{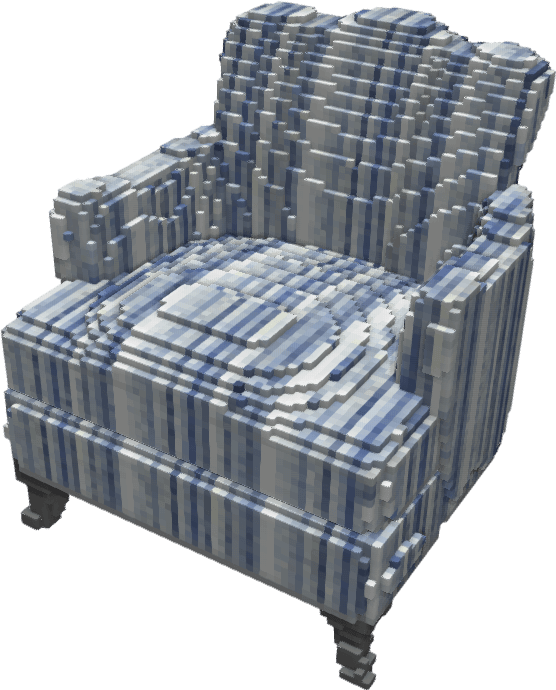} &
        \includegraphics[width=0.22\textwidth]{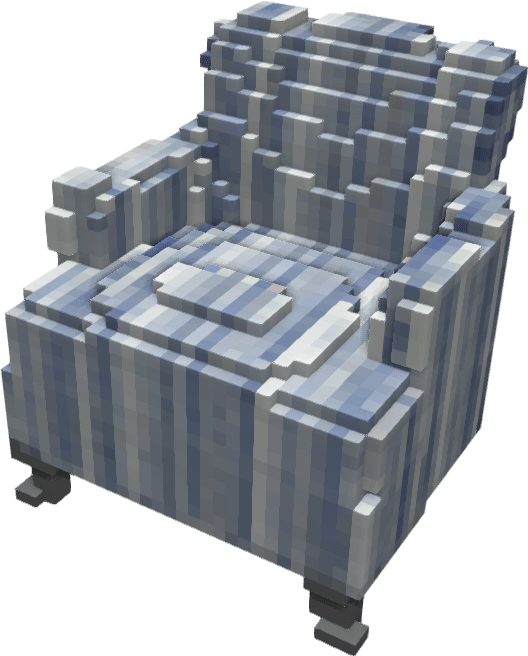}\\
    \end{tabular}
    \caption{Color voxelization of textured 3D chair mesh. \textbf{Left column}: original 3D mesh (top) and slice from solid voxelization (bottom). Columns to the right show unfiltered (top) and filtered (bottom) versions of the solid voxelization at resolutions of $128^3$, $64^3$ and $32^3$. The low-pass filtering of voxel colors during downsampling is important to mitigate aliasing artifacts due to sampling of the high frequency patterns in the chair fabric.}
    \label{fig:filtering}
\end{figure*}

%% file: fig/voxelization-examples.tex
\begin{figure*}
	\captionsetup[subfigure]{justification=centering,labelformat=empty}
	\footnotesize\centering
    \vspace{-1em}
    \begin{subfigure}[t]{0.24\linewidth}
        \includegraphics[width=\textwidth]{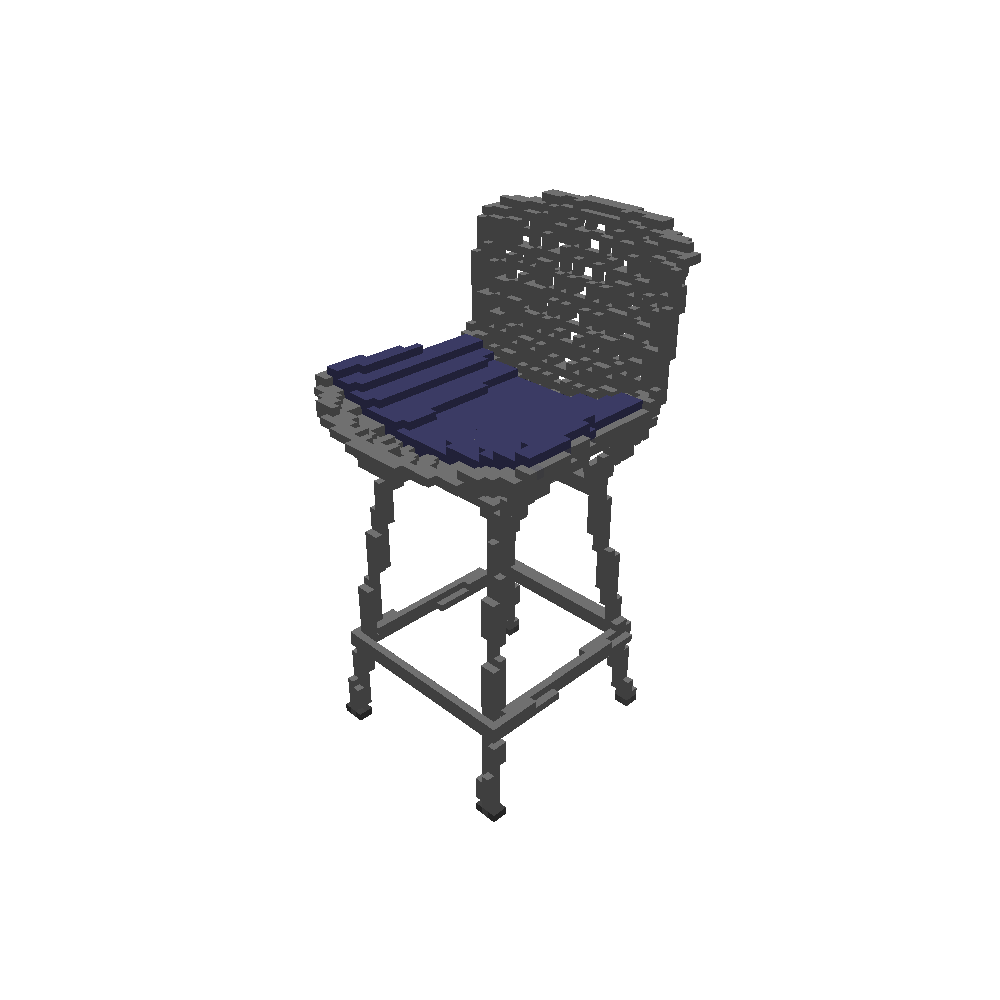}
        \vspace{-3em}
        \caption{\emph{A violet seated chair with netted seat rest and four legs}}
    \end{subfigure}
    \begin{subfigure}[t]{0.24\linewidth}
        \includegraphics[width=\textwidth]{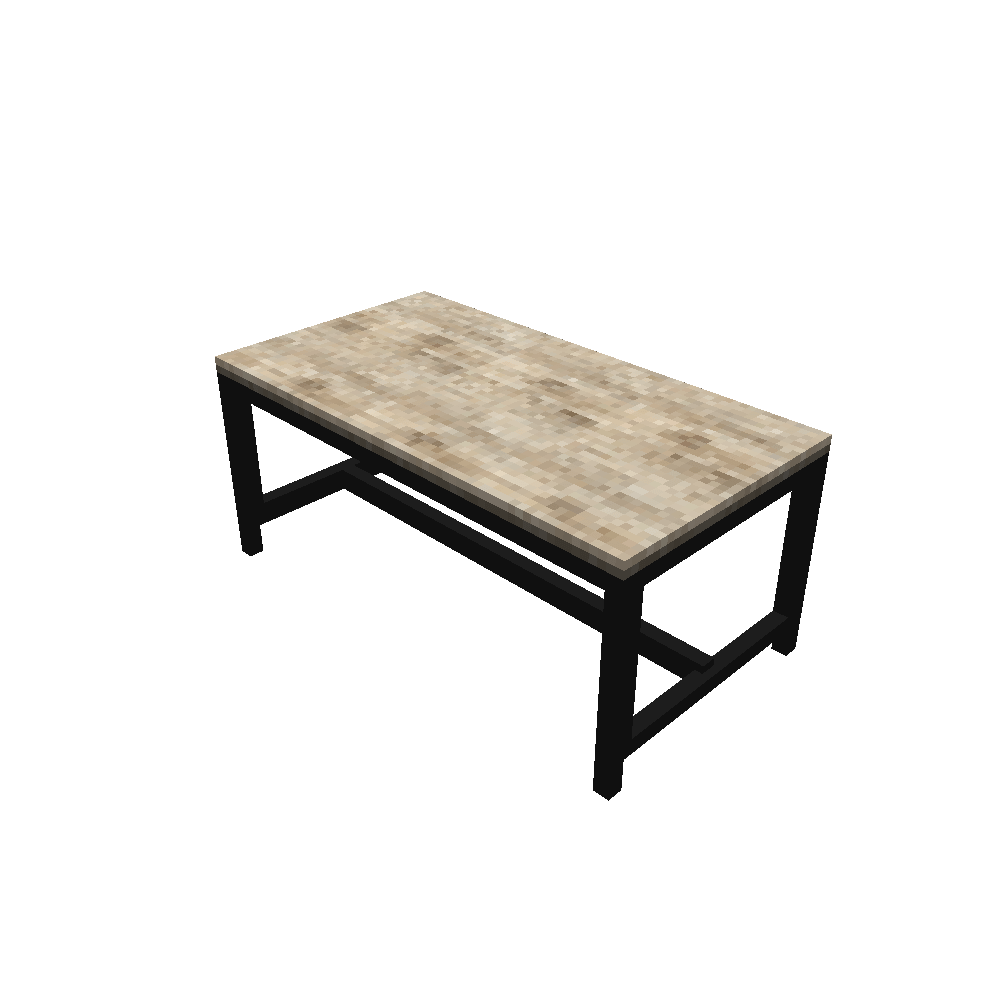}
        \vspace{-3em}
        \caption{\emph{A rectangular wooden coffee table with an iron base}}
    \end{subfigure}
    \begin{subfigure}[t]{0.24\linewidth}
        \includegraphics[width=\textwidth]{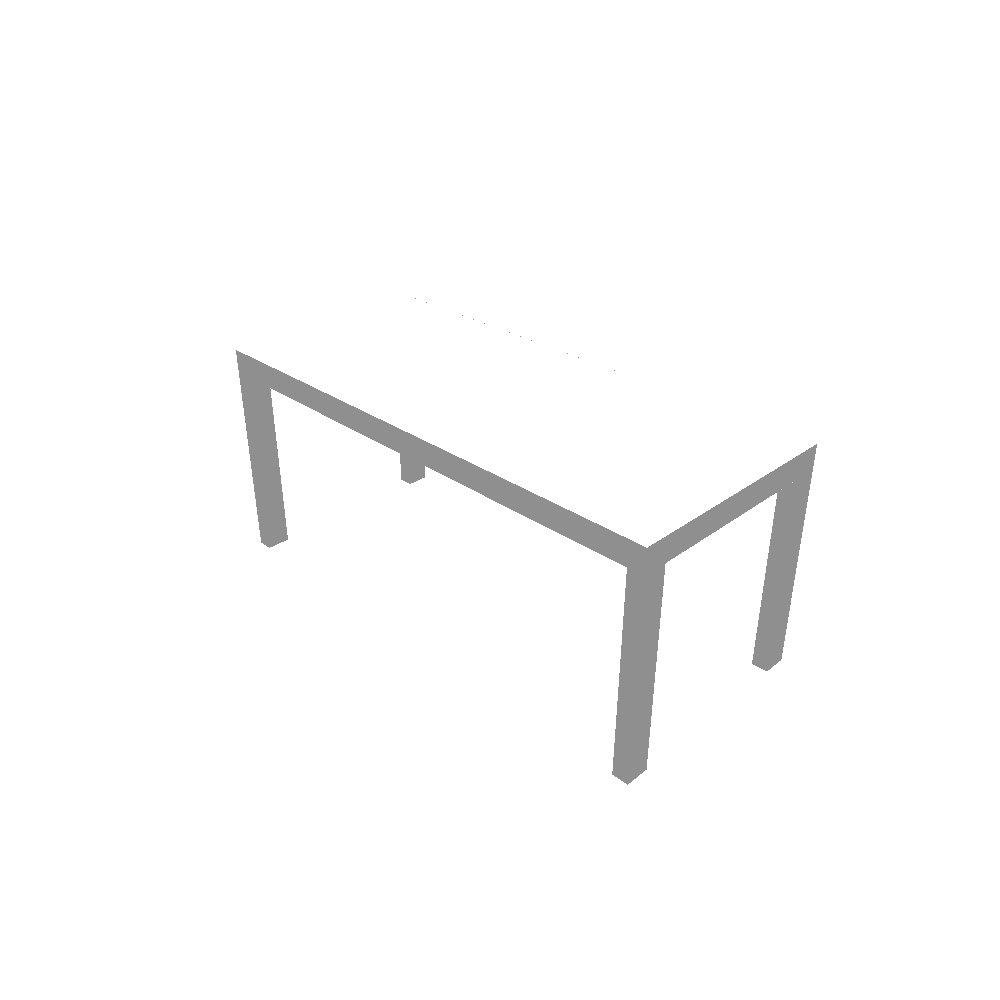}
        \vspace{-3em}
        \caption{\emph{A silver colored rectangular steel four legged table.}}
    \end{subfigure}
    \begin{subfigure}[t]{0.24\linewidth}
        \includegraphics[width=\textwidth]{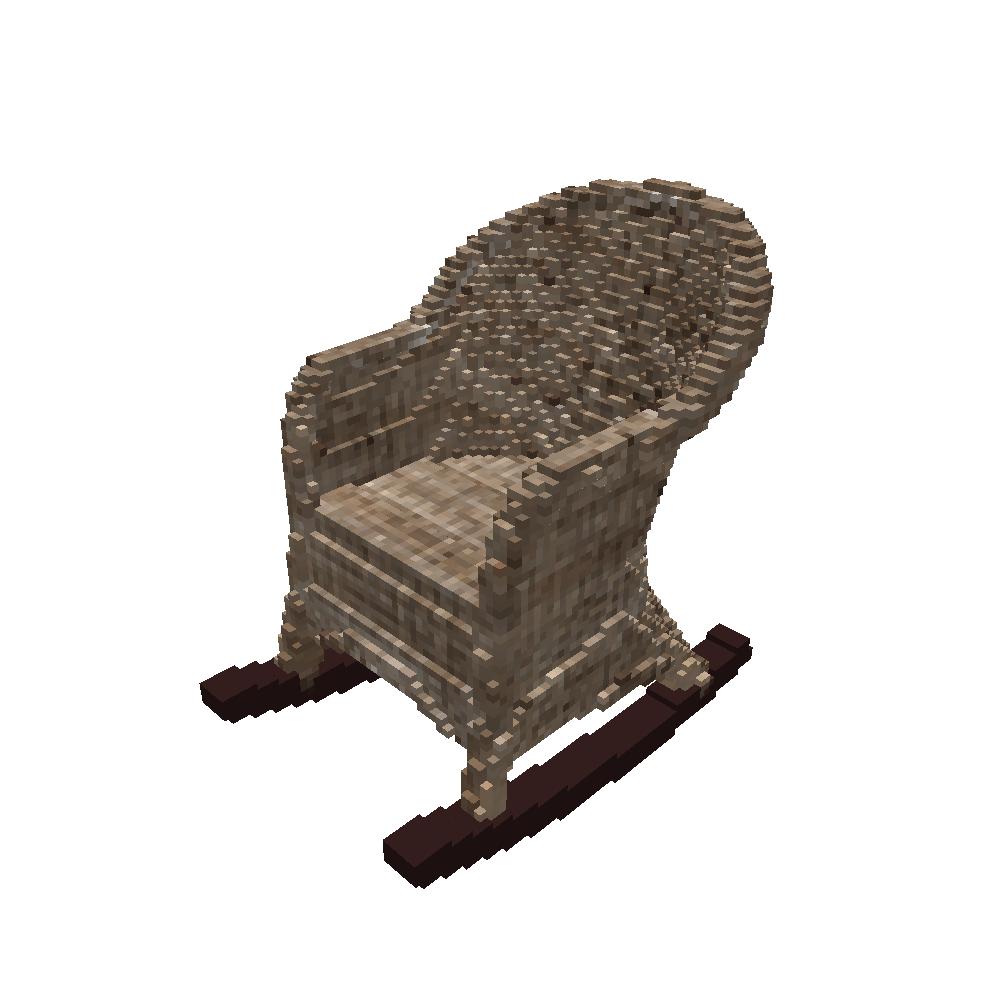}
        \vspace{-3em}
        \caption{\emph{Light brown rattan wicker rocking chair with rounded corners}}
    \end{subfigure}
    \vspace{-2em}
    \begin{subfigure}[t]{0.24\linewidth}
        \includegraphics[width=\textwidth]{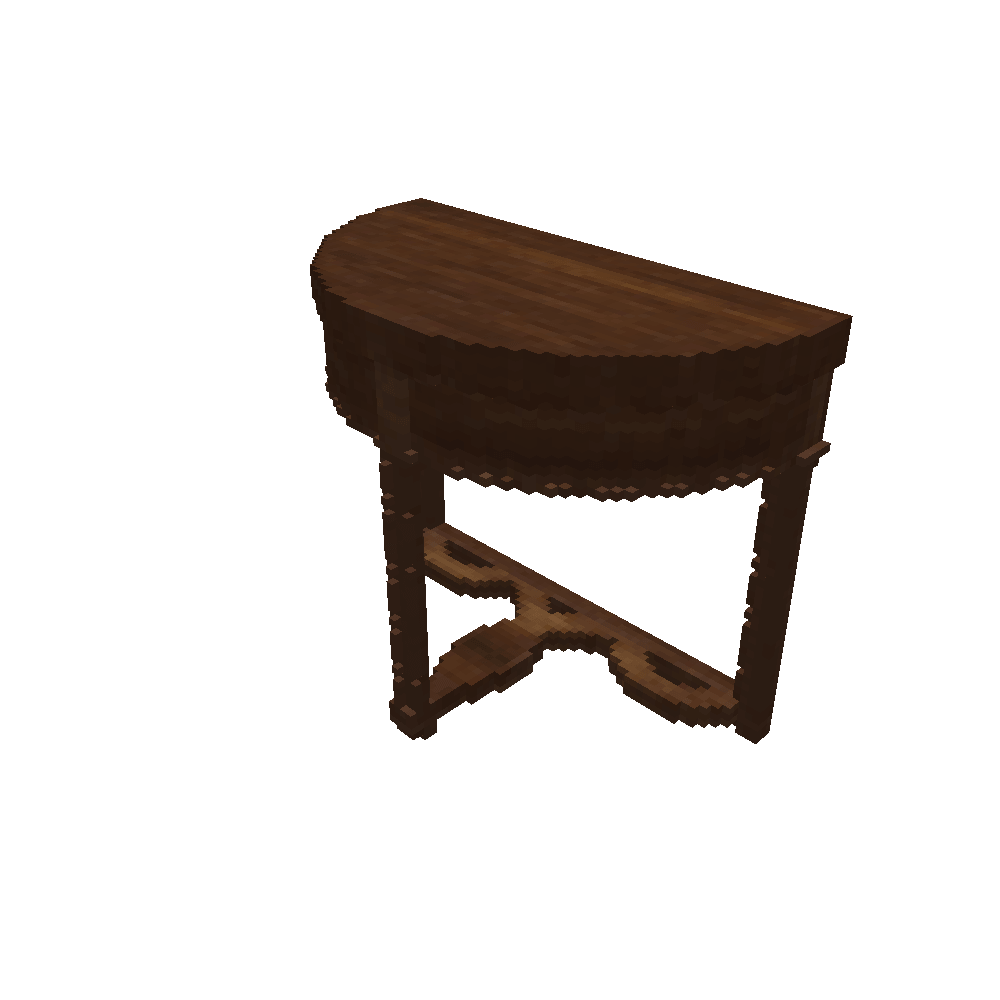}
        \vspace{-2.5em}
        \caption{\emph{A brown wooden moon shaped table with three decorative legs with a wooden vine shaped decoration base...}}
    \end{subfigure}
    \begin{subfigure}[t]{0.24\linewidth}
        \includegraphics[width=\textwidth]{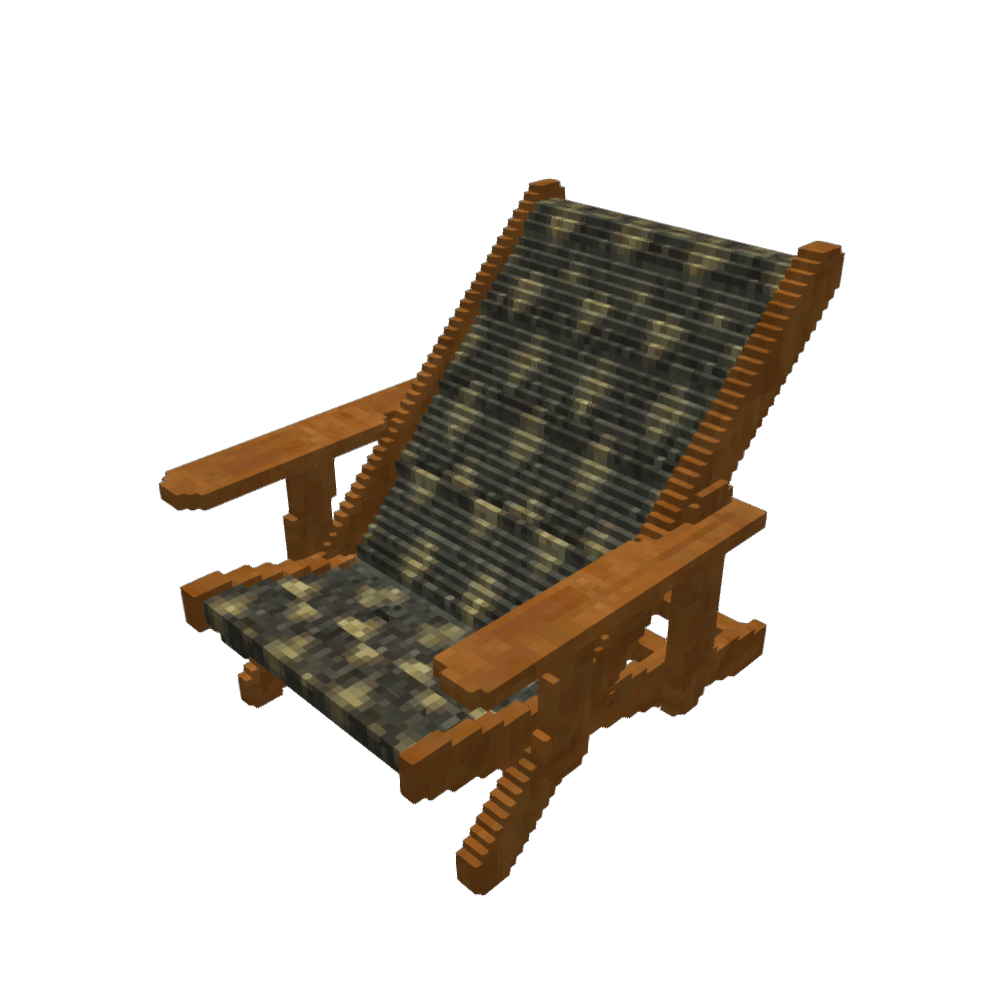}
        \vspace{-2.5em}
        \caption{\emph{Dark brown wooden chair with adjustable back rest and gold printed upholestry. Designed for comfort.}}
    \end{subfigure}
    \begin{subfigure}[t]{0.24\linewidth}
        \includegraphics[width=\textwidth]{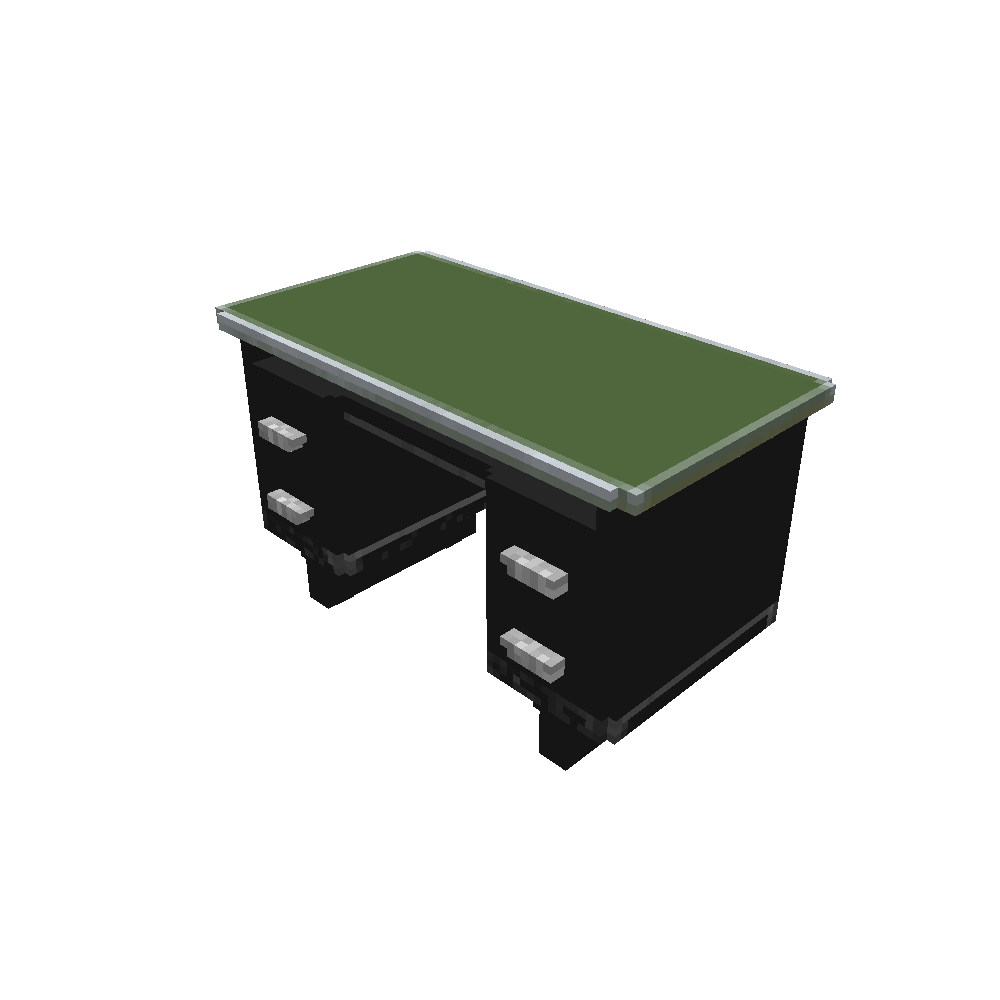}
        \vspace{-2.5em}
        \caption{\emph{A black coloured table with green table top, has two drawers on each side and supported by two legs}}
    \end{subfigure}
    \begin{subfigure}[t]{0.24\linewidth}
        \includegraphics[width=\textwidth]{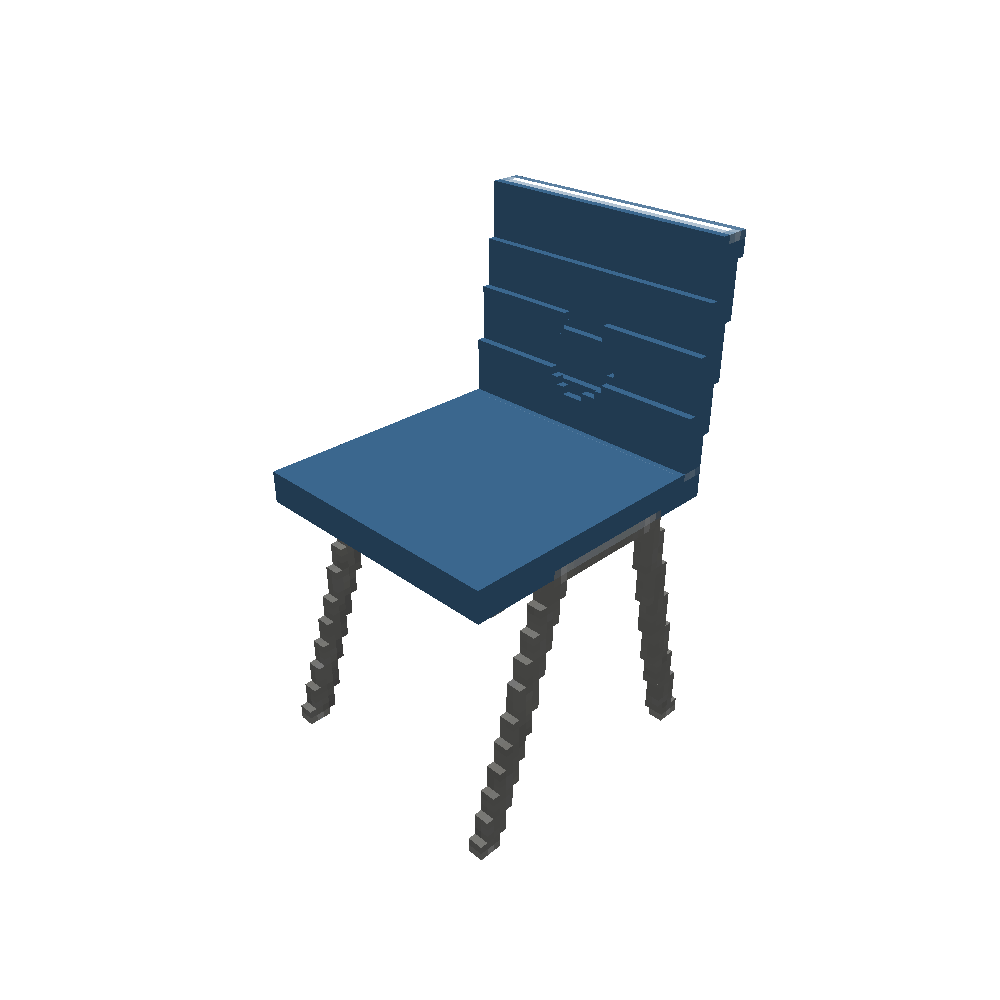}
        \vspace{-2.5em}
        \caption{\emph{Rectangular shaped metal chair with blue color, back rest is provided but no armrest, legs are splayed.}}
    \end{subfigure}
    \vspace{-1em}
    \begin{subfigure}[t]{0.24\linewidth}
        \includegraphics[width=\textwidth]{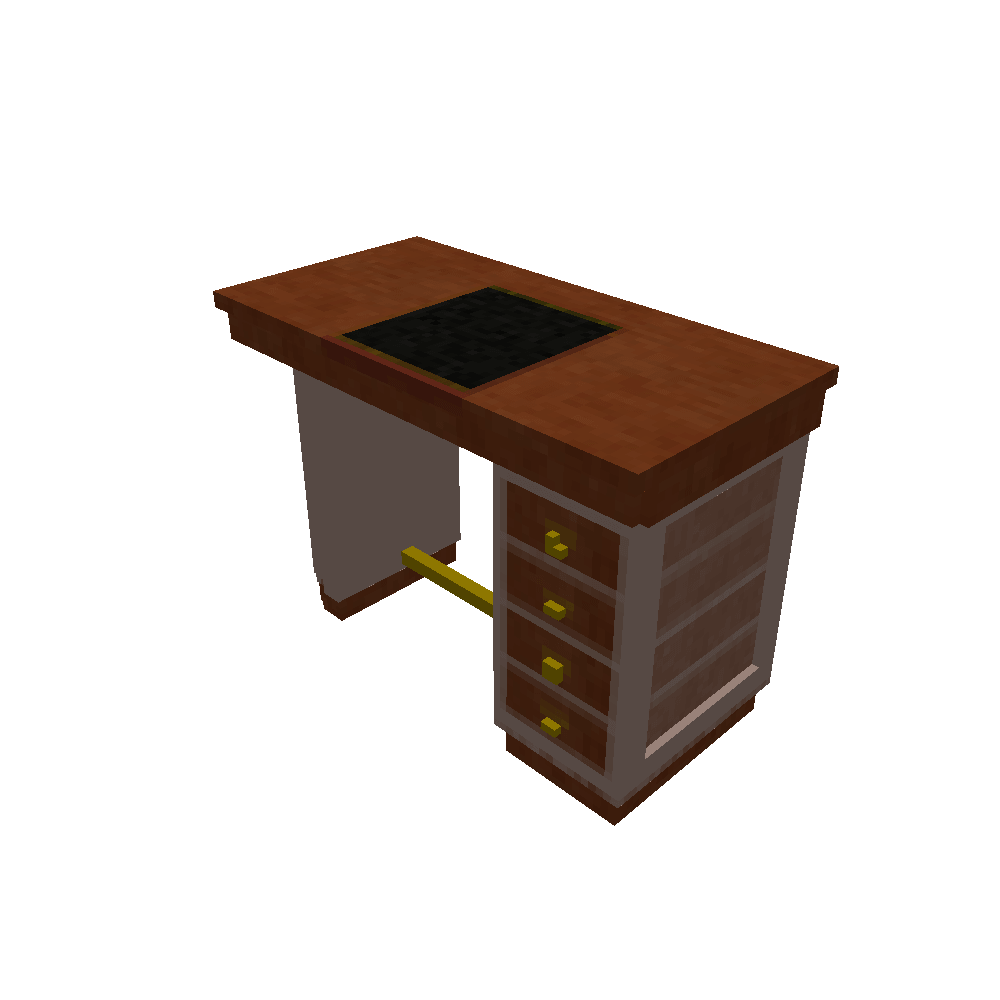}
        \vspace{-3em}
        \caption{\emph{Office desk that looks to be constructed by multiple materials such as brown wood for the top and drawers and grey...}}
    \end{subfigure}
    \begin{subfigure}[t]{0.24\linewidth}
        \includegraphics[width=\textwidth]{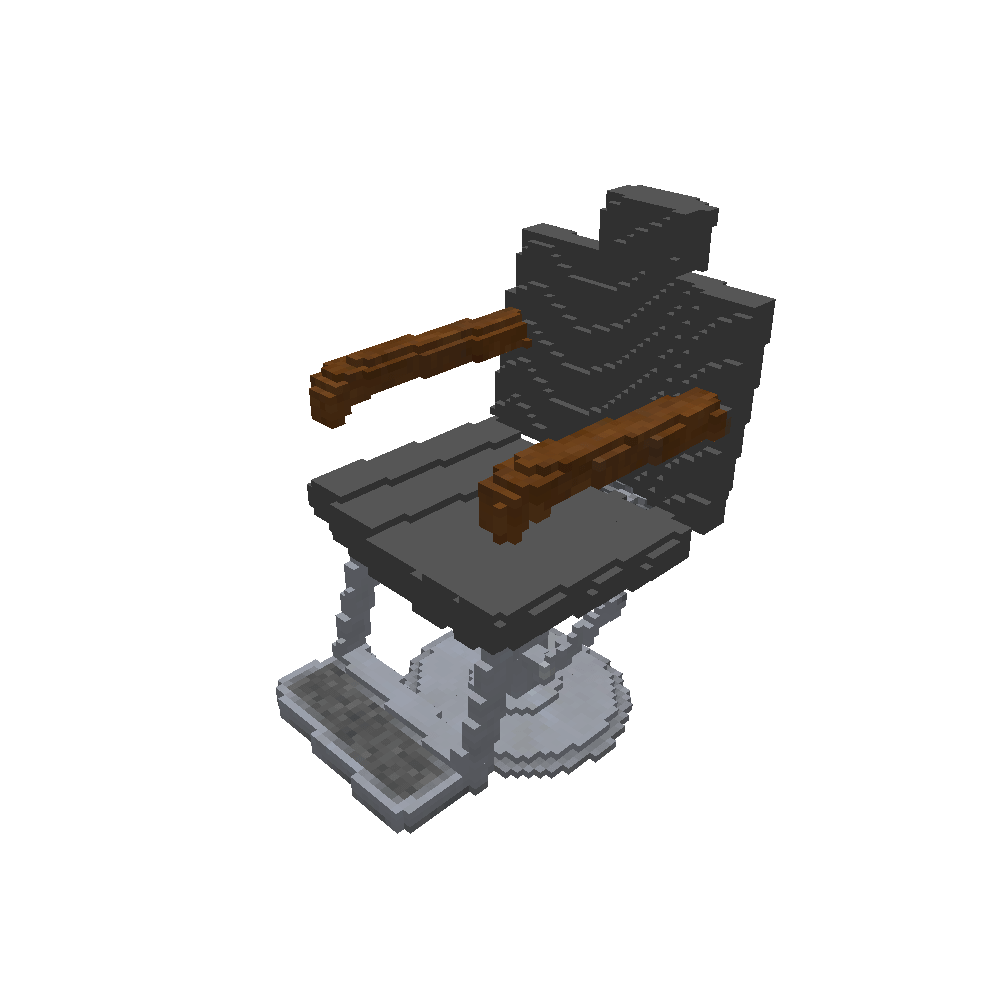}
        \vspace{-3em}
        \caption{\emph{Hairstylist's chair with metal base, foot pump and foot rest.  The seat area looks like it is a hard plastic material and has a head rest...}}
    \end{subfigure}
    \begin{subfigure}[t]{0.24\linewidth}
        \includegraphics[width=\textwidth]{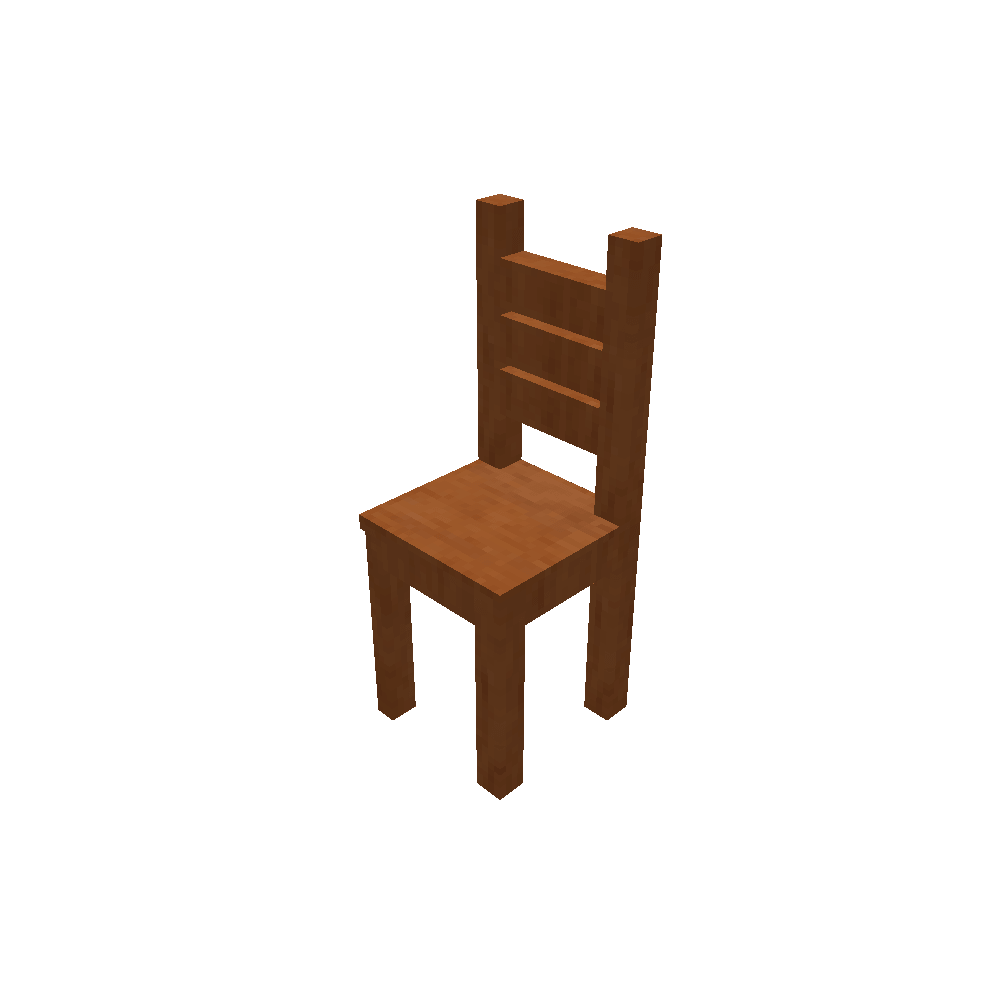}
        \vspace{-3em}
        \caption{\emph{A four legged, wooden high back chair. The backing has two vertical supports at the sides, with three thicker slats...}}
    \end{subfigure}
    \begin{subfigure}[t]{0.24\linewidth}
        \includegraphics[width=\textwidth]{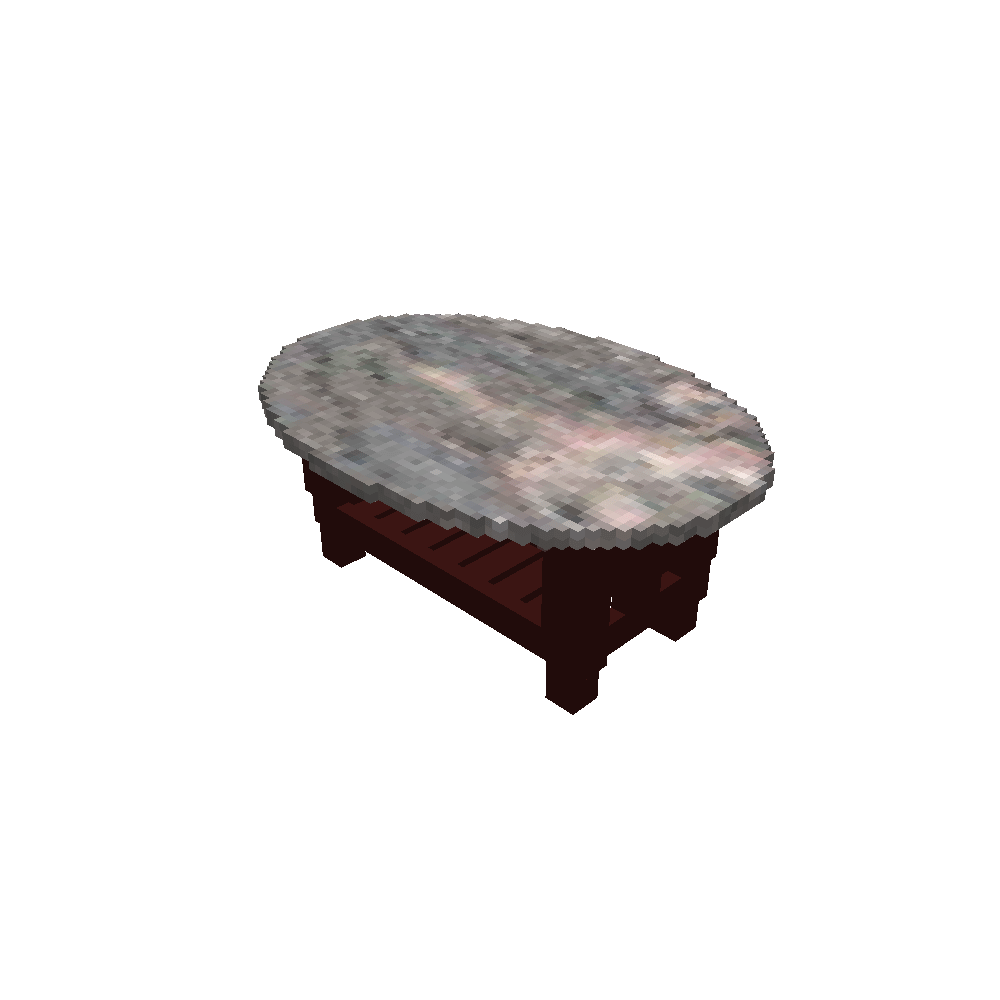}
        \vspace{-3em}
        \caption{\emph{Oval shaped marble-top coffee table with greys and beige colours. Four dark wooden legs with wooden rungs under table for...}}
    \end{subfigure}
    \vspace{-2em}
    \begin{subfigure}[t]{0.24\linewidth}
        \includegraphics[width=\textwidth]{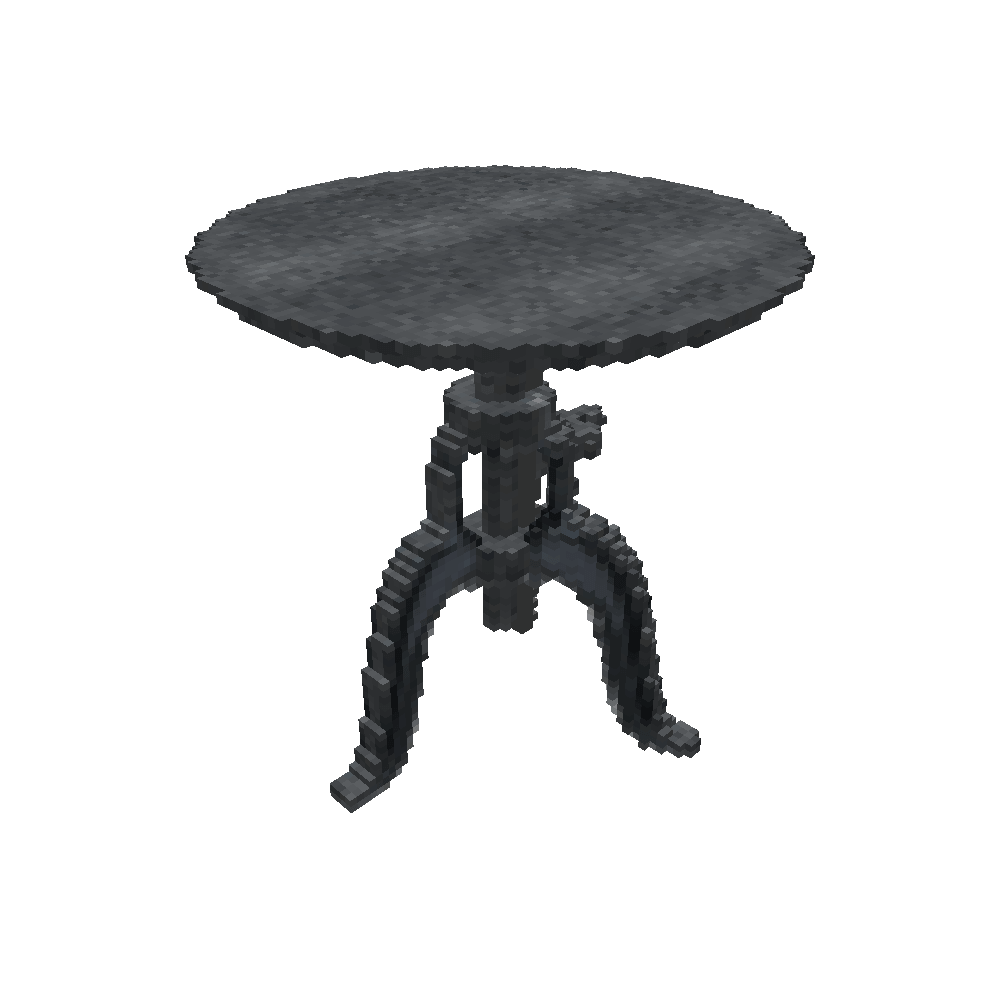}
        \vspace{-2.5em}
        \caption{\emph{A grey colored round wooden table with three curve shaped legs.}}
    \end{subfigure}
    \begin{subfigure}[t]{0.24\linewidth}
        \includegraphics[width=\textwidth]{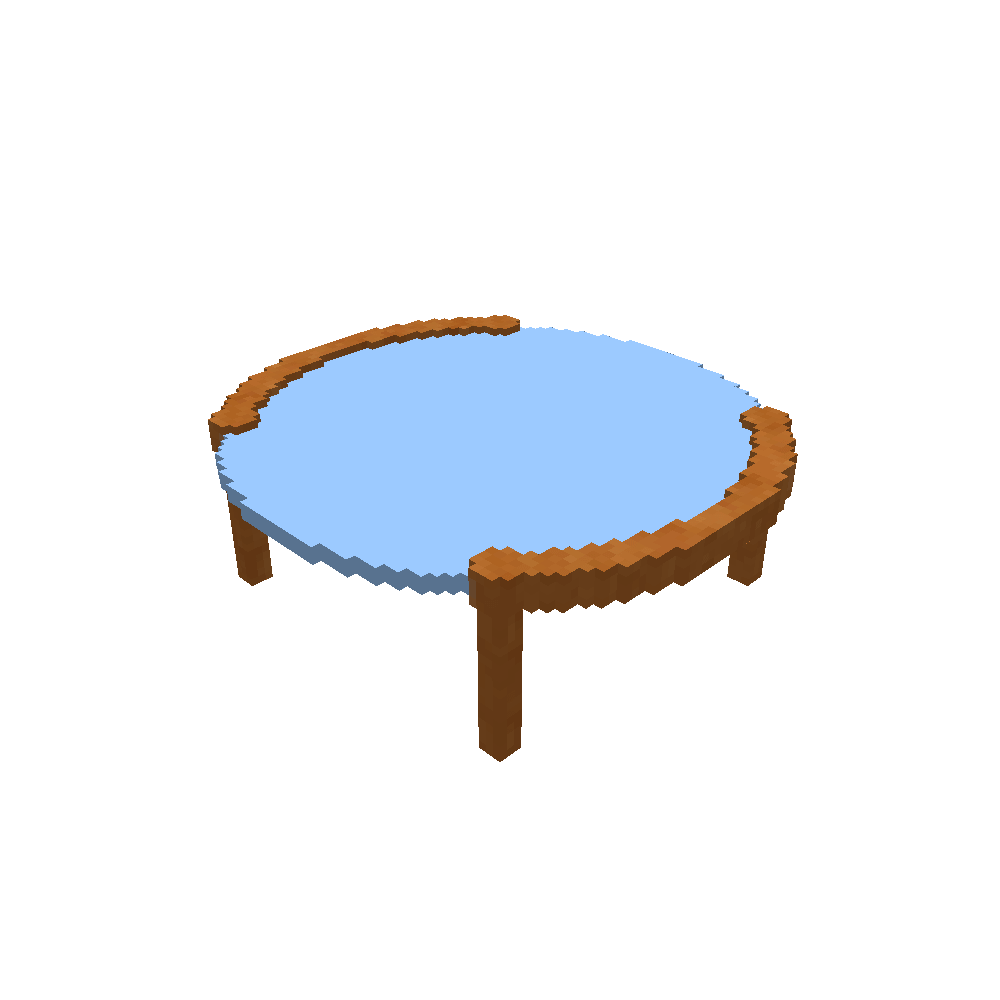}
        \vspace{-2.5em}
        \caption{\emph{Circular glass coffee table with two sets of wooden legs that clasp over the...}}
    \end{subfigure}
    \begin{subfigure}[t]{0.24\linewidth}
        \centering
        \includegraphics[width=0.85\textwidth]{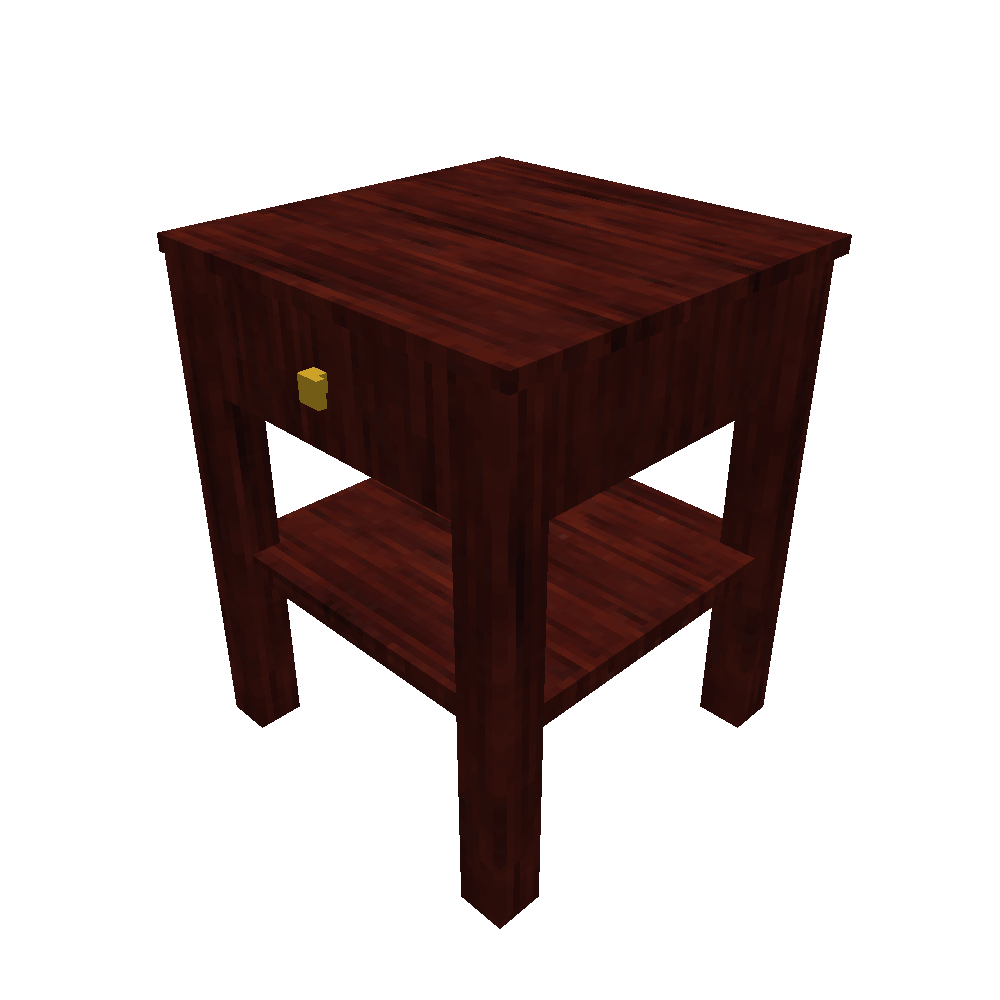}
        \vspace{-1.5em}
        \caption{\emph{A dark brown rectangular table four long legs with gold color knob...}}
    \end{subfigure}
    \begin{subfigure}[t]{0.24\linewidth}
        \includegraphics[width=\textwidth]{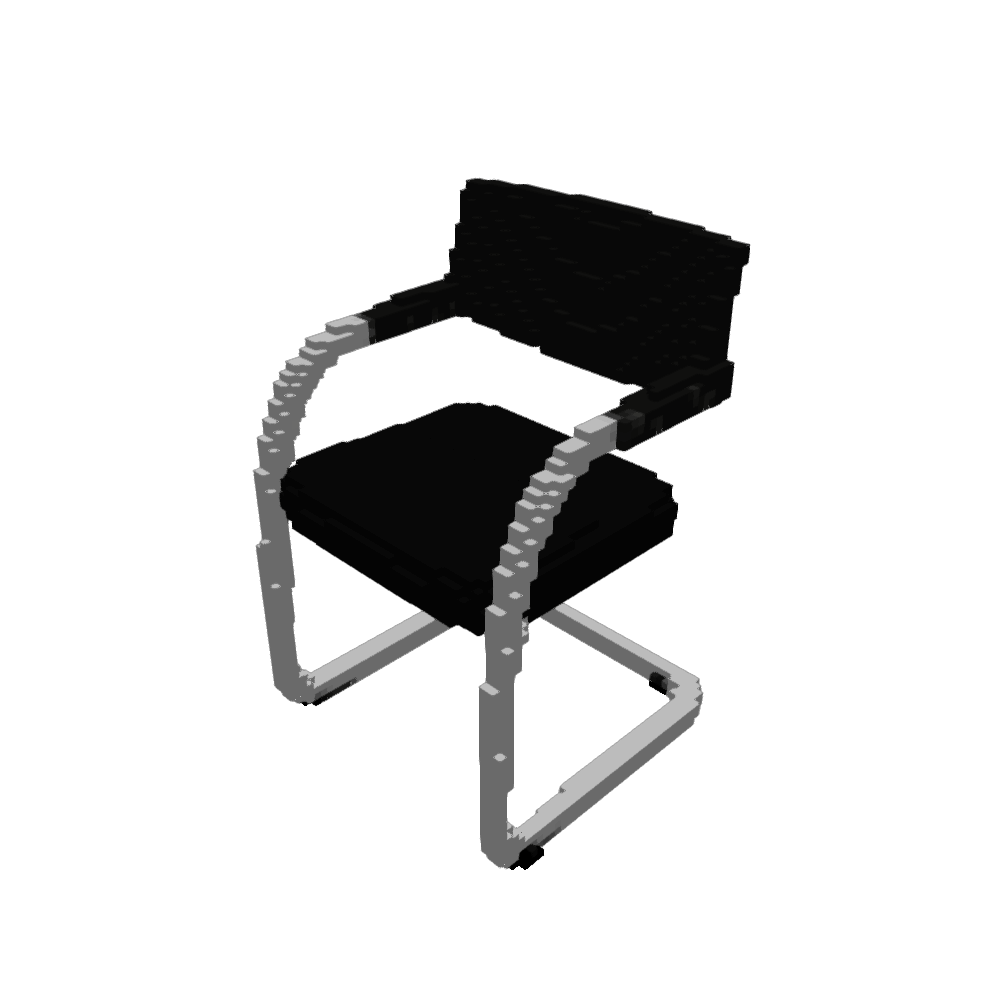}
        \vspace{-2.5em}
        \caption{\emph{A relaxing office chair with single metal finish having black color seat}}
    \end{subfigure}
    \vspace{1em}
    \caption{Example color voxelizations for a variety of chair and table instances from ShapeNetCore.  The voxelizations are computed at a resolution of $256^3$ and downsampled to $64^3$.  Below each shape is one example of a natural language description that was provided by a crowdworker.}
    \label{fig:voxelizations}
\end{figure*}

%% file: tab/primitives-same-modal-retrieval.tex
\begin{table*}
\vspace{-2em}
\begin{center}
\caption{\textbf{Same-modal retrieval results on the primitives dataset.}}
\label{tab:retrieval-same-modal}
\small
\begin{tabular}{lcccccc}
\toprule
 & \multicolumn{3}{c}{Text-to-text} & \multicolumn{3}{c}{Shape-to-shape}\\
\cmidrule(l){2-4} \cmidrule(l){5-7}   
Method & RR@1 & RR@5 & NDCG@5 & RR@1 & RR@5 & NDCG@5 \\
\midrule
Random & 6.31 & 17.94 & 5.27 & 1.33 & 1.6 & 1.35 \\
DS-SJE~\cite{reed2016learning} & 35.41 & 43.38 & 35.05 & 37.87 & 46.00 & 35.48 \\
LBA-TST~\cite{haeusser2017learning} & 23.16 & 24.18 & 23.21 & 13.47 & 29.07 & 12.06 \\
\midrule
ML & 24.01 & 28.31 & 23.73 & 30.93 & 64.27 & 29.60 \\
LBA-MM & 61.84 & 62.71 & 62.03 & 20.80 & 28.00 & 18.94 \\
Full-TST & \textbf{100} & \textbf{100} & \textbf{100} & \textbf{96.53} & 98.80 & 95.30 \\
Full-MM & \textbf{100} & \textbf{100} & \textbf{100} & 96.13 & \textbf{99.20} & \textbf{95.59} \\
\bottomrule
\end{tabular}
\end{center}
\vspace{-2em}
\end{table*}

%% file: tab/primitives-cross-modal-retrieval.tex
\begin{table*}
\vspace{-2em}
\begin{center}
\caption{\textbf{Cross-modal retrieval results on the primitives dataset.}}
\label{tab:retrieval-cross-modal}
\small
\begin{tabular}{lcccccc}
\toprule
 & \multicolumn{3}{c}{Shape-to-text} & \multicolumn{3}{c}{Text-to-shape}\\
\cmidrule(l){2-4} \cmidrule(l){5-7}
Method & RR@1 & RR@5 & NDCG@5 & RR@1 & RR@5 & NDCG@5 \\
\midrule
Random & 2.8 & 6.53 & 1.84 & 0.24 & 0.76 & 0.27 \\
DS-SJE~\cite{reed2016learning} & 80.5 & 85.87 & 80.36 & 81.77 & 90.70 & 81.29 \\
LBA-TST~\cite{haeusser2017learning} & 5.20 & 6.13 & 5.25 & 5.06 & 15.29 & 5.92 \\
\midrule
ML & 24.67 & 29.87 & 24.38 & 25.93 & 57.24 & 25.00 \\
LBA-MM & 89.2 & 90.53 & 89.48 & 91.13 & 98.27 & 91.90 \\
Full-TST & 92.00 & 92.40 & 91.98 & 94.24 & 97.55 & 95.20 \\
Full-MM & \textbf{93.47} & \textbf{93.47} & \textbf{93.47} & \textbf{95.07} & \textbf{99.08} & \textbf{95.51} \\
\bottomrule
\end{tabular}
\end{center}
\vspace{-2em}
\end{table*}

%% file: tab/shapenet-retrieval.tex
\begin{table*}
\vspace{-2em}
\small\centering
\caption{\textbf{Retrieval results on the ShapeNet dataset.}}
\begin{tabular}{lccccccccc}
\toprule
 & \multicolumn{3}{c}{Text-to-text} & \multicolumn{3}{c}{Shape-to-text} & \multicolumn{3}{c}{Text-to-shape}\\
\cmidrule(l){2-4} \cmidrule(l){5-7} \cmidrule(l){8-10}   
Method & RR@1 & RR@5 & NDCG@5 & RR@1 & RR@5 & NDCG@5 & RR@1 & RR@5 & NDCG@5 \\
\midrule
Random & 0.08 & 0.38 & 0.08 & 0.07 & 0.34 & 0.06 & 0.11 & 0.35 & 0.23 \\
DS-SJE~\cite{reed2016learning} & 0.08 & 0.52 & 0.11 & 0.13 & 0.60 & 0.13 & 0.12 & 0.65 & 0.38 \\
LBA-TST~\cite{haeusser2017learning} & 0.82 & 3.75 & 0.82 & 0.20 & 0.80 & 0.12 & 0.04 & 0.20 & 0.12 \\
\midrule
ML & 0.11 & 0.48 & 0.10 & 0.13 & 0.47 & 0.11 & 0.13 & 0.61 & 0.36 \\
LBA-MM & 0.86 & 3.35 & 0.73 & 0.067 & 0.34 & 0.07 & 0.08 & 0.34 & 0.21 \\
Full-TST & 0.74 & 3.22 & 0.70 & \textbf{0.94} & \textbf{3.69} & \textbf{0.85} & 0.22 & 1.63 & 0.87 \\
Full-MM & \textbf{1.74} & \textbf{6.05} & \textbf{1.43} & 0.83 & 3.37 & 0.73 & \textbf{0.40} & \textbf{2.37} & \textbf{1.35} \\
\bottomrule
\end{tabular}
\label{tab:shapenet-retrieval}
\vspace{-2em}
\end{table*}


%% file: fig/text-text-retrieval.tex
\begin{figure}
    \includegraphics[width=\linewidth]{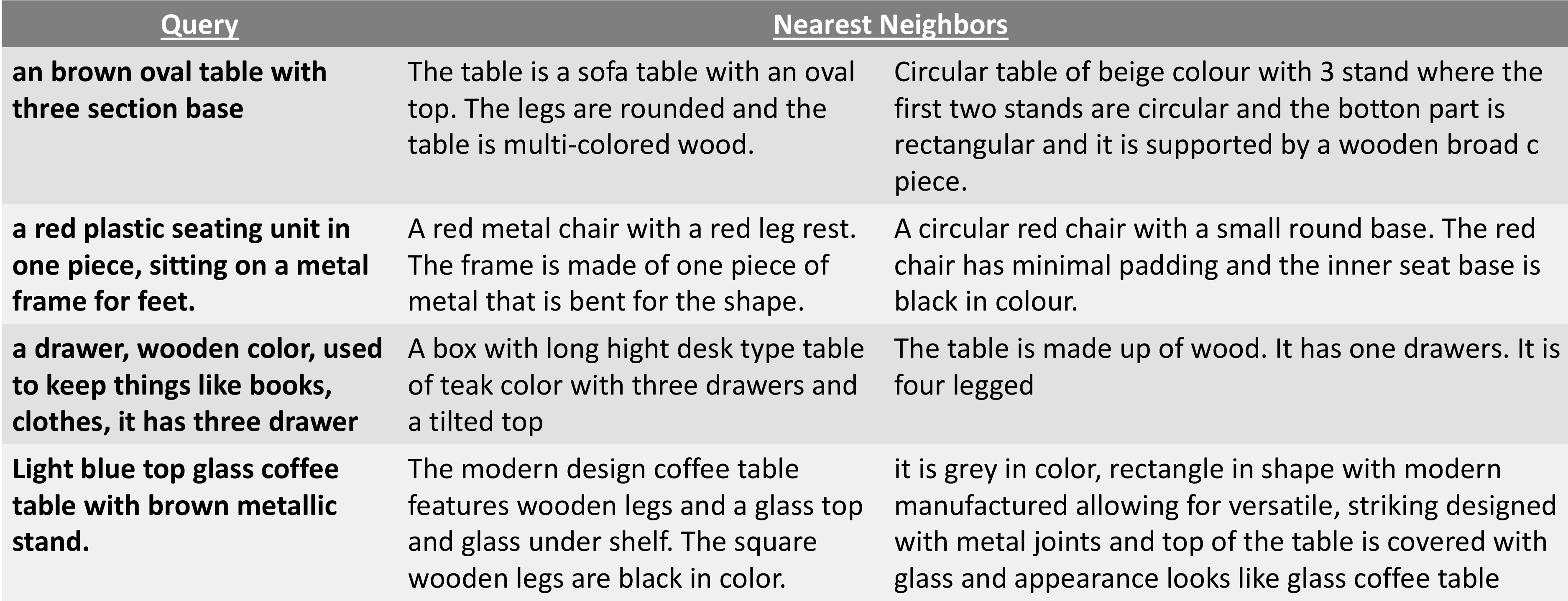}
    \vspace{-1em}
    \caption{\textbf{Text-to-text retrieval.} The retrieved descriptions are semantically similar to the query despite originating from different shape instances. 
    }
    \label{fig:text-text-retrieval}
\end{figure}

%% file: fig/shape-shape-retrieval.tex
\begin{figure}
    \centering
    \includegraphics[width=0.9\linewidth]{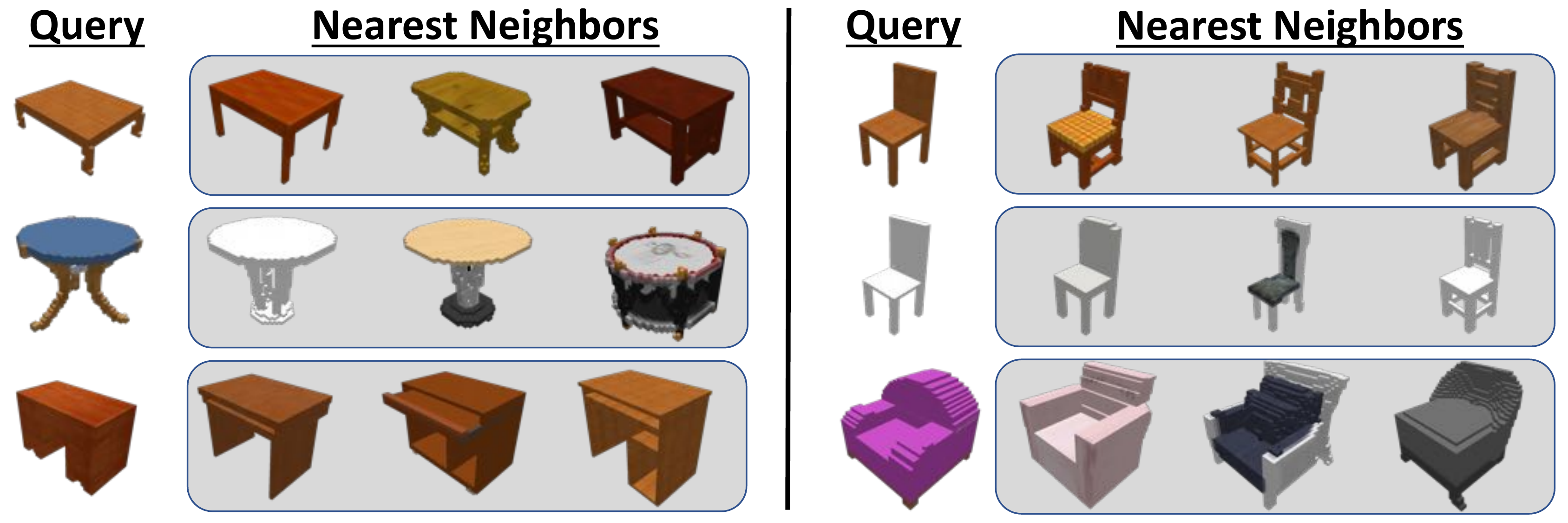}
    \vspace{-1em}
    \caption{\textbf{Shape-to-shape retrieval examples.} The nearest neighbor shapes for each query have closely matching attributes.}
    \label{fig:shape-shape-retrieval}
    \vspace{-1em}
\end{figure}

%% file: fig/shape-shape-retrieval-comparison.tex
\begin{figure}
    \centering
    \includegraphics[width=\linewidth]{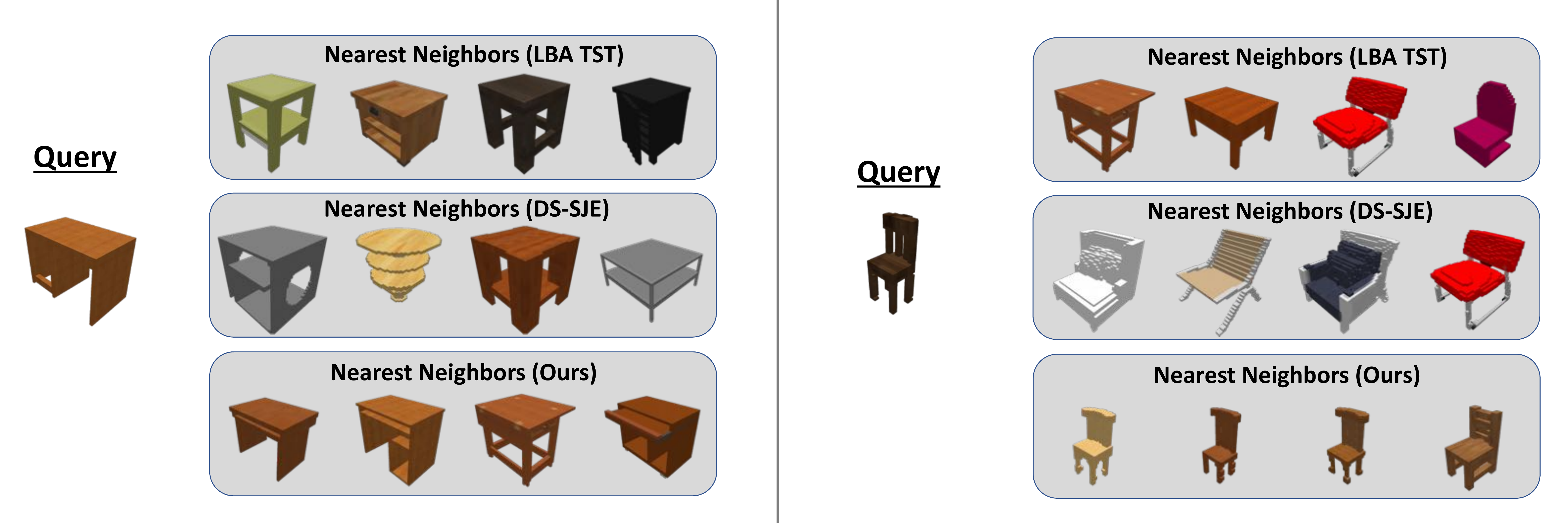}
    \vspace{-1em}
    \caption{\textbf{Shape-to-shape retrieval comparison.} We compare with the LBA (TST) and DS-SJE~\cite{reed2016learning} baselines. Ours retrieves the most reasonable nearest neighbors.}
    \label{fig:shape-shape-retrieval-comparison}
    \vspace{-1em}
\end{figure}

%% file: fig/cross-modal-retrieval-comparison.tex
\begin{figure}
    \centering
    \includegraphics[width=0.7\linewidth]{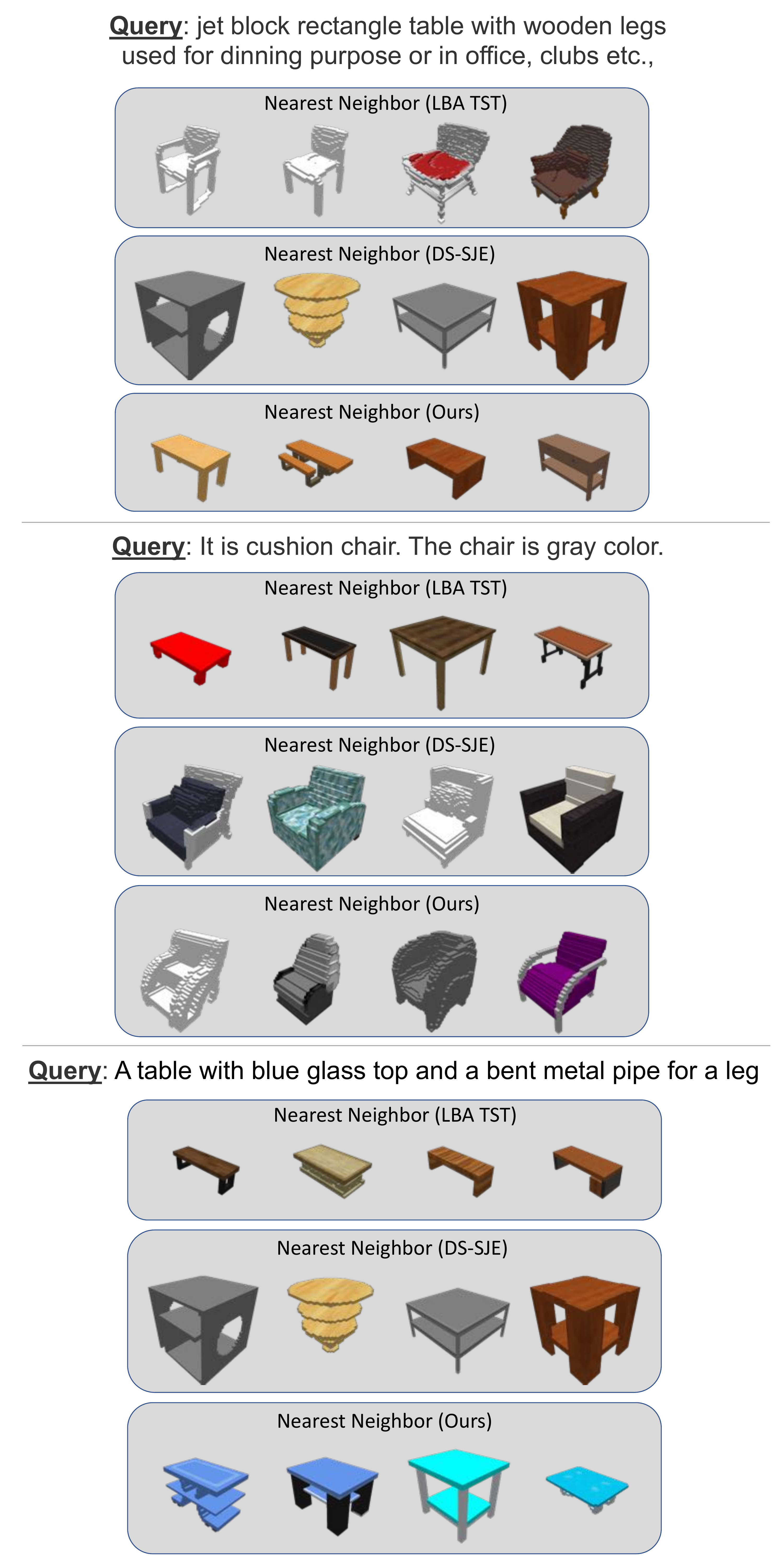}
    \caption{\textbf{Text-to-shape retrieval.} Each row shows the five nearest neighbors to the text in our learned embedding, which match in category, color, and shape. We compare with the LBA (TST) and DS-SJE \cite{reed2016learning} baselines. Ours is the only method to return reasonable text-to-shape retrieval results.}
    \label{fig:cross-modal-retrieval-comparison}
\end{figure}

%% file: fig/text2shape-results-supplementary.tex
\begin{figure}
    \vspace{-2em}
    \centering\scriptsize
    \begin{tabular}{m{6cm}M{1.3cm}M{1.3cm}M{1.3cm}M{1.3cm}}
    Input Text & GAN-INT-CLS~\cite{reed2016generative} & Ours CGAN & Ours CWGAN & GT
    \\\toprule
    \textbf{1)} modern dining table chair with nice and sturdy for wooden legs. looks comfortable &
    \adjincludegraphics[width=\linewidth,trim={{.05\width} {.05\height} {.05\width} {.05\height}},clip]{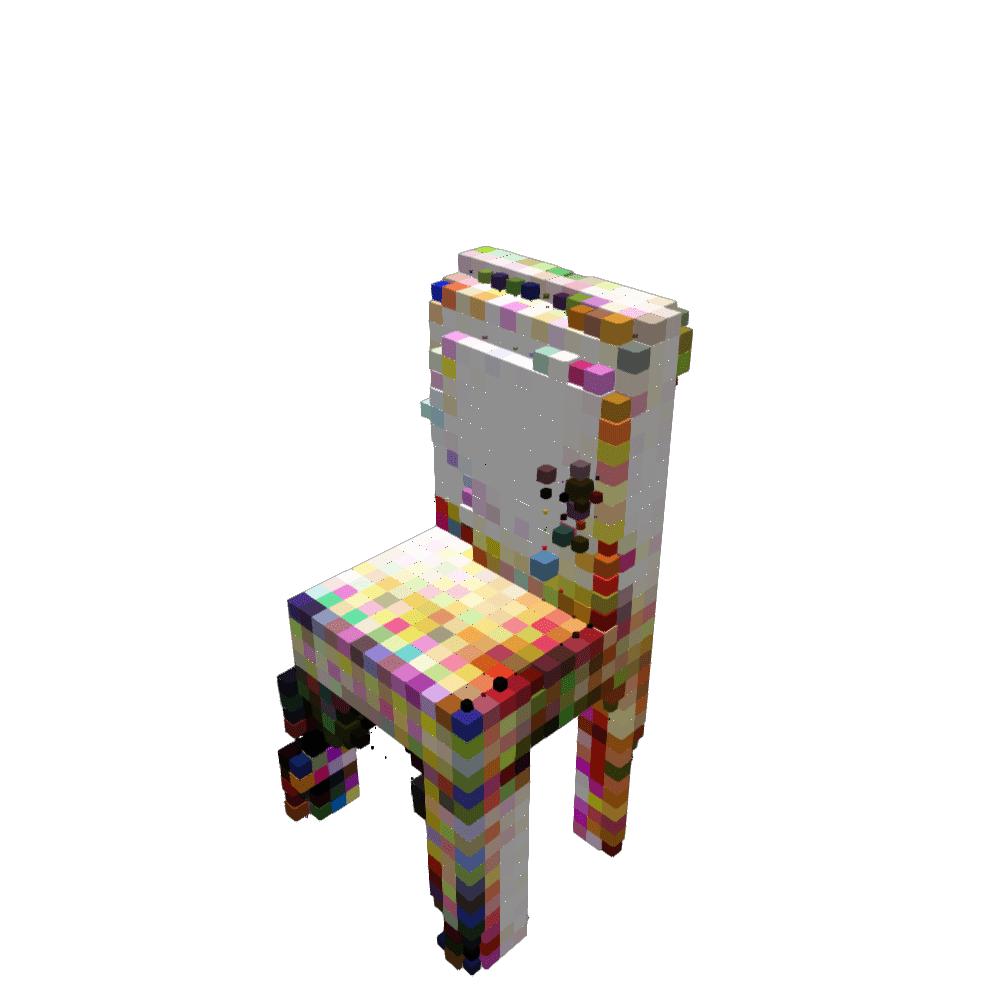} &
    \adjincludegraphics[width=\linewidth,trim={{.05\width} {.05\height} {.05\width} {.05\height}},clip]{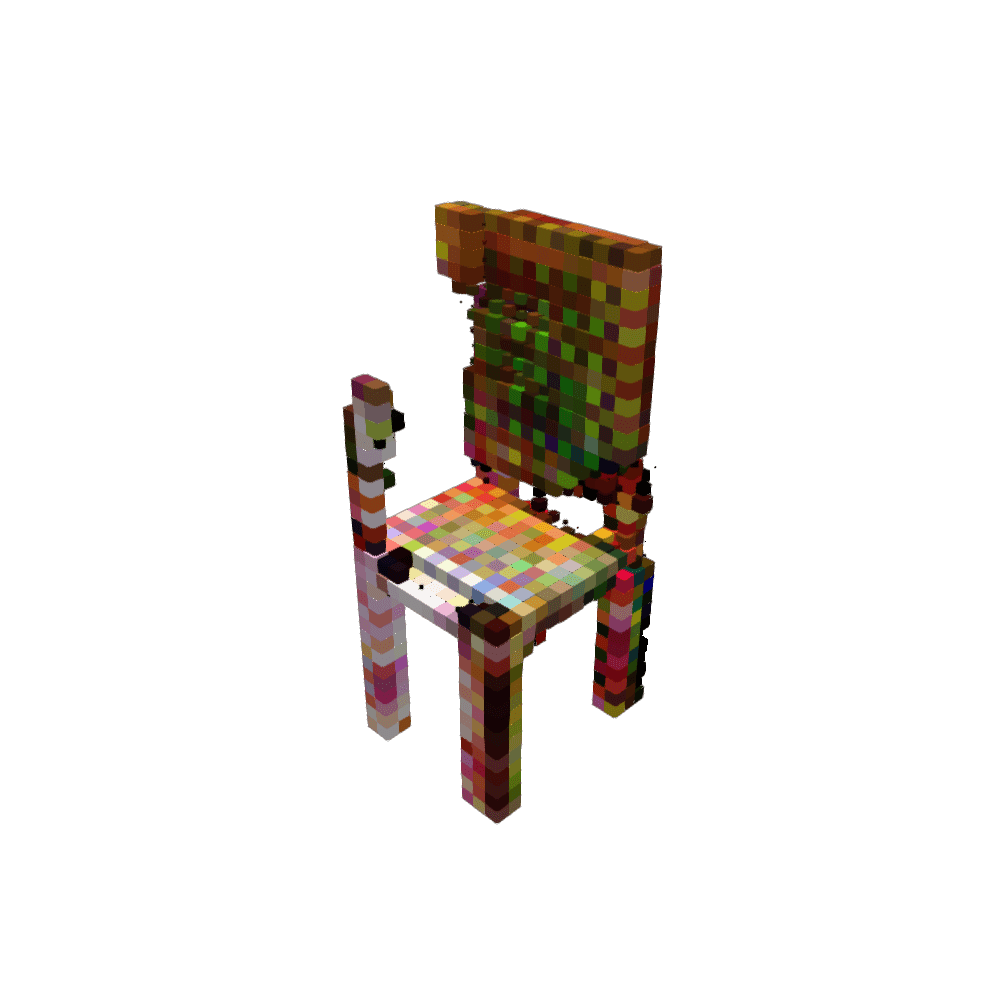} &
    \adjincludegraphics[width=\linewidth,trim={{.05\width} {.05\height} {.05\width} {.05\height}},clip]{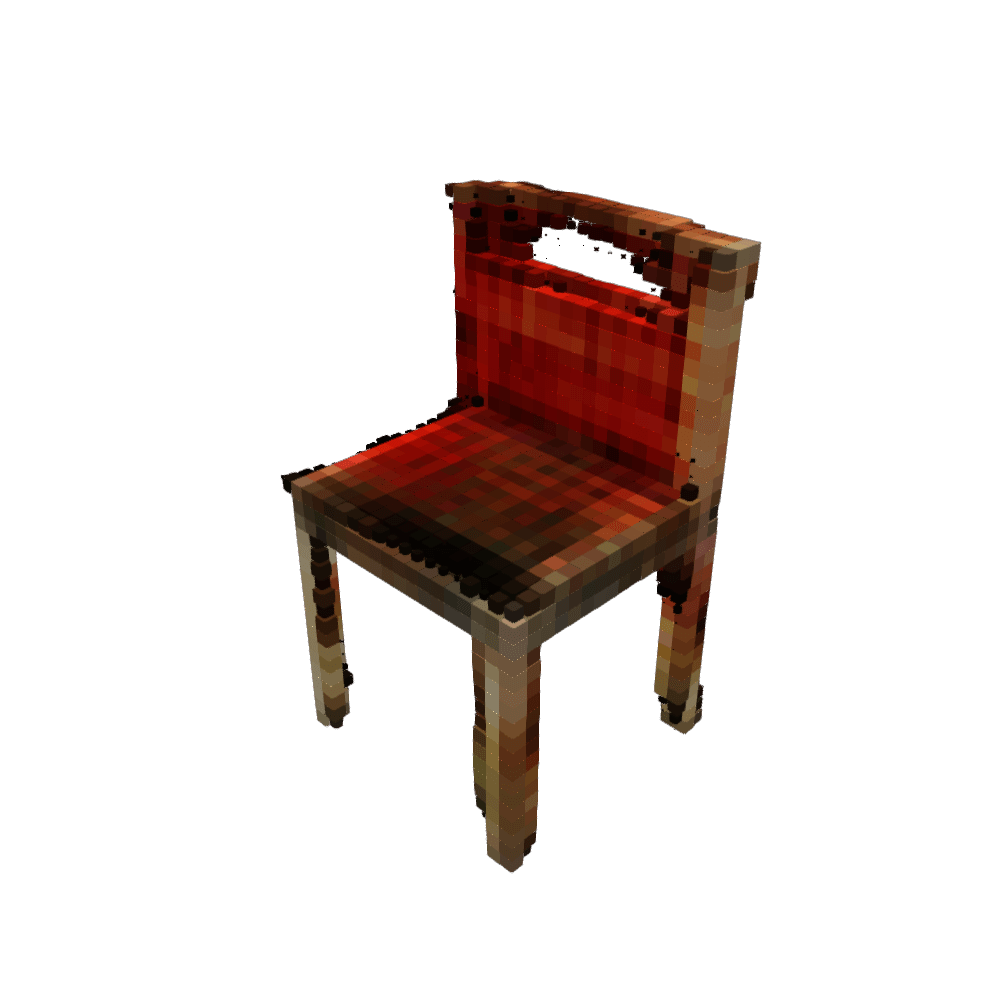} &
    \adjincludegraphics[width=\linewidth,trim={{.05\width} {.05\height} {.05\width} {.05\height}},clip]{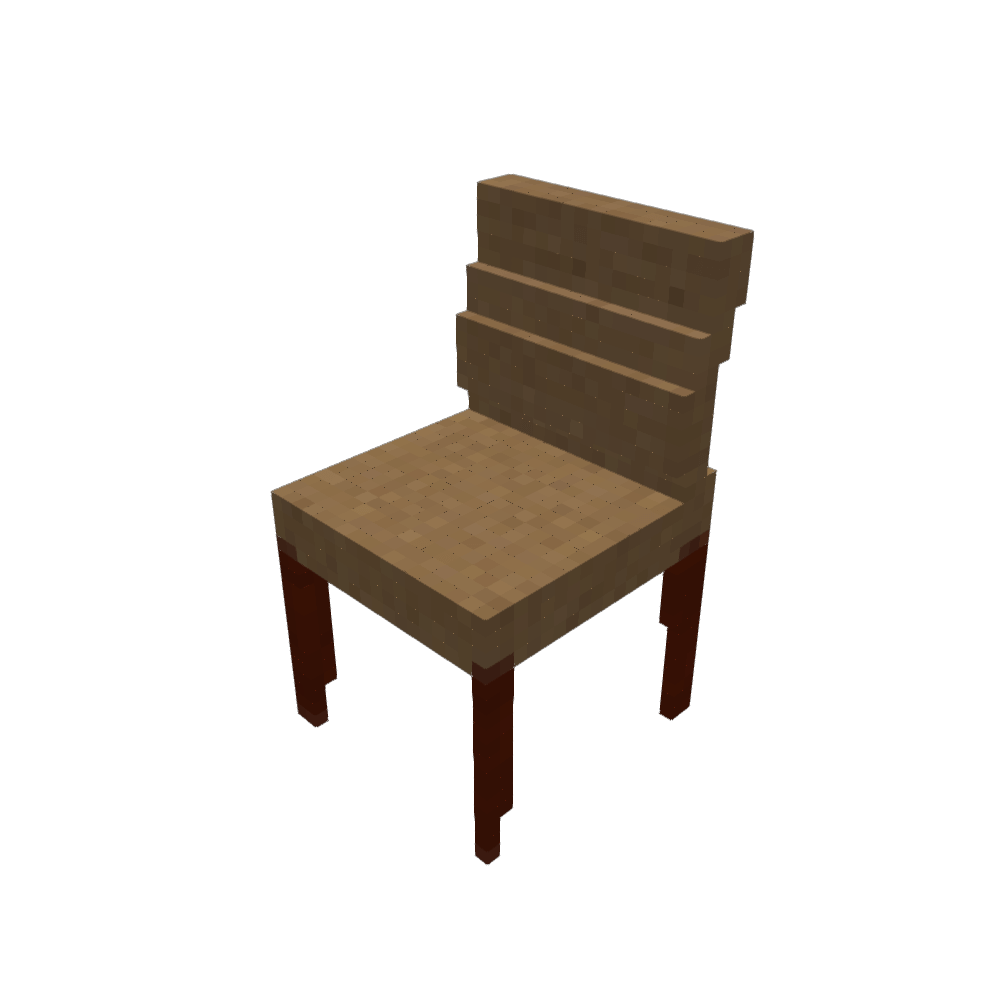}
    \\\midrule
    \textbf{2)} Chair with wave back. Seat is light grey and legs and armrests are curved wood. &
    \adjincludegraphics[width=\linewidth,trim={{.05\width} {.05\height} {.05\width} {.05\height}},clip]{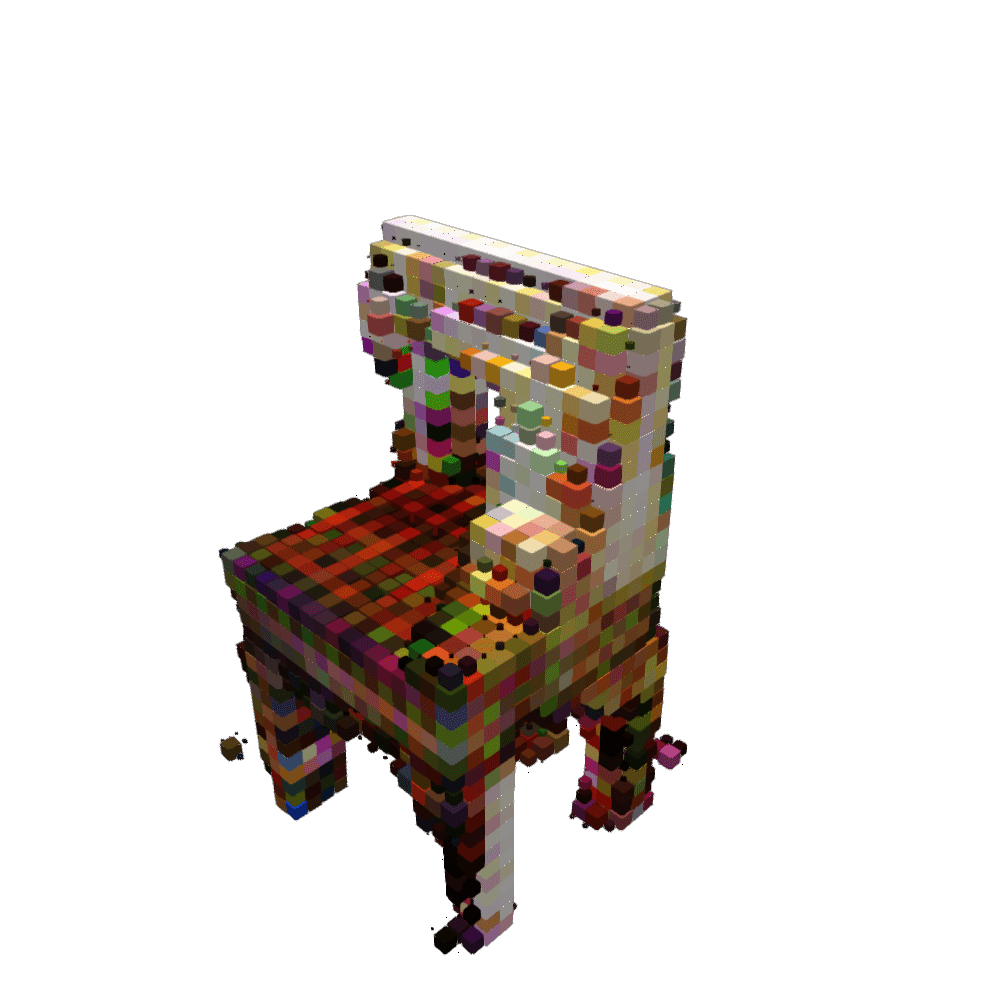} &
    \adjincludegraphics[width=\linewidth,trim={{.05\width} {.05\height} {.05\width} {.05\height}},clip]{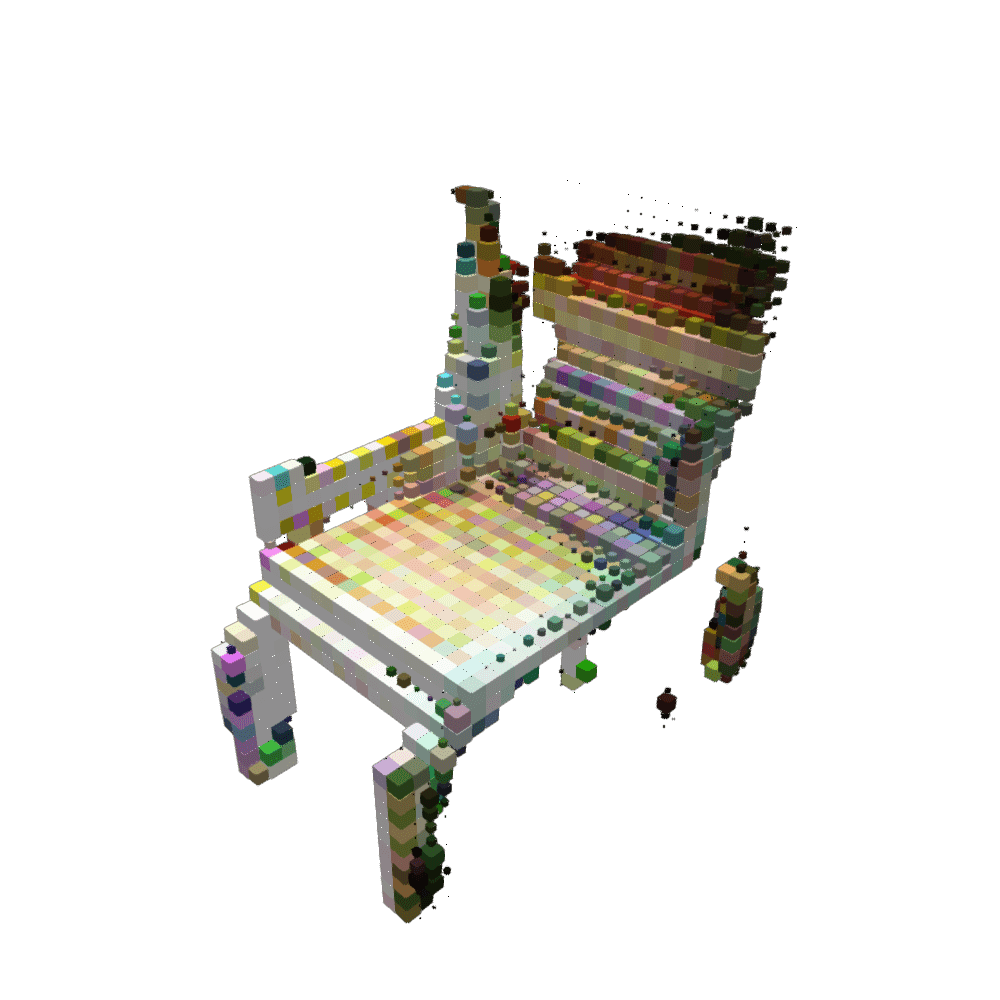} &
    \adjincludegraphics[width=\linewidth,trim={{.05\width} {.05\height} {.05\width} {.05\height}},clip]{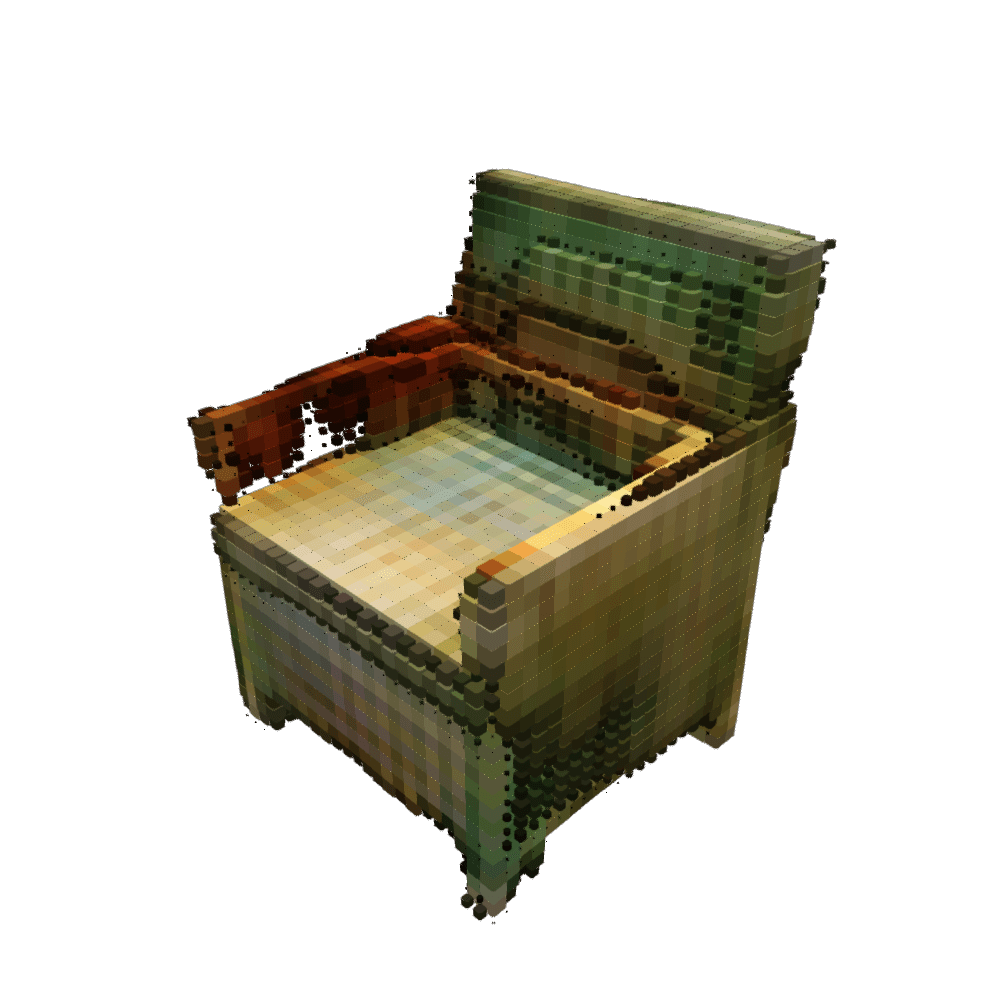} &
    \adjincludegraphics[width=\linewidth,trim={{.05\width} {.05\height} {.05\width} {.05\height}},clip]{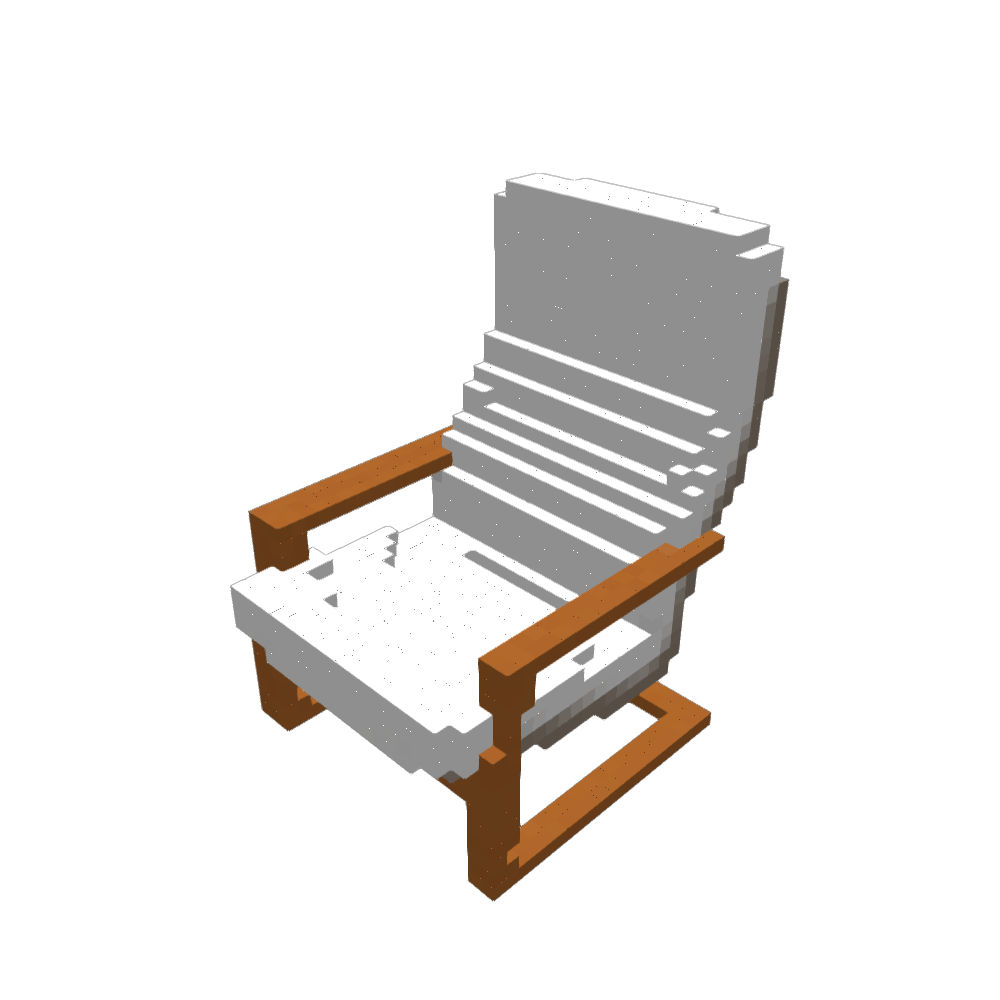}
    \\\midrule
    \textbf{3)} brown wooden chair with fabric on the seat and the backrest &
    \adjincludegraphics[width=\linewidth,trim={{.05\width} {.05\height} {.05\width} {.05\height}},clip]{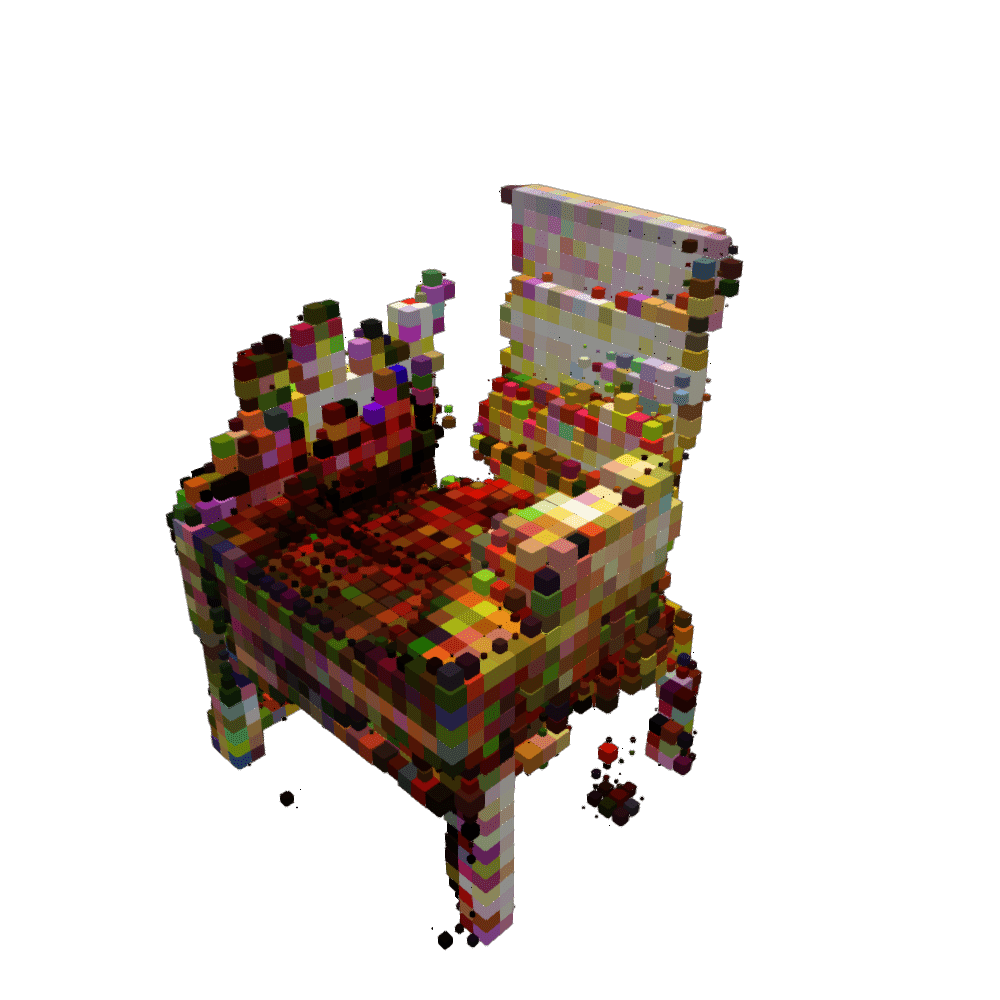} &
    \adjincludegraphics[width=\linewidth,trim={{.05\width} {.05\height} {.05\width} {.05\height}},clip]{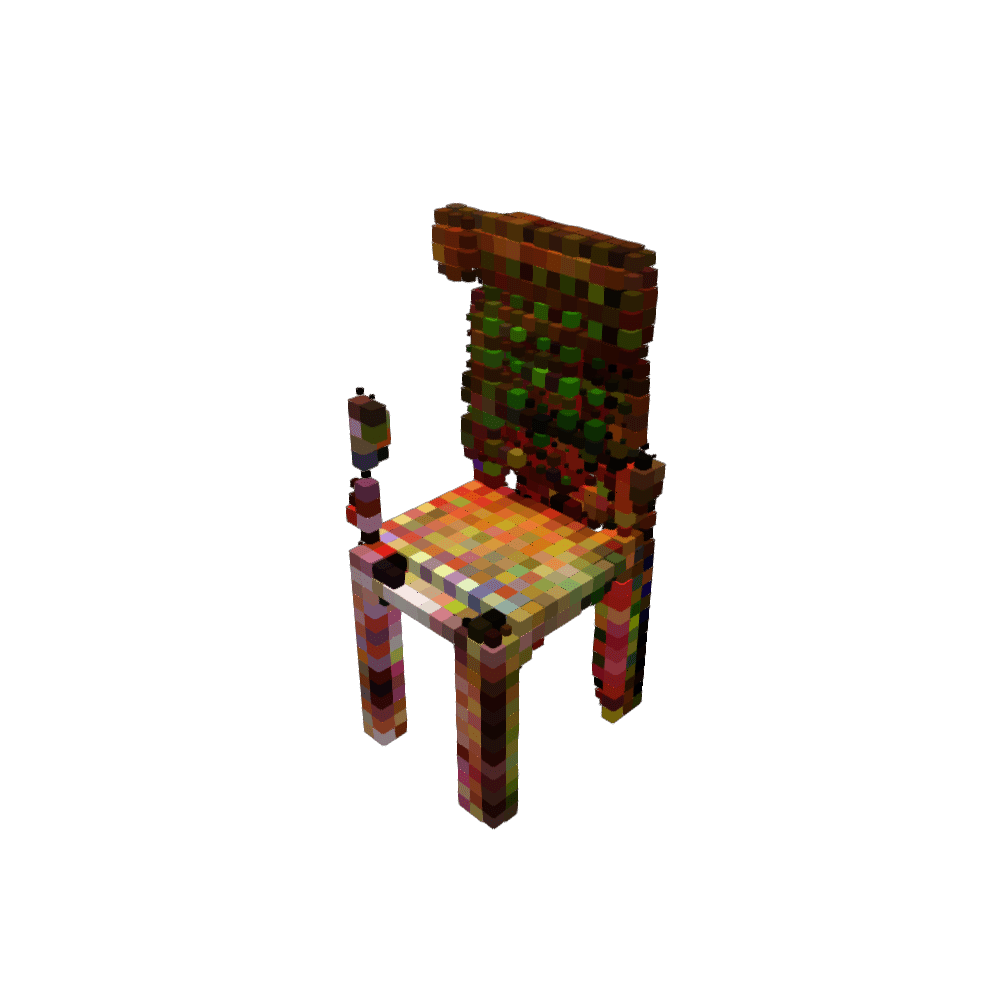} &
    \adjincludegraphics[width=\linewidth,trim={{.05\width} {.05\height} {.05\width} {.05\height}},clip]{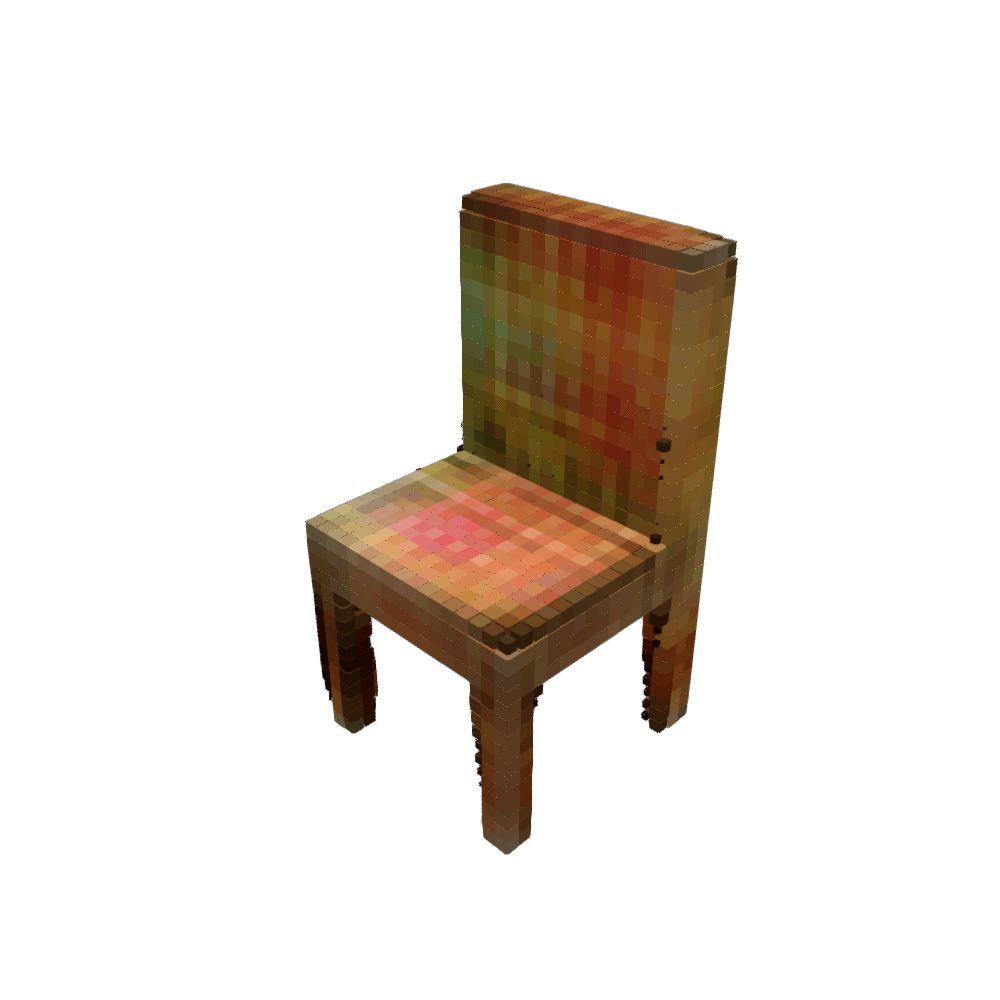} &
    \adjincludegraphics[width=\linewidth,trim={{.05\width} {.05\height} {.05\width} {.05\height}},clip]{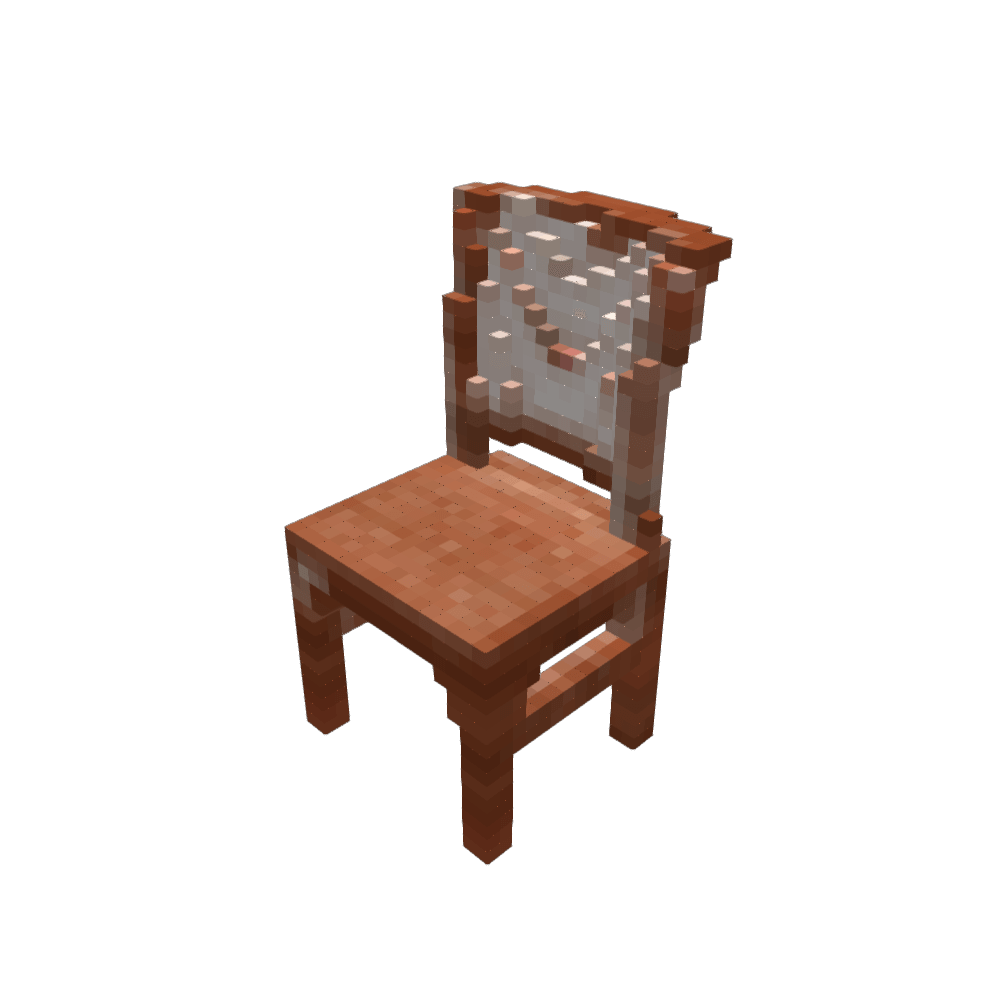}
    \\\midrule
    \textbf{4)} A desk, made out of what appears to be wood, with a white top &
    \adjincludegraphics[width=\linewidth,trim={{.05\width} {.05\height} {.05\width} {.05\height}},clip]{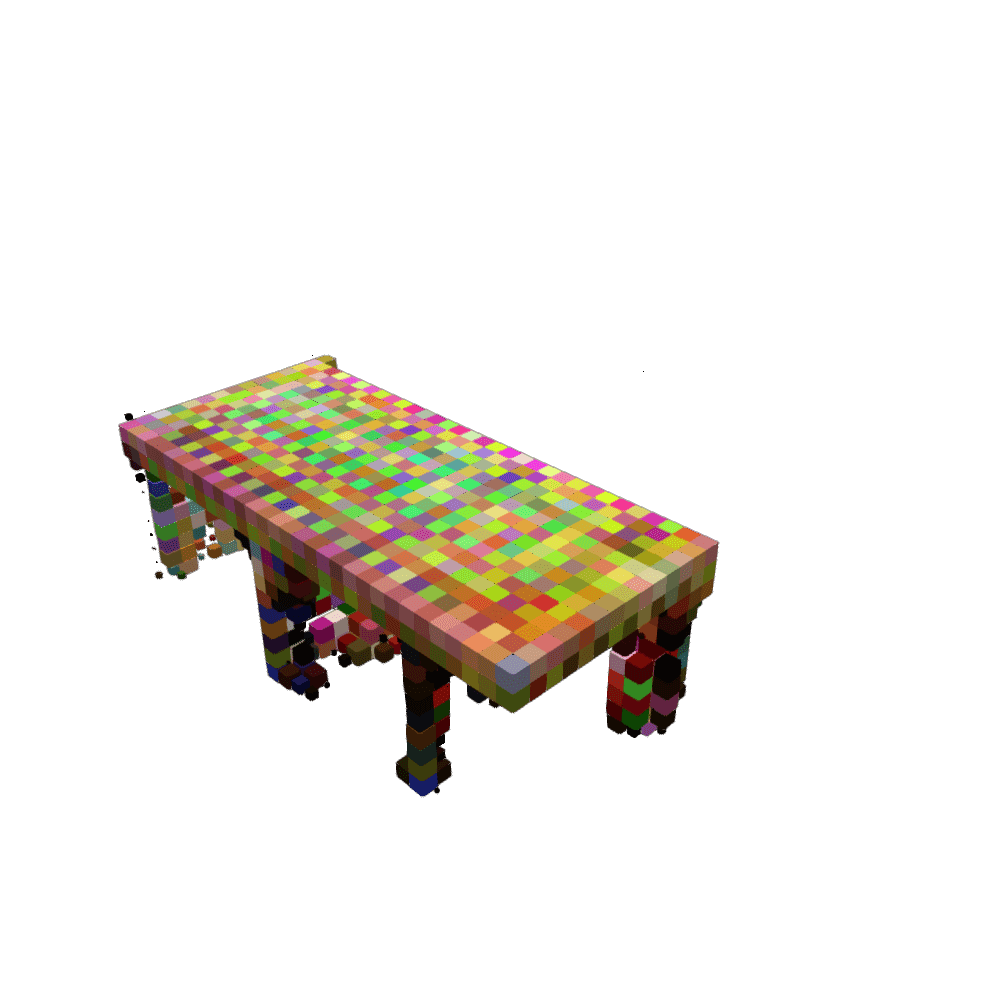} &
    \adjincludegraphics[width=\linewidth,trim={{.05\width} {.05\height} {.05\width} {.05\height}},clip]{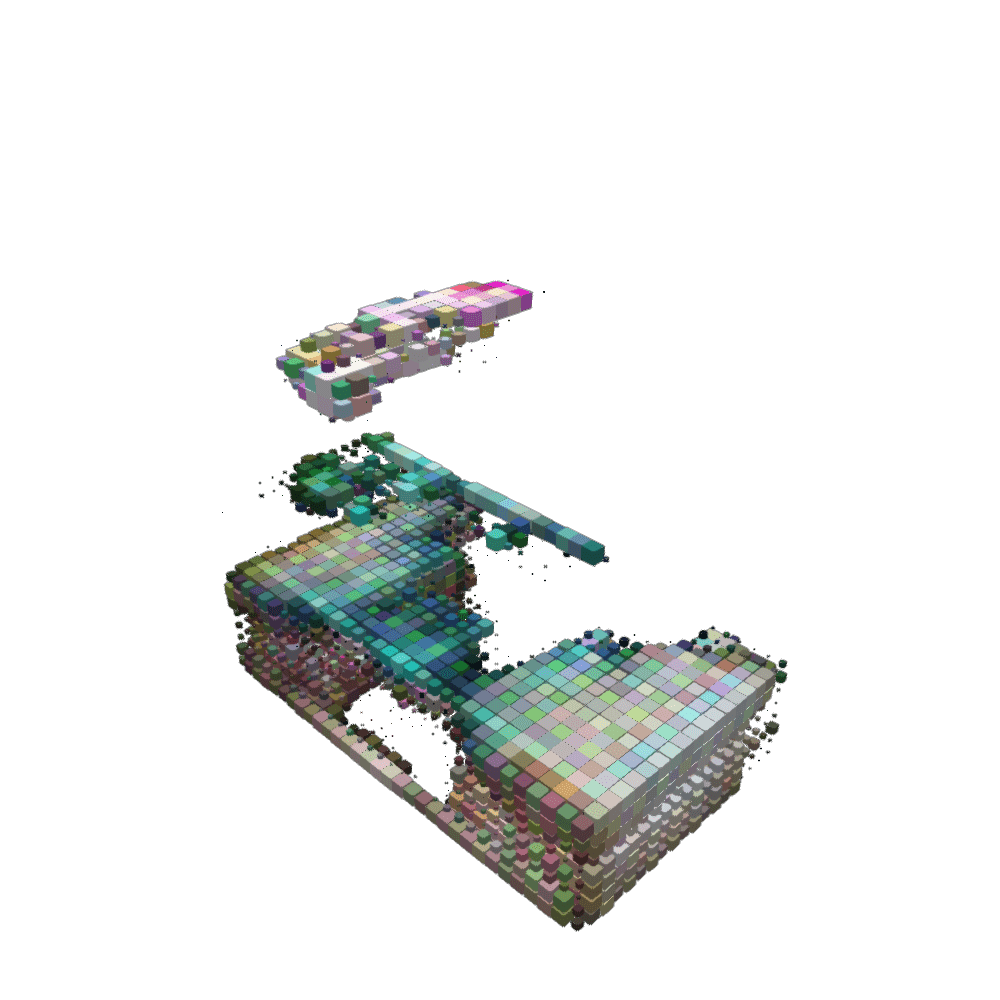} &
    \adjincludegraphics[width=\linewidth,trim={{.05\width} {.05\height} {.05\width} {.05\height}},clip]{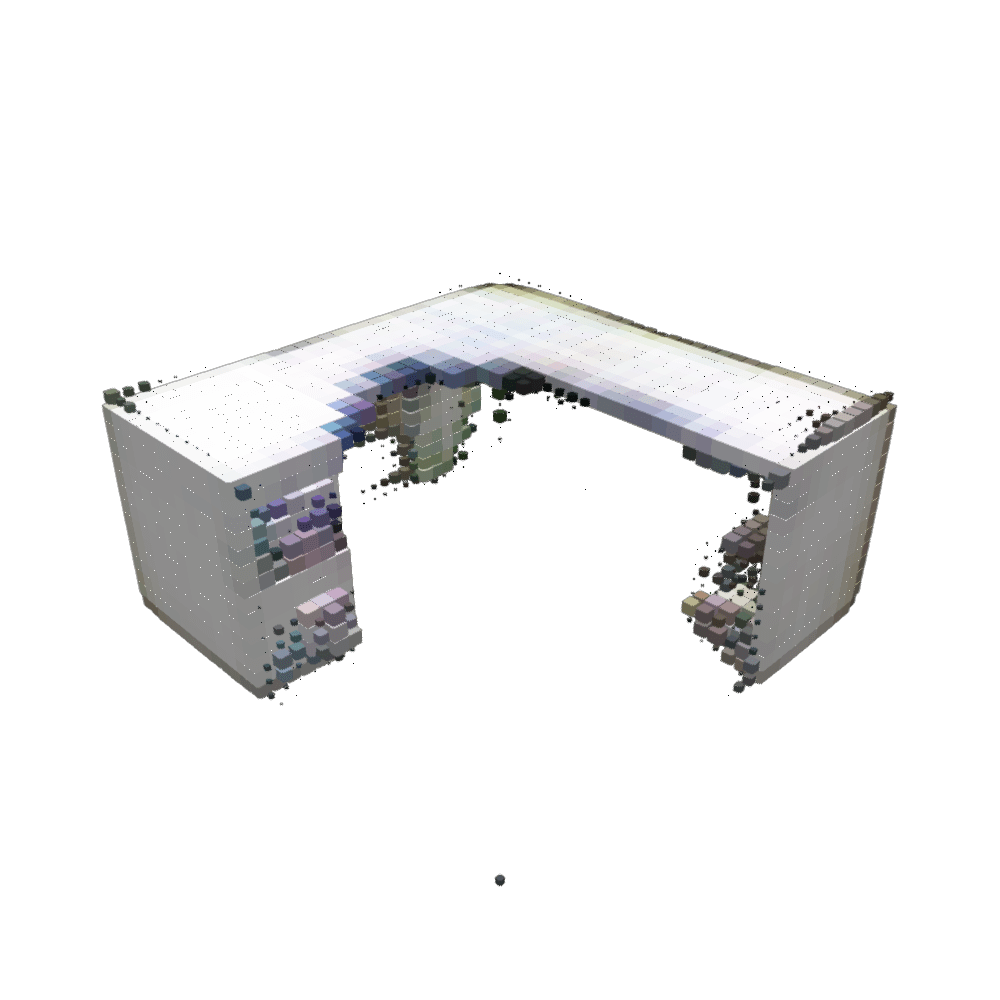} &
    \adjincludegraphics[width=\linewidth,trim={{.05\width} {.05\height} {.05\width} {.05\height}},clip]{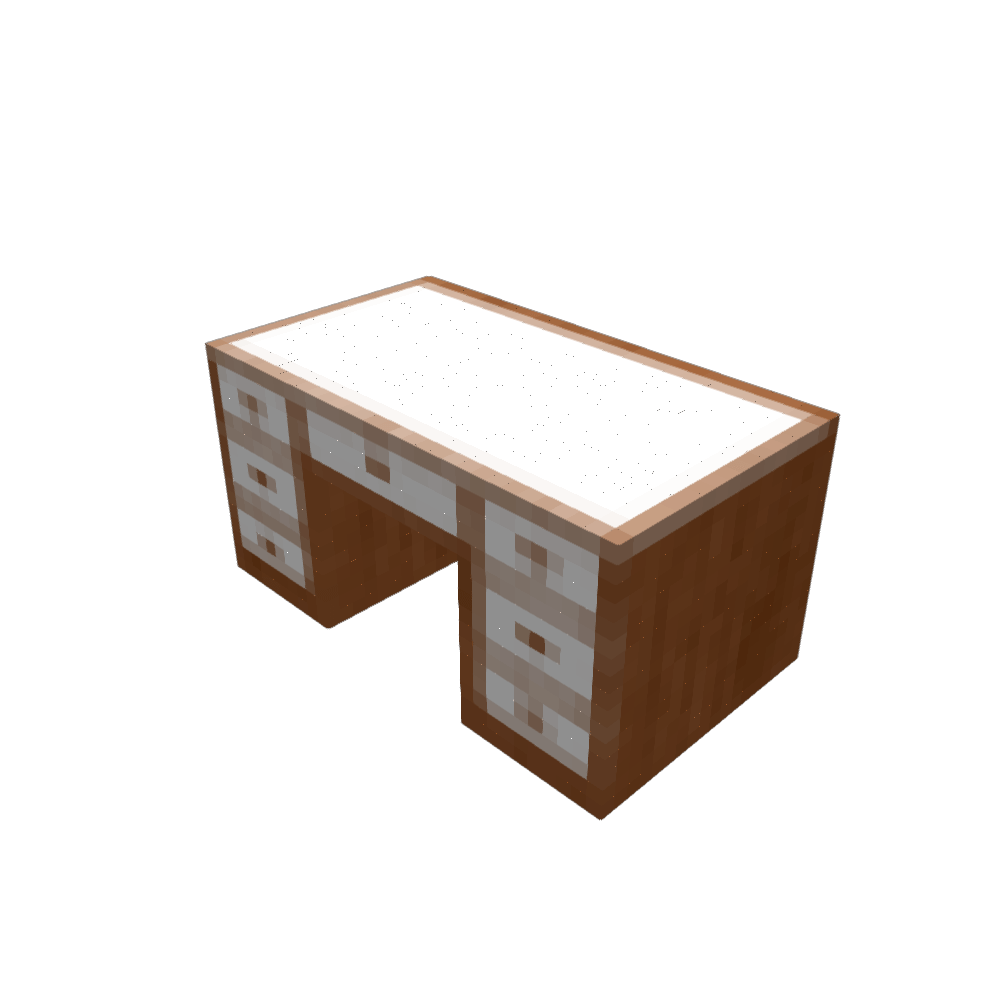}
    \\\midrule
    \textbf{5)} Brown color, Rectangular shape, made in wood, four legs slightly shorter. &
    \adjincludegraphics[width=\linewidth,trim={{.05\width} {.05\height} {.05\width} {.05\height}},clip]{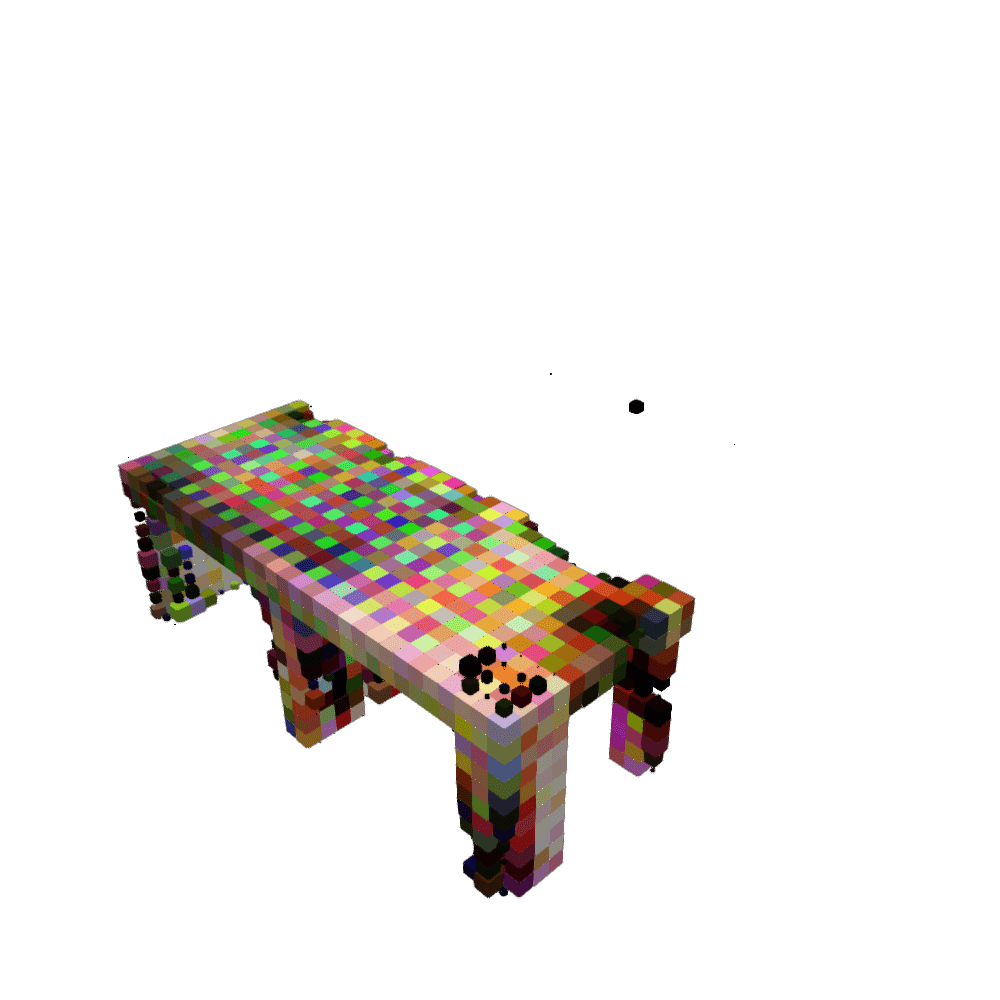} &
    \adjincludegraphics[width=\linewidth,trim={{.05\width} {.05\height} {.05\width} {.05\height}},clip]{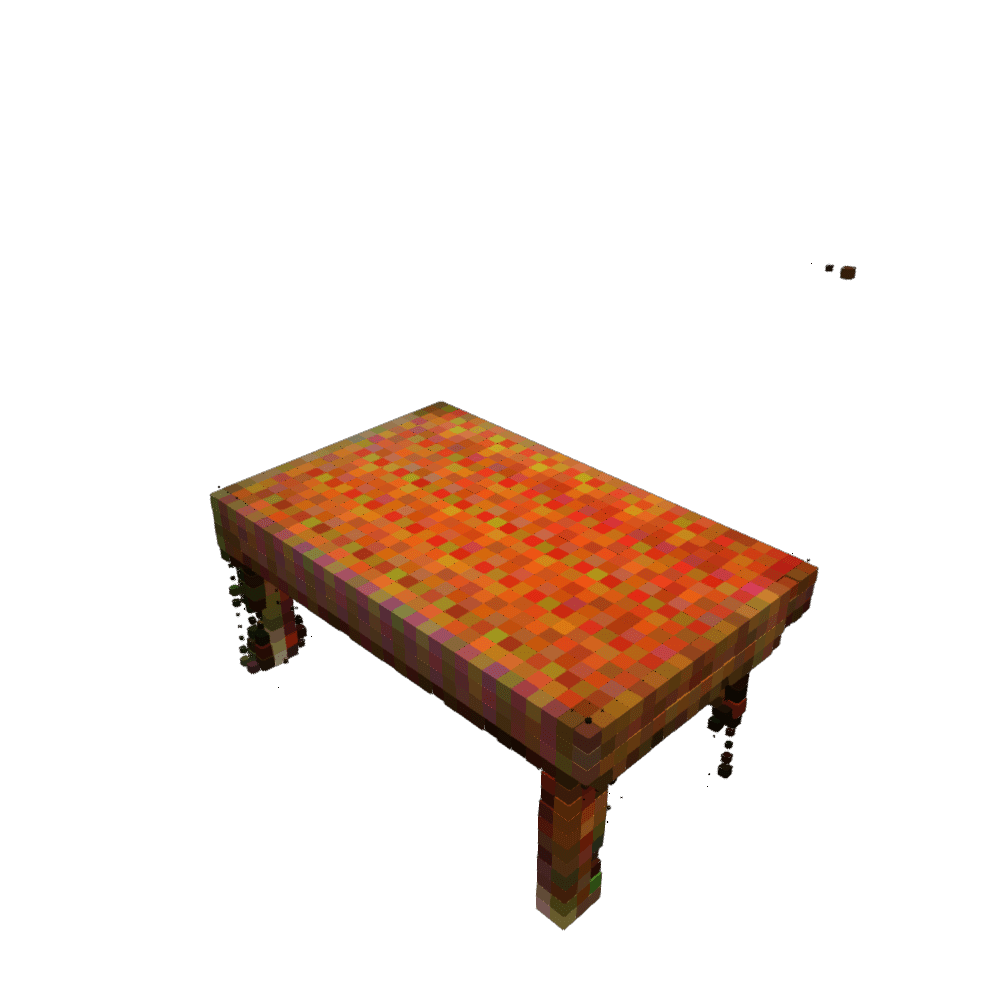} &
    \adjincludegraphics[width=\linewidth,trim={{.05\width} {.05\height} {.05\width} {.05\height}},clip]{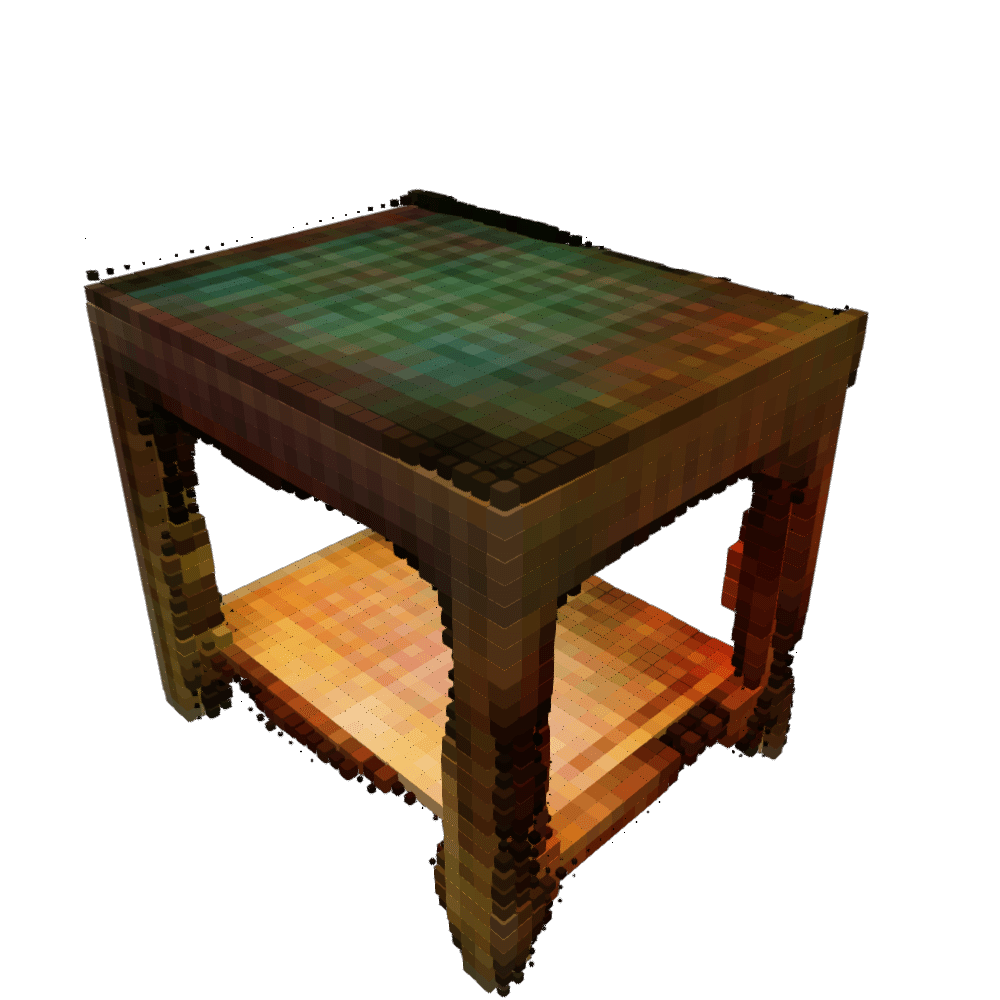} &
    \adjincludegraphics[width=\linewidth,trim={{.05\width} {.05\height} {.05\width} {.05\height}},clip]{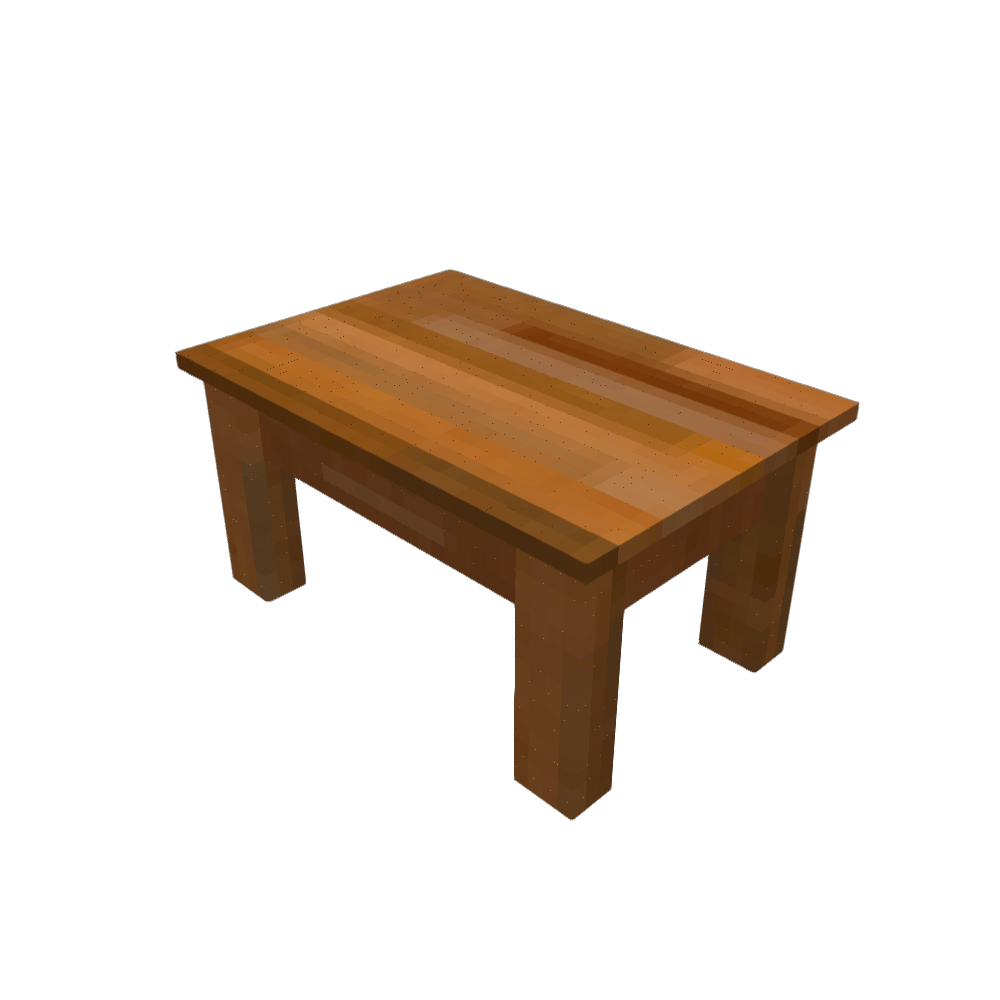}
    \\\midrule
    \textbf{6)} a simple and arrogant looking white square table. &
    \adjincludegraphics[width=\linewidth,trim={{.05\width} {.05\height} {.05\width} {.05\height}},clip]{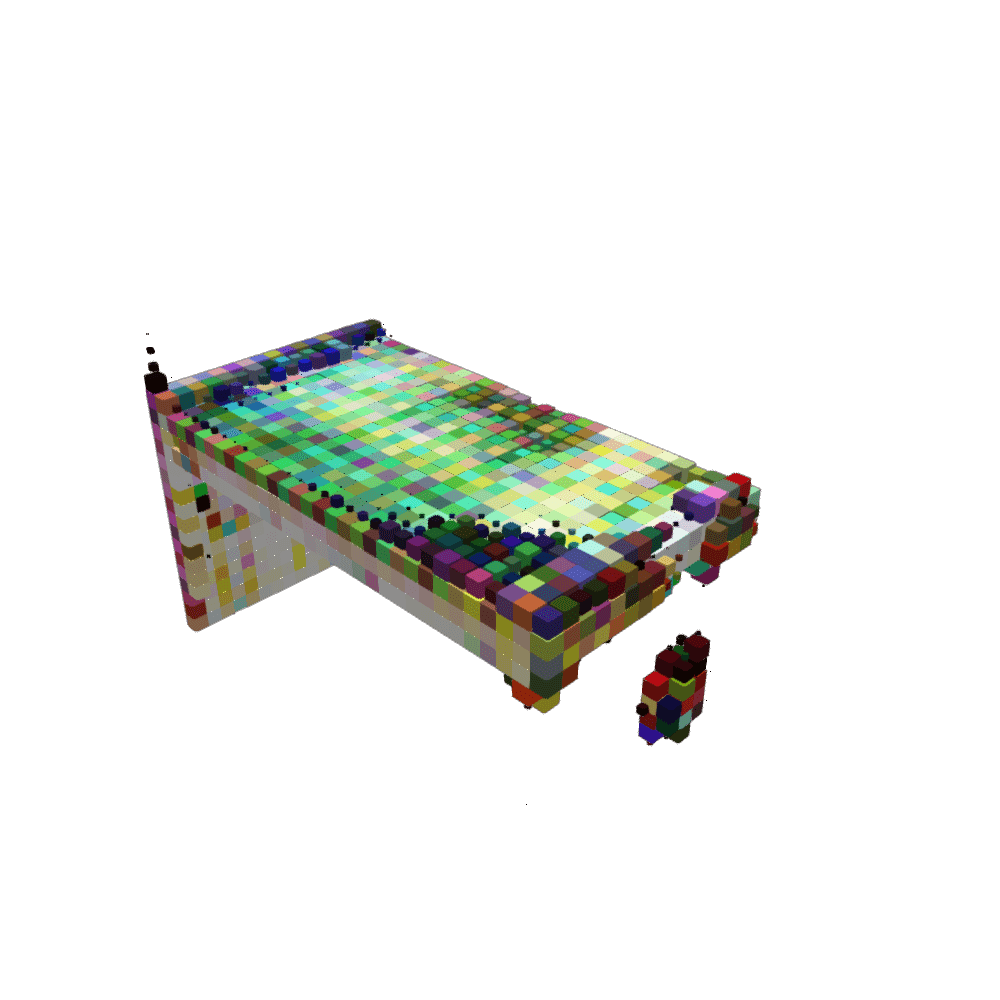} &
    \adjincludegraphics[width=\linewidth,trim={{.05\width} {.05\height} {.05\width} {.05\height}},clip]{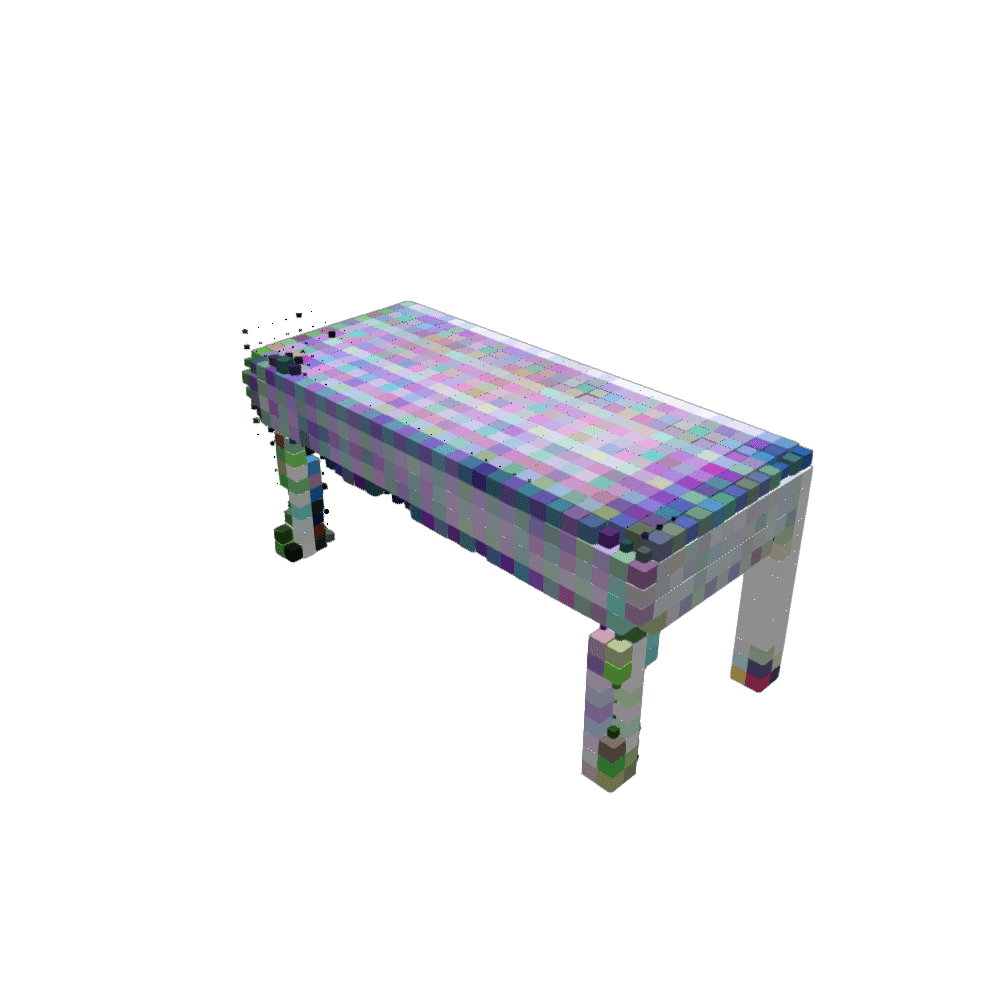} &
    \adjincludegraphics[width=\linewidth,trim={{.05\width} {.05\height} {.05\width} {.05\height}},clip]{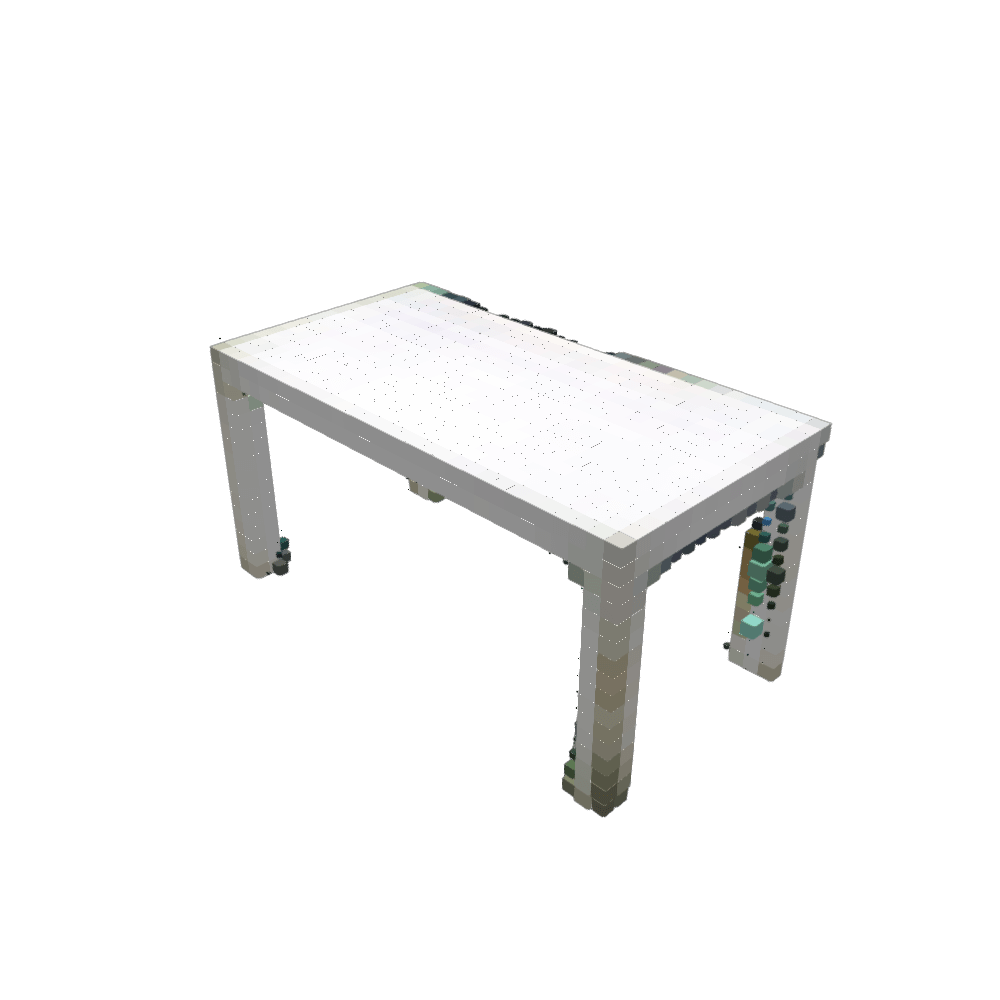} &
    \adjincludegraphics[width=\linewidth,trim={{.05\width} {.05\height} {.05\width} {.05\height}},clip]{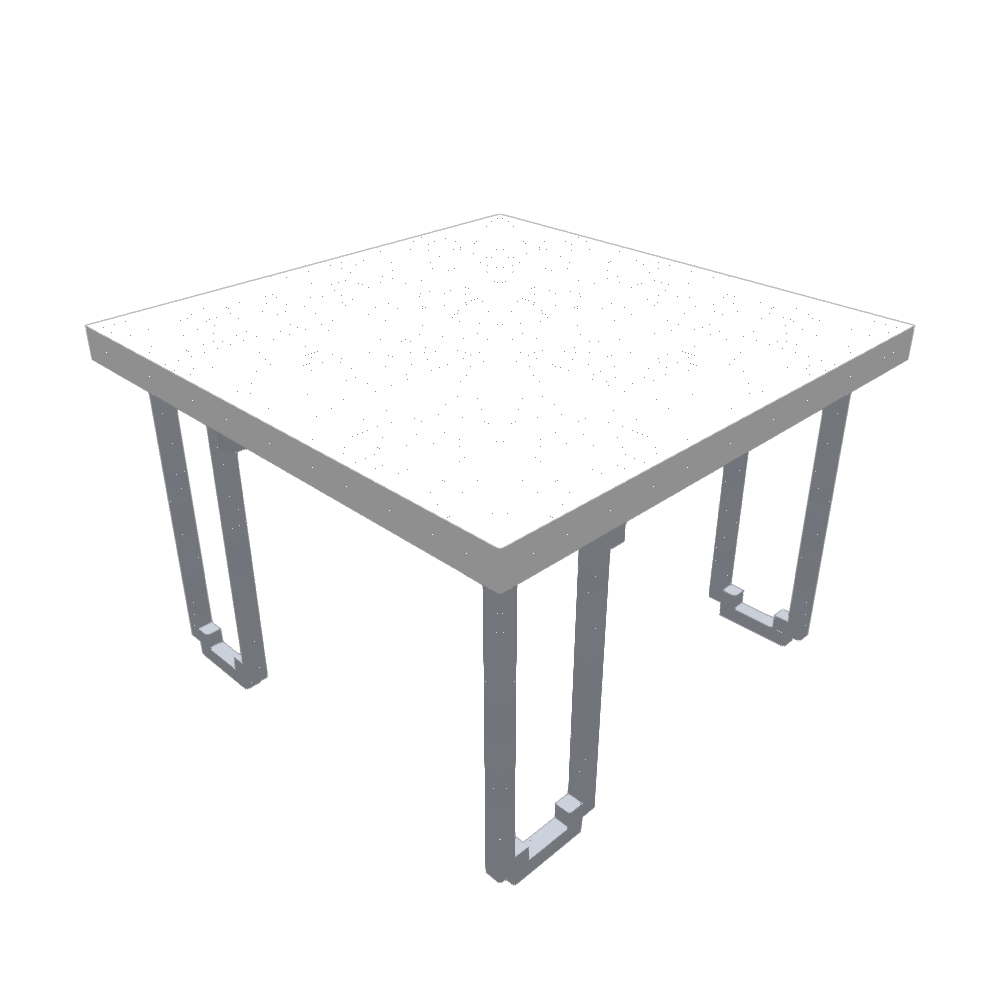}
    \\\midrule
    \textbf{7)} A table with square wood legs. The tabletop is blue. &
    \adjincludegraphics[width=\linewidth,trim={{.05\width} {.05\height} {.05\width} {.05\height}},clip]{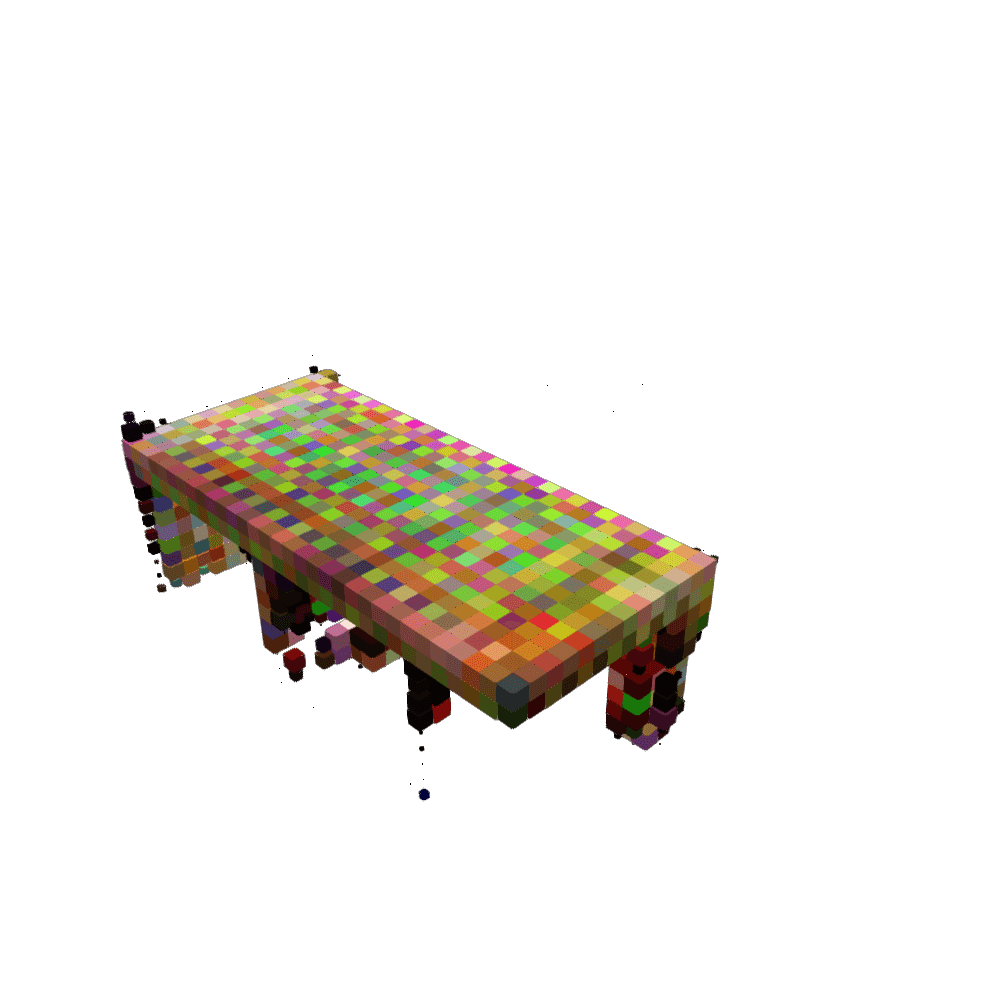} &
    \adjincludegraphics[width=\linewidth,trim={{.05\width} {.05\height} {.05\width} {.05\height}},clip]{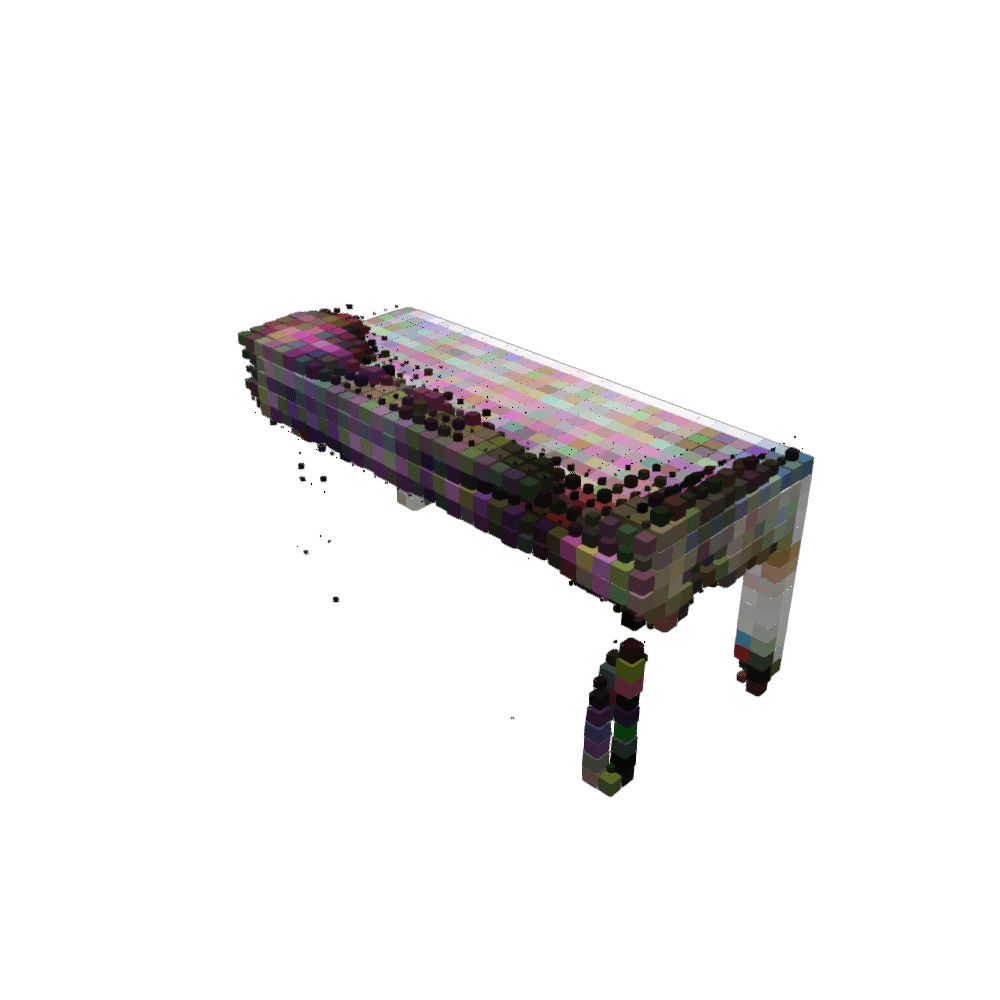} &
    \adjincludegraphics[width=\linewidth,trim={{.05\width} {.05\height} {.05\width} {.05\height}},clip]{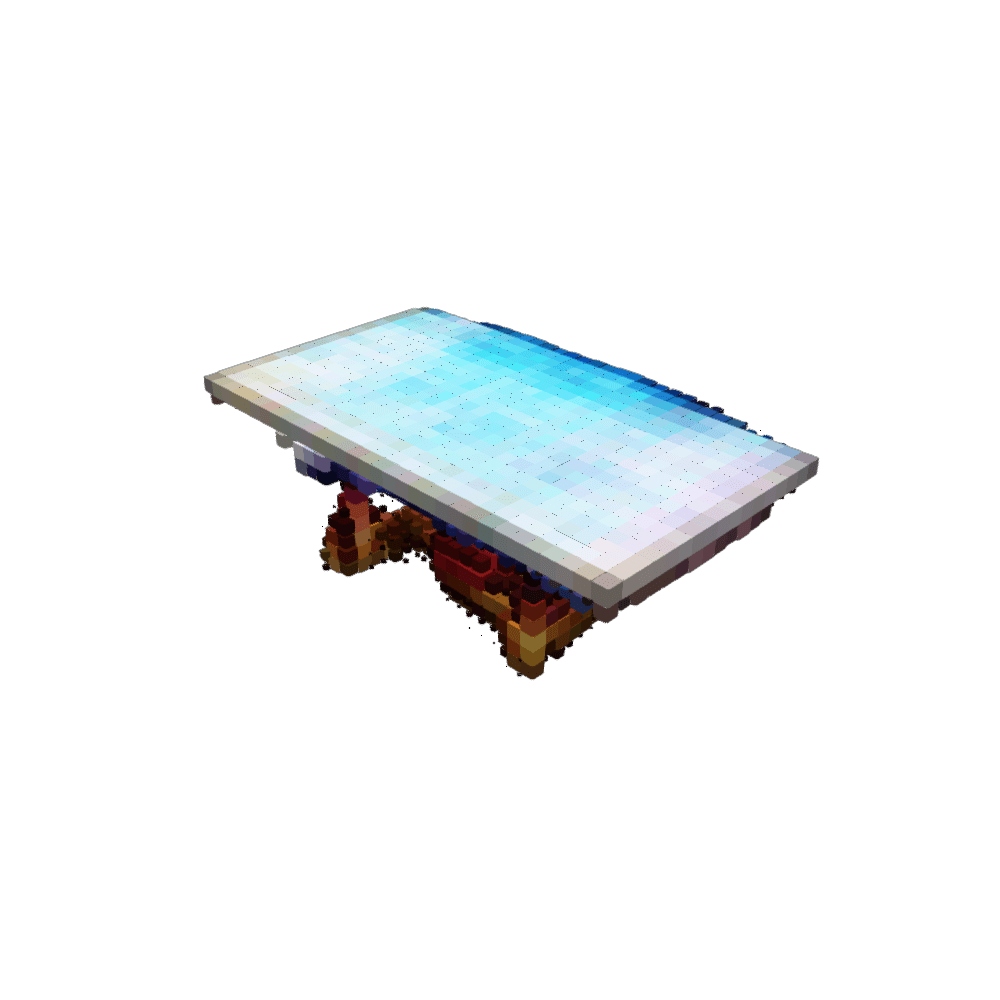} &
    \adjincludegraphics[width=\linewidth,trim={{.05\width} {.05\height} {.05\width} {.05\height}},clip]{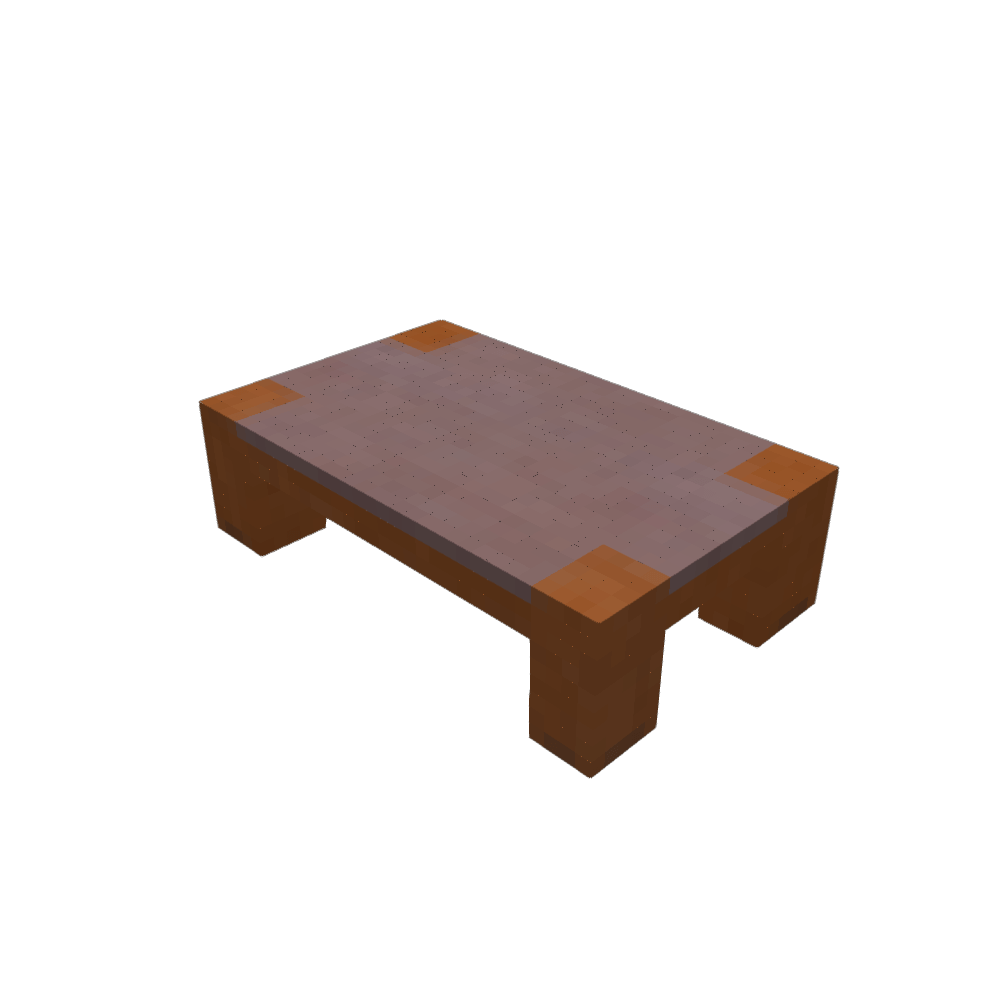}
    \\\midrule
    \textbf{8)} this table for constructing tbale color has brown, made from wooden sturcutre, it appears look practical appliances wooden frame,shape has rectangle, &
    \adjincludegraphics[width=\linewidth,trim={{.05\width} {.05\height} {.05\width} {.05\height}},clip]{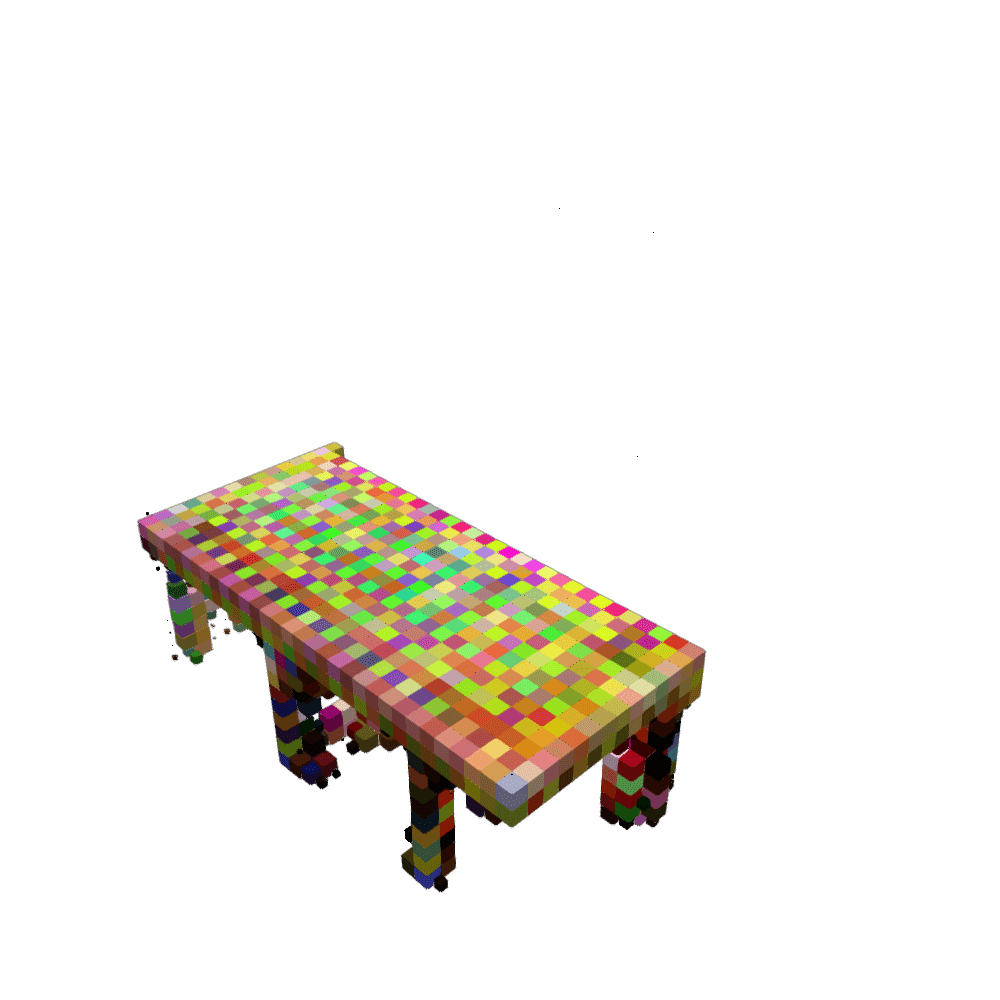} &
    \adjincludegraphics[width=\linewidth,trim={{.05\width} {.05\height} {.05\width} {.05\height}},clip]{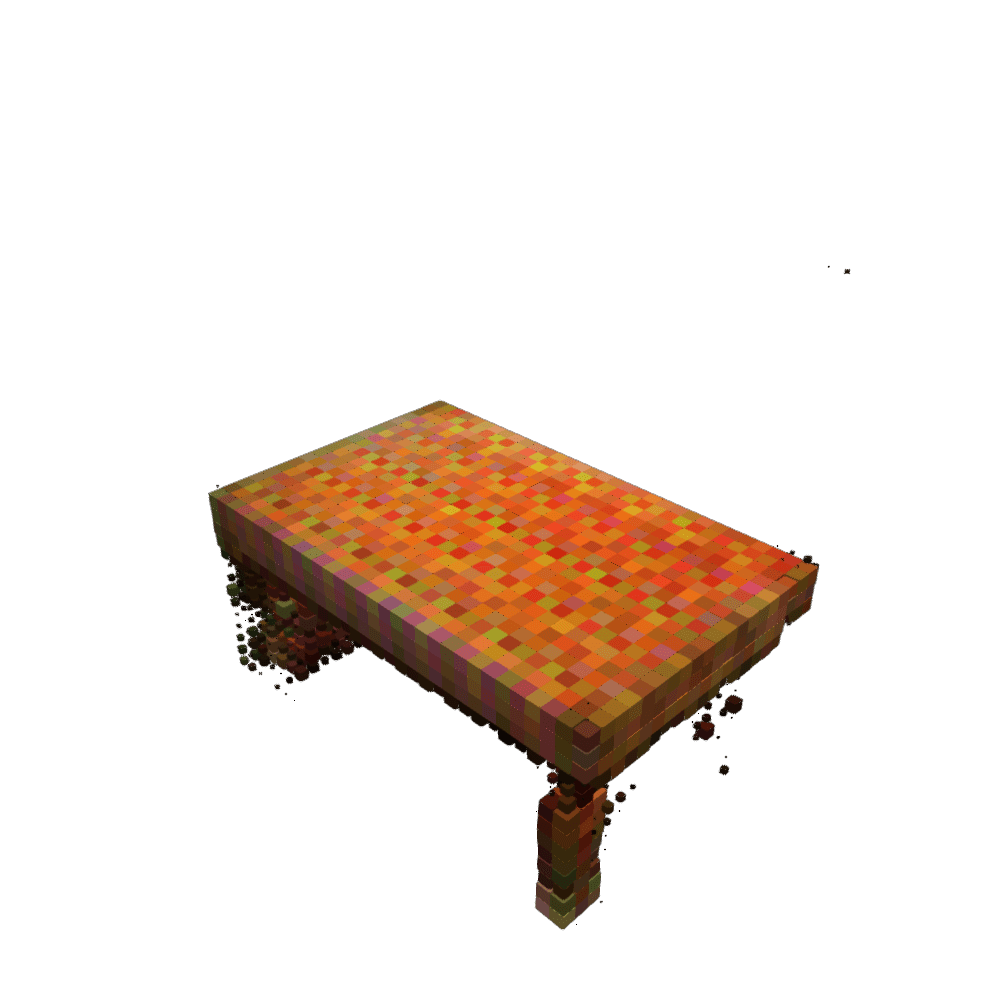} &
    \adjincludegraphics[width=\linewidth,trim={{.05\width} {.05\height} {.05\width} {.05\height}},clip]{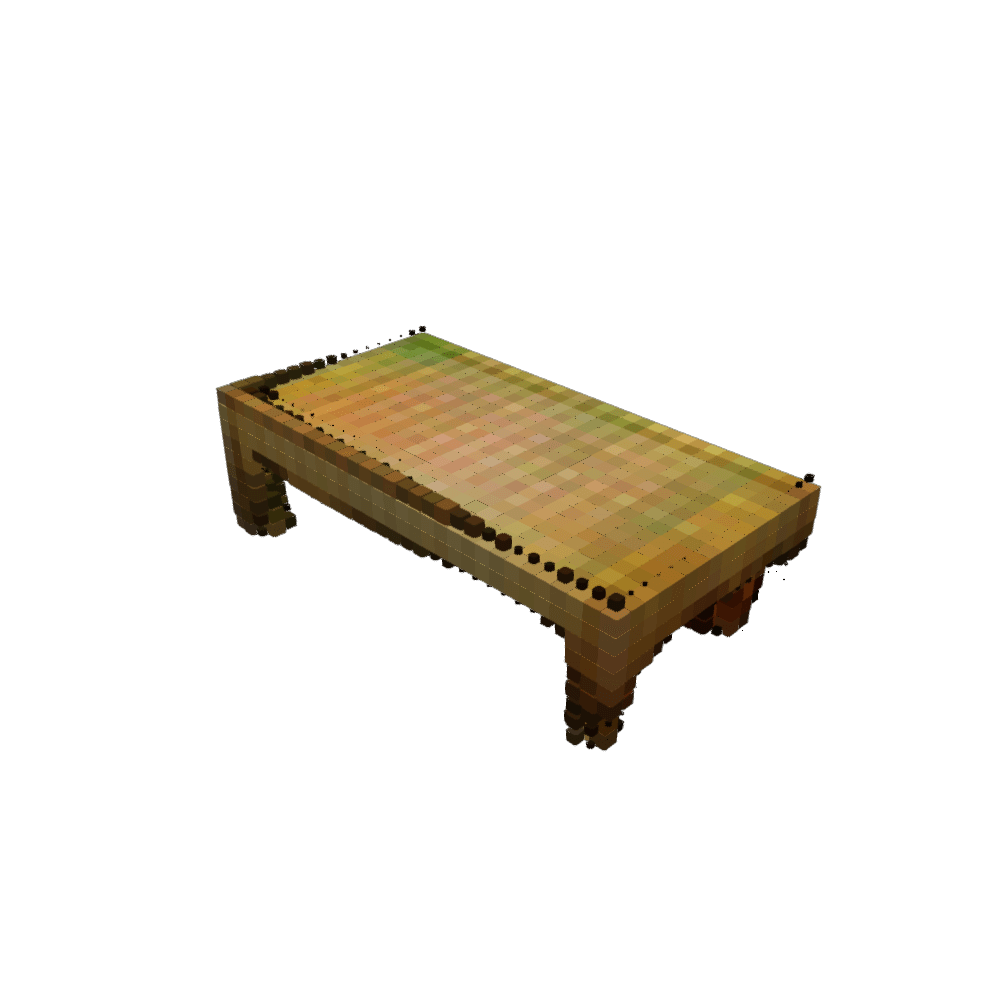} &
    \adjincludegraphics[width=\linewidth,trim={{.05\width} {.05\height} {.05\width} {.05\height}},clip]{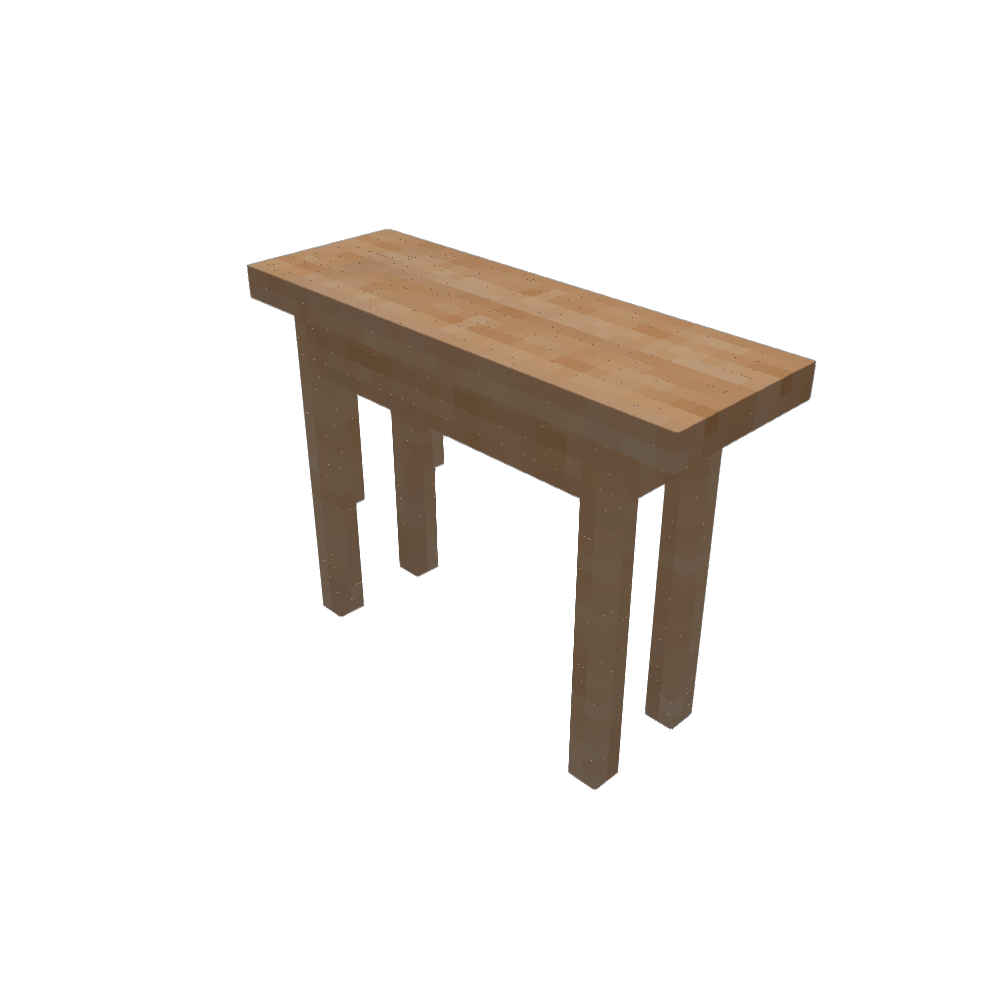}
    \\\midrule
    \textbf{9)} A rectangle shaped dining room table with metal legs. &
    \adjincludegraphics[width=\linewidth,trim={{.05\width} {.05\height} {.05\width} {.05\height}},clip]{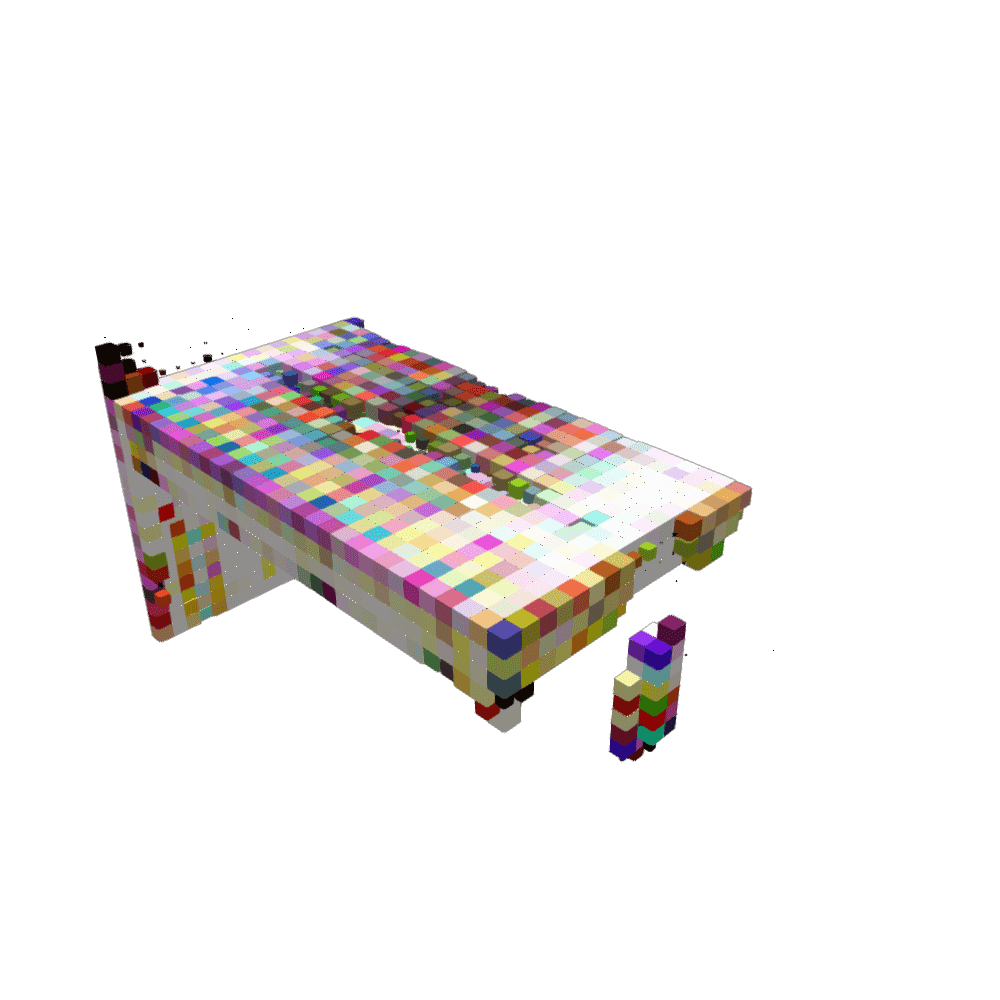} &
    \adjincludegraphics[width=\linewidth,trim={{.05\width} {.05\height} {.05\width} {.05\height}},clip]{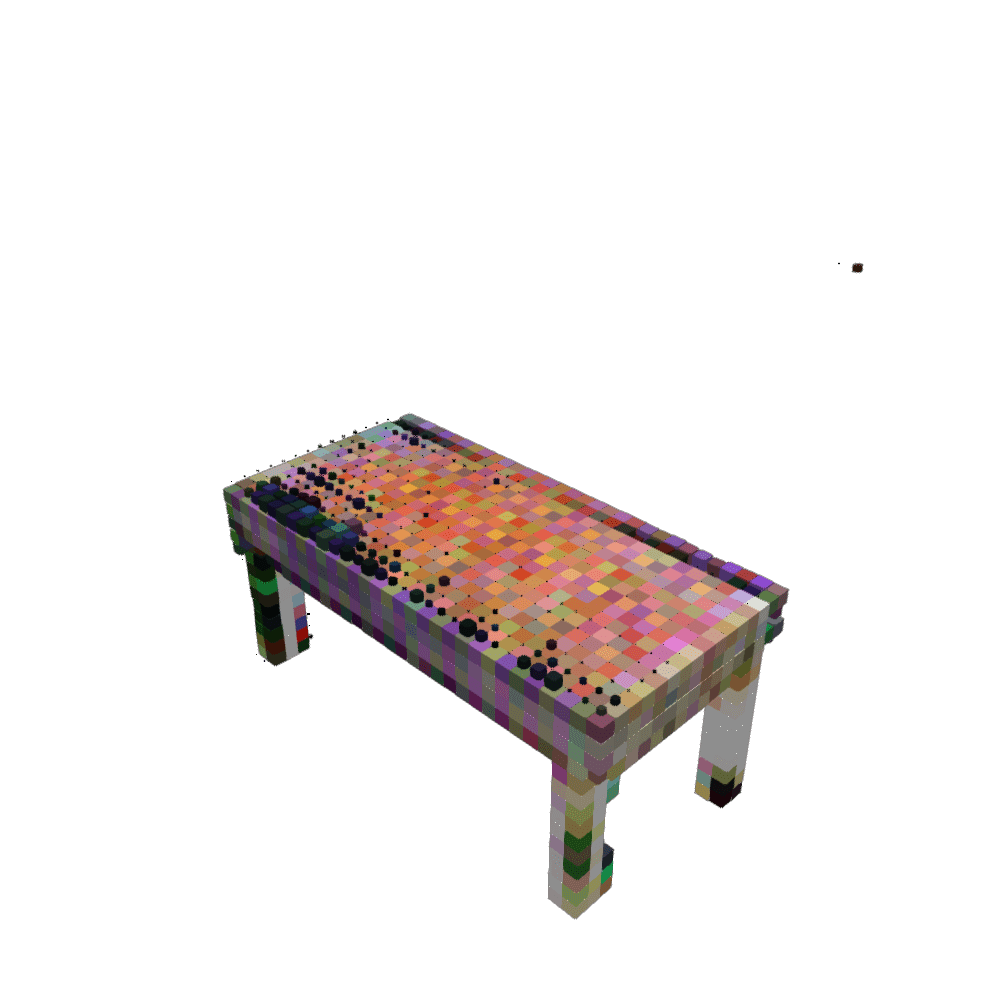} &
    \adjincludegraphics[width=\linewidth,trim={{.05\width} {.05\height} {.05\width} {.05\height}},clip]{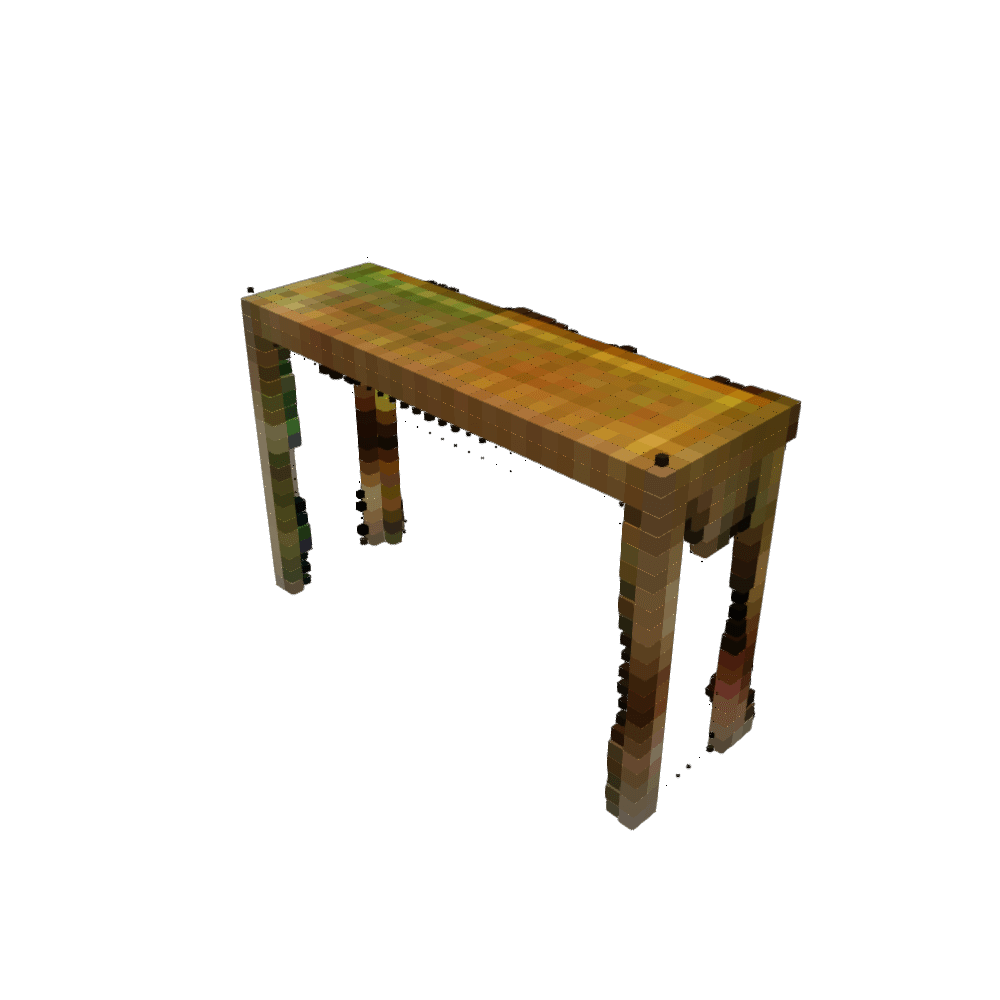} &
    \adjincludegraphics[width=\linewidth,trim={{.05\width} {.05\height} {.05\width} {.05\height}},clip]{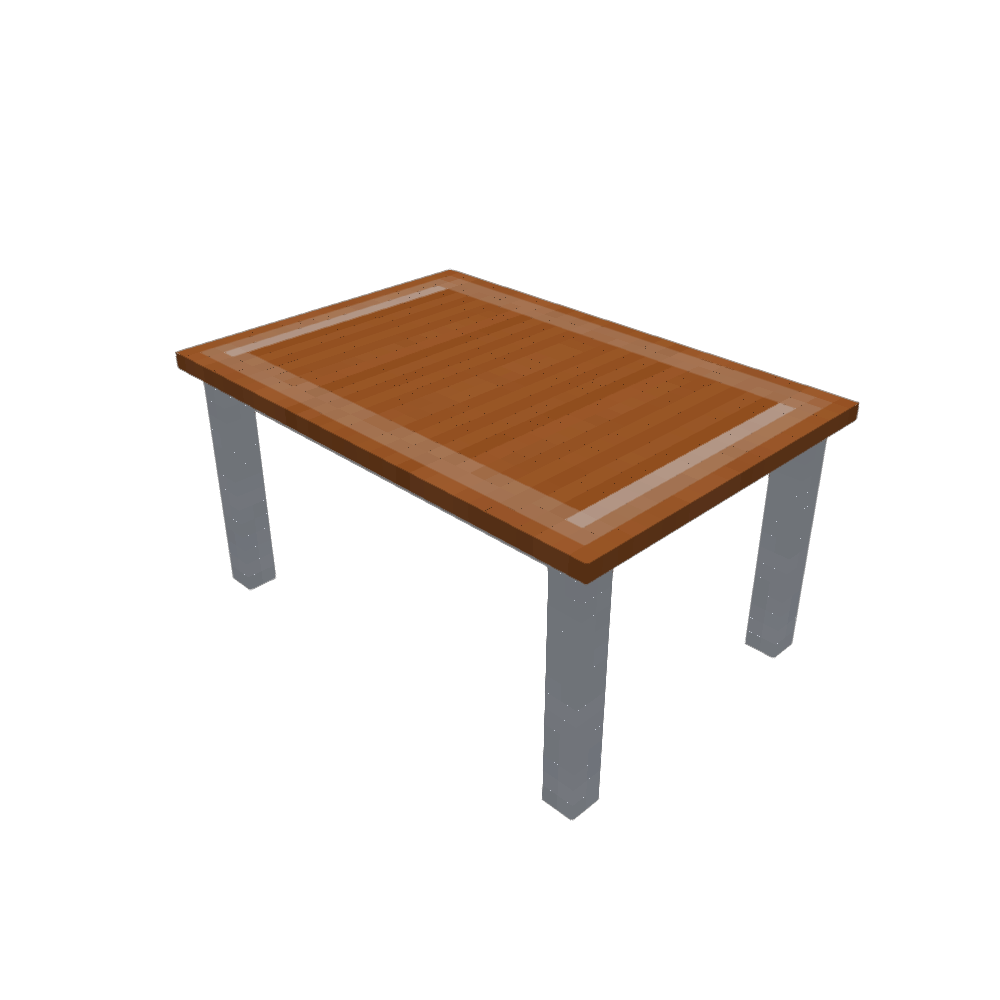}
    \\\midrule
    \textbf{10)} Wooden chair made up of cuisine seat and back which is in silver in colour good appearnce. &
    \adjincludegraphics[width=\linewidth,trim={{.05\width} {.05\height} {.05\width} {.05\height}},clip]{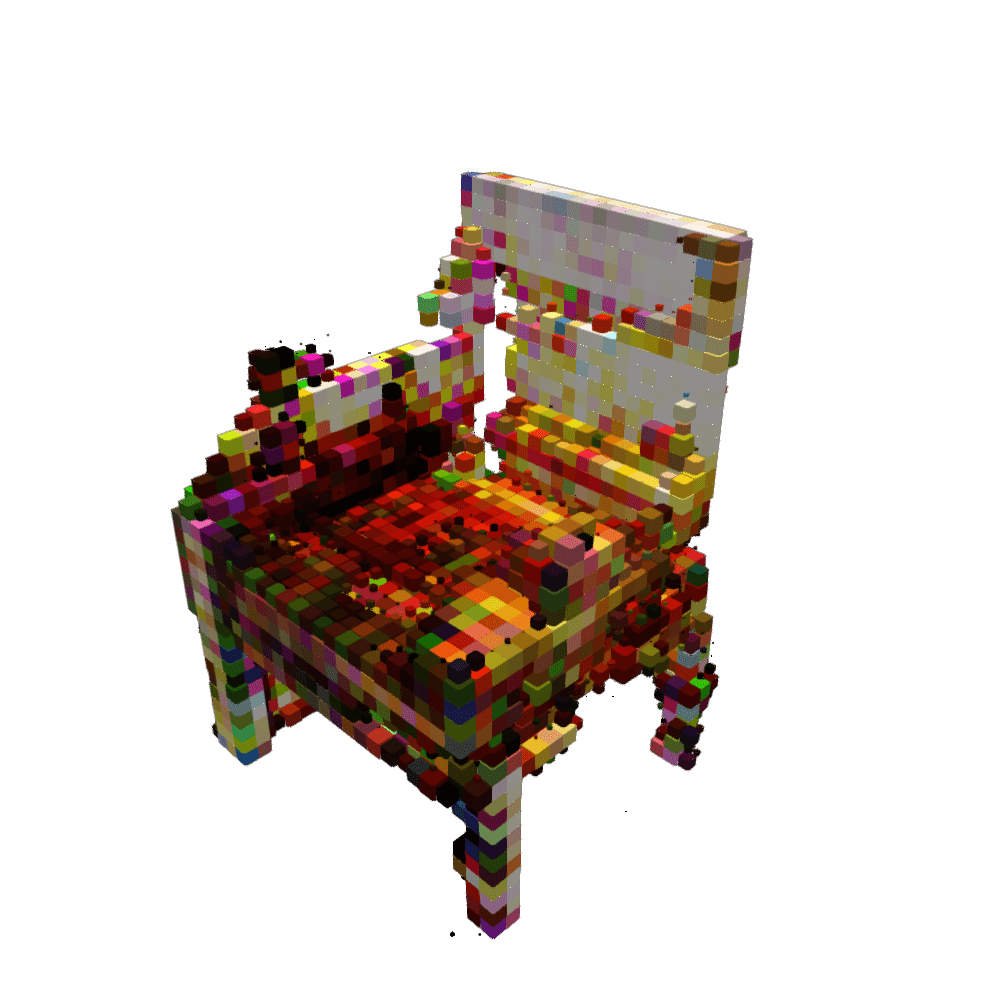} &
    \adjincludegraphics[width=\linewidth,trim={{.05\width} {.05\height} {.05\width} {.05\height}},clip]{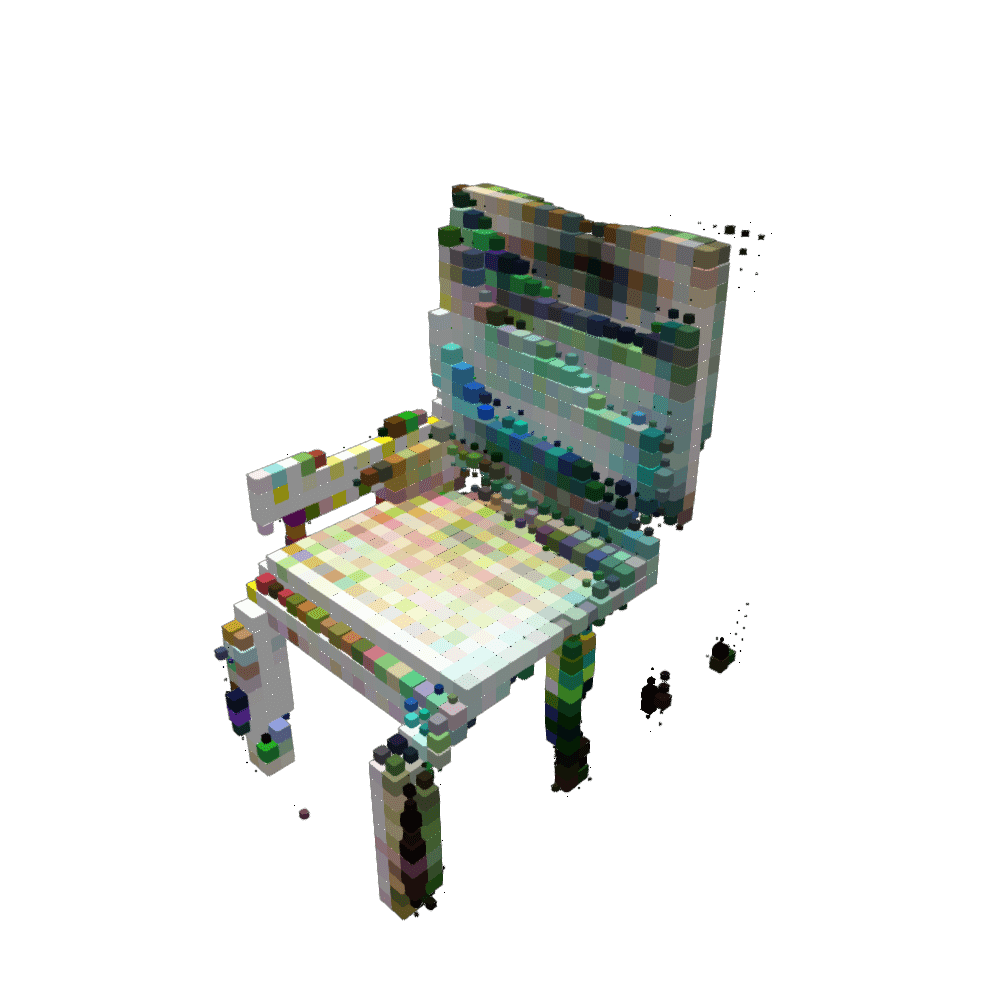} &
    \adjincludegraphics[width=\linewidth,trim={{.05\width} {.05\height} {.05\width} {.05\height}},clip]{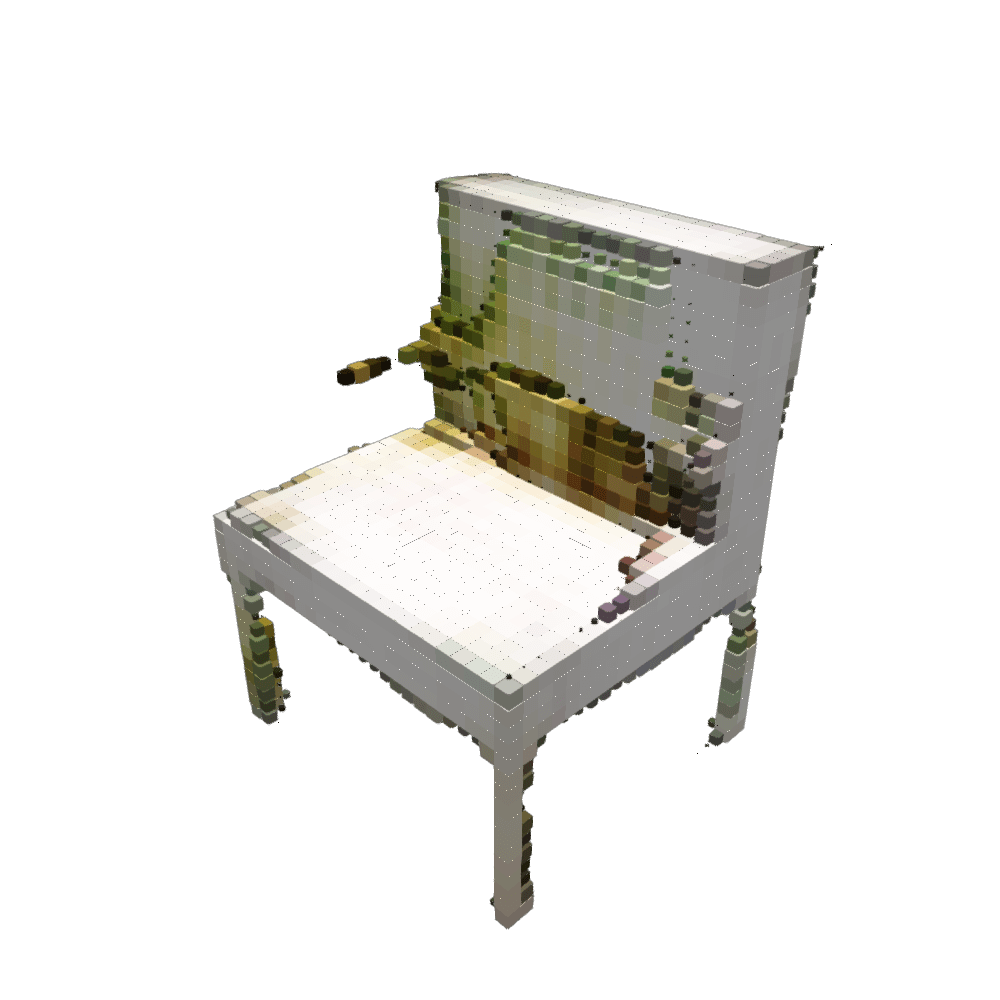} &
    \adjincludegraphics[width=\linewidth,trim={{.05\width} {.05\height} {.05\width} {.05\height}},clip]{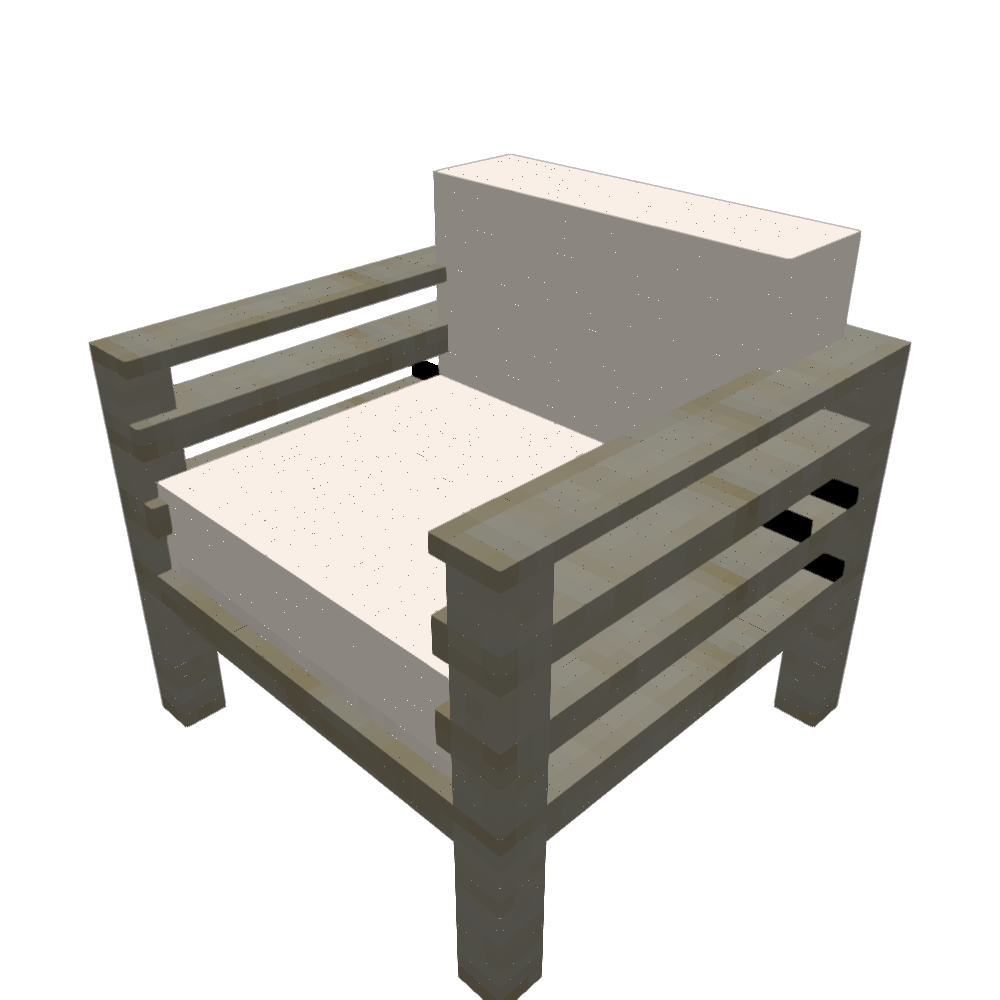}
    \end{tabular}
    \vspace{-1em}
    \caption{\textbf{Example text-to-shape generations}. Our CWGAN approach is correctly conditioned by the input text, generating shapes that match the described attributes. The baseline approach struggles to generate plausible shapes. Additional results can be found in the supplementary material.}
    \vspace{-1.5em}
    \label{fig:text2shape-results-supplementary}
\end{figure}

%% file: fig/text2shape-results-supplementary-2.tex
\begin{figure}
    \vspace{-2em}
    \centering\scriptsize
    \begin{tabular}{m{6cm}M{1.3cm}M{1.3cm}M{1.3cm}M{1.3cm}}
    Input Text & GAN-INT-CLS~\cite{reed2016generative} & Ours CGAN & Ours CWGAN & GT
    \\\toprule
    \textbf{11)} A table with a glass table top, and four U-shaped legs. &
    \adjincludegraphics[width=\linewidth,trim={{.05\width} {.05\height} {.05\width} {.05\height}},clip]{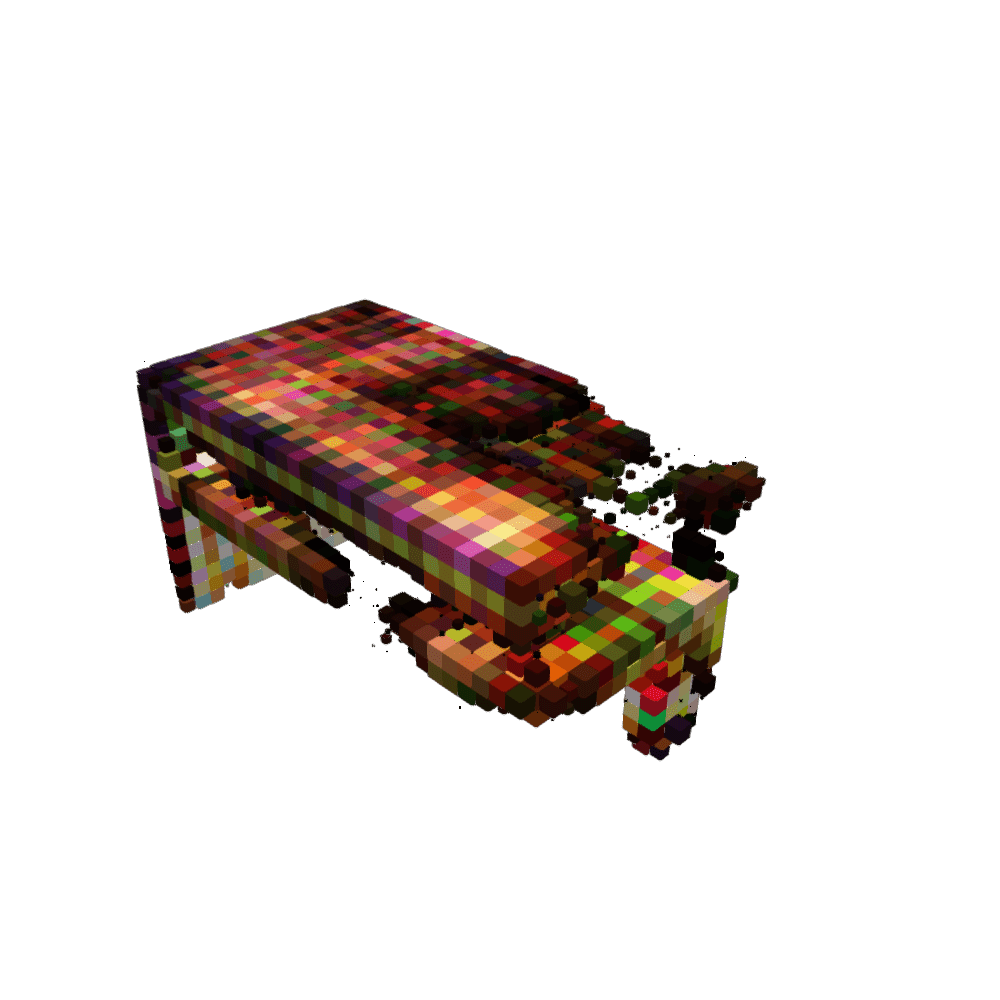} &
    \adjincludegraphics[width=\linewidth,trim={{.05\width} {.05\height} {.05\width} {.05\height}},clip]{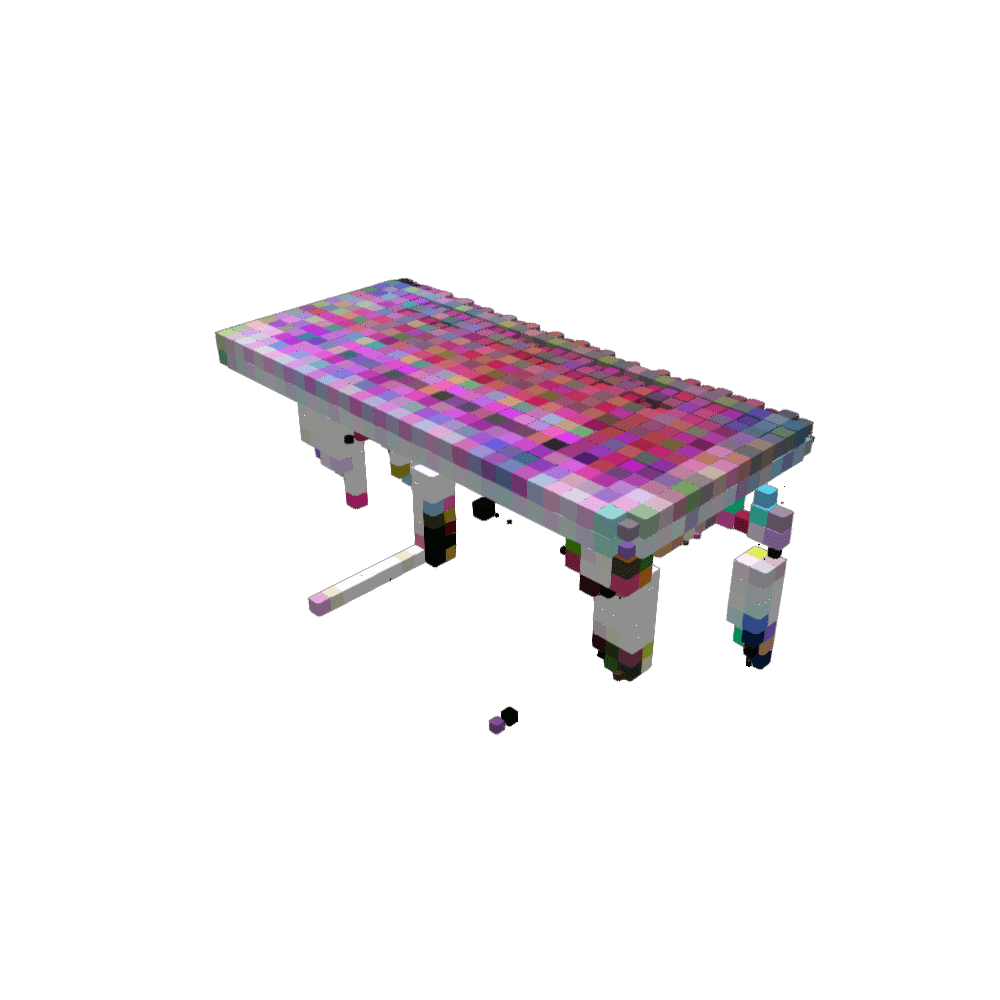} &
    \adjincludegraphics[width=\linewidth,trim={{.05\width} {.05\height} {.05\width} {.05\height}},clip]{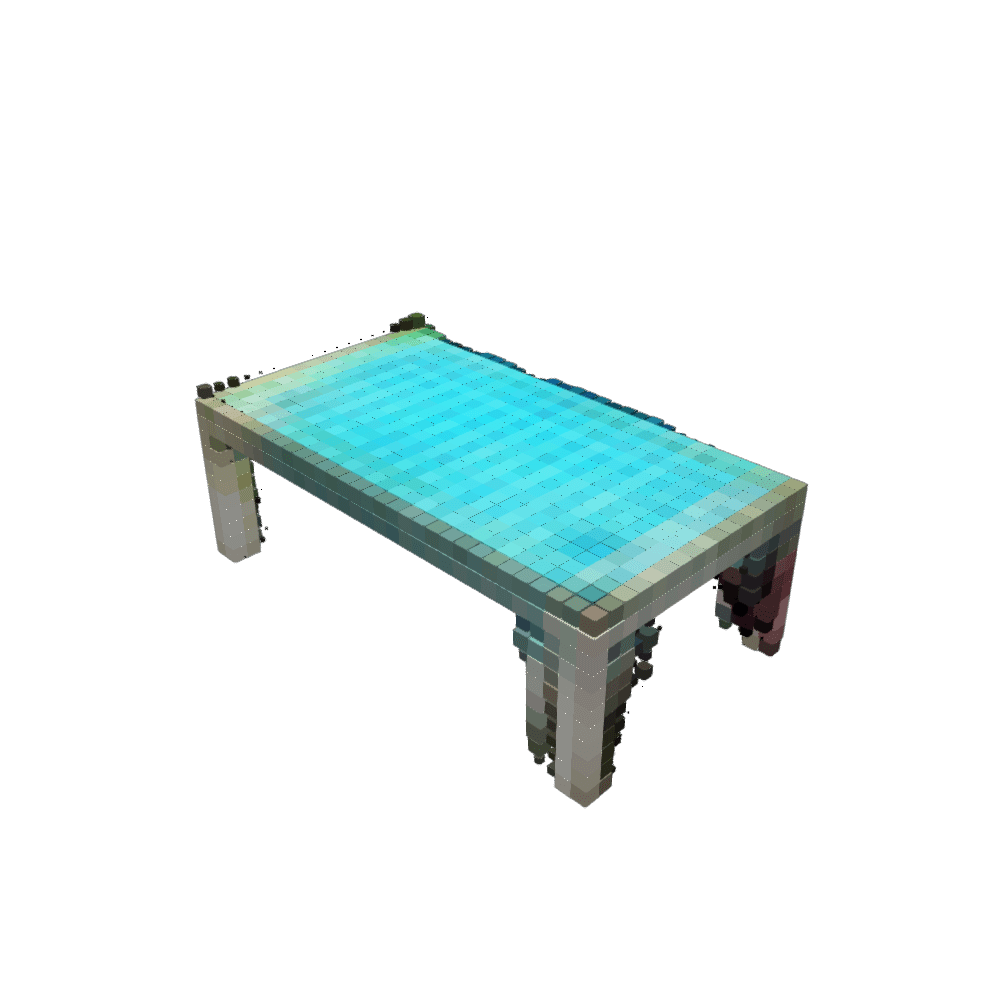} &
    \adjincludegraphics[width=\linewidth,trim={{.05\width} {.05\height} {.05\width} {.05\height}},clip]{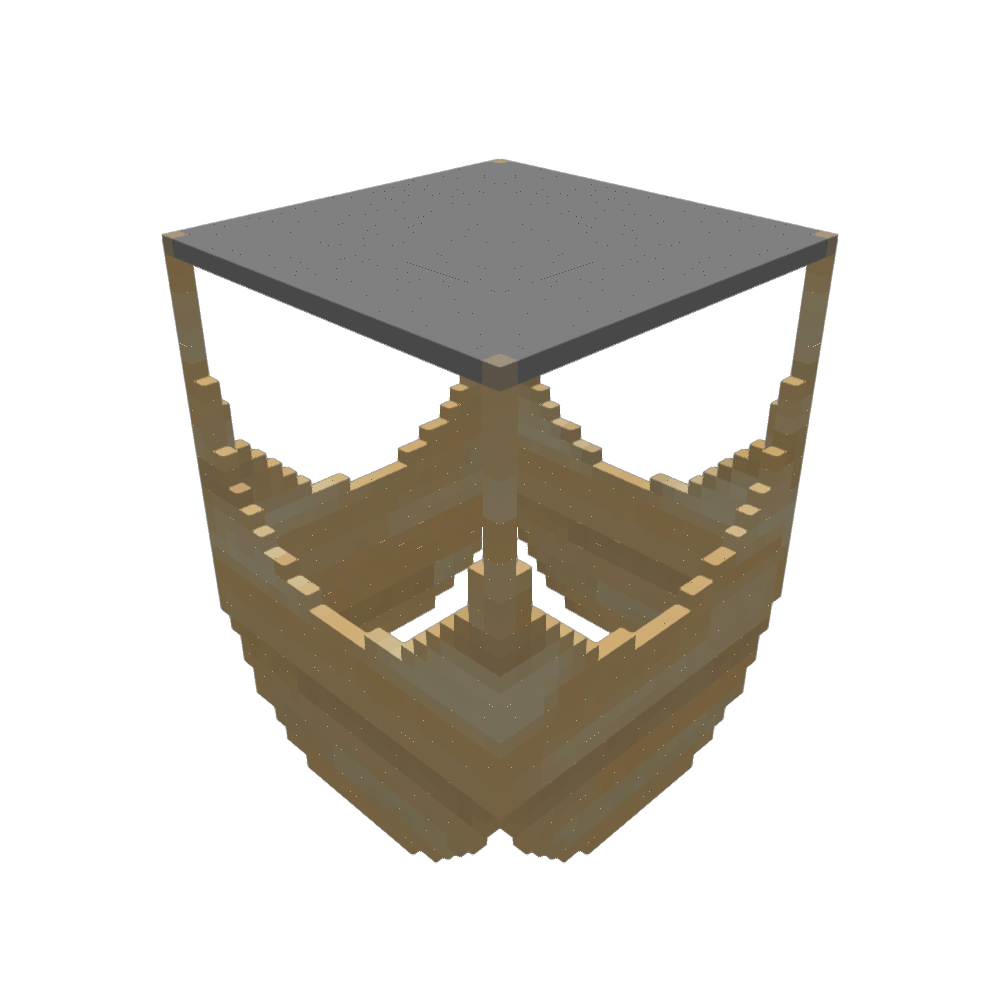}
    \\\midrule
    \textbf{12)} A round wooden coffee table with three legs and a raised edge around the table top &
    \adjincludegraphics[width=\linewidth,trim={{.05\width} {.05\height} {.05\width} {.05\height}},clip]{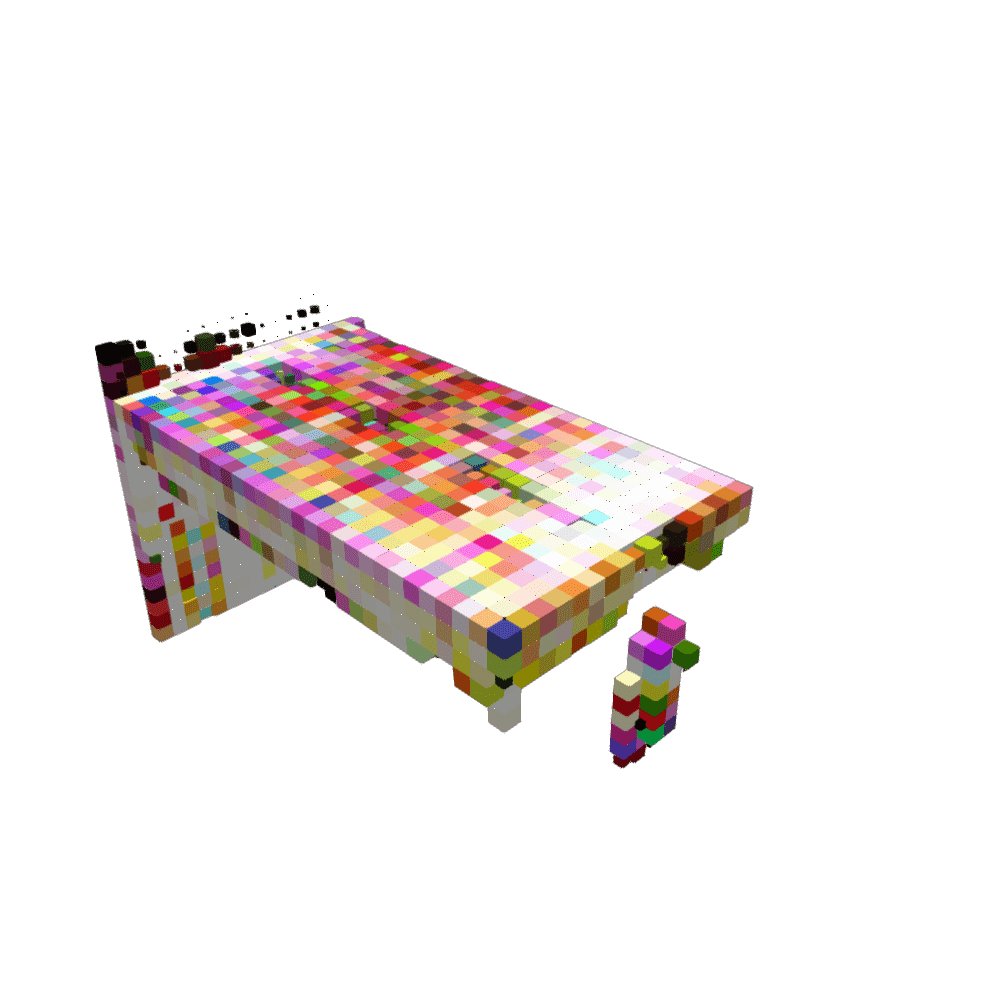} &
    \adjincludegraphics[width=\linewidth,trim={{.05\width} {.05\height} {.05\width} {.05\height}},clip]{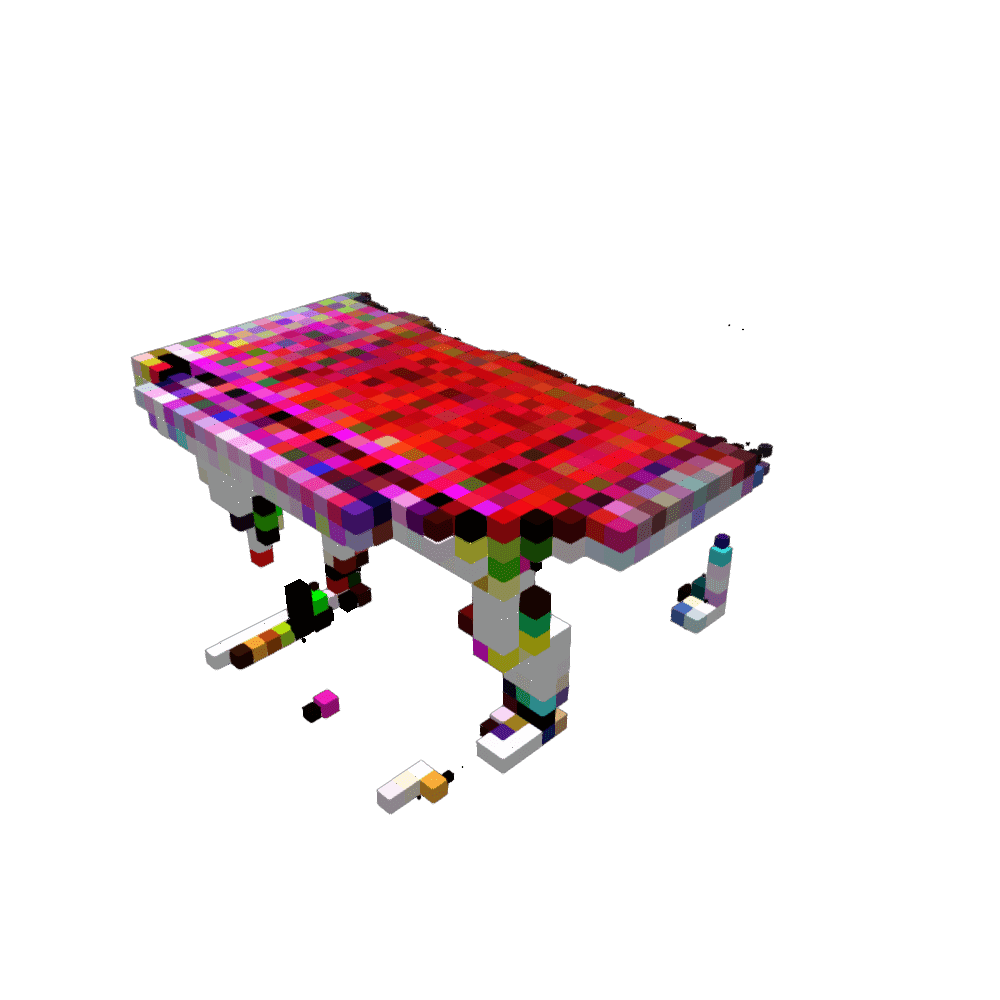} &
    \adjincludegraphics[width=\linewidth,trim={{.05\width} {.05\height} {.05\width} {.05\height}},clip]{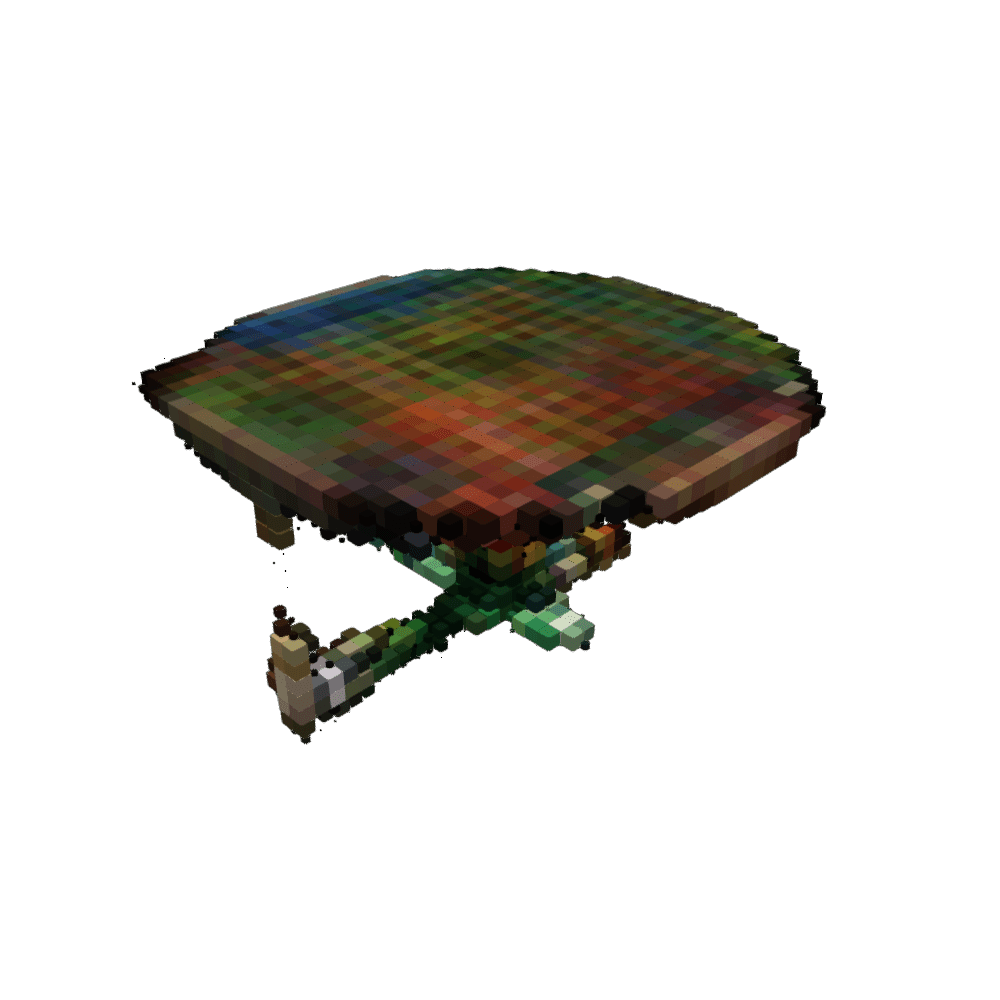} &
    \adjincludegraphics[width=\linewidth,trim={{.05\width} {.05\height} {.05\width} {.05\height}},clip]{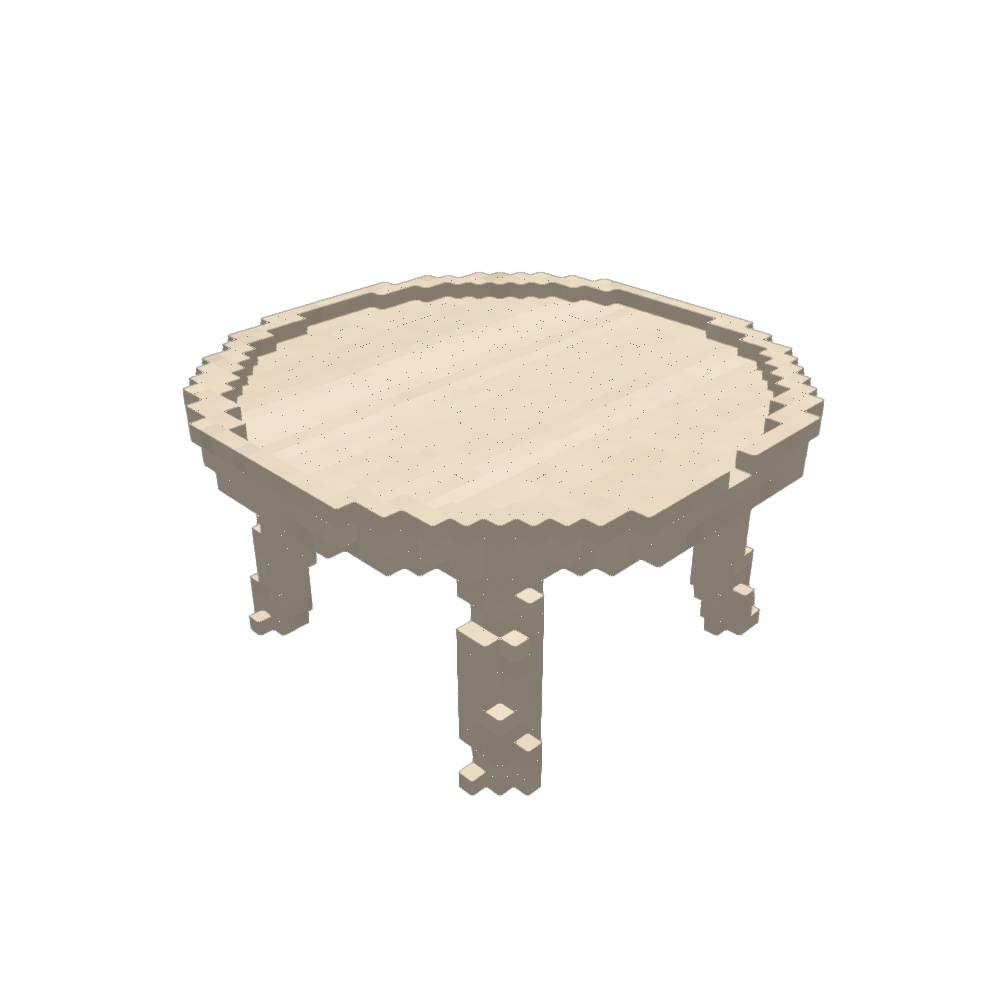}
    \\\midrule
    \textbf{13)} Broad rectangular shaped surface table with four short legs. &
    \adjincludegraphics[width=\linewidth,trim={{.05\width} {.05\height} {.05\width} {.05\height}},clip]{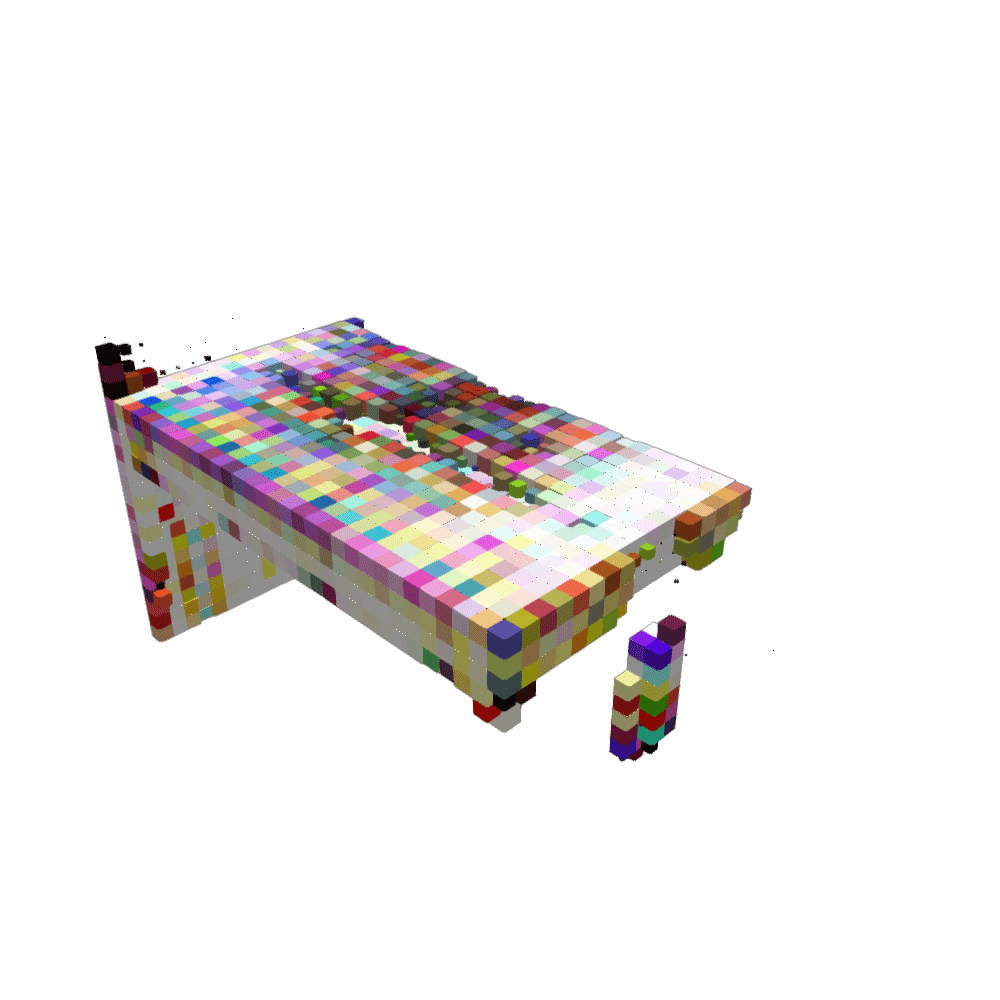} &
    \adjincludegraphics[width=\linewidth,trim={{.05\width} {.05\height} {.05\width} {.05\height}},clip]{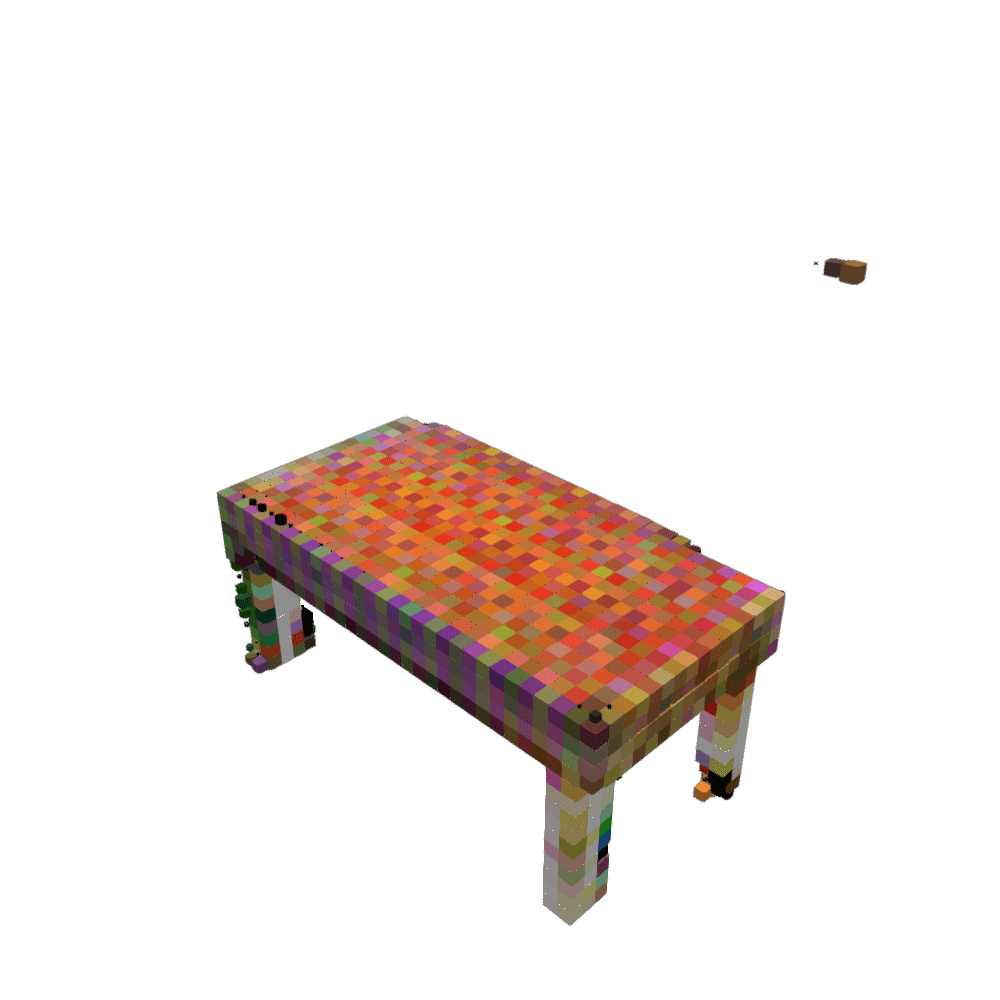} &
    \adjincludegraphics[width=\linewidth,trim={{.05\width} {.05\height} {.05\width} {.05\height}},clip]{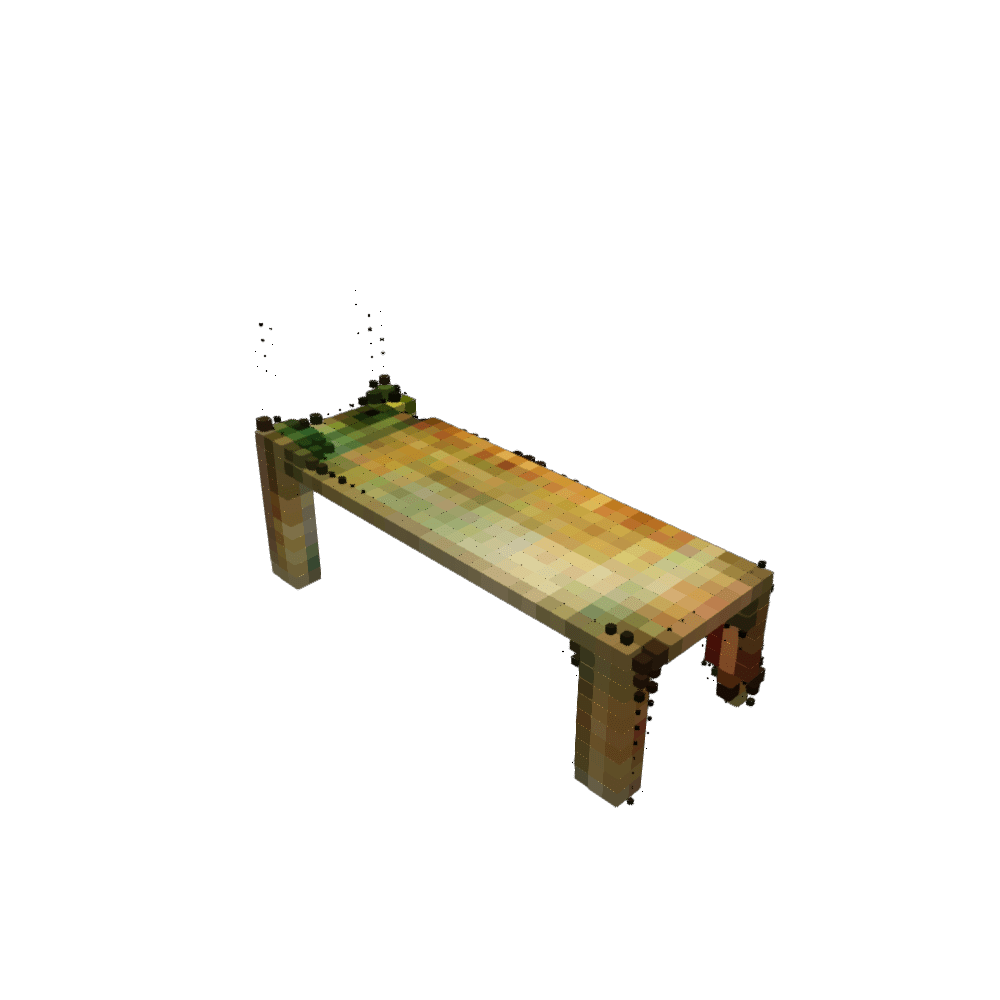} &
    \adjincludegraphics[width=\linewidth,trim={{.05\width} {.05\height} {.05\width} {.05\height}},clip]{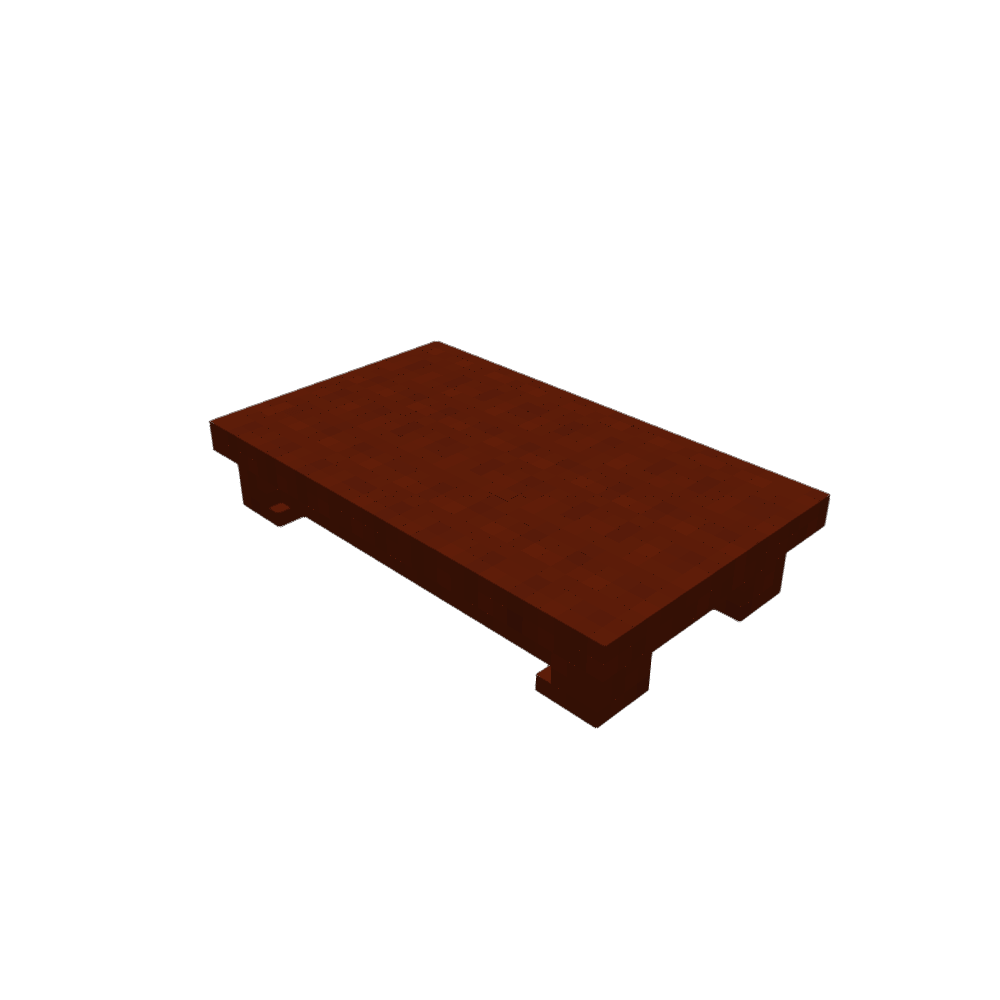}
    \\\midrule
    \textbf{14)} It is a gray coffee table with lined patterns on the top and undershelf. &
    \adjincludegraphics[width=\linewidth,trim={{.05\width} {.05\height} {.05\width} {.05\height}},clip]{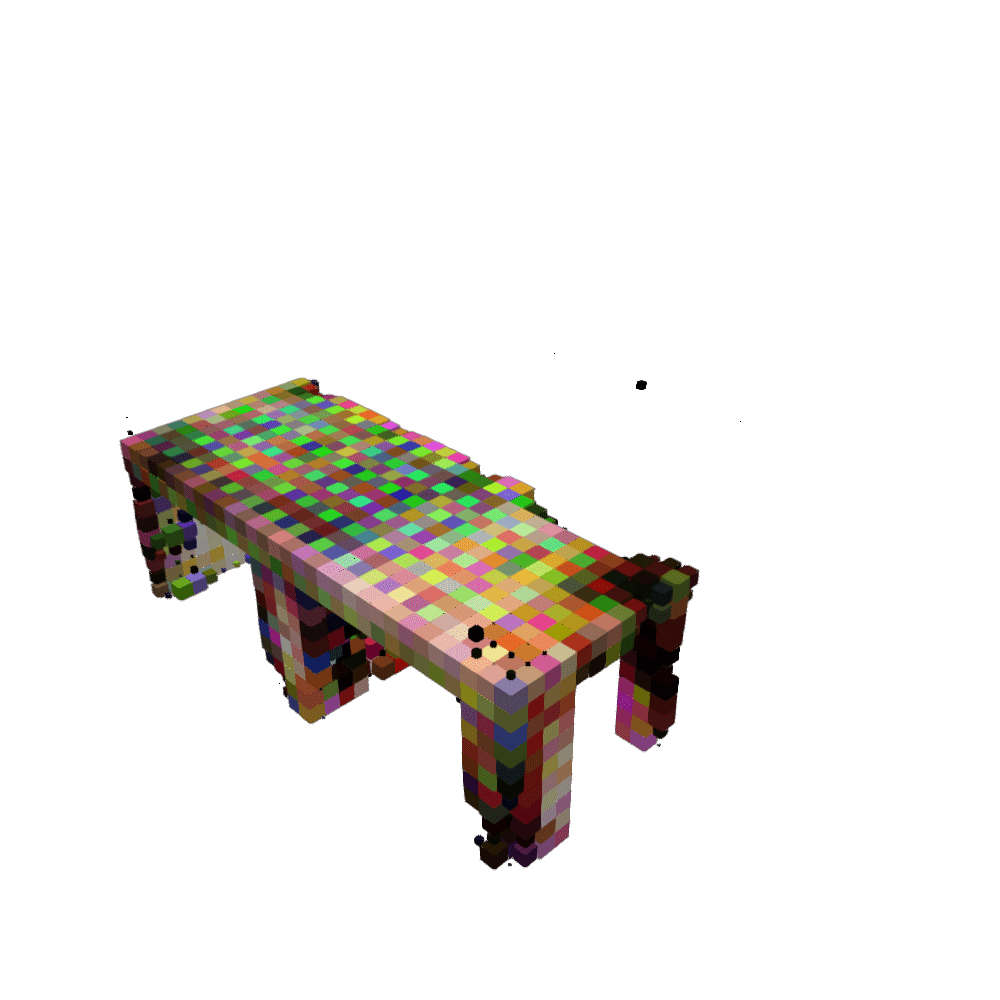} &
    \adjincludegraphics[width=\linewidth,trim={{.05\width} {.05\height} {.05\width} {.05\height}},clip]{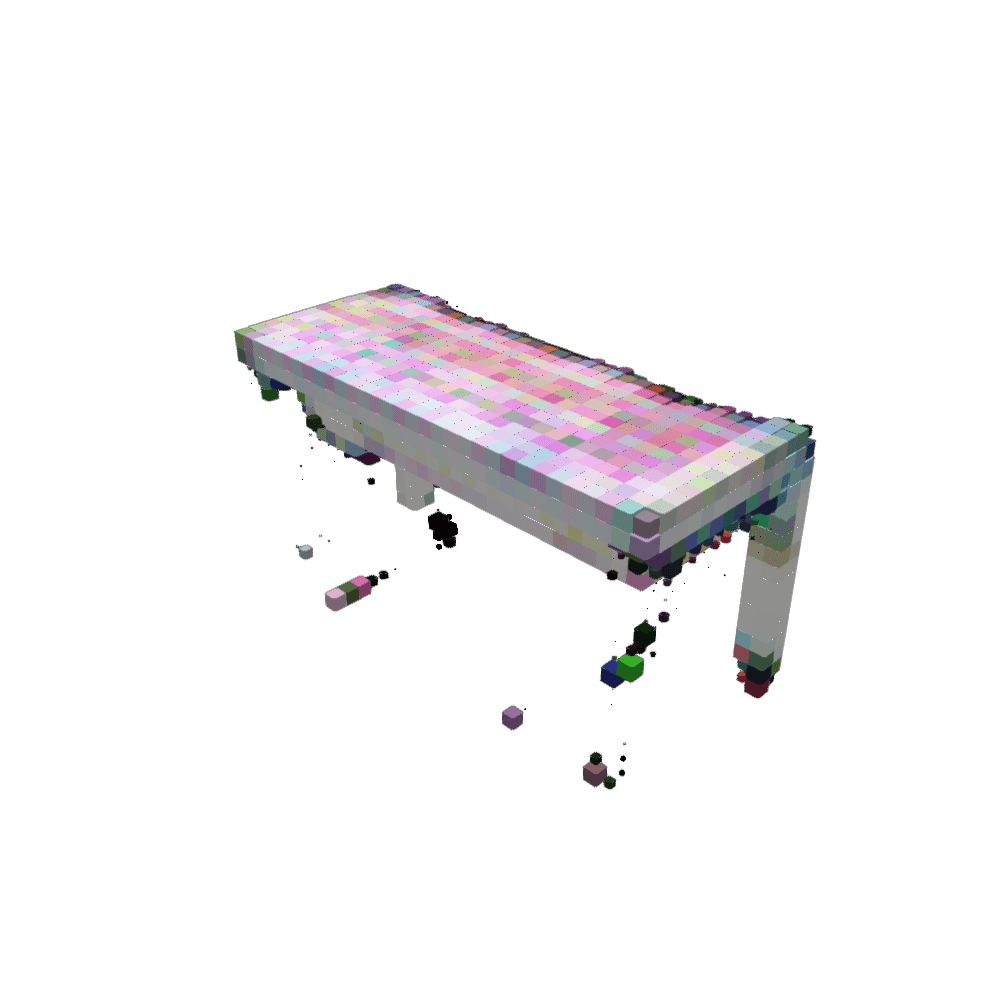} &
    \adjincludegraphics[width=\linewidth,trim={{.05\width} {.05\height} {.05\width} {.05\height}},clip]{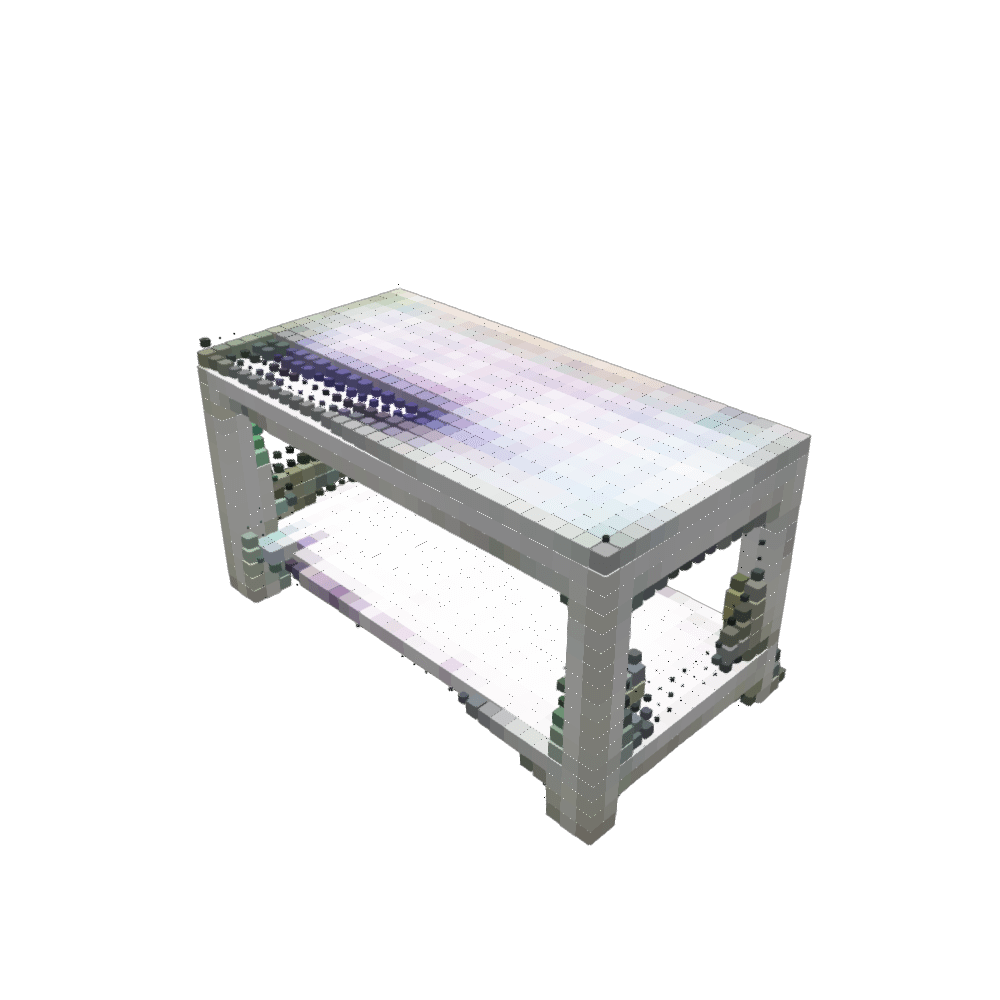} &
    \adjincludegraphics[width=\linewidth,trim={{.05\width} {.05\height} {.05\width} {.05\height}},clip]{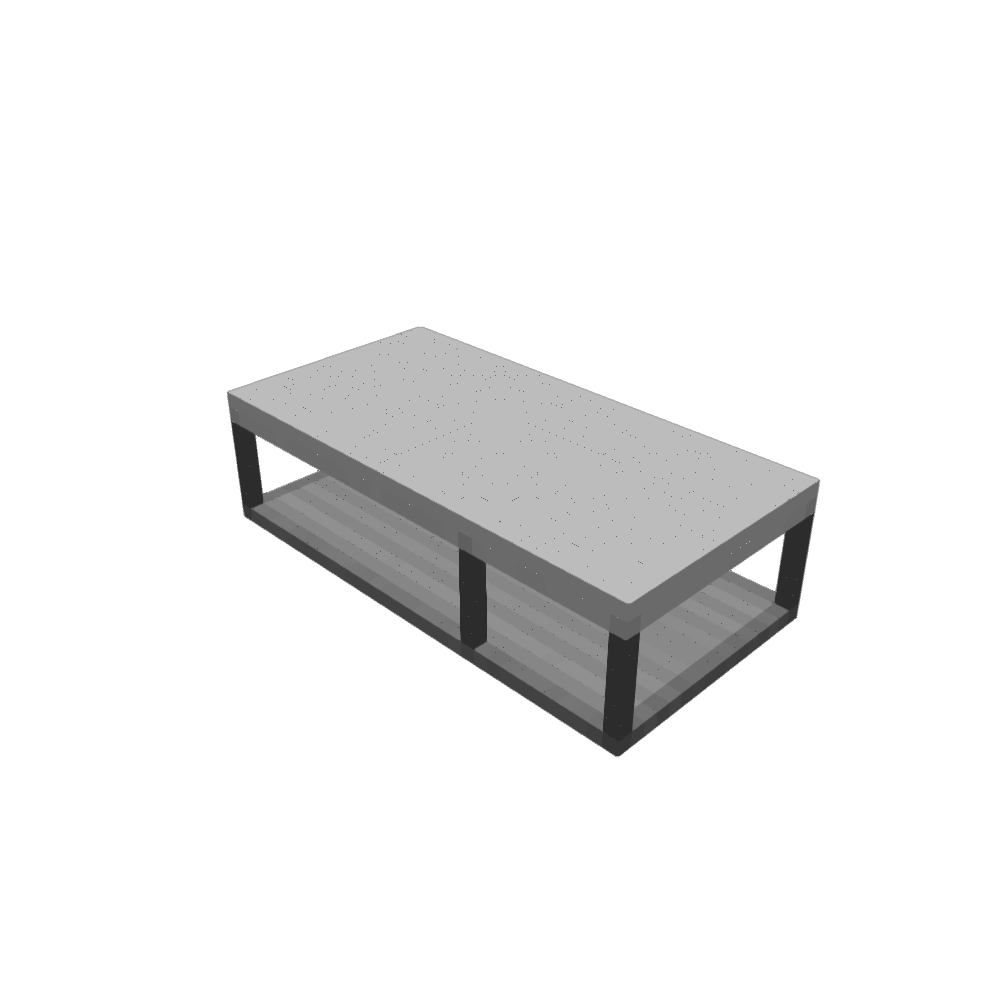}
    \\\midrule
    \textbf{15)} a brown folding chair &
    \adjincludegraphics[width=\linewidth,trim={{.05\width} {.05\height} {.05\width} {.05\height}},clip]{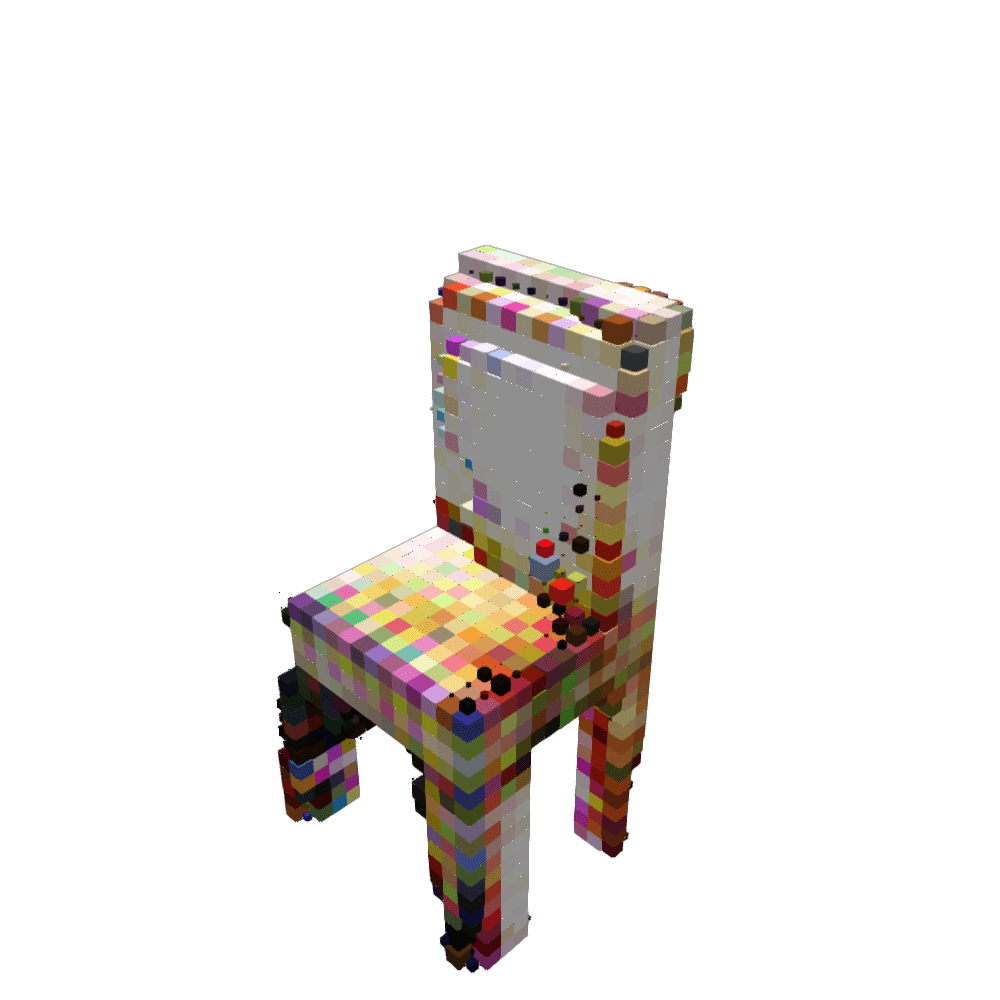} &
    \adjincludegraphics[width=\linewidth,trim={{.05\width} {.05\height} {.05\width} {.05\height}},clip]{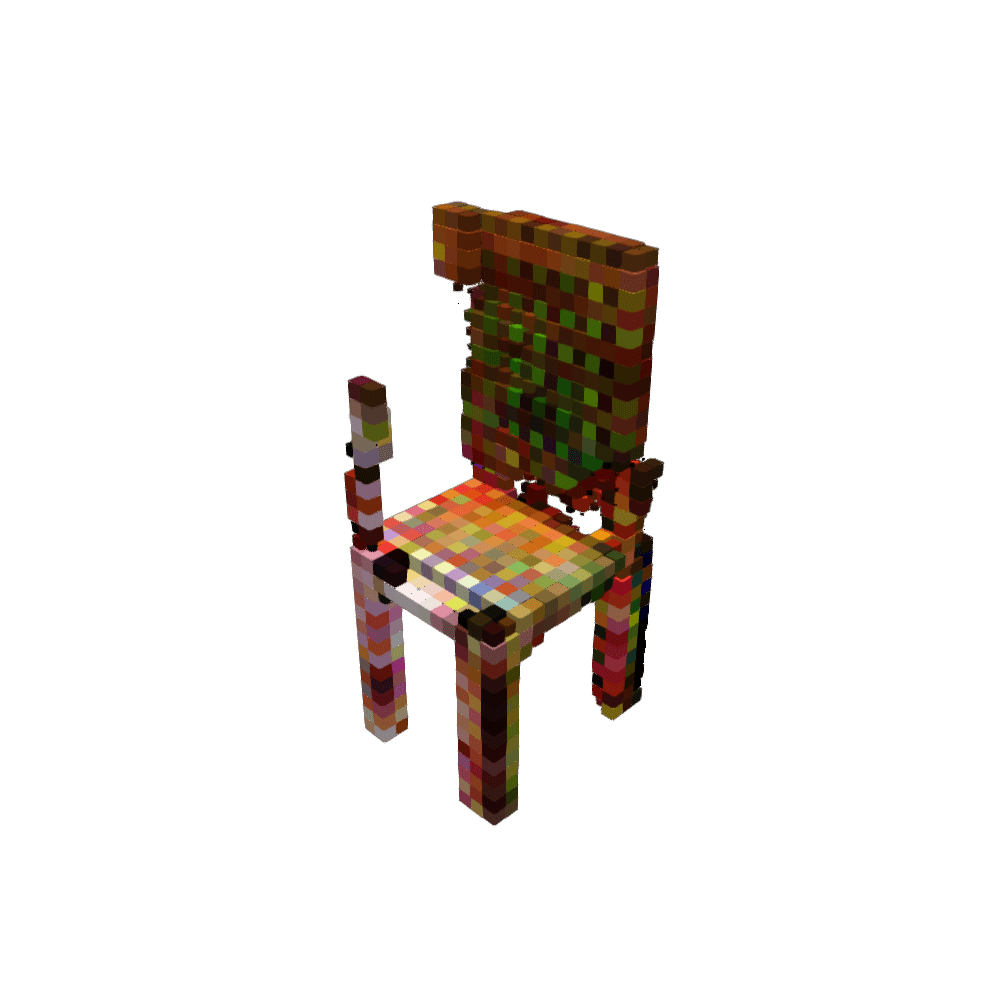} &
    \adjincludegraphics[width=\linewidth,trim={{.05\width} {.05\height} {.05\width} {.05\height}},clip]{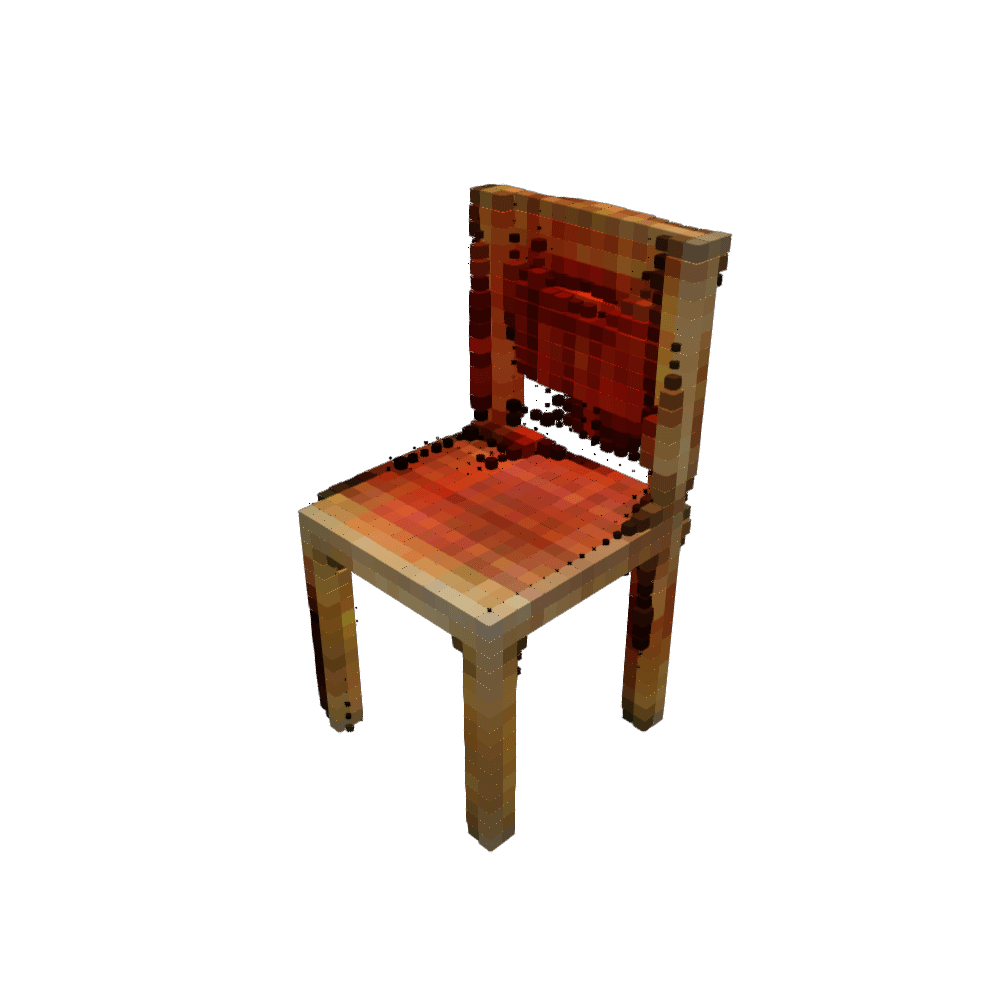} &
    \adjincludegraphics[width=\linewidth,trim={{.05\width} {.05\height} {.05\width} {.05\height}},clip]{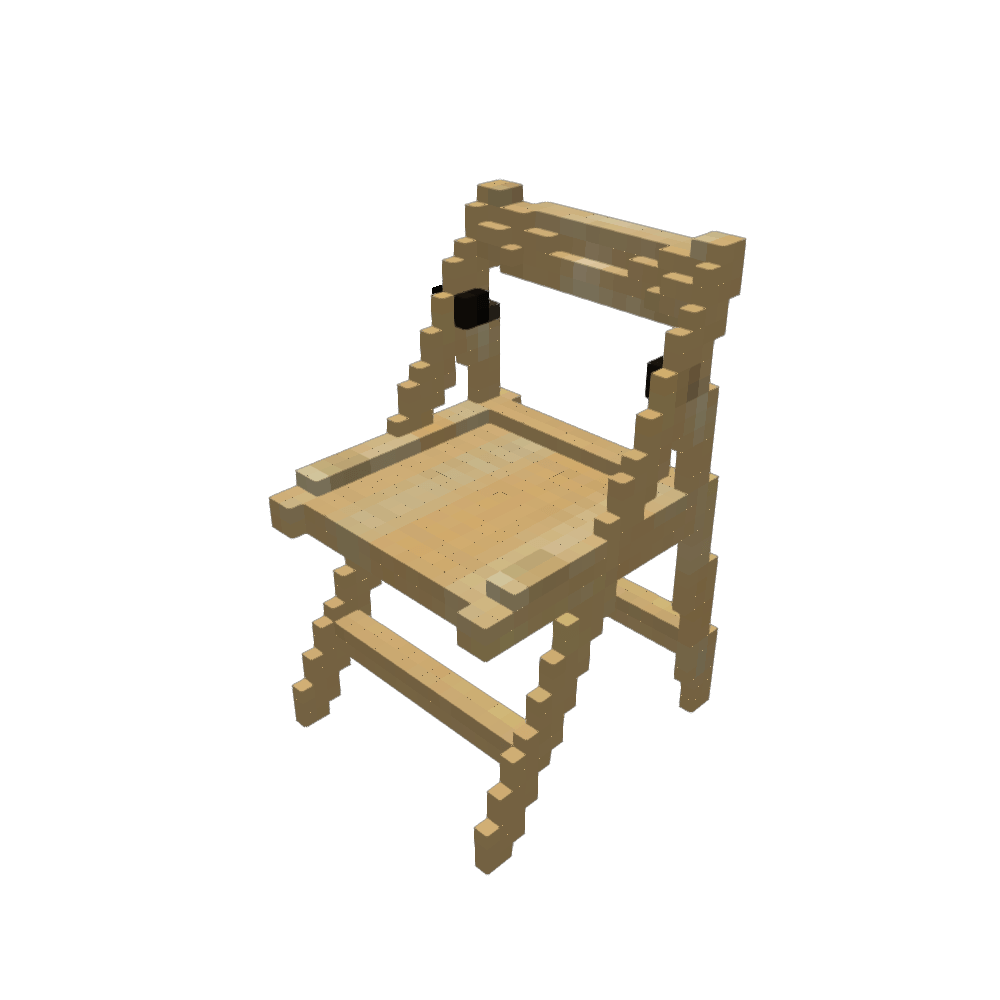}
    \\\midrule
    \textbf{16)} Modern, brown color, square shape, wooden table with a curve at bottom &
    \adjincludegraphics[width=\linewidth,trim={{.05\width} {.05\height} {.05\width} {.05\height}},clip]{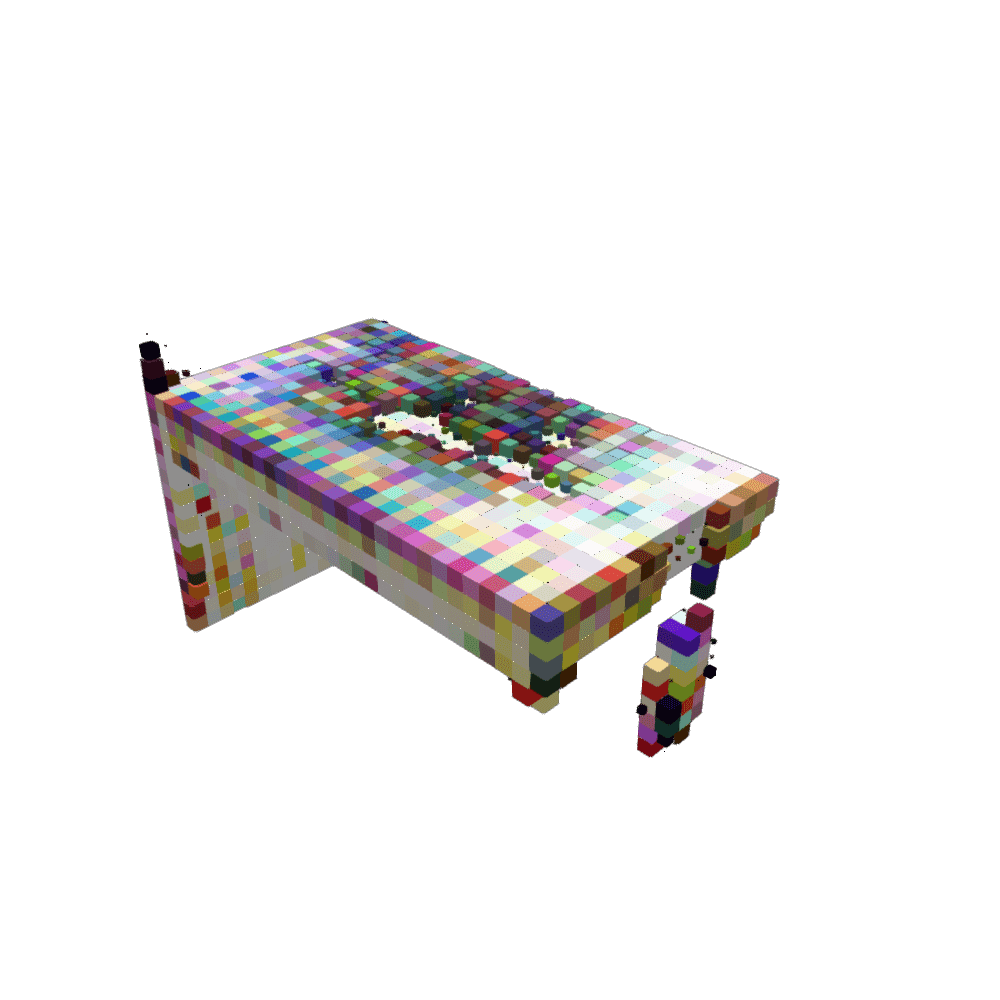} &
    \adjincludegraphics[width=\linewidth,trim={{.05\width} {.05\height} {.05\width} {.05\height}},clip]{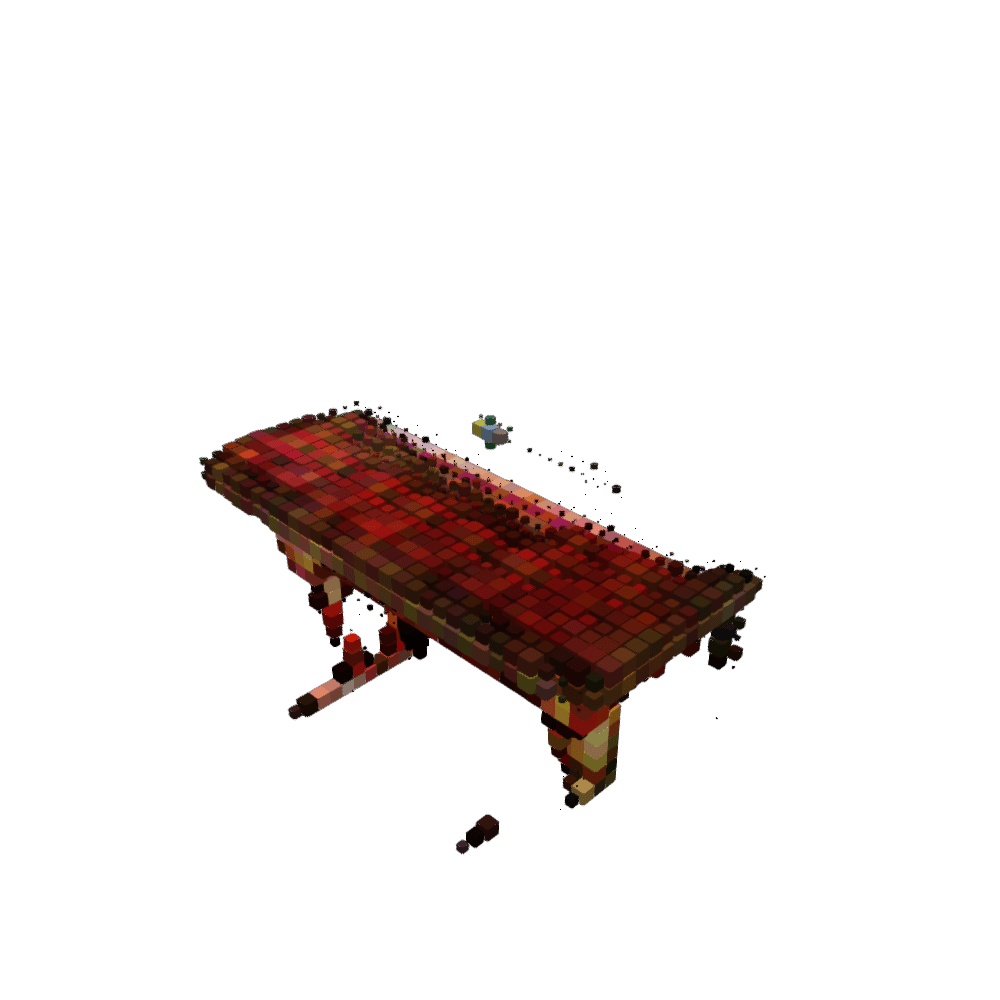} &
    \adjincludegraphics[width=\linewidth,trim={{.05\width} {.05\height} {.05\width} {.05\height}},clip]{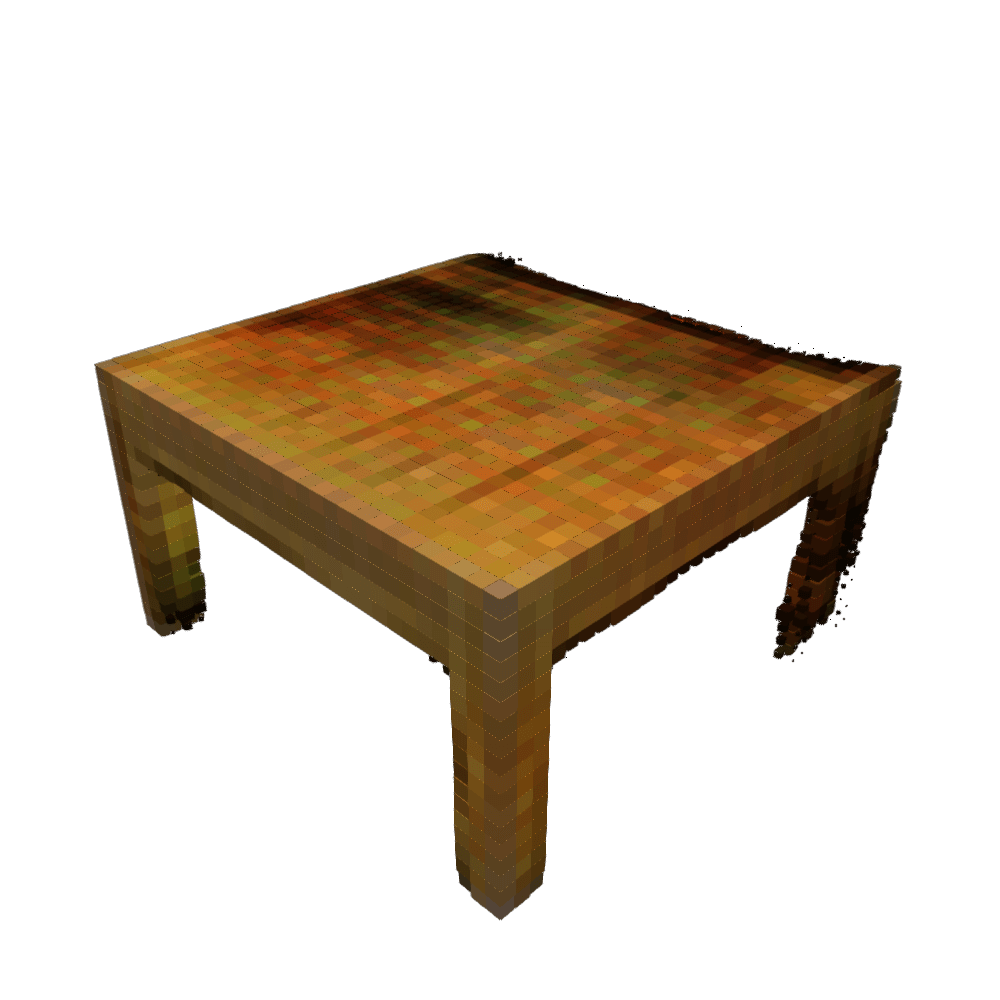} &
    \adjincludegraphics[width=\linewidth,trim={{.05\width} {.05\height} {.05\width} {.05\height}},clip]{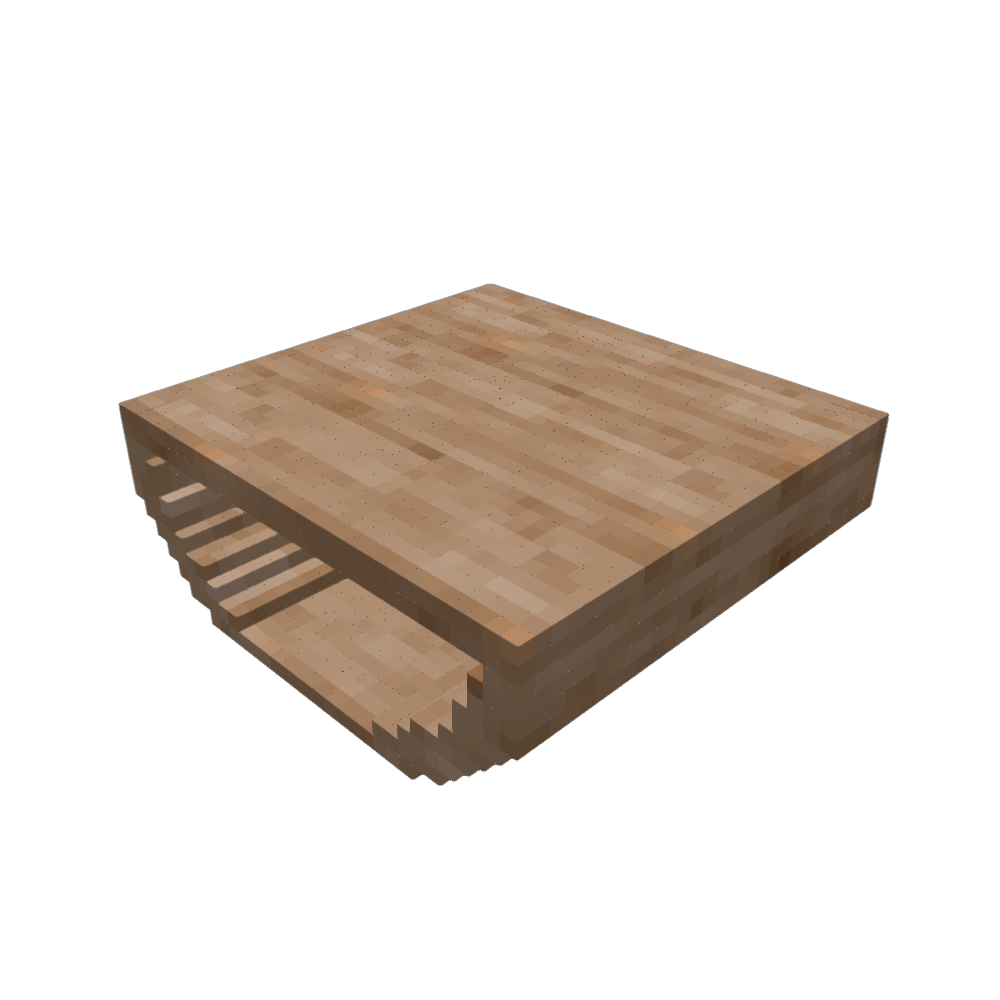}
    \\\midrule
    \textbf{17)} Wooden office desk with lamp. &
    \adjincludegraphics[width=\linewidth,trim={{.05\width} {.05\height} {.05\width} {.05\height}},clip]{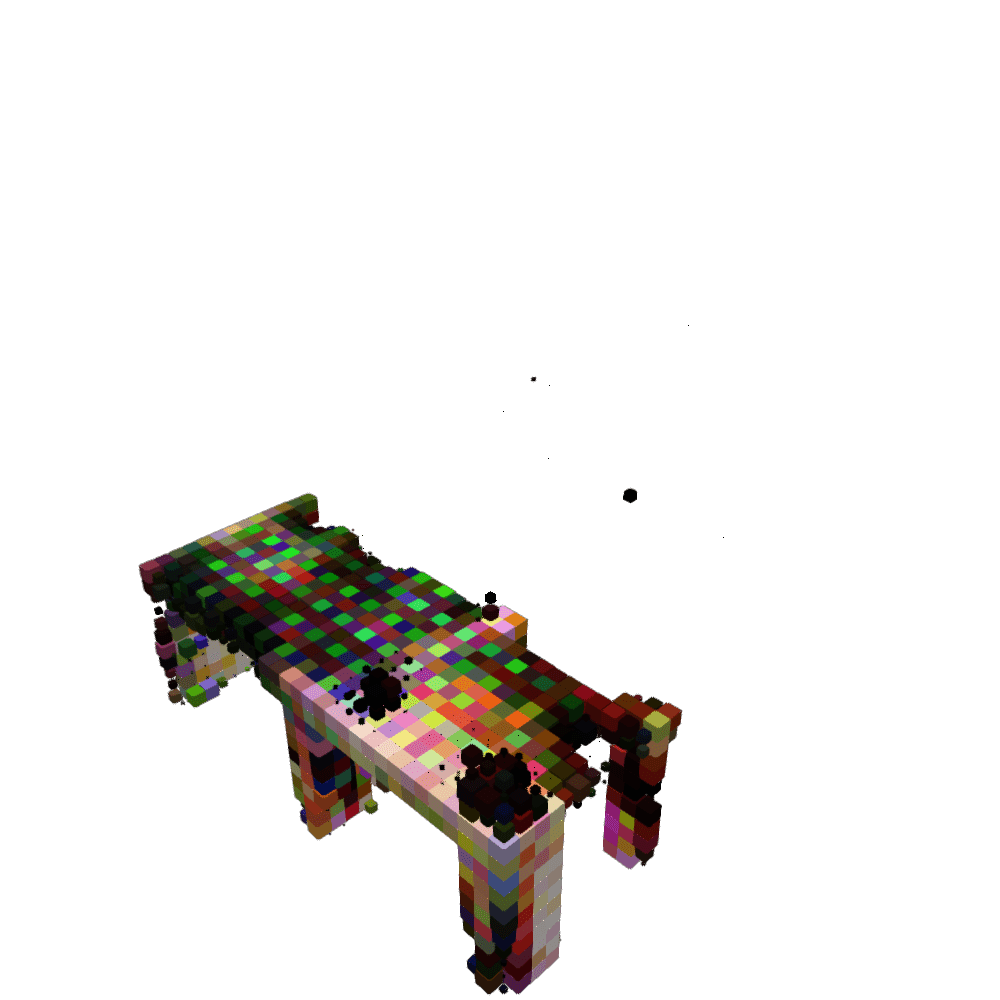} &
    \adjincludegraphics[width=\linewidth,trim={{.05\width} {.05\height} {.05\width} {.05\height}},clip]{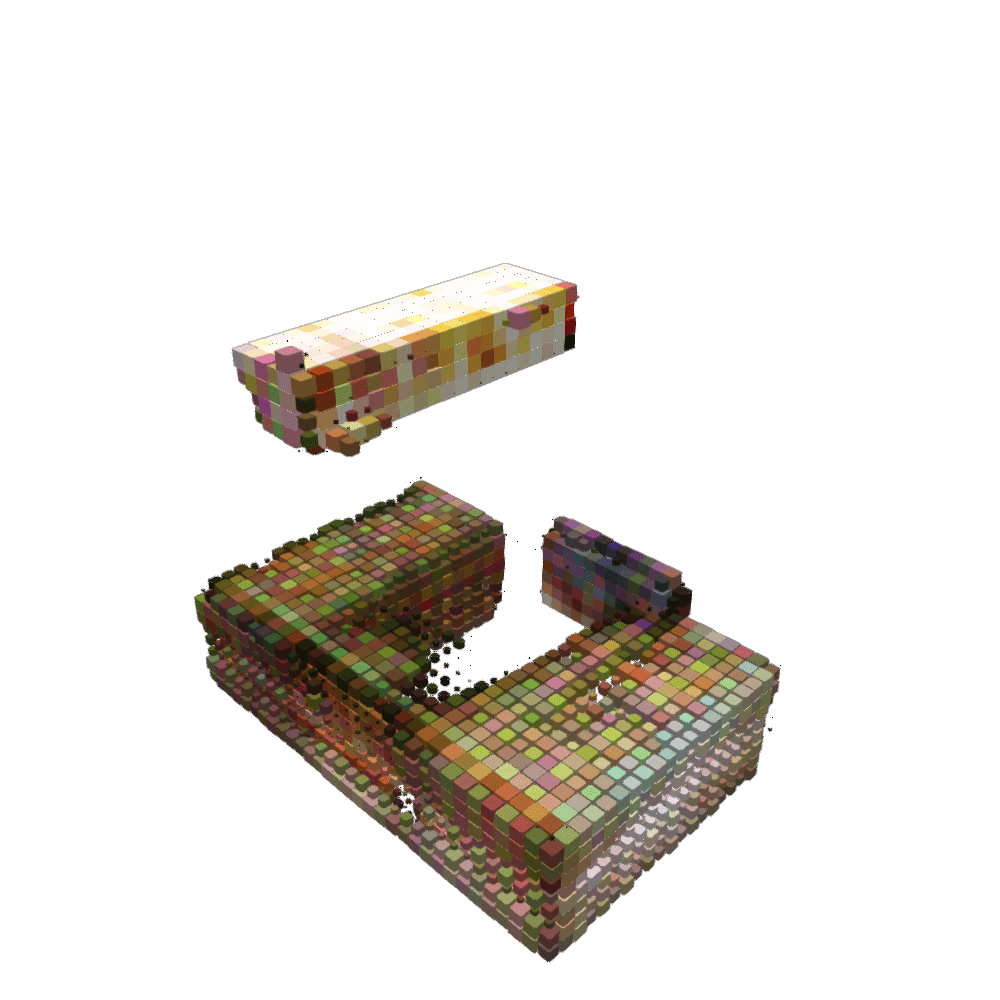} &
    \adjincludegraphics[width=\linewidth,trim={{.05\width} {.05\height} {.05\width} {.05\height}},clip]{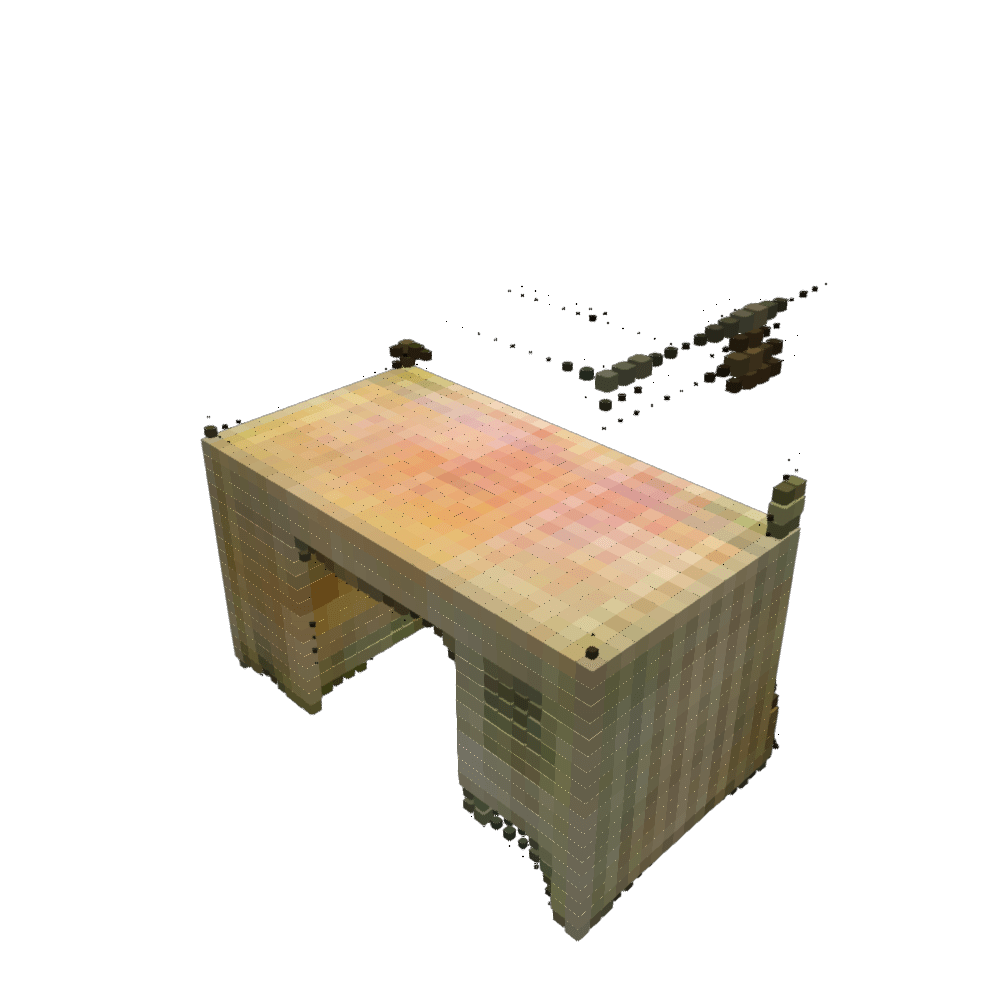} &
    \adjincludegraphics[width=\linewidth,trim={{.05\width} {.05\height} {.05\width} {.05\height}},clip]{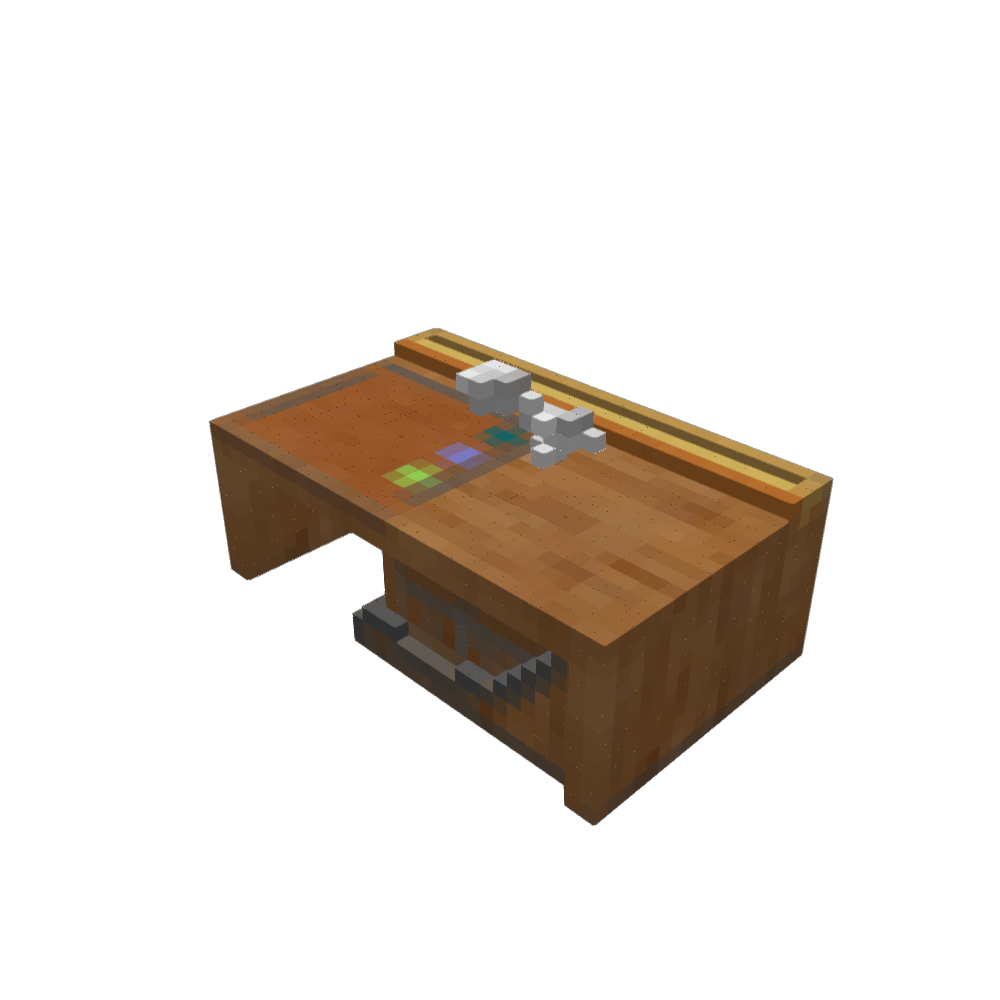}
    \\\midrule
    \textbf{18)} A rectangle glass table with a unique base design. &
    \adjincludegraphics[width=\linewidth,trim={{.05\width} {.05\height} {.05\width} {.05\height}},clip]{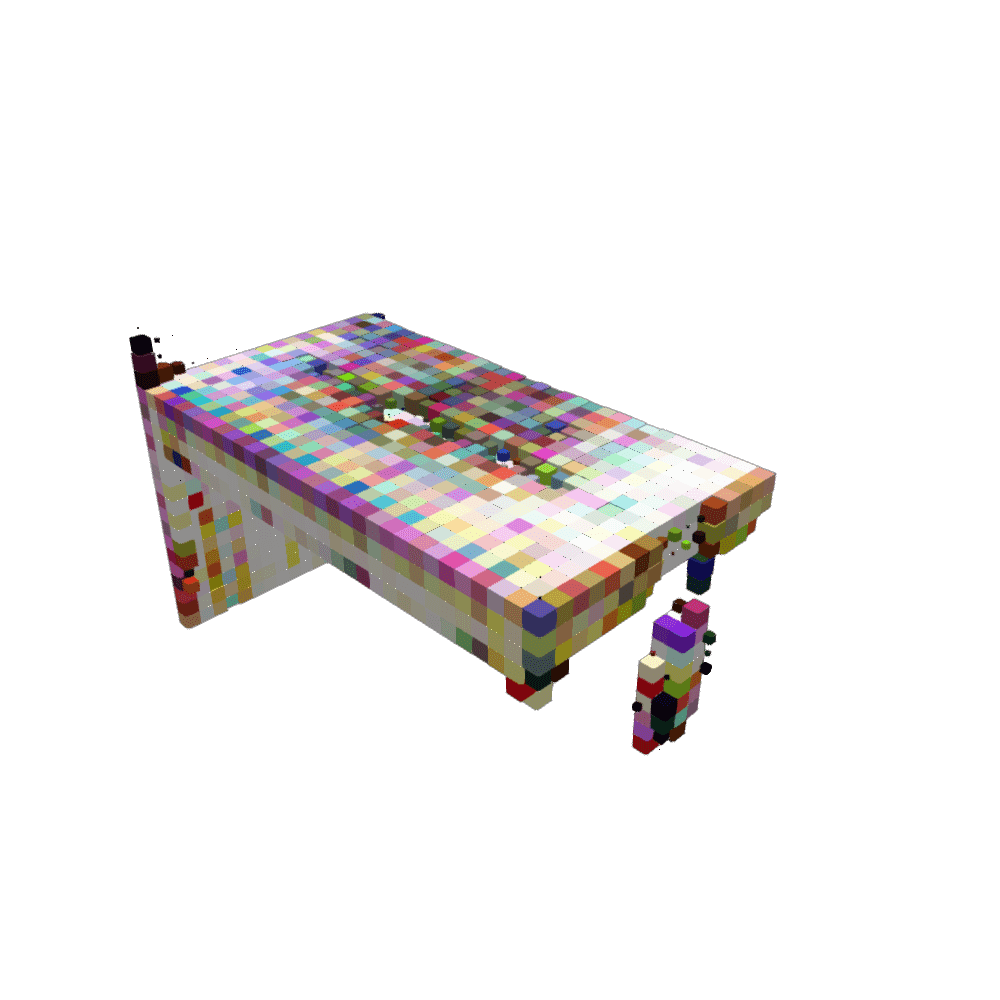} &
    \adjincludegraphics[width=\linewidth,trim={{.05\width} {.05\height} {.05\width} {.05\height}},clip]{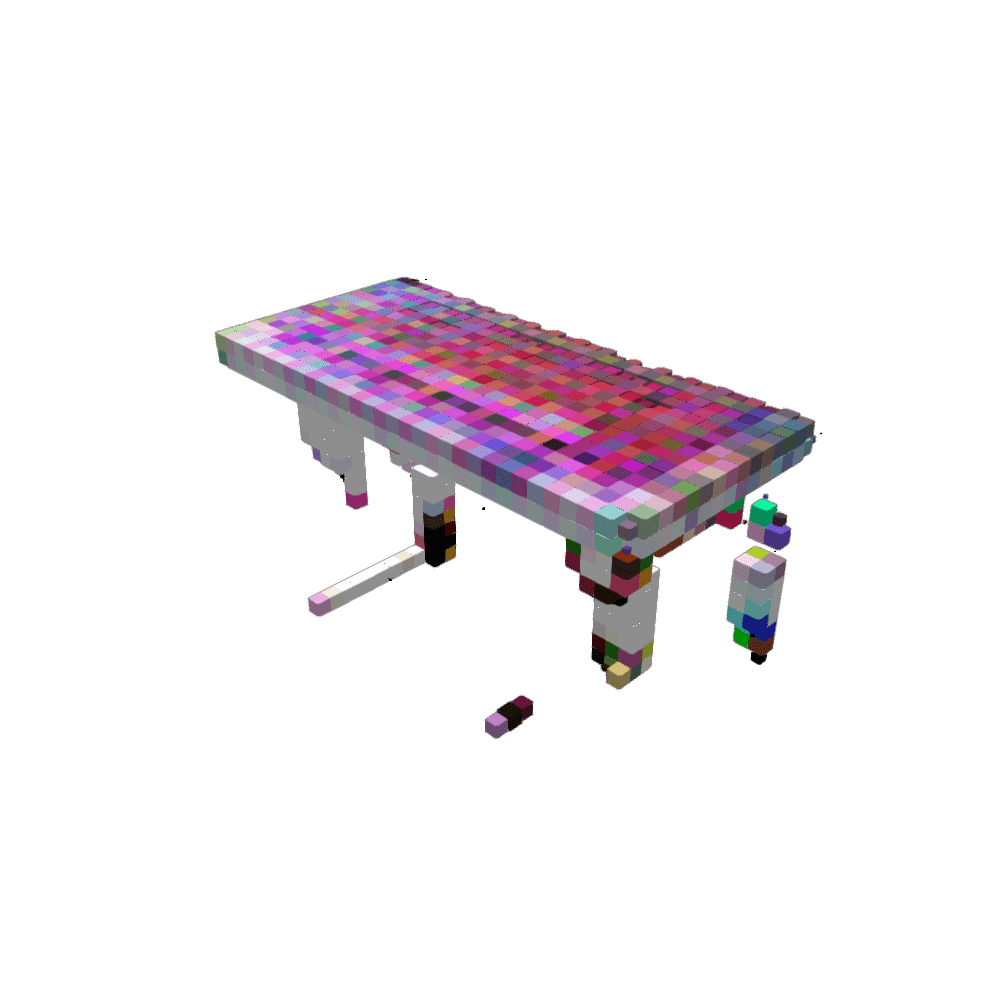} &
    \adjincludegraphics[width=\linewidth,trim={{.05\width} {.05\height} {.05\width} {.05\height}},clip]{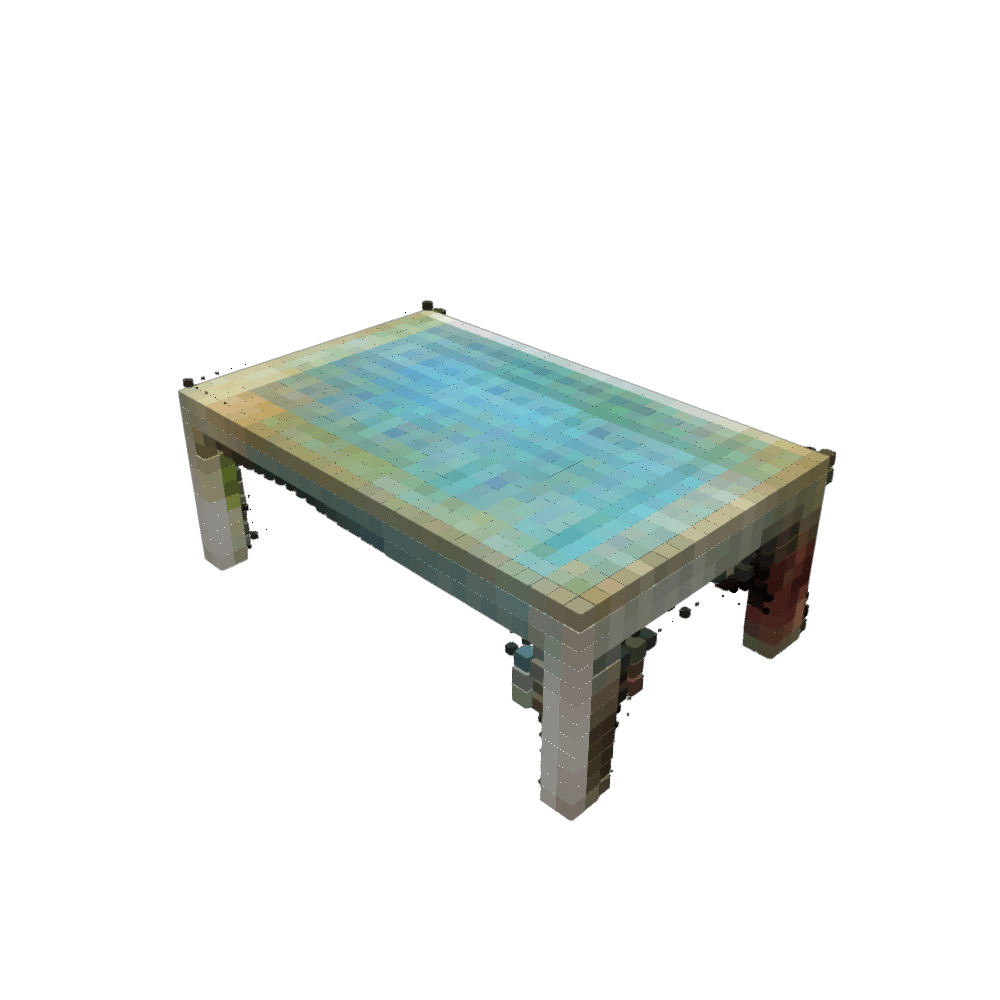} &
    \adjincludegraphics[width=\linewidth,trim={{.05\width} {.05\height} {.05\width} {.05\height}},clip]{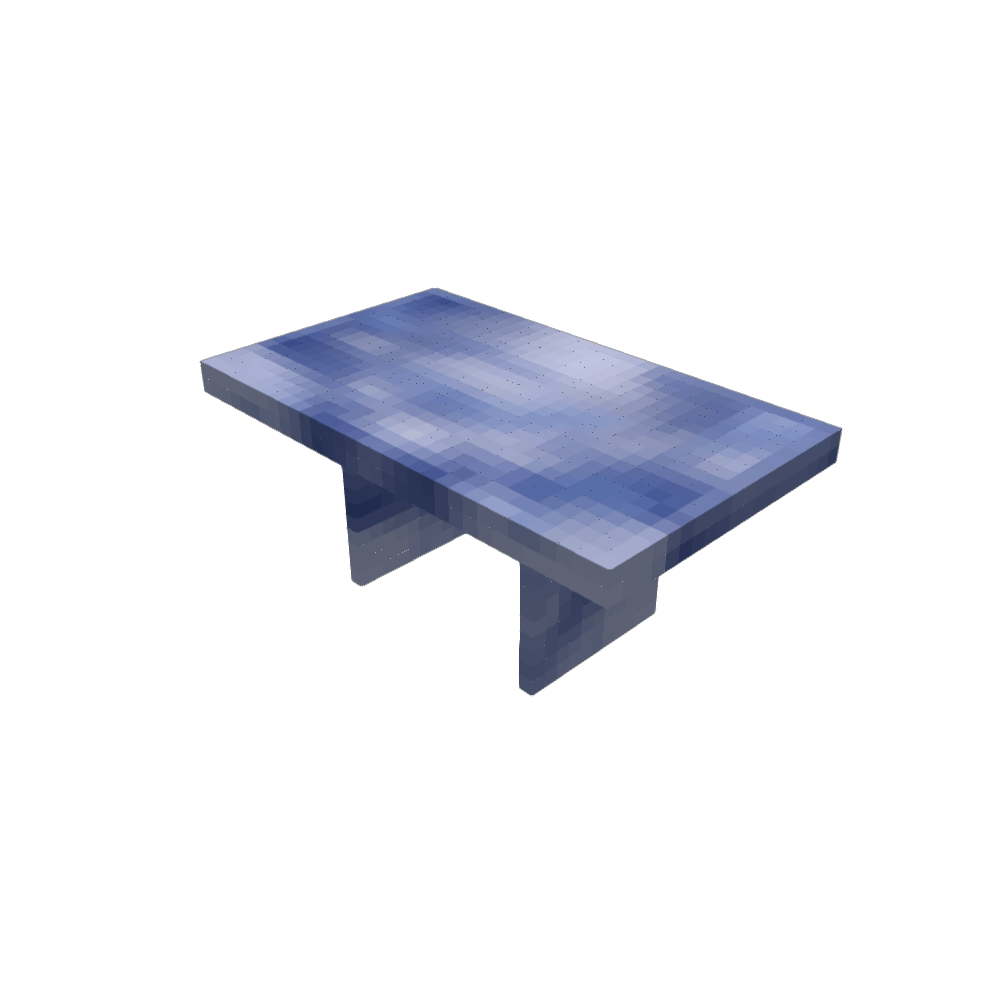}
    \\\midrule
    \textbf{19)} The object is grey in color. It is rectangular and flat in shape with four legs supporting it. Usual material is wood. &
    \adjincludegraphics[width=\linewidth,trim={{.05\width} {.05\height} {.05\width} {.05\height}},clip]{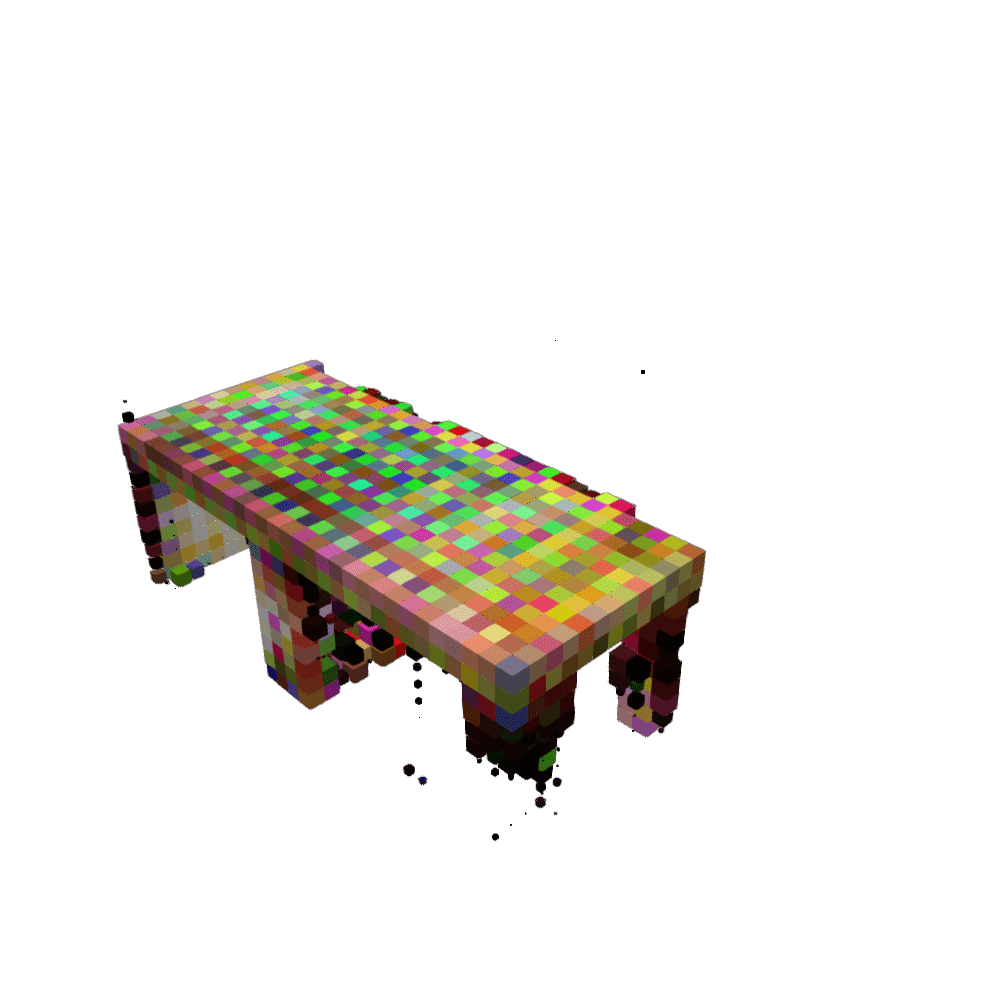} &
    \adjincludegraphics[width=\linewidth,trim={{.05\width} {.05\height} {.05\width} {.05\height}},clip]{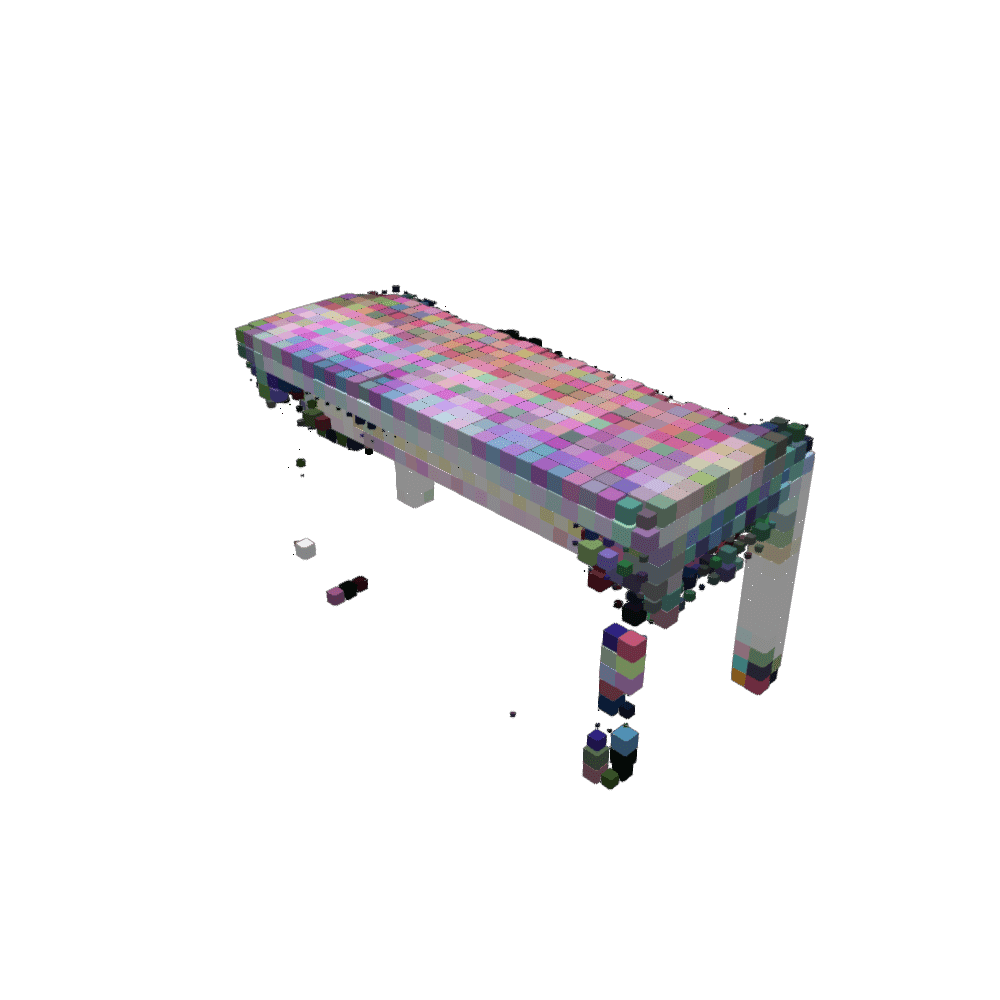} &
    \adjincludegraphics[width=\linewidth,trim={{.05\width} {.05\height} {.05\width} {.05\height}},clip]{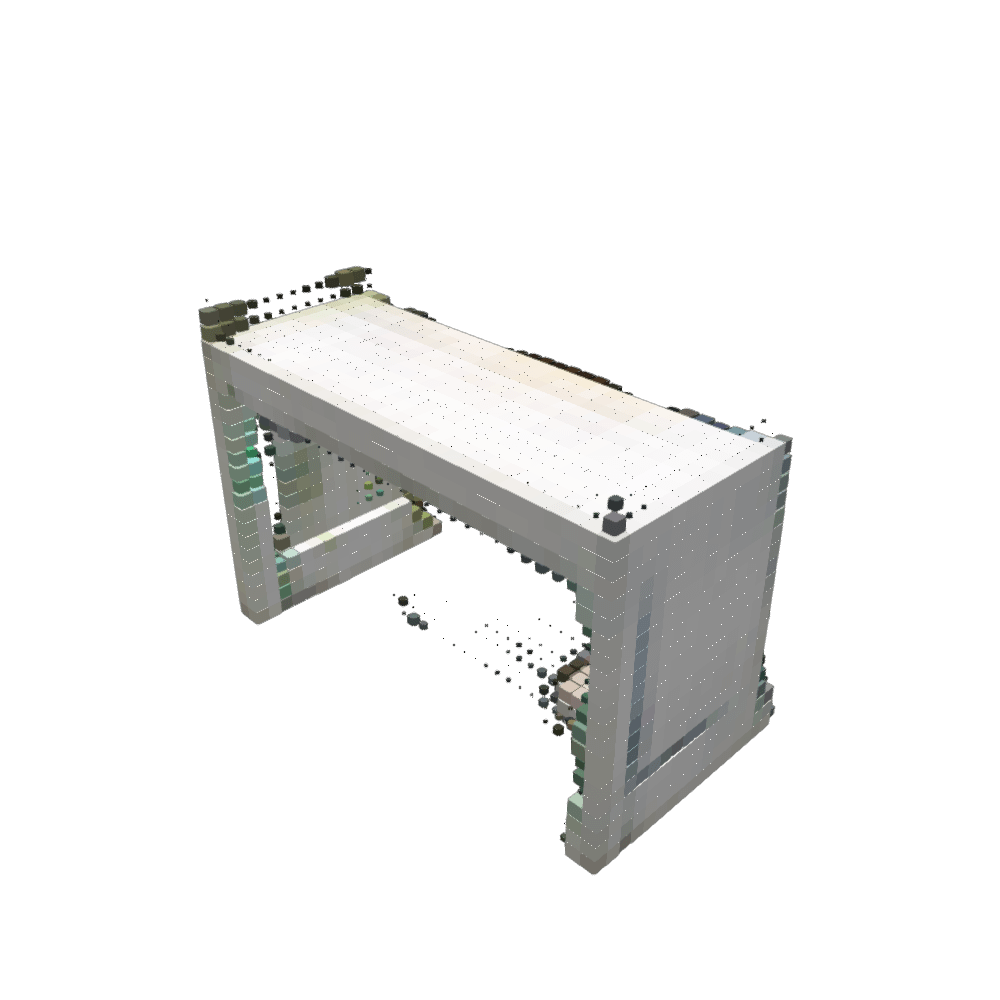} &
    \adjincludegraphics[width=\linewidth,trim={{.05\width} {.05\height} {.05\width} {.05\height}},clip]{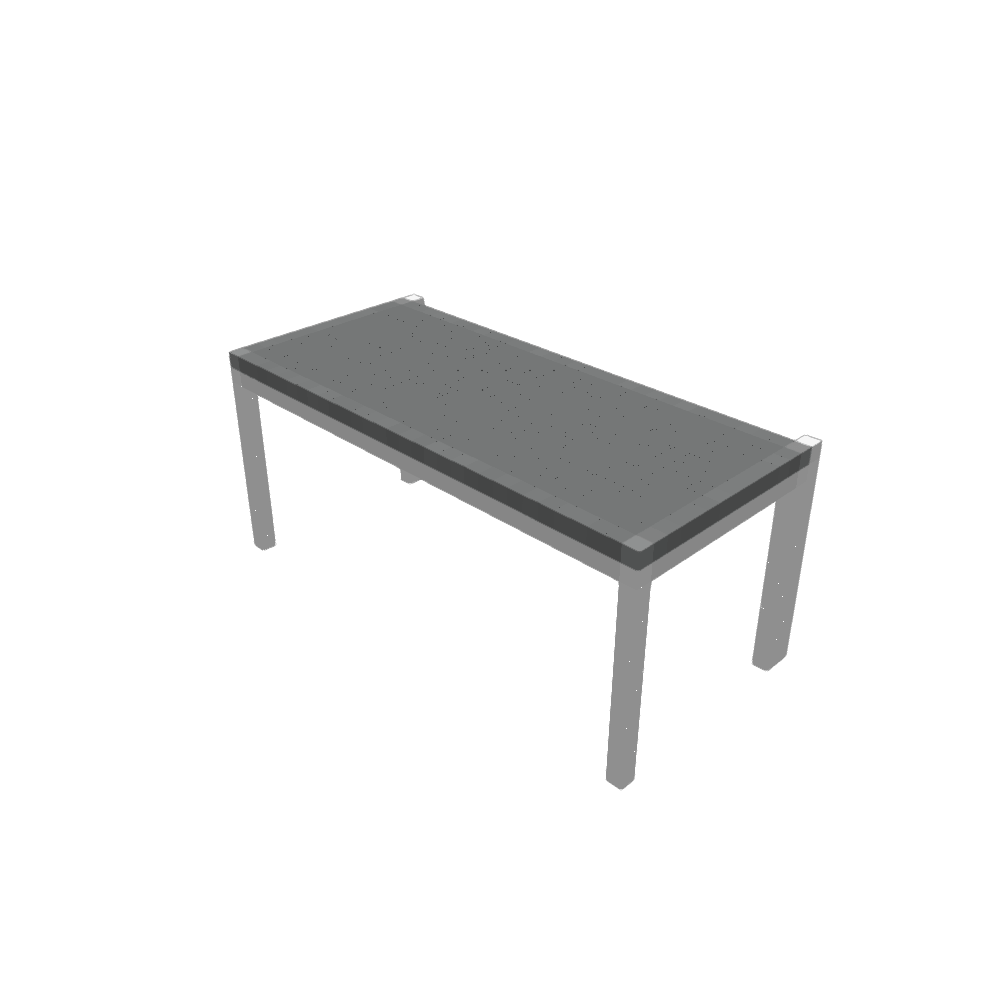}
    \\\midrule
    \textbf{20)} A wood square table with four legs. It is painted brown. &
    \adjincludegraphics[width=\linewidth,trim={{.05\width} {.05\height} {.05\width} {.05\height}},clip]{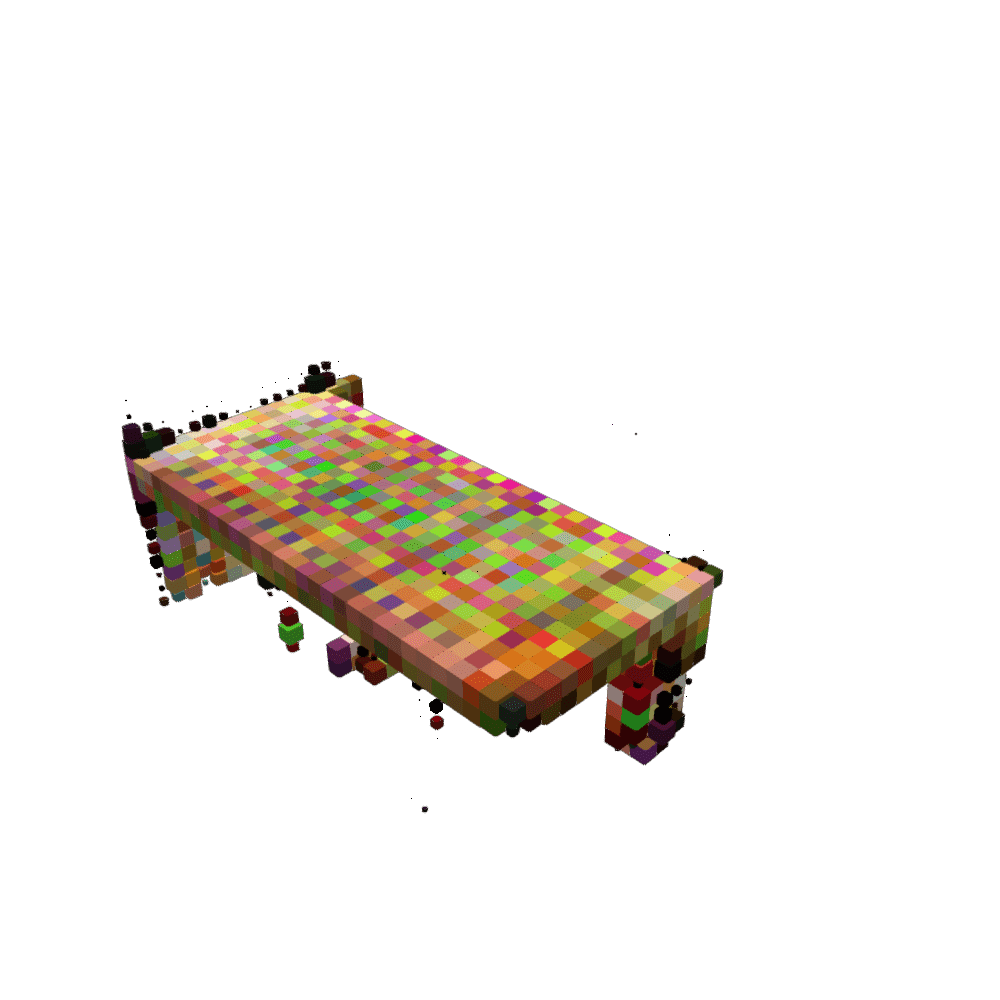} &
    \adjincludegraphics[width=\linewidth,trim={{.05\width} {.05\height} {.05\width} {.05\height}},clip]{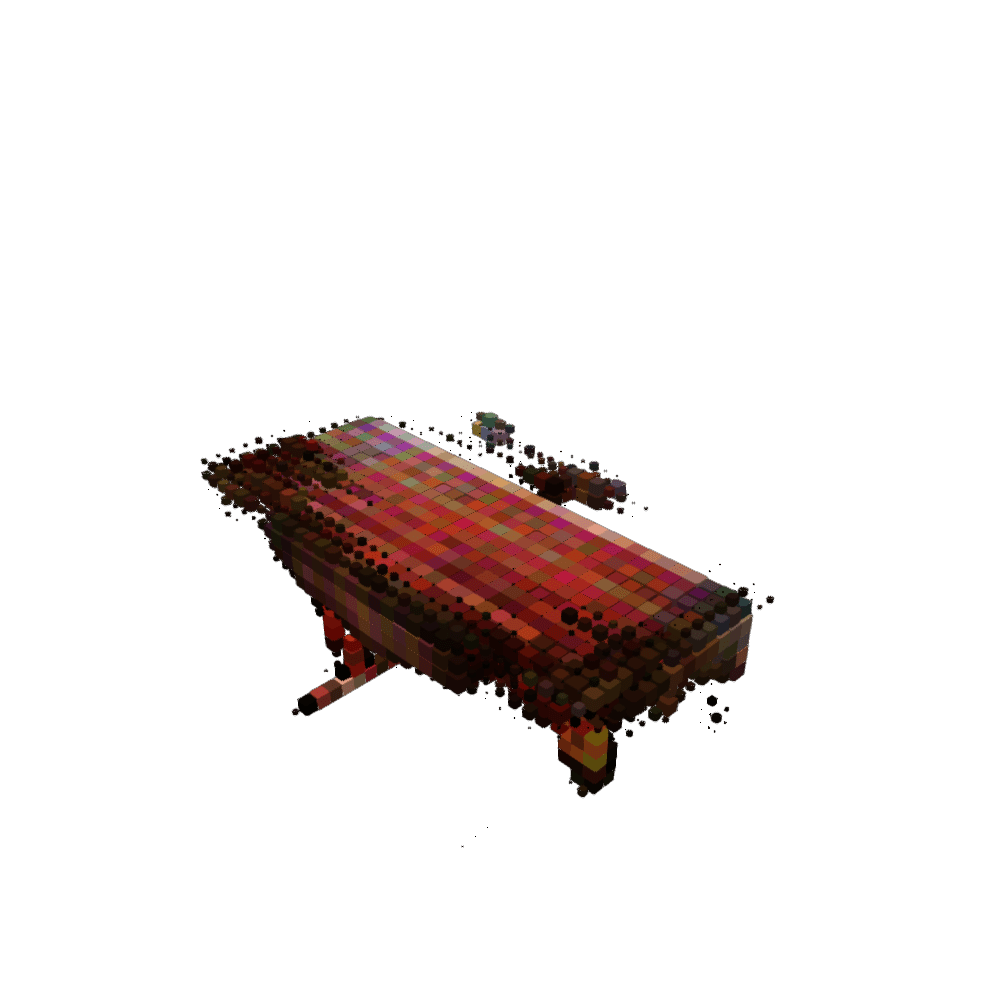} &
    \adjincludegraphics[width=\linewidth,trim={{.05\width} {.05\height} {.05\width} {.05\height}},clip]{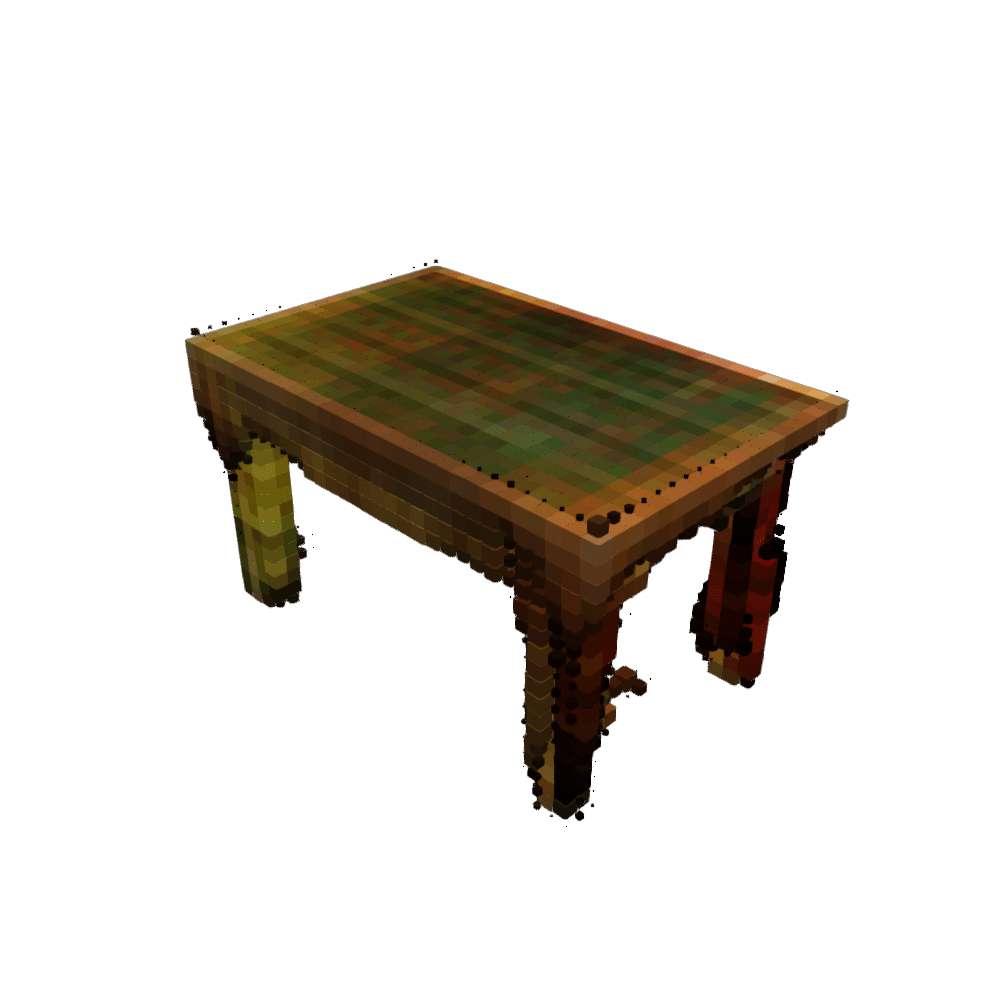} &
    \adjincludegraphics[width=\linewidth,trim={{.05\width} {.05\height} {.05\width} {.05\height}},clip]{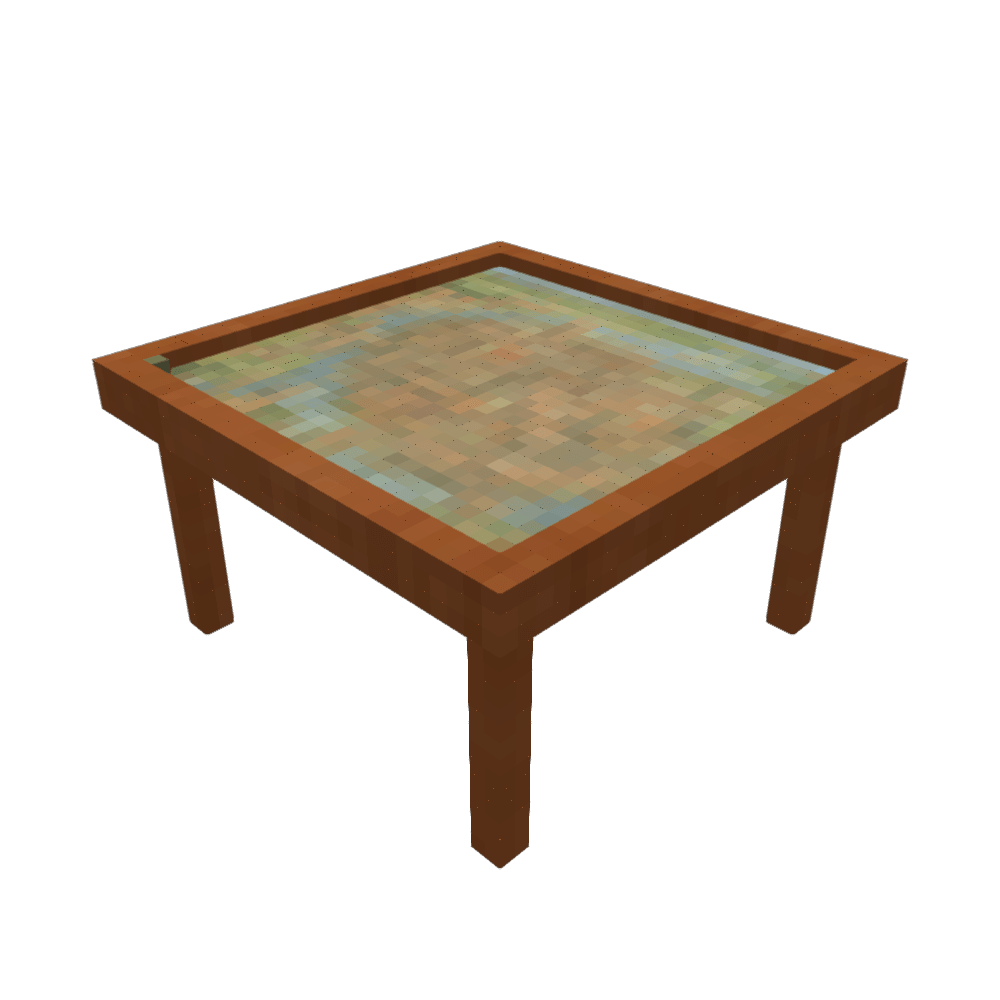}
    \end{tabular}
    \vspace{-1em}
    \caption{\textbf{Example text-to-shape generations}. Our CWGAN approach is correctly conditioned by the input text, generating shapes that match the described attributes. The baseline approach struggles to generate plausible shapes. Additional results can be found in the supplementary material.}
    \vspace{-1.5em}
    \label{fig:text2shape-results-supplementary-2}
\end{figure}

%% file: fig/vector-arithmetic-additional.tex
\begin{figure}
	\captionsetup[subfigure]{justification=centering,labelformat=empty}
	\centering
	\begin{subfigure}[t]{0.7\linewidth}
	    \includegraphics[width=\linewidth]{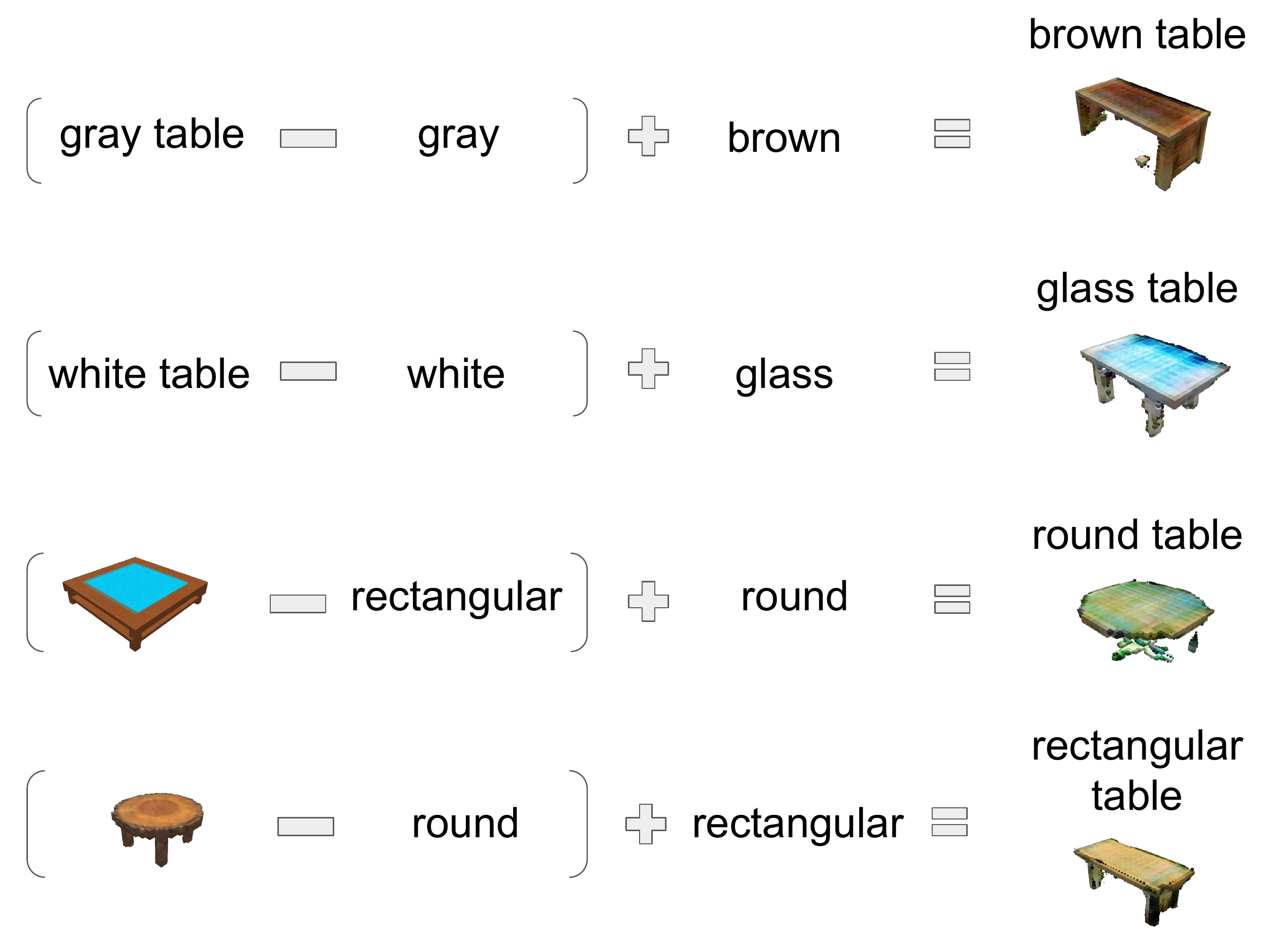}
	\end{subfigure}
	
	\hrule
	
	\begin{subfigure}[t]{0.7\linewidth}
		\centering\includegraphics[width=\linewidth]{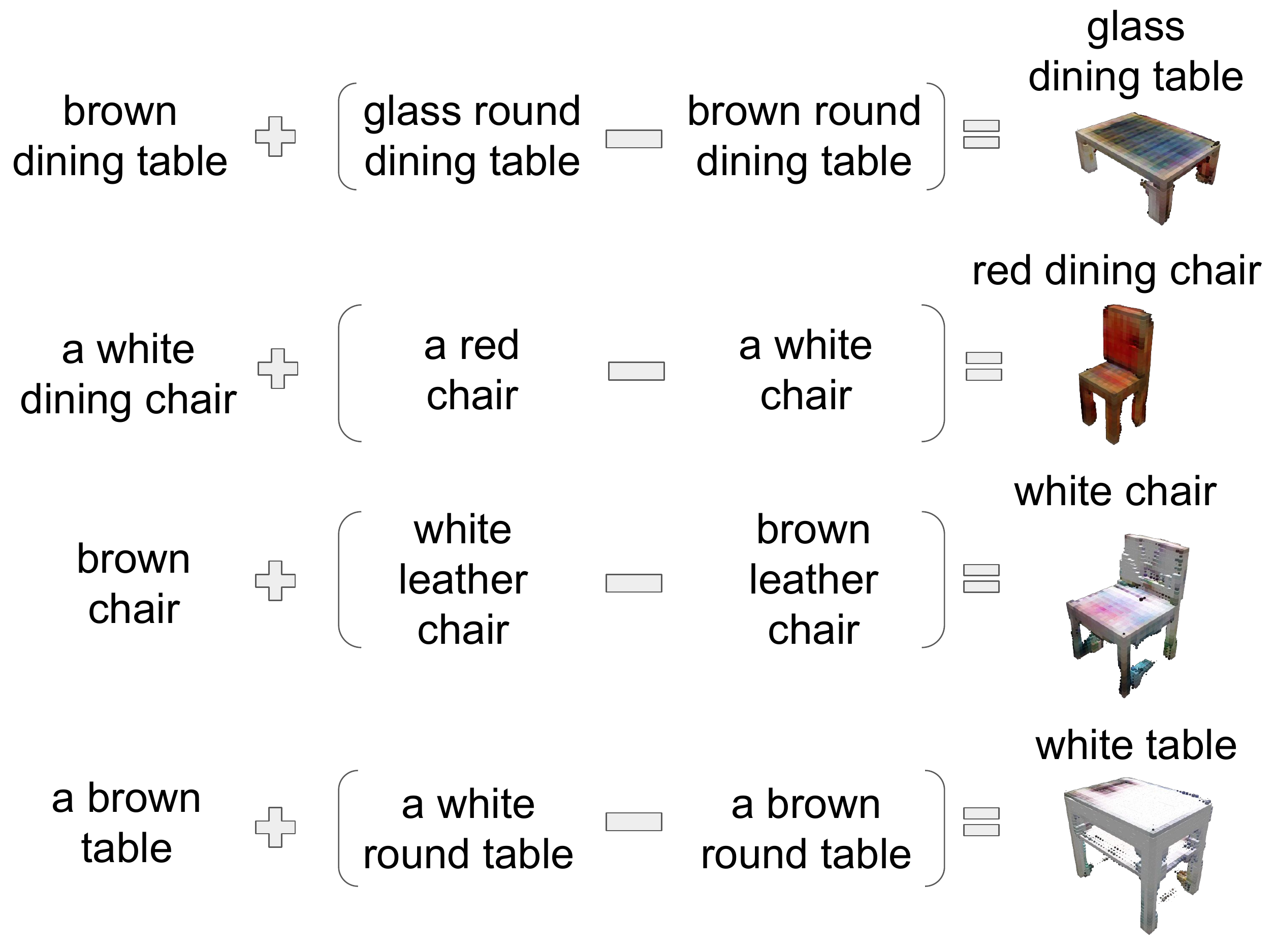}
	\end{subfigure}
	\vspace{-1em}
	
	\caption{\textbf{Vector arithmetic on joint text-shape embeddings}. Similar to the results from the main paper, our embeddings admit composition and transfer of attributes to control shape generation.}
	\label{fig:vector-arithmetic-additional}
\end{figure}